\documentclass[twoside,11pt]{article}

\usepackage{blindtext}
\usepackage{amsmath}
\usepackage{placeins}
\usepackage{enumitem}
\usepackage{arydshln}

\usepackage[preprint]{jmlr2e}

\usepackage{csvsimple}
\usepackage{subfig}
\usepackage{todonotes}
\usepackage{listings}
\usepackage{xcolor}
\usepackage{amsmath}
\usepackage{amssymb}
\usepackage{mathtools}
\usepackage{algorithm}
\usepackage{algpseudocode}
\usepackage{graphicx}
\usepackage{booktabs}
\usepackage{multirow}
\usepackage[most]{tcolorbox}

\usepackage{lastpage}
\jmlrheading{26}{2026}{1-\pageref{LastPage}}{2/26}{}{}{Samhruth Ananthanarayanan and Ayan Sengupta and Tanmoy Chakraborty}

\ShortHeadings{Physics of KV Compression in LLMs}{Samhruth Ananthanarayanan and Ayan Sengupta and Tanmoy Chakraborty}
\firstpageno{1}

\begin{document}

\title{Understanding the Physics of Key-Value Cache Compression for LLMs through Attention Dynamics}

\author{
       \name Samhruth Ananthanarayanan\thanks{Authors contributed equally.}
       \email samhruth@gmail.com \\
       \addr IIT Delhi, India\\
       \AND
       \name Ayan Sengupta\footnotemark[1]
       \email ayan.sengupta@ee.iitd.ac.in \\
       \addr IIT Delhi, India \\
       \AND
       \name Tanmoy Chakraborty 
       \email tanchak@iitd.ac.in \\
       \addr IIT Delhi, India
       }
       
\editor{My editor}

\maketitle

\begin{abstract}
As the context window in large language models (LLMs) expands into the hundreds of thousands of tokens, the key--value (KV) cache has become the dominant memory bottleneck in autoregressive decoding, motivating a surge of KV compression methods that report 80\%--90\% memory savings with minimal degradation on standard long-context benchmarks. We argue that such evaluations obscure a deeper structural issue: attention functions not only as a storage mechanism but as a routing mechanism, and retaining key–value pairs does not guarantee semantic accessibility during generation. We introduce a physics-inspired framework that treats KV compression as a controlled perturbation of token-level routing within self-attention and distinguish between information retention, accessibility, and utilization. Using a suite of controlled synthetic datasets probing multi-entity tracking, instance disambiguation, coreference consistency, and multi-hop reasoning, we uncover three key empirical phenomena. First, moderate compression substantially degrades internal representations while leaving task accuracy largely intact, revealing large redundancy margins. Second, all evaluated architectures exhibit a sharp hallucination ``safety cliff'' near 90\% compression, strongly correlated with a spike in our proposed Global Eviction Ratio (GER), indicating a phase transition in semantic reachability when answer-critical tokens are globally erased. Third, architectural differences produce distinct routing dynamics: LLaMA models exhibit early-layer consensus and late diversification, while Qwen models show funnel-like behavior with late-stage convergence, leading to different compression resilience profiles. Beyond erasure, we identify a second failure mode -- representational rigidity -- where tokens survive but excessive head-level consensus collapses routing flexibility, degrading performance despite nominal retention. Together, these findings provide empirical evidence for sparse token-route structures whose survival governs compression tolerance, reframing KV compression as a structural probe of attention geometry, and linking long-context scalability with sparsity and the ``lottery ticket hypothesis'' in self-attention.
\end{abstract}


\section{Introduction}

The recent evolution of large language models (LLMs) has been defined as much by context length as by parameter scale~\citep{grattafiori2024LLaMA, yang2025qwen251mtechnicalreport, geminiteam2024gemini15unlockingmultimodal}. Context windows have expanded from a few thousand tokens to hundreds of thousands and beyond, enabling document-level reasoning, repository-scale code analysis, and persistent conversational memory~\citep{liu2025comprehensive}. This trend mirrors earlier scaling laws in parameters and data~\citep{kaplan2020scaling, hoffmann2022training, JMLR:v26:24-1000}, reinforcing the intuition that larger horizons yield greater capability. Yet unlike parameters, which are shared across tokens, the key–value (KV) cache grows linearly with sequence length during autoregressive decoding. For a transformer with $L$ layers, $H_{KV}$ heads of dimension $D_h$, batch size $B$, sequence length $S$, and precision $b$, the memory footprint scales as
\begin{equation}
\mathrm{Memory} = B \times S \times L \times H_{KV} \times D_h \times b.
\label{eq:KVCacheFormula}
\end{equation}
As $S$ increases, KV storage rapidly dominates inference memory, often exceeding parameter storage itself. In practice, memory, not compute, becomes the principal bottleneck~\citep{meng2025understanding} for long-context LLMs.

KV cache compression~\citep{li2024survey} emerged as an engineering solution to this constraint. Methods based on quantization~\citep{hooper2024kvquant, liu2024kivi, kang2024gear}, token eviction~\citep{wan2025textdtexto, cai2025pyramidkv, li_snapkv_2024}, adaptive head pruning~\citep{feng_ada-kv_2025}, or clustering~\citep{liu2025clusterkvmanipulatingllmkv, zhang-etal-2025-clusterattn} report striking results: up to 80\%–90\% KV reduction with limited degradation on benchmarks such as LongBench and RULER~\citep{bai2023longbench, hsieh2024ruler, zhang-etal-2024-bench}. These findings have fostered a prevailing assumption: that the KV cache contains substantial redundancy, and that careful pruning can remove most of it without altering model behavior. We argue that this assumption overlooks a deeper structural question concerning attention as a routing system rather than a passive memory store.

Self-attention constructs dynamic token-level computational pathways across heads and layers~\citep{clark2019does, voita2019bottom, elhage2021mathematical, olsson2022context}. Induction head analyses and circuit-level studies demonstrate that reasoning depends on structured cross-layer routes rather than isolated token storage~\citep{elhage2021mathematical, olsson2022context}. From this perspective, retaining a key–value pair does not guarantee functional accessibility; tokens may remain stored yet become unreachable~\citep{lee2025understanding} if routing pathways are severed. Conversely, theoretical work on sparse subnetworks and the ``Lottery Ticket Hypothesis'' suggests that a relatively small structured subset of attention components may suffice to preserve behavior when critical routes remain intact~\citep{frankle2018the, otsuka2025strong}. These insights motivate a reframing of KV compression as a structural perturbation of semantic reachability.

\begin{figure*}[!htb]
    \centering
    \includegraphics[width=\linewidth]{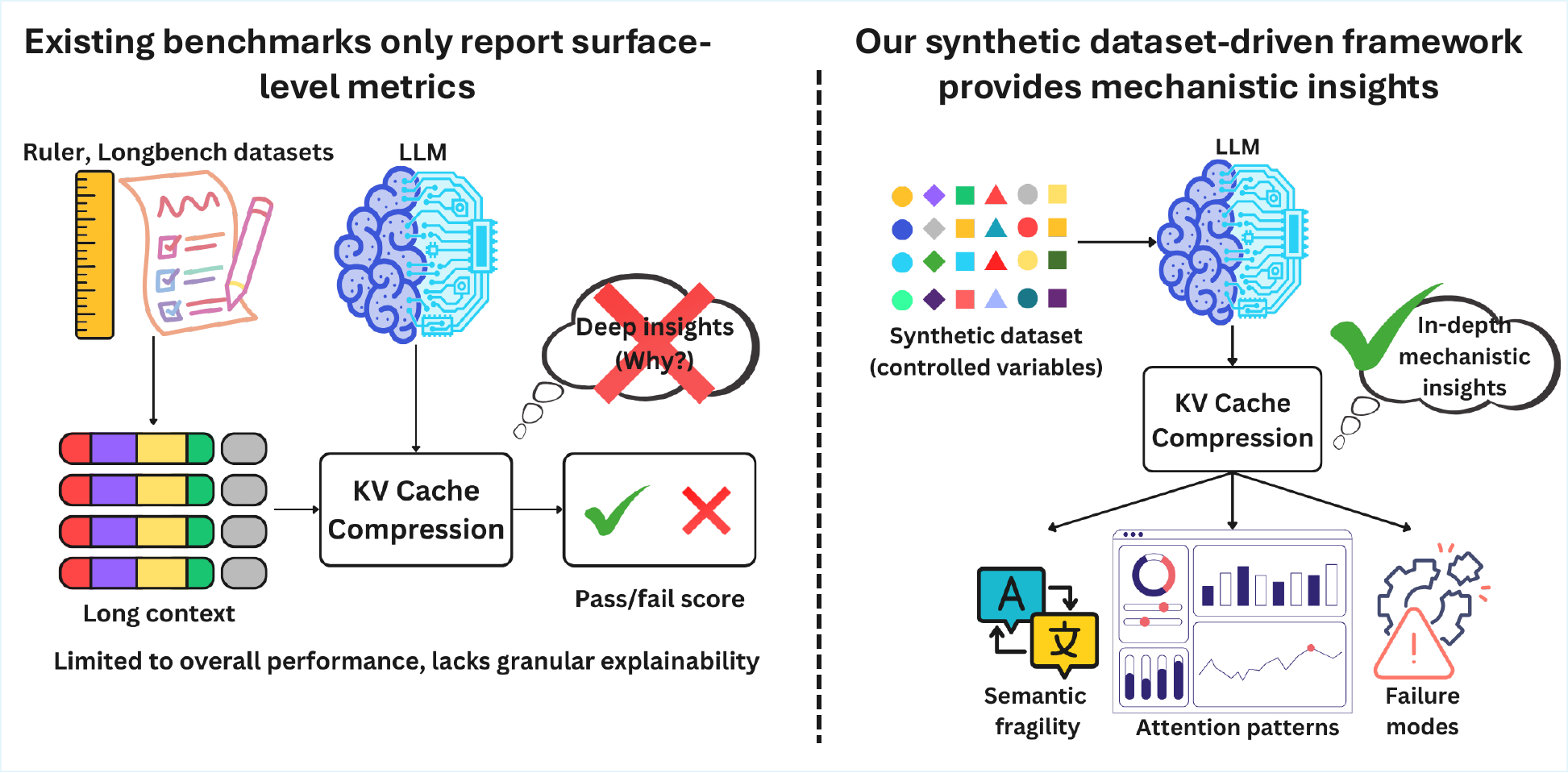}
    \caption{Comparison of traditional versus synthetic evaluation frameworks for KV cache compression. Existing benchmarks (left) primarily report coarse-grained task accuracy, whereas our synthetic, controlled framework (right) probes structural reachability, routing collapse, and semantic fragility under compression.}
    \label{fig:schematic}
\end{figure*}

This leads to our central hypothesis: within dense attention layers exist sparse \emph{token-route lottery tickets} (TR-LTs) -- minimal cross-head and cross-layer pathways that preserve semantic reachability necessary for correct generation. Mild compression may expose these latent routes, while extreme compression may destroy them. If true, KV compression should induce not merely gradual degradation, but a structural phase transition in semantic accessibility~\citep{ma2026phase, ersoy2025phase}.

\begin{tcolorbox}[colback=white,colframe=blue!50,title=Central Research Question]
Does KV compression merely remove redundant memory, or does it reveal and eventually destroy latent sparse token routes that govern reasoning?
\end{tcolorbox}

To investigate this question, we develop a physics-inspired evaluation framework~\citep{AllenZhu-icml2024-tutorial} built upon controlled synthetic datasets and structural metrics (Figure~\ref{fig:schematic}). Rather than evaluating only answer accuracy, we introduce reachability-aware measures such as the Global Eviction Ratio (GER), which quantifies cross-head erasure of answer-critical tokens, and head-level consensus, which measures routing coordination and flexibility. Across multiple architectures (LLaMA and Qwen families) and compression levels up to 90\%, several striking patterns emerge.

First, moderate compression induces substantial representational degradation while leaving task accuracy largely intact, revealing large redundancy margins. Second, near 90\% compression, all models exhibit a sharp \emph{safety cliff} in hallucination rates, strongly correlated with spikes in GER, indicating a reachability threshold. Third, architectural depth dynamics differ fundamentally: LLaMA models exhibit an inverted funnel-shaped pattern (early consensus, late diversification), whereas Qwen models exhibit a funnel-like pattern (early exploration, late convergence), implying architecture-dependent sparsity allocation. Fourth, probing analyses reveal a semantic hierarchy in compression robustness, with subject entities like `Person' and `Thing' representations remaining stable, while `Organization', `Location', and especially `Creature' degrade rapidly.

\begin{tcolorbox}[colback=white,colframe=blue!50,title=Main Finding]
KV compression induces a structural phase transition in semantic reachability. Reasoning persists while sparse token-route lottery tickets survive; collapse occurs when global eviction or excessive consensus destroys routing flexibility.
\end{tcolorbox}

These findings expose two distinct failure modes under compression: (i) \emph{representational erasure}, where answer tokens are globally evicted across all heads, and (ii) \emph{representational rigidity}, where tokens survive, but excessive head consensus prevents flexible utilization. Compression tolerance, therefore, depends not on raw token count but on effective route capacity within attention.

By connecting empirical compression dynamics with sparsity theory in self-attention~\citep{huang2022lotterytickethypothesisselfattention, otsuka2025on, otsuka2025strong}, this work reframes KV compression from an engineering optimization problem into a structural probe of attention. Rather than asking how much of the KV cache can be discarded without harming accuracy, we ask a deeper question: \textit{what minimal sparse routing structure must survive for reasoning to remain possible?} The answer, we argue, lies in understanding attention as a fragile yet powerful network of latent token-route lottery tickets embedded within dense transformers.

\section{Background}

\subsection{Self-Attention and the KV Cache in Autoregressive Decoding}

Transformers~\citep{vaswani_attention_2017-1} are built upon the self-attention mechanism. Given an input sequence $X \in \mathbb{R}^{S \times d}$, each layer computes queries, keys, and values:
\[
Q = XW_Q, \quad K = XW_K, \quad V = XW_V,
\]
where $W_Q, W_K, W_V \in \mathbb{R}^{d \times d_k}$.
Scaled dot-product attention is then defined as
\[
\mathrm{Attention}(Q,K,V)
=
\mathrm{softmax}\!\left(\frac{QK^\top}{\sqrt{d_k}}\right)V.
\]
In multi-head attention (MHA), this operation is replicated across $H$ heads and concatenated. In autoregressive decoding, recomputing $K$ and $V$ for all previous tokens at each time step would incur $\mathcal{O}(S^2)$ cost. 
Instead, models cache previously computed keys and values, forming the \emph{KV cache}. 
At decoding step $t$, attention is computed between the current query $q_t$ and all stored keys $K_{1:t}$, while values $V_{1:t}$ are reused. Unlike parameters, which are shared across tokens, the KV cache grows linearly in $S$ and $B$ as seen in Equation \ref{eq:KVCacheFormula}, making it the dominant memory consumer in long-context inference \citep{li2024survey}. 

\subsection{Efficient Attention Mechanisms}

The quadratic $\mathcal{O}(S^2)$ time and memory complexity of vanilla self-attention has motivated a broad spectrum of efficiency-oriented architectures that trade off exactness, expressivity, or inductive bias for improved scalability \citep{sun2026efficientattentionmechanismslarge}. One line of work enforces structured sparsity by restricting attention patterns to predefined or learned subsets of tokens, thereby reducing pairwise interactions; representative examples include Sparse Transformers \citep{child2019generating}, Longformer \citep{beltagy_longformer_2020}, and BigBird \citep{zaheer2020big}, which exploit locality, block structure, or random connectivity to preserve partial global context while lowering cost. A second family approximates the attention matrix itself: low-rank projections such as Linformer \citep{wang2020linformer} reduce dimensionality prior to softmax computation, while kernelized or feature-map approaches such as Performer \citep{choromanski2021rethinking} replace the softmax kernel with linearizable approximations, achieving $\mathcal{O}(S)$ complexity at the expense of approximation error. Related locality-based methods constrain attention to sliding windows or neighborhood structures \citep{beltagy_longformer_2020}, reinforcing inductive biases toward short-range dependencies but potentially limiting long-range reasoning. More recently, recurrence- or state-space–based models, such as Mamba \citep{gu2024mamba}, abandon explicit attention altogether, replacing it with structured sequence operators that maintain implicit memory in linear time. Collectively, these approaches illustrate distinct compromises between global expressivity, computational efficiency, approximation fidelity, and hardware alignment, underscoring that attention scalability can be achieved either by structural sparsification, matrix approximation, architectural redesign, or alternative sequence modeling paradigms.

\subsection{IO-Aware Exact Attention and KV Sharing Mechanisms}

In contrast to approximation-based efficiency methods, FlashAttention \citep{dao2022flashattention} demonstrates that substantial performance gains can be achieved without altering the mathematical definition of softmax attention. Rather than materializing the full $S \times S$ attention matrix in high-bandwidth memory (HBM), FlashAttention reorders computation via tiling and fusion, performing block-wise softmax normalization and accumulation entirely within on-chip SRAM. This IO-aware reformulation minimizes memory reads and writes, which are often the true bottleneck in modern accelerators, and achieves significant speedups while preserving exactness. FlashAttention-2 \citep{dao2024flashattention} further improves work partitioning, parallelism, and kernel utilization, delivering higher throughput and better scaling across GPUs. These results underscore a crucial insight: attention efficiency is frequently constrained more by memory bandwidth and data movement than by arithmetic complexity.

Orthogonal to kernel-level optimization, architectural modifications such as Multi-Query Attention (MQA) \citep{shazeer2019fasttransformerdecodingwritehead} and Grouped-Query Attention (GQA) \citep{ainslie2023gqa} reduce inference memory by sharing key and value projections across heads. In standard multi-head attention, each of the $H_Q$ query heads maintains its own $K$ and $V$ projections; MQA collapses these into a single shared KV head, while GQA generalizes this by allowing $H_Q$ query heads to share a smaller number $H_{KV} < H_Q$ of KV heads. This reduces KV cache memory proportionally by a factor of $H_{KV}/H_Q$, yielding substantial savings during autoregressive decoding with minimal degradation in model quality. Modern large-scale LLMs widely adopt GQA or MQA variants due to their favorable trade-offs between memory footprint, decoding speed, and representational flexibility \citep{li2024survey}. Together, IO-aware exact kernels and KV-sharing architectures illustrate complementary strategies for improving attention efficiency: one targets hardware-level data movement, while the other reduces structural redundancy in key–value representations.

\subsection{KV Cache Compression}

Beyond modifying attention kernels or architectural structure, a rapidly growing line of work directly compresses  KV cache during inference, treating it as a memory management problem under long-context decoding. Existing methods can be organized along several orthogonal axes. \emph{Quantization-based approaches}~\citep{hooper2024kvquant,yang2024tokenleftbehindreliable} reduce the bit precision $b$ of stored keys and values -- via post-training quantization, mixed-precision schemes, or dynamic per-layer strategies -- achieving linear memory savings with modest degradation in perplexity. \emph{Eviction and heavy-hitter methods} selectively discard tokens deemed unlikely to influence future predictions, using heuristics such as recency bias, cumulative attention mass, token norms, or learned importance predictors (e.g., H2O \citep{zhang_h_2o_2023}, StreamingLLM~\citep{xiao_efficient_2024}, Scissorhand~\citep{liu2023scissorhands}, or CurDKV~\citep{sengupta2025valueguided}); these approaches implicitly assume that attention contributions are highly skewed and that only a small subset of tokens dominates downstream influence. \emph{Chunking and semantic summarization} methods~\citep{liu2025chunkkv, bai-etal-2024-citrus} cluster contiguous or semantically related tokens and replace them with compressed representatives, attempting to preserve coverage while reducing storage. \emph{Query-agnostic compression} mechanisms~\citep{kim2025kvzipqueryagnostickvcache, chari2025compactorcalibratedqueryagnostickv} pre-compress KV tensors independent of the current query, exploiting structural redundancy across tokens or layers. Finally, \emph{adaptive and predictive methods}~\citep{ge2024model,zhou-etal-2025-dynamickv,feng_ada-kv_2025, qin2025cake} frame compression as a forecasting problem, dynamically estimating which tokens will be required in future decoding steps (e.g., expected-attention–based~\citep{devoto2025expectedattentionkvcache} or AdaKV-style~\citep{feng_ada-kv_2025} approaches), thereby aligning memory retention with anticipated routing demand.

Across these categories, evaluation typically reports compression ratio, latency overhead, perplexity or downstream accuracy degradation, robustness under long or adversarial contexts, and whether the method is training-free or requires fine-tuning. Empirical evidence suggests substantial redundancy in KV storage, with many methods achieving 70\%--90\% reduction under standard benchmarks. However, prevailing evaluations predominantly measure aggregate retrieval accuracy or task-level performance, implicitly equating token retention with functional preservation. They rarely distinguish between storage, accessibility, and utilization, leaving open the deeper question of whether compressed caches maintain the structural routing pathways required for semantic reachability during autoregressive generation.

\section{Methodology}

\subsection{Datasets and Controlled KV Compression Evaluation}
\label{sec:datasets}

Standard long-context benchmarks such as LongBench~\citep{bai2023longbench}, InfiniteBench~\citep{zhang-etal-2024-bench}, and RULER~\citep{hsieh2024ruler} evaluate downstream task accuracy under extended contexts, but they are not designed to isolate the structural effects of KV compression on internal attention routing. In particular, they conflate three distinct phenomena: (i) whether information remains stored in the cache, (ii) whether it remains accessible through viable attention pathways, and (iii) whether it is effectively utilized during generation. A model may answer correctly because redundancy compensates for routing damage, or fail due to unrelated reasoning limitations rather than compression-induced structural collapse. As emphasized in the Physics of LLMs line of work~\citep{allen-zhu2025physics,allenzhu2024physicslanguagemodels31}, understanding scaling behavior requires carefully controlled synthetic settings that expose internal mechanisms rather than relying solely on naturalistic benchmarks. Motivated by this perspective, we design a suite of synthetic datasets that treat KV compression as a structural perturbation of token routes within self-attention.

Our dataset construction follows two complementary paradigms. First, we employ controlled generative synthesis, prompting frontier LLMs with tightly structured instructions that enforce explicit entity–attribute linkage, mention frequency constraints, and bidirectional querying. Second, we use deterministic template-based construction with slot-filling to eliminate linguistic confounders and ensure maximal experimental control. Together, these approaches balance realism and controllability while maintaining route sensitivity. 

\paragraph{Design principles.}
Each dataset is constructed to satisfy four constraints: (1) \emph{route sensitivity}, where correct answers depend on preserving specific cross-layer token pathways; (2) \emph{minimal redundancy}, preventing accidental recoverability after aggressive eviction; (3) \emph{bidirectional querying}, enforcing both subject$\rightarrow$attribute and attribute$\rightarrow$subject tracing; and (4) \emph{failure interpretability}, enabling classification into representational erasure or rigidity. Under these principles, compression functions as a controlled ablation of token-route structures.

\paragraph{Generation framework.}
For generative datasets, we use structured prompting to enforce attribute-entity bindings and controlled semantic relationships. The generic procedure is formalized in Algorithm~\ref{alg:data_gen}.

\begin{algorithm}[H]
\caption{Controlled Synthetic Example Generation}
\begin{algorithmic}[1]
\Require Subject $s$, structural constraints $C$, length range $L$
\State Prompt LLM with instructions enforcing $C$
\State Replace entity names with synthetic identifiers
\State Generate passage $P$
\State Generate QA pairs under bidirectional linkage rules
\State Validate explicit answer grounding
\State Store $(P,\{(q_i,a_i)\})$
\end{algorithmic}
\label{alg:data_gen}
\end{algorithm}

For template-based datasets, examples are produced via deterministic slot substitution from pre-generated entity pools, optionally followed by controlled perturbations (e.g., pronoun swaps). This removes surface variability while preserving structural dependencies.

\paragraph{Dataset suite.}
The resulting suite spans short to long contexts (150–1300+ tokens) and targets distinct routing stressors:

\begin{itemize}[leftmargin=*, itemsep=0pt, parsep=0pt, topsep=0pt]
    \item \textbf{Base task:} Short passages with one subject and tightly bound attributes; probes precision of eviction under minimal redundancy.
    \item \textbf{Knowledge manipulation:} Slot-filled biographies testing minimally distributed factual structures and systematic eviction biases.
    \item \textbf{Multi presence:} Repeated entity mentions requiring instance disambiguation; evaluates positional robustness under pruning.
    \item \textbf{Multi entity:} Multiple semantically linked entities; tests cross-entity and cross-head routing integrity.
    \item \textbf{Long context:} Extended passages requiring distributed reasoning across distant spans; probes route capacity at scale.
    \item \textbf{Coreference:} Controlled pronoun perturbations with ``I don't know'' ground truth; detects fine-grained routing failures and hallucination under compression.
    \item \textbf{Hops:} Chain-structured entities requiring multi-hop reasoning; evaluates preservation of sequential semantic routes.
\end{itemize}

Representative prompt and template structures are summarized below to ensure reproducibility. Table~\ref{tab:ContextSamples} highlights samples of each task while Table~\ref{tab:ForRevSamples} shows forward and reverse samples in Appendix~\ref{appx:data} 

\begin{lstlisting}[language=Python]
# Base task
Generate a passage about <subject> (150-200 words)
- >=4 directly linked attributes
- Fake entity names
Generate 6 QA pairs linking subject <-> attributes.
\end{lstlisting}

\begin{lstlisting}
# Knowledge manipulation (Template)
<Firstname Lastname> was born on <Date>.
They studied in <University>.
They work as <Occupation>.
They live in <City>.
\end{lstlisting}

\begin{lstlisting}[language=Python]
# Multi presence / Multi entity / Long context
Generate passage (400-1300 words)
- Multiple entities or repeated mentions
- Explicit attribute linkage
- Bidirectional QA generation
\end{lstlisting}

\begin{lstlisting}
# Coreference
<Firstname Lastname> was born on <Date> in <Location>.
<He/She> studied at <University>.
...
# Pronoun-swapped queries -> answer = "I don't know"
\end{lstlisting}

\begin{lstlisting}[language=Python]
# Hops
Generate subject + 3 linked entities
- Sequential semantic linkage
- 16 QA pairs spanning multi-hop chains
\end{lstlisting}

\paragraph{Dataset statistics.}
Table~\ref{tab:dataset_info} summarizes the structural characteristics of the synthetic dataset suite, including average passage length, number of queries per passage, total passages, and total query count. The datasets span a broad range of context regimes, from short, tightly controlled passages (e.g., Base task and Knowledge manipulation) to extended multi-hop narratives exceeding one thousand words (Long context and Hops). This deliberate variation in passage length and structural complexity enables systematic evaluation across compression ratios, from mild pruning to extreme eviction. Importantly, the suite balances LLM-generated passages, which introduce controlled semantic diversity, with deterministic template-based constructions, which maximize experimental precision. This combination ensures that compression behavior can be analyzed both under naturalistic variability and under strictly controlled structural constraints.

\begin{table}[!t]
\centering
\resizebox{\columnwidth}{!}{
\begin{tabular}{l|c c c c c}
\toprule
\textbf{Name} & \textbf{Passage Length (words)} & \textbf{\# Queries} & \textbf{Total Passages} & \textbf{Total Queries} & \textbf{LLM Generated} \\
\midrule
Base task & 175 $\pm$ 25 & 6 & 200 & 1199 & Yes \\ 
Knowledge manipulation & 32 & 13 & 4000 & 52000 & No \\ 
Multi presence & 500 $\pm$ 100 & 10 & 100 & 1000 & Yes \\ 
Multi entity & 500 $\pm$ 100 & 10 & 100 & 1000 & Yes \\ 
Long context & 1100 $\pm$ 200 & 10 & 100 & 1000 & Yes \\ 
Coreference & 44 & 9 & 8000 & 72000 & No \\ 
Hops & 1100 $\pm$ 20 & 16 & 100 & 1600 & Yes \\
\bottomrule
\end{tabular}}
\caption{Structural summary of the synthetic dataset suite used for controlled KV compression evaluation.}
\label{tab:dataset_info}
\end{table}

\paragraph{Significance of the generated synthetic datasets.}
Unlike conventional long-context benchmarks that primarily report aggregate accuracy, our framework enables causal attribution of compression failures to specific structural mechanisms within attention. The Base task and Knowledge manipulation datasets isolate eviction precision and entity–attribute binding stability. Multi presence and Multi entity expose positional sensitivity and cross-head routing fragility. Long context stresses route capacity under large token budgets, while Hops directly probes preservation of multi-step semantic chains. Coreference evaluates fine-grained representational consistency and hallucination sensitivity. This layered structure permits systematic classification of failures into representational erasure, where all viable token routes to critical information are removed, and representational rigidity, where tokens survive but routing flexibility collapses. Compression is thus analyzed as a structural phase transition in semantic reachability rather than as a monotonic decline in accuracy.

\subsection{Tagging Framework}
To move beyond aggregate performance metrics, each question–answer pair is annotated using a structured tagging framework that enables fine-grained behavioral analysis under compression. The objective is to transform accuracy statistics into interpretable signals about internal representational robustness.

Tagging operates along two orthogonal axes: answer type and question difficulty. The answer-type axis categorizes responses according to semantic domain, allowing us to examine whether certain conceptual representations are more resilient to compression. The answer categories include \textit{Person}, \textit{Thing}, \textit{Organization}, \textit{Creature}, \textit{Location}, \textit{Numerals}, \textit{Date/Time}, and \textit{Event} (Table \ref{tab:AnsTagSamples} in Appendix~\ref{appx:data}). By decomposing performance along these semantic dimensions, we can detect systematic degradation patterns, identify hierarchies of representational stability, and evaluate whether compression disproportionately affects particular classes of concepts.

The second axis captures reasoning demand through question-difficulty tags. Questions are classified as \textit{Standard} (direct retrieval of explicitly stated information), \textit{Manipulated} (requiring implicit contextual interpretation or transformation), or \textit{Part} (requiring aggregation of information distributed across multiple textual regions). This dimension separates failures caused by token loss from those arising due to reasoning complexity (Table \ref{tab:QuesTagSamples} in Appendix~\ref{appx:data}). A model may retain the necessary tokens yet fail on high-difficulty queries that require multi-step inference; conversely, it may fail simple retrieval tasks due to representational erasure.

Together, these tagging axes define a multidimensional evaluation space. Performance can be decomposed not only by dataset and compression ratio but also by semantic content and reasoning structure. This structured analysis enables systematic identification of compression-sensitive domains and facilitates deeper interpretation of model-specific robustness profiles.

\subsection{Experimental Setup}

All compression experiments were conducted using NVIDIA's KVPress library\footnote{\url{https://github.com/NVIDIA/kvpress/}}, which provides a unified interface for scoring-based and pruning-based KV cache compression during autoregressive decoding. For all models, KV compression was applied at inference time without additional fine-tuning. We systematically vary compression ratios from mild pruning (10\%) to aggressive eviction (up to 90\% removal of KV entries) and report performance as a function of the retained KV budget.

\paragraph{Scoring and Pruning Strategies.}
We adopt Expected Attention~\citep{devoto2025expectedattentionkvcache} as the scoring function, which estimates token importance by aggregating expected attention weights across decoding steps and layers. This per-layer scoring mechanism allows high-salience key–value pairs, particularly those residing in structurally significant heads, to be preferentially preserved. Two complementary pruning strategies are employed:

\begin{itemize}[leftmargin=*, itemsep=0pt, parsep=0pt, topsep=0pt]
    \item \textbf{FINCH (Chunk) Press}~\citep{finch2024tacl} performs chunk-level pruning by partitioning the context into contiguous segments and removing the lowest-scoring tokens within each segment. This approach is computationally efficient and preserves coarse-grained structural coverage across the document.
    \item \textbf{AdaKV Press}~\citep{feng_ada-kv_2025} performs head-wise global pruning by ranking KV entries across all heads simultaneously and removing the lowest-scoring entries irrespective of positional grouping. This provides finer-grained adaptive compression at the cost of increased overhead.
\end{itemize}

The combination of FINCH and AdaKV enables evaluation under both coarse structural pruning and globally adaptive head-aware compression, allowing us to analyze whether routing collapse depends on local chunk structure or global head-level importance.

\paragraph{Inference Regimes.}
Compression is evaluated under two inference settings:

\begin{itemize}[leftmargin=*, itemsep=0pt, parsep=0pt, topsep=0pt]
    \item \textbf{Question-agnostic:} The model first receives the full context and performs KV pruning without access to the downstream question. The query is provided only after compression. This setting ensures that eviction decisions are independent of retrieval demands and reflects realistic deployment scenarios.
    \item \textbf{Question-aware:} The list of candidate questions is provided prior to pruning, allowing compression decisions to condition on anticipated query structure. This setting measures the upper bound of compression tolerance when routing can be optimized for known retrieval targets.
\end{itemize}

Comparing these regimes isolates whether compression robustness arises from intrinsic route redundancy or from query-conditioned pruning strategies.

\paragraph{Models.}
We evaluate five instruction-tuned LLMs spanning two architectural families:

\begin{itemize}[leftmargin=*, itemsep=0pt, parsep=0pt, topsep=0pt]
    \item LLaMA-3.2 3B Instruct and LLaMA-3 8B Instruct~\citep{grattafiori2024LLaMA},
    \item Qwen-2.5 3B Instruct, Qwen-2.5 7B Instruct, and Qwen-2.5 14B Instruct~\citep{yang2024qwen2}.
\end{itemize}

These models cover a range of parameter scales (3B–14B) and architectural variations in attention design and training data, enabling analysis of model-dependent compression tolerance. All checkpoints were obtained from the Hugging Face model hub\footnote{\url{https://huggingface.co/models}} and used without modification.

\paragraph{Implementation Details.}
All experiments were executed on a single NVIDIA RTX A6000 GPU using the KVPress text generation pipeline. The default configuration enforces question-agnostic pruning unless explicitly overridden. Generation is performed using greedy decoding to avoid variability introduced by sampling. Performance is evaluated at the token level, with answers matched using the LLaMA-3 8B tokenizer to ensure consistent segmentation across models and compression settings. 

This setup ensures that performance differences arise solely from KV compression behavior rather than decoding randomness or tokenizer inconsistencies, enabling controlled analysis of structural failure modes under varying compression budgets.

\section{Results}

We now present a consolidated analysis of compression behavior across tasks, architectures, and tagging dimensions. Rather than interpreting degradation as a single monotonic decline, the results reveal structured transitions in semantic reachability. Across datasets, three recurring phenomena emerge: localized performance spikes consistent with sparse route selection, architecture-dependent divergence between retention and manipulation, and systematic collapse of multi-hop reasoning under aggressive compression.

\subsection{Aggregate Performance Trends}

Table~\ref{tab:main_results} reports the base (0\% compression) F1 scores across all tasks under question-agnostic (AGN) and question-aware (AWR) settings.

\begin{table}[h]
\centering
\resizebox{\columnwidth}{!}{
\begin{tabular}{l|l|c c c c c c c}
\toprule
Model & Setup & Base & Know & Multi-P & Multi-E & Long & Coref & Hops \\
\midrule
\multirow{2}{*}{LLaMA-3 8B} 
 & AGN & 70.05 & 91.28 & 79.64 & 80.33 & 45.93 & 75.63 & 31.54 \\
 & AWR & 69.86 & 92.09 & 77.66 & 81.93 & 45.54 & 60.90 & 31.25 \\
\midrule
\multirow{2}{*}{Qwen-2.5 7B} 
 & AGN & 68.80 & 92.20 & 73.63 & 77.46 & 38.68 & 78.89 & 27.13 \\
 & AWR & 69.63 & 90.93 & 71.88 & 79.38 & 40.20 & 52.00 & 26.35 \\
\bottomrule
\end{tabular}
}
\caption{Base (0\% compression) F1 scores across datasets.}
\label{tab:main_results}
\end{table}

\begin{figure*}[!t]
    \centering
    \subfloat[LLaMA-3 8B Instruct]{%
        \includegraphics[width=0.5\columnwidth]{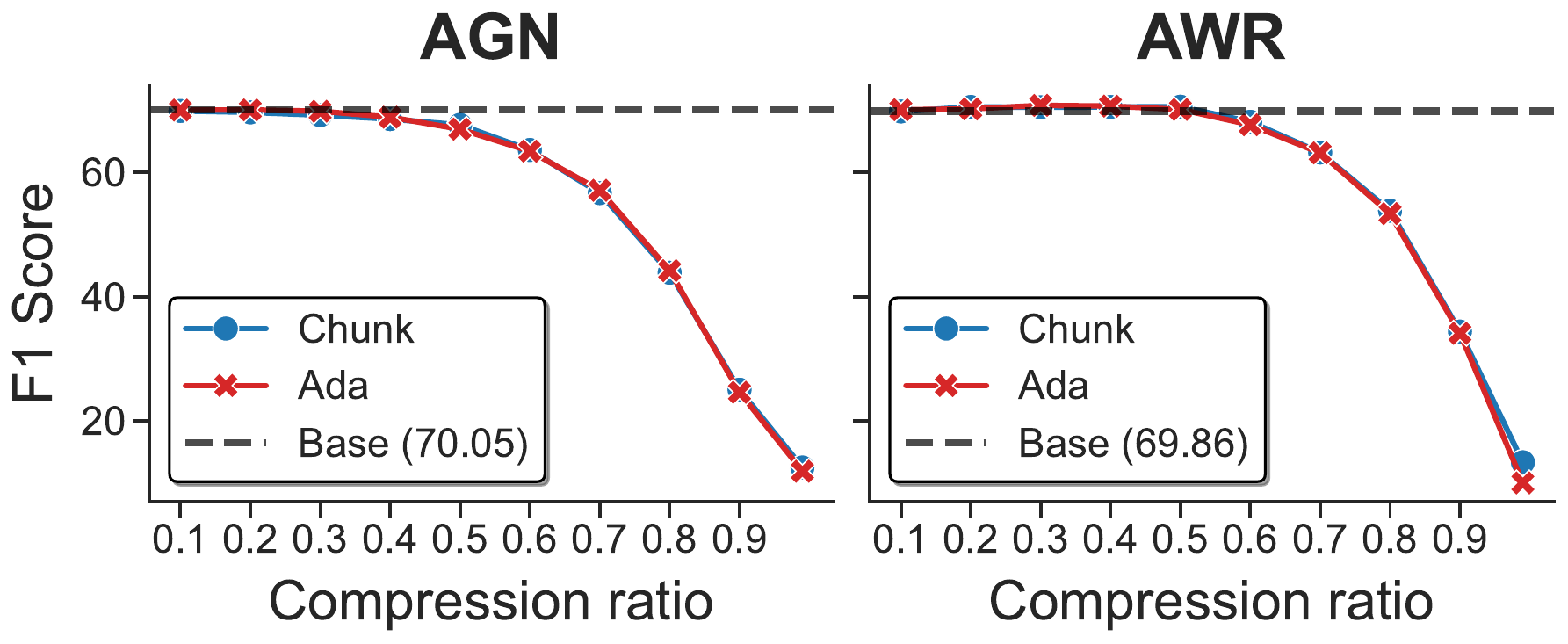}
    }
    \subfloat[Qwen-2.5 7B Instruct]{%
        \includegraphics[width=0.5\columnwidth]{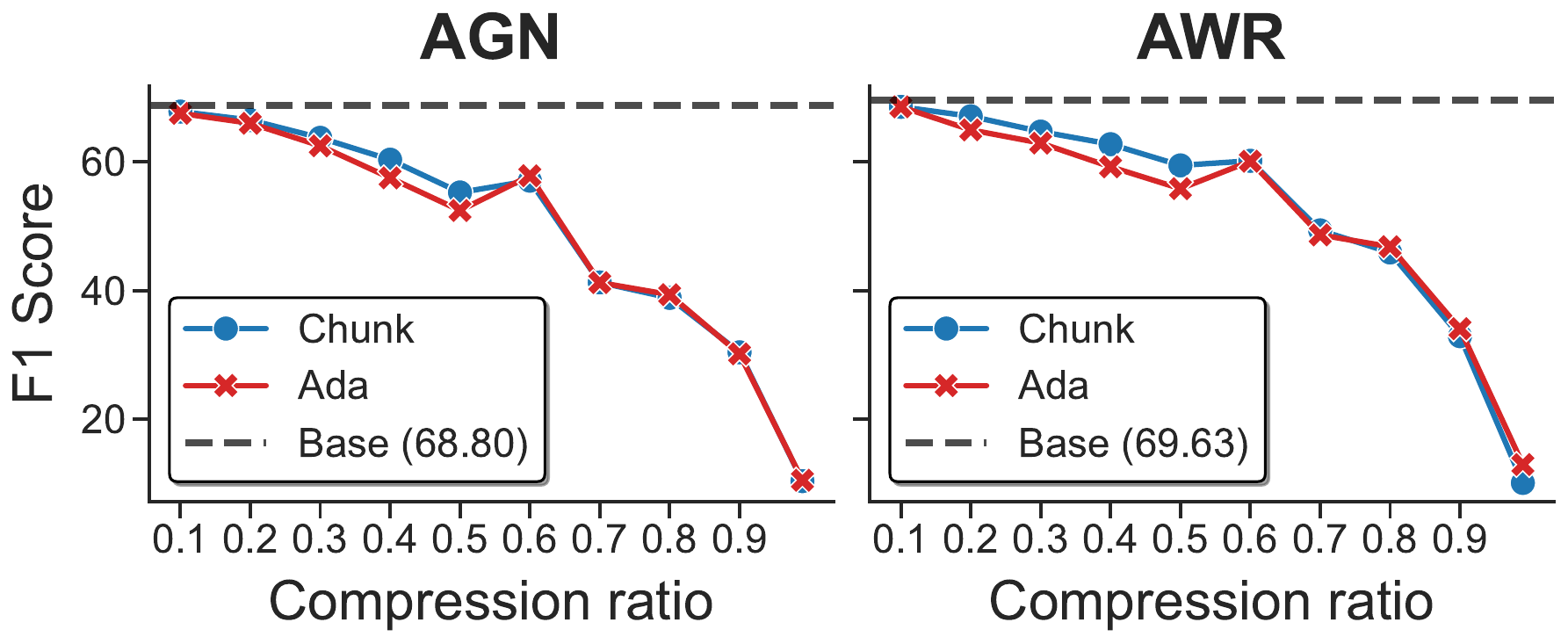}
    }
    \caption{Base task performance across compression levels. Localized spikes under moderate compression suggest sparse substructure effects. Corresponding results for LLaMA-3.2 3B, Qwen-2.5 3B, and Qwen-2.5 14B are shown in Figure \ref{fig:BaseTaskPerfApp} of Appendix~\ref{appx:result}.}
    \label{fig:BaseTaskPerf}
\end{figure*}

\noindent \textbf{Base task.} On the Base task (Figure~\ref{fig:BaseTaskPerf}), both models operate near 70\% F1 (LLaMA: 70.05 AGN, 69.86 AWR; Qwen: 68.80 AGN, 69.63 AWR). Under increasing compression, performance generally declines; however, the question-aware setup exhibits localized non-monotonic spikes, particularly for Qwen around intermediate compression levels. This suggests that moderate pruning can remove interfering or redundant KV routes, temporarily improving alignment. Such behavior is consistent with the existence of sparse token-route~\citep{zhu2025lottery} substructures within attention that remain functionally intact under partial eviction.

\begin{figure*}[!t]
    \centering
    \subfloat[LLaMA-3 8B Instruct]{%
        \includegraphics[width=0.49\linewidth]{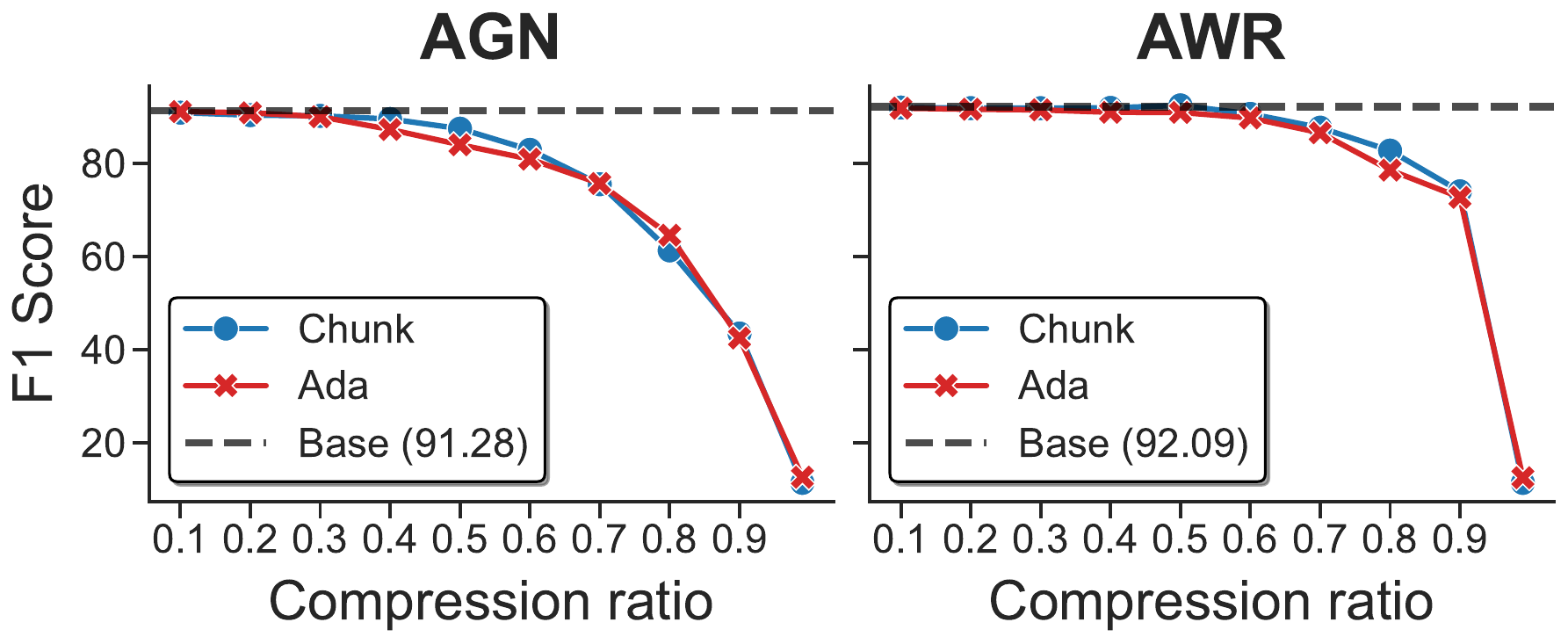}
    }
    \subfloat[Qwen-2.5 7B Instruct]{%
        \includegraphics[width=0.49\linewidth]{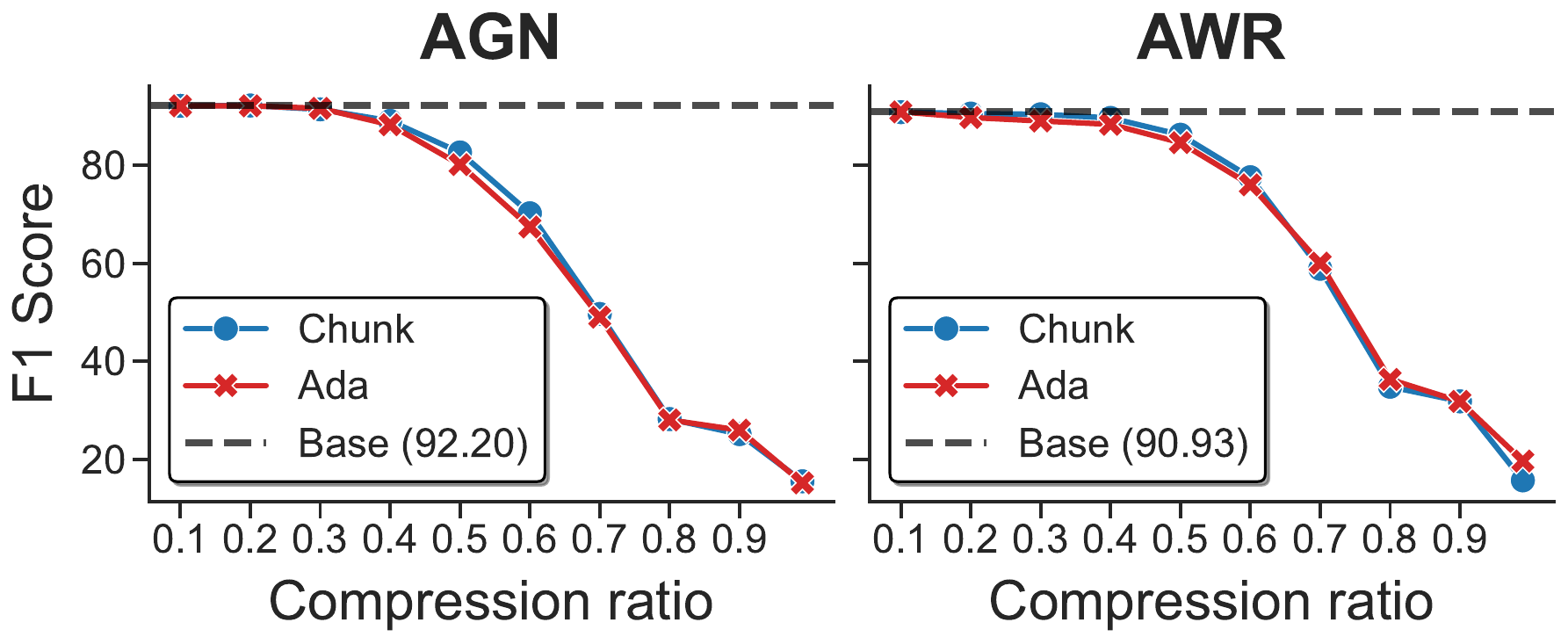}
    }
    \caption{Knowledge manipulation results. Qwen exhibits a more gradual rate of degradation under compression, particularly in the question-aware setting.}
    \label{fig:KnowledgeManPerf}
\end{figure*}

\begin{figure*}[!htb]
    \centering
    \subfloat[LLaMA-3 8B Instruct]{
    \includegraphics[width=0.49\linewidth]{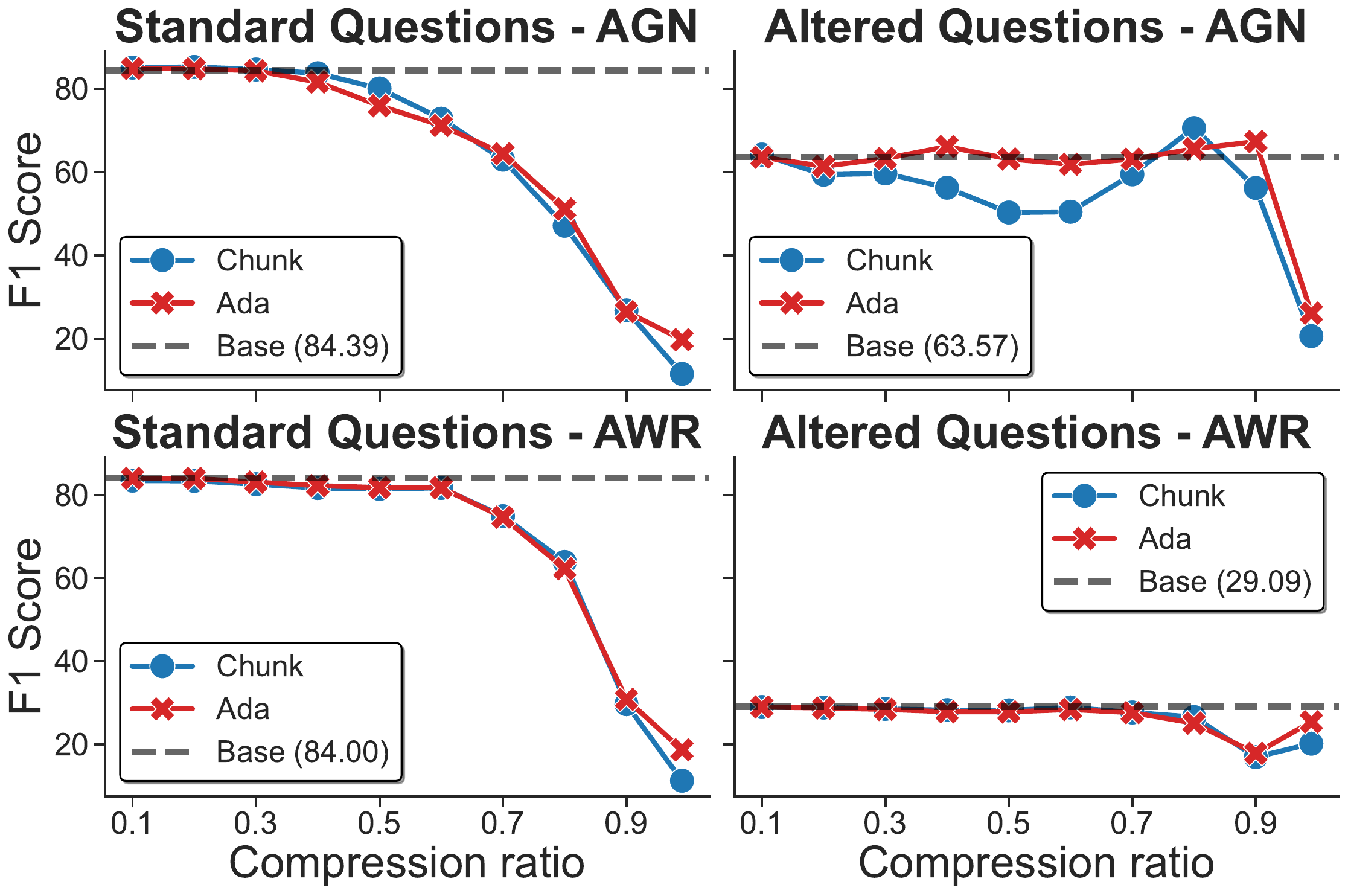}
    }
    \subfloat[Qwen-2.5 7B Instruct]{
    \includegraphics[width=0.49\linewidth]{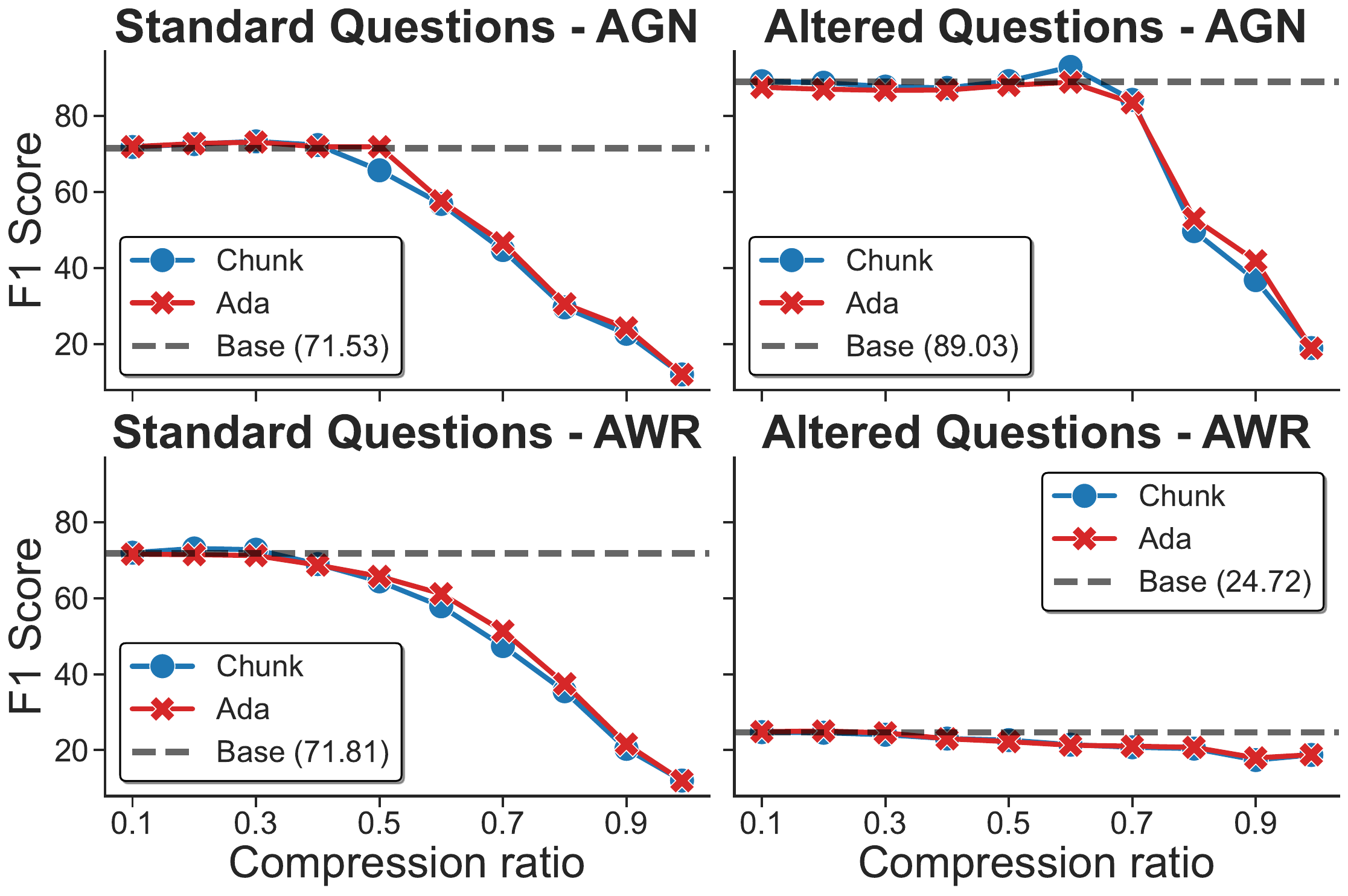}
    }
    \caption{Coreference performance across setups. Question-aware pruning significantly increases overconfident errors.}
    \label{fig:CoreferencePerf}
\end{figure*}

\noindent \textbf{Knowledge manipulation} presents a contrasting pattern (Figure~\ref{fig:KnowledgeManPerf}). Qwen slightly surpasses LLaMA in the question-agnostic setting (92.20 vs. 91.28), and while LLaMA is marginally stronger in question-aware (92.09 vs. 90.93), Qwen degrades more gradually under higher compression. This task emphasizes structured transformation rather than pure retrieval, indicating that Qwen's robustness derives from instruction-conditioned reasoning, whereas LLaMA's strength lies in stable factual retention. Notably, both models remain above 90\% F1 until approximately 40\% compression.

\noindent \textbf{Coreference} (Figure~\ref{fig:CoreferencePerf}) exposes a pronounced divergence between AGN and AWR. LLaMA drops from an aggregate performance of 75.63 (AGN) to 60.90 (AWR), while Qwen drops from aggregate performance of 78.89 (AGN) to 52.00 (AWR), a 26.89-point reduction. Conditioning pruning on anticipated queries appears to bias models toward premature commitment, reducing their ability to correctly abstain when inconsistencies are introduced, supplementing the results obtained by~\citet{jin2025long}.

\begin{figure*}[!t]
    \centering
    \subfloat[LLaMA-3 8B Instruct]{
    \includegraphics[width=0.49\linewidth]{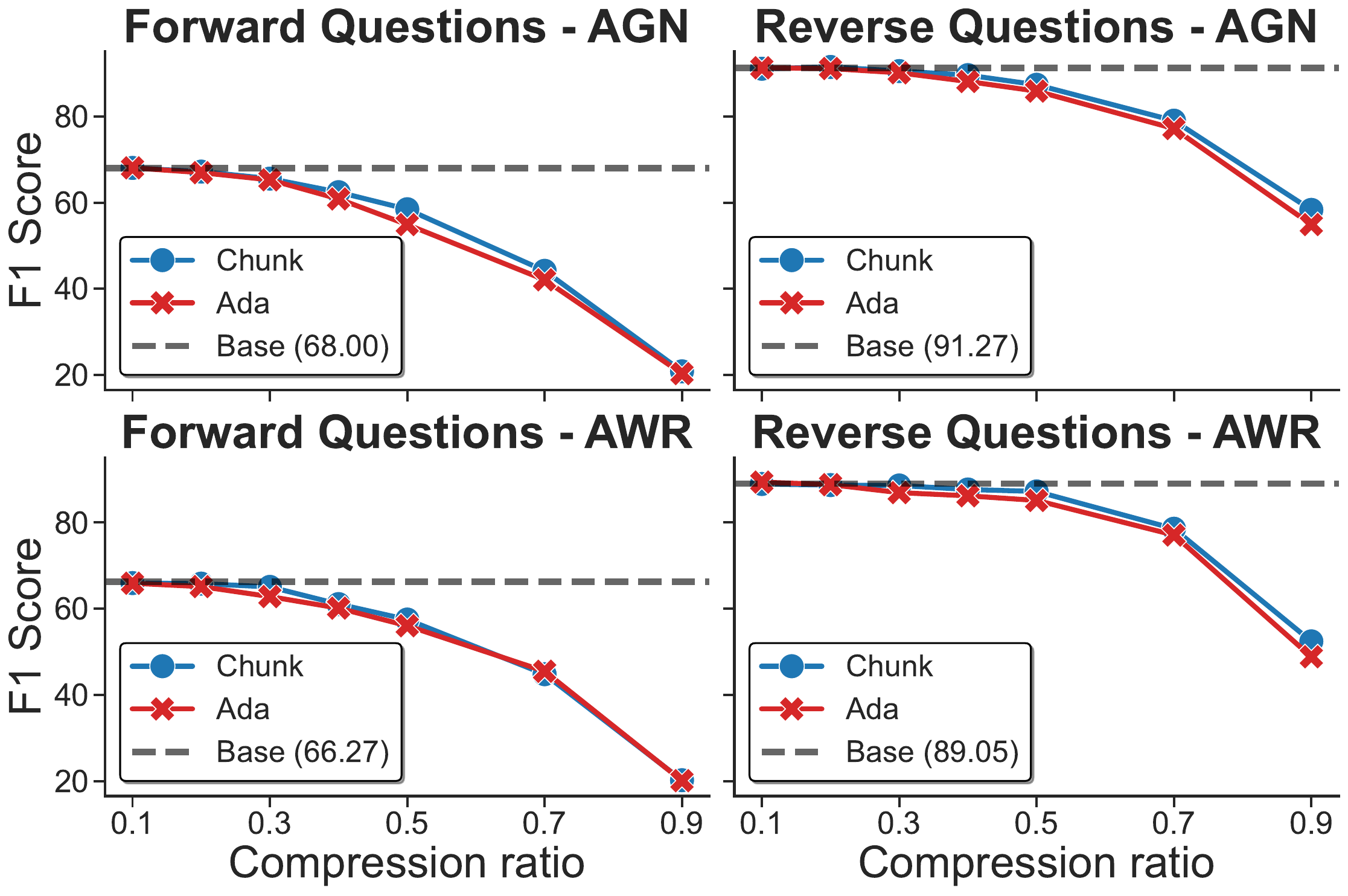}
    }
    \subfloat[Qwen-2.5 7B Instruct]{
    \includegraphics[width=0.49\linewidth]{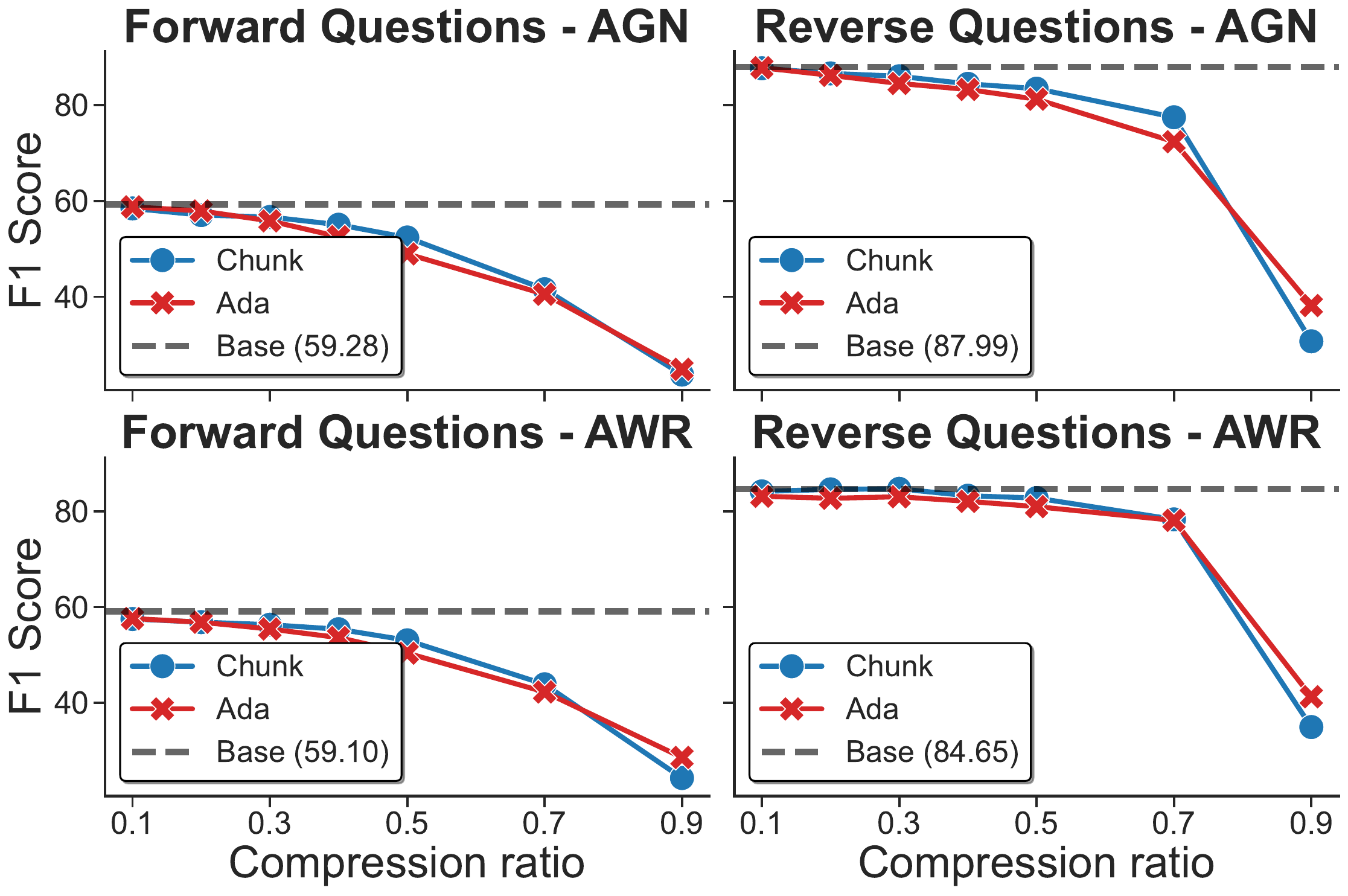}
    }
    \caption{Multi presence forward vs. reverse asymmetry.}
    \label{fig:MultiPresenceForRev}
\end{figure*}

\noindent \textbf{Multi presence} in  Figure~\ref{fig:MultiPresenceForRev} illustrates that LLaMA outperforms Qwen (79.64 aggregate vs. 73.63 aggregate AGN), but forward queries degrade more sharply than reverse queries under compression (68 forward and 91.27 reverse AGN vs. 66.27 forward and 89.05 reverse AWR), indicating asymmetric routing fragility when entities are repeated.

\noindent \textbf{Multi entity}, in contrast, shows more balanced behavior (LLaMA 80.33 aggregate AGN; Qwen 77.46 aggregate AGN), suggesting that distributing attributes across distinct entities mitigates interference (79.94 forward and 80.72 reverse AGN vs. 80.88 forward and 82.98 reverse AWR).

\begin{figure*}[!t]
    \centering
    \subfloat[LLaMA-3 8B Instruct]{
    \includegraphics[width=0.49\linewidth]{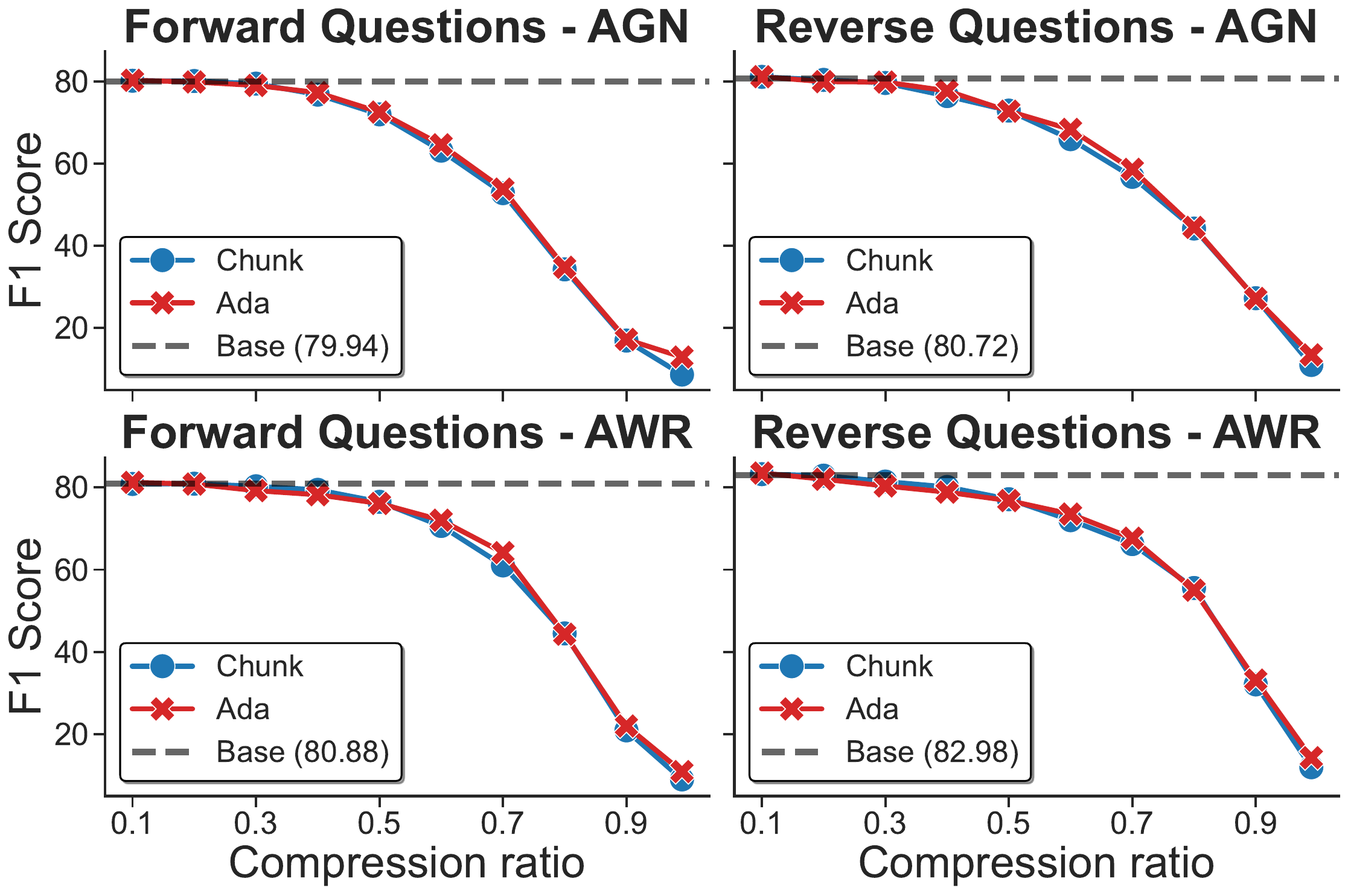}
    }
    \subfloat[Qwen-2.5 7B Instruct]{
    \includegraphics[width=0.49\linewidth]{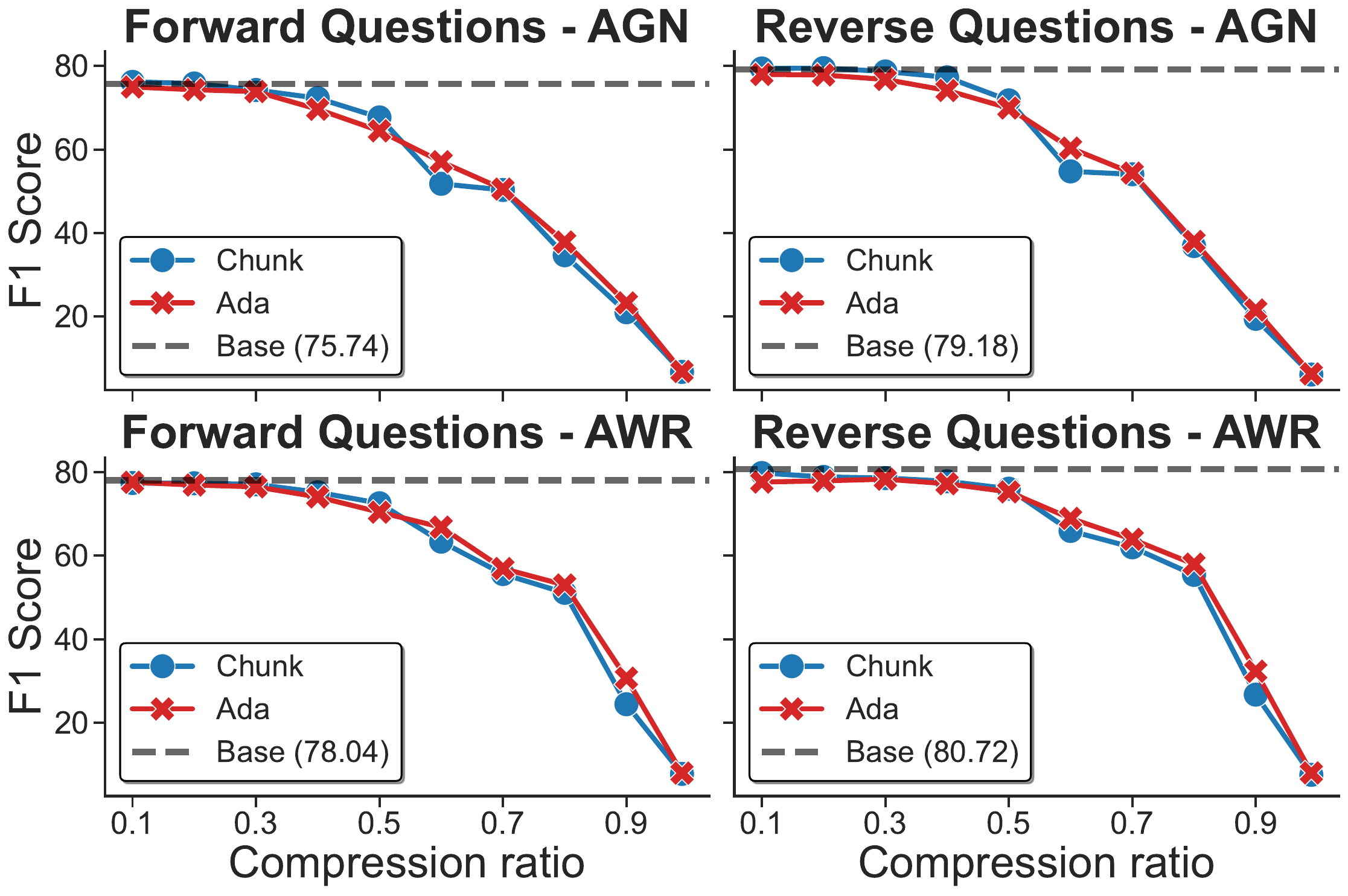}
    }
    \caption{Multi entity results showing reduced directional asymmetry. Corresponding results for LLaMA-3.2 3B, Qwen-2.5 3B, and Qwen-2.5 14B are shown in Figure \ref{fig:MultiEntityPerfAppendix} of Appendix~\ref{appx:result}.}
    \label{fig:MultiEntityForRev}
\end{figure*}

\begin{figure*}[!htb]
    \centering
    \subfloat[LLaMA-3 8B Instruct]{
    \includegraphics[width=0.49\linewidth]{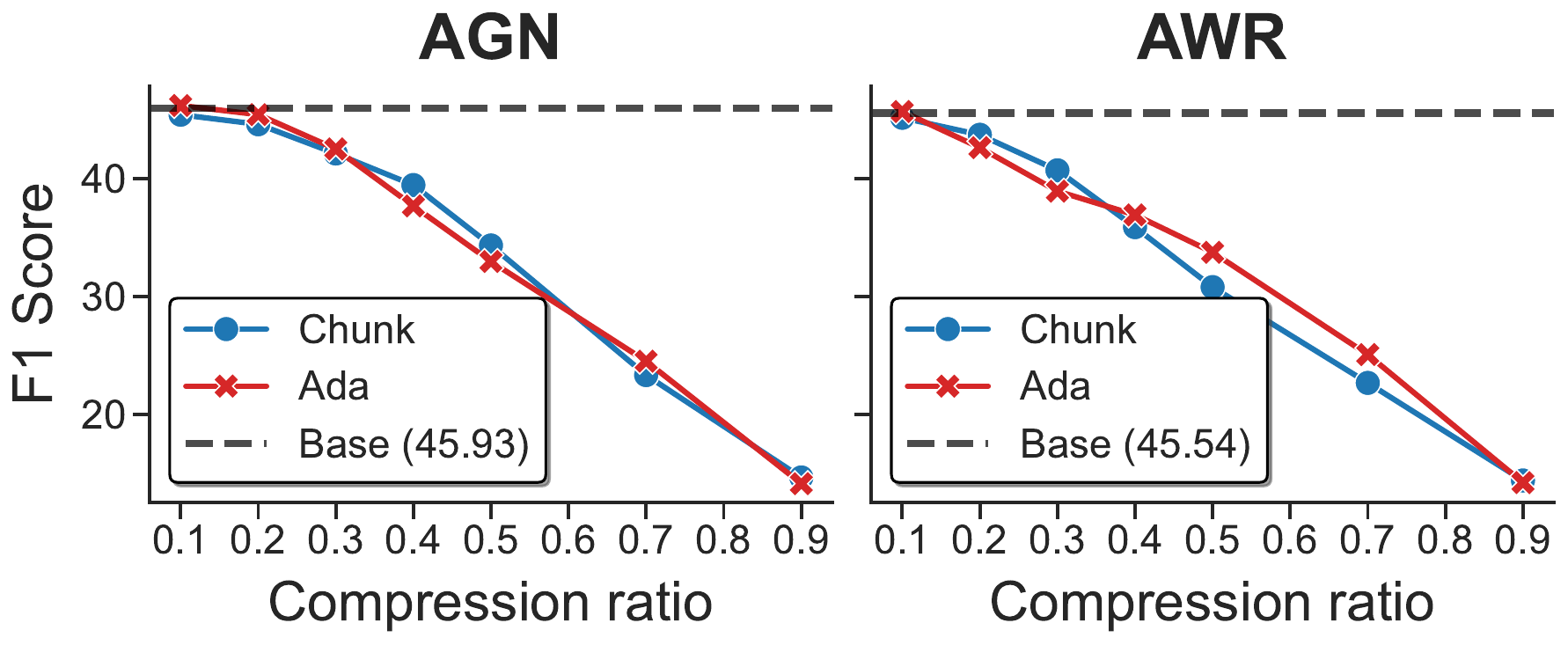}
    }
    \subfloat[Qwen-2.5 7B Instruct]{
    \includegraphics[width=0.49\linewidth]{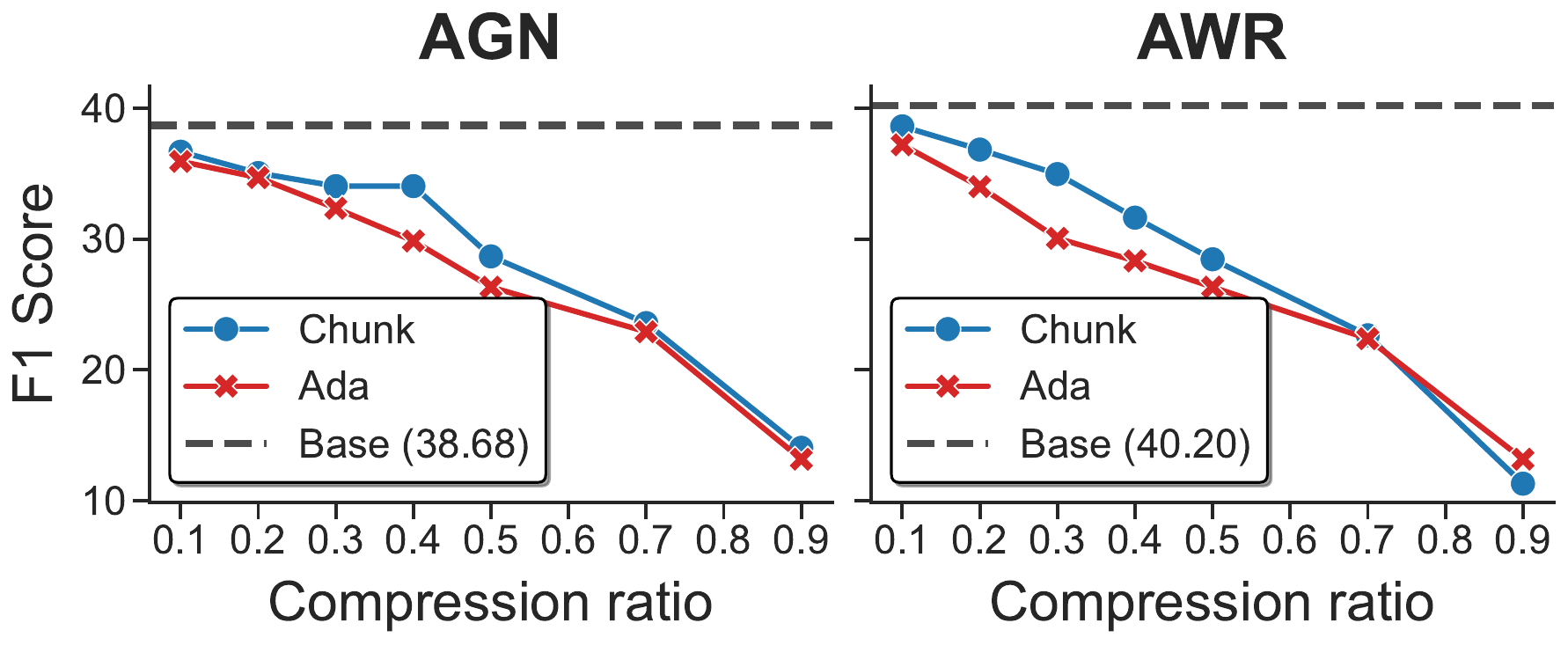}
    }
    \caption{Long context degradation across compression levels.}
    \label{fig:LongContextPerf}
\end{figure*}

\begin{figure*}[!htb]
    \centering
    \subfloat[LLaMA-3 8B Instruct]{
    \includegraphics[width=0.49\linewidth]{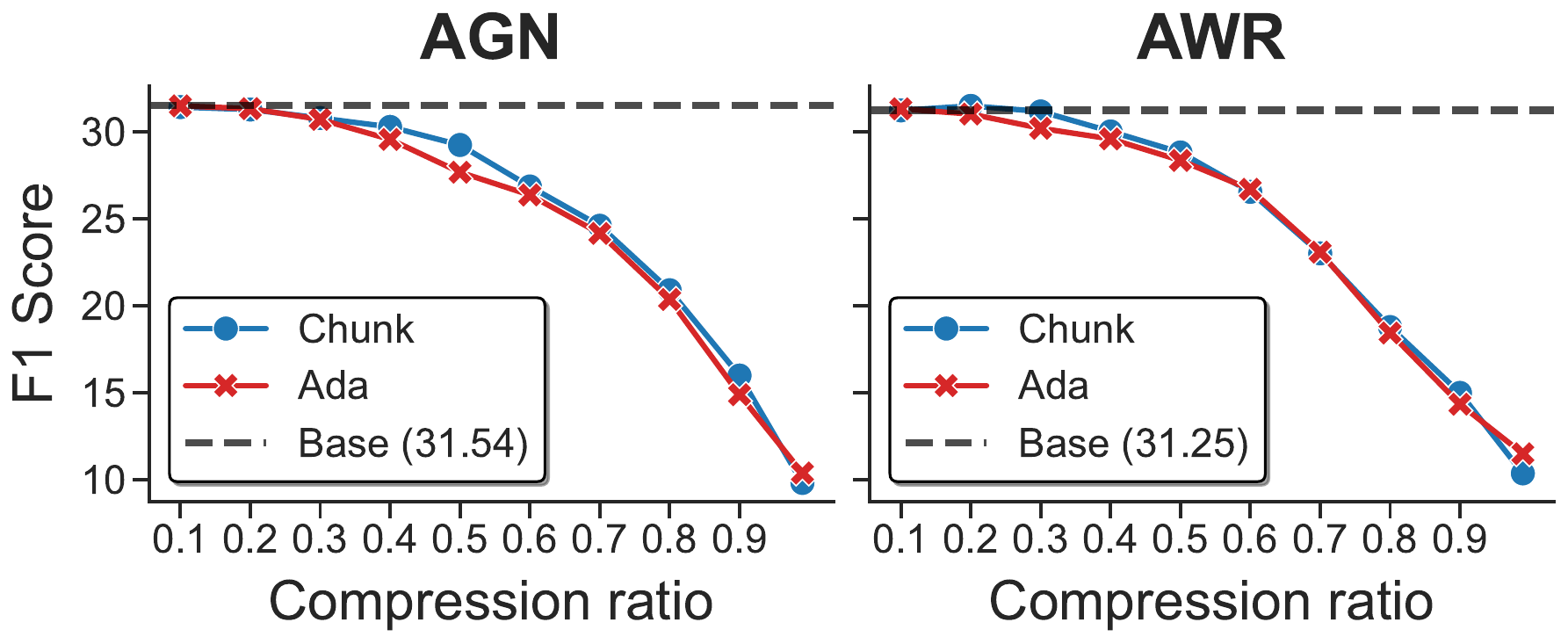}
    }
    \subfloat[Qwen-2.5 7B Instruct]{
    \includegraphics[width=0.49\linewidth]{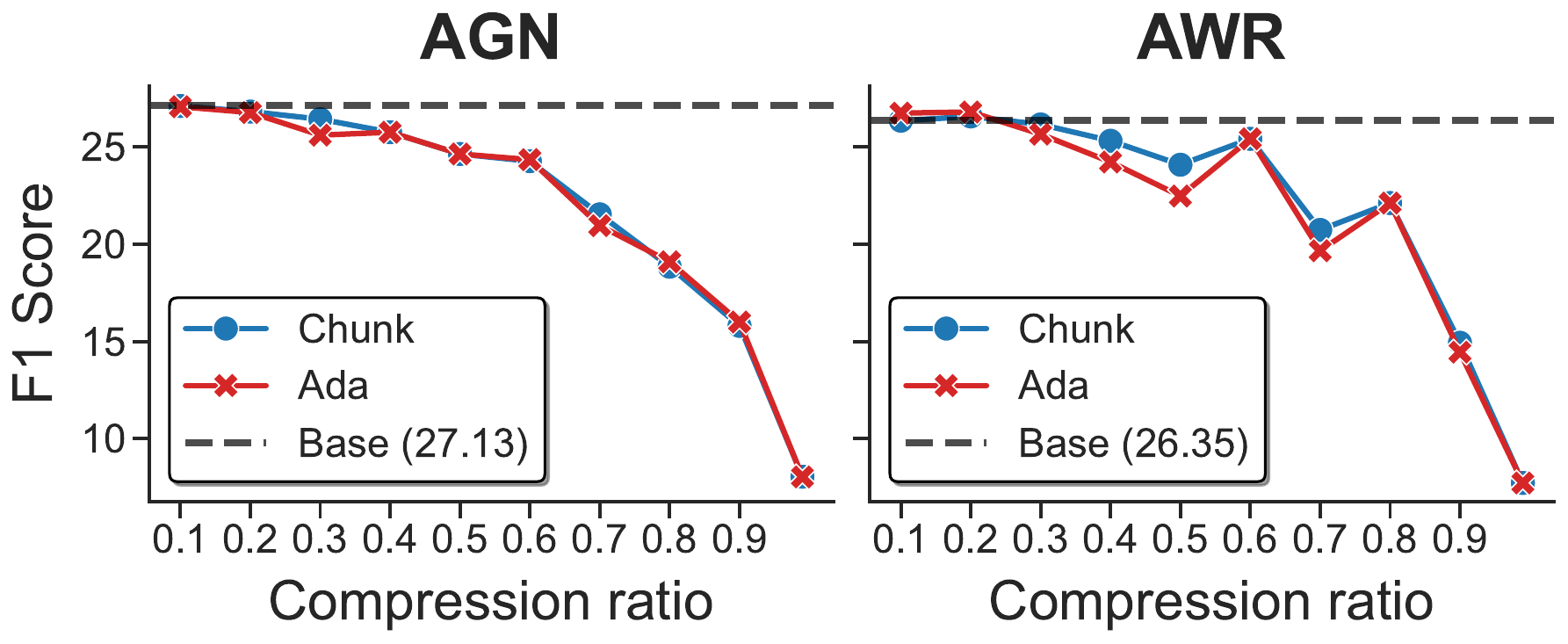}
    }
    \caption{Hops task performance. Multi-step semantic chaining collapses rapidly under compression.}
    \label{fig:HopsGeneralPerf}
\end{figure*}

\begin{figure*}[!htb]
    \centering
    \subfloat[LLaMA-3 8B Instruct]{
    \includegraphics[width=0.9\textwidth]{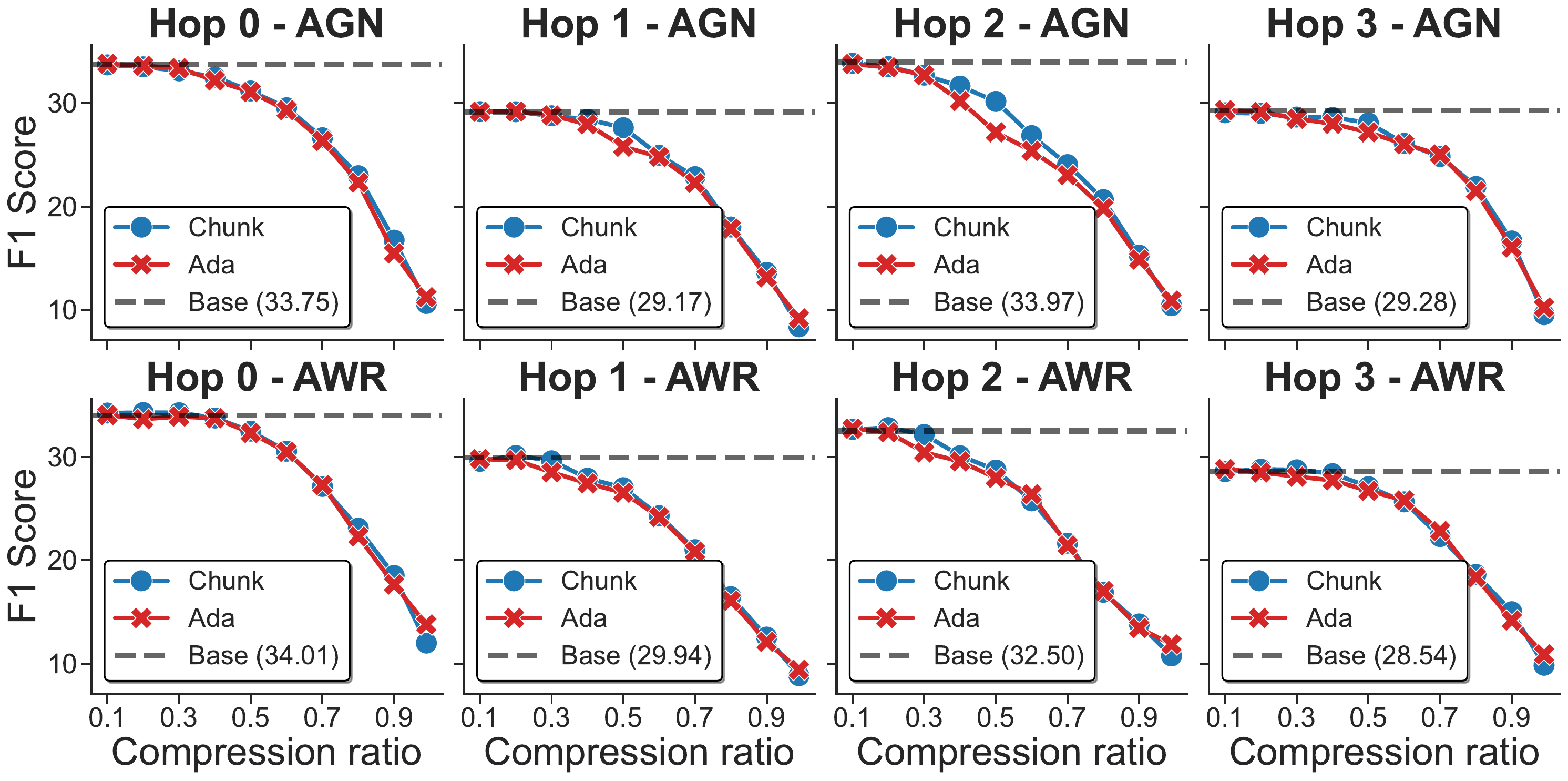}
    }\\
    \subfloat[Qwen-2.5 7B Instruct]{
    \includegraphics[width=0.9\textwidth]{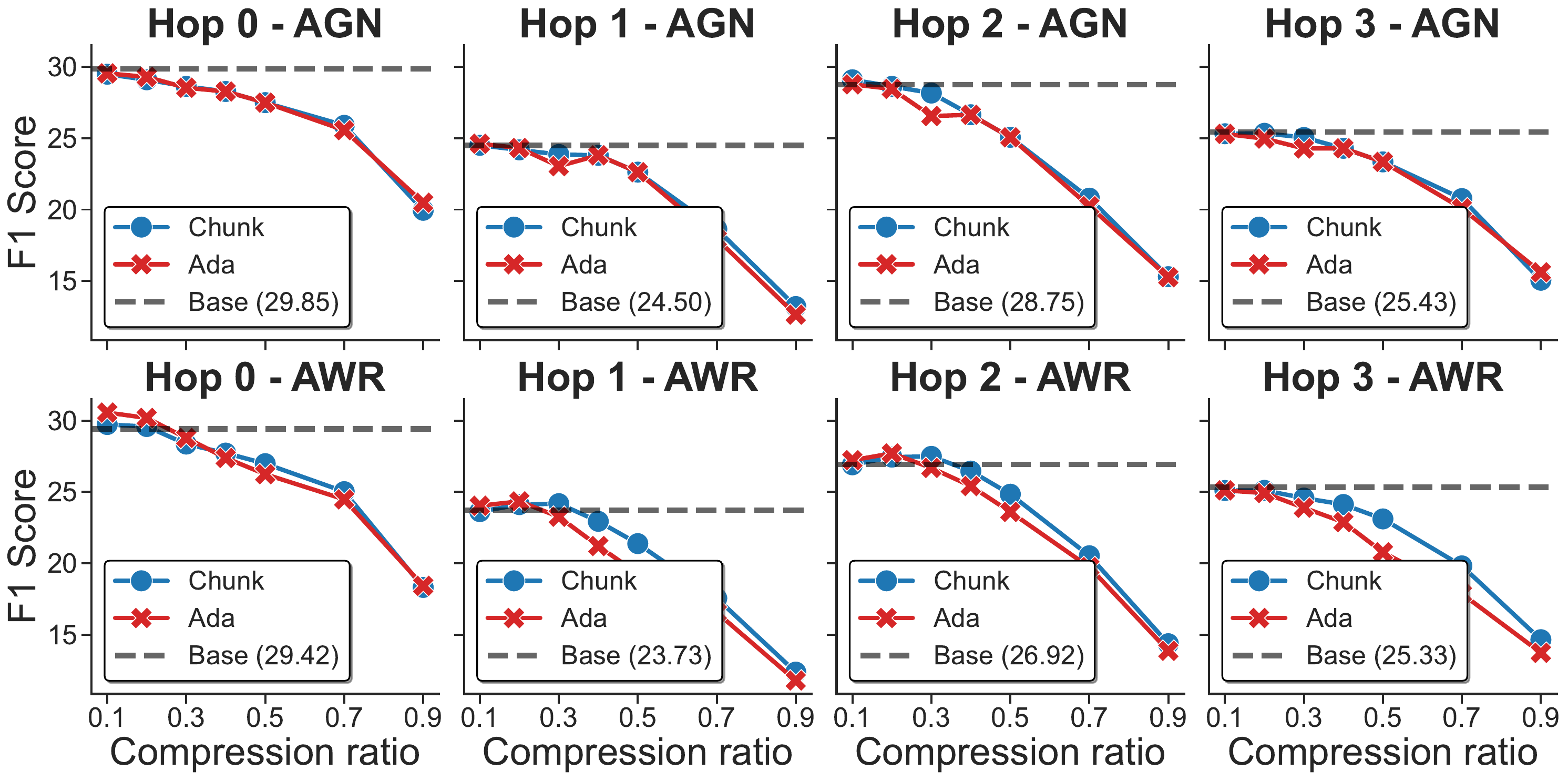}
    }
    \caption{Individual hop breakdown showing instability in intermediate semantic links.}
    \label{fig:HopsIndividualPerf}
\end{figure*}

\noindent \textbf{Long context and Hops} reveal (Figures~\ref{fig:LongContextPerf}, ~\ref{fig:HopsGeneralPerf} and~\ref{fig:HopsIndividualPerf}) deeper structural limits. Even without compression, Long context F1 remains modest (LLaMA 45.93 AGN; Qwen 38.68 AGN), and Hops is lower still (LLaMA 31.54 AGN; Qwen 27.13 AGN). Multi-hop reasoning degrades more rapidly than direct retrieval~\citep{li2025brief} as compression increases, suggesting that relational pathways are more fragile than token presence.

\begin{tcolorbox}[colback=white,colframe=blue!50,title=Structural Summary]
(1) Moderate compression can transiently improve performance by isolating sparse token-route substructures.  
(2) Architectural differences determine whether robustness emerges from dense retention (LLaMA) or instruction-conditioned reasoning (Qwen).  
(3) Multi-hop and long-range reasoning collapse structurally before complete token loss, indicating failure of semantic connectivity rather than simple memory removal.
\end{tcolorbox}

\subsection{Tag-Level Analysis}

While aggregate F1 trends reveal structural degradation under compression, tag-level analysis exposes the semantic and cognitive dimensions along which this degradation unfolds. By decomposing performance according to answer-type and question-type tags, we can distinguish between simple token loss and deeper representational fragility. Across datasets, compression does not affect all semantic categories uniformly; instead, it selectively erodes relational, hierarchical, and morphologically complex structures.

\paragraph{Answer-Type Tags.}

\begin{figure*}[!htb]
    \centering
    \subfloat[LLaMA-3 8B Instruct]{
    \includegraphics[width=\linewidth]{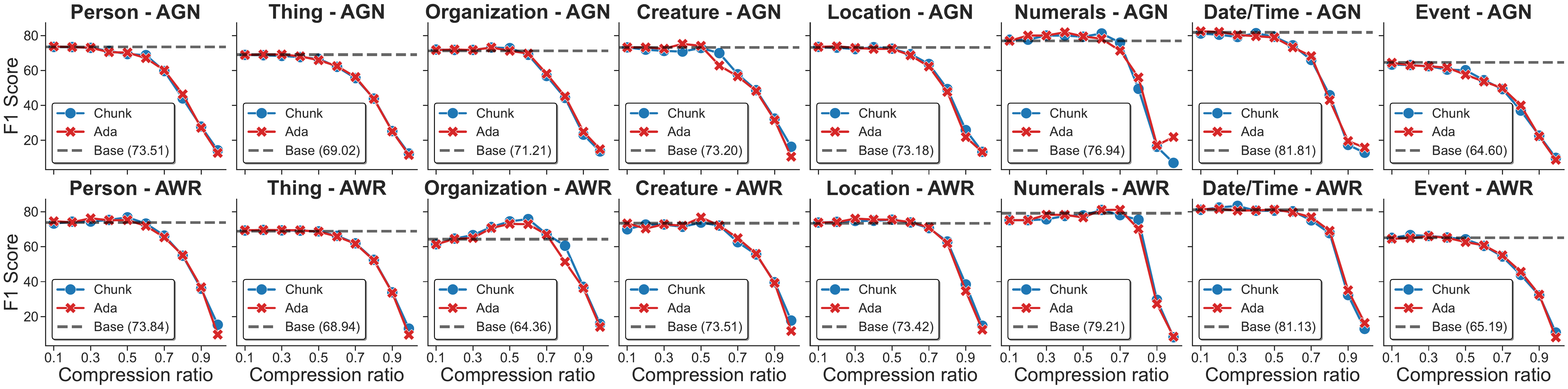}
    }\\
    \subfloat[Qwen-2.5 7B Instruct]{
    \includegraphics[width=\linewidth]{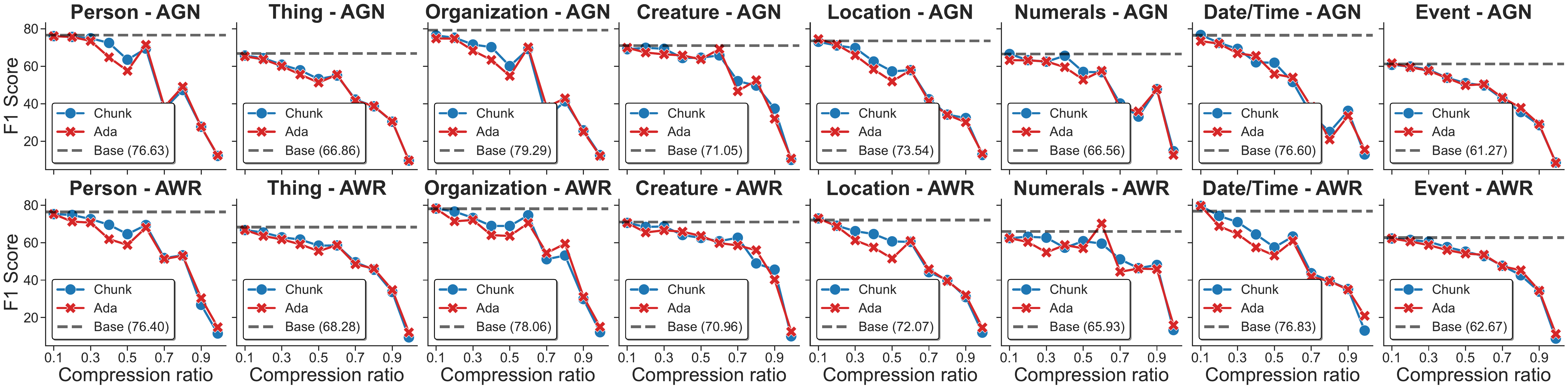}
    }
    \caption{Answer-type tag performance in Base task. Corresponding results for LLaMA-3.2 3B, Qwen-2.5 3B, and Qwen-2.5 14B are shown in Figure \ref{fig:BaseTaskTextApp} of Appendix~\ref{appx:result}.}
    \label{fig:BaseTaskTextPerf}
\end{figure*}

Figure~\ref{fig:BaseTaskTextPerf} shows answer-type performance for the Base task. Although the overall F1 score remains near 70\% at 0\% compression, tag-level trends reveal systematic variation. Categories such as \textit{Person} and \textit{Location} are relatively stable across compression levels, complementing the similar results obtained by ~\citet{liu2025clusterkv}. These entities are typically realized as concrete nouns with consistent surface forms, enabling direct lexical anchoring in the KV cache. Even when pruning removes a subset of tokens, surviving mentions are often sufficient for correct retrieval.

In contrast, \textit{Event} exhibits significantly sharper degradation. Event answers require normalization across verb conjugations, aspect markers, and paraphrased action phrases. Unlike atomic nouns, verbs are distributed across morphologically-varied tokens and syntactic contexts. Under compression, this distributed encoding fragments, leading to incomplete semantic reconstruction. Performance curves for the Event tag consistently show earlier inflection points compared to noun-based categories, indicating that relational and process-level representations are more fragile than entity storage.

\begin{figure*}[!htb]
    \centering
    \subfloat[LLaMA-3 8B Instruct]{
    \includegraphics[width=\linewidth]{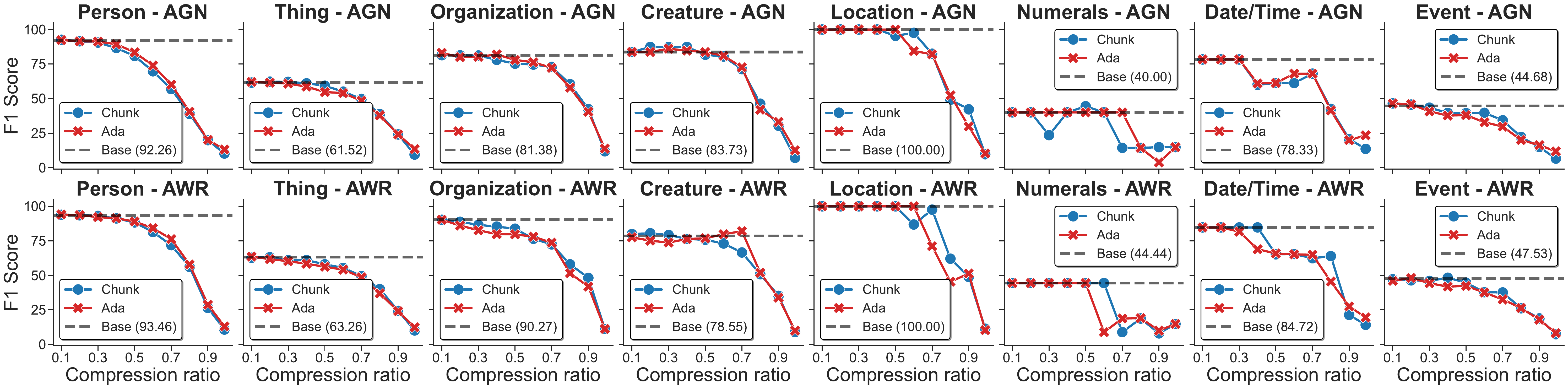}
    }\\
    \subfloat[Qwen-2.5 7B Instruct]{
    \includegraphics[width=\linewidth]{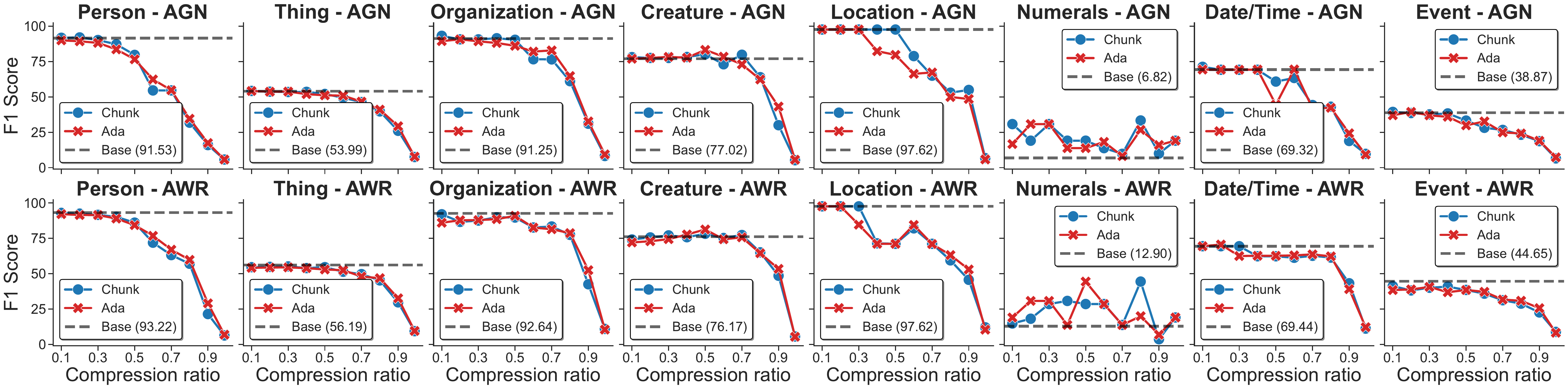}
    }
    \caption{Answer-type tag behavior in Multi entity. Corresponding results for LLaMA-3.2 3B, Qwen-2.5 3B, and Qwen-2.5 14B are shown in Figure 
    \ref{fig:MultiEntityTextApp} of Appendix~\ref{appx:result}.}
    \label{fig:MultiEntityTextPerf}
\end{figure*}

\begin{figure*}[!htb]
    \centering
    \subfloat[LLaMA-3 8B Instruct]{
    \includegraphics[width=\linewidth]{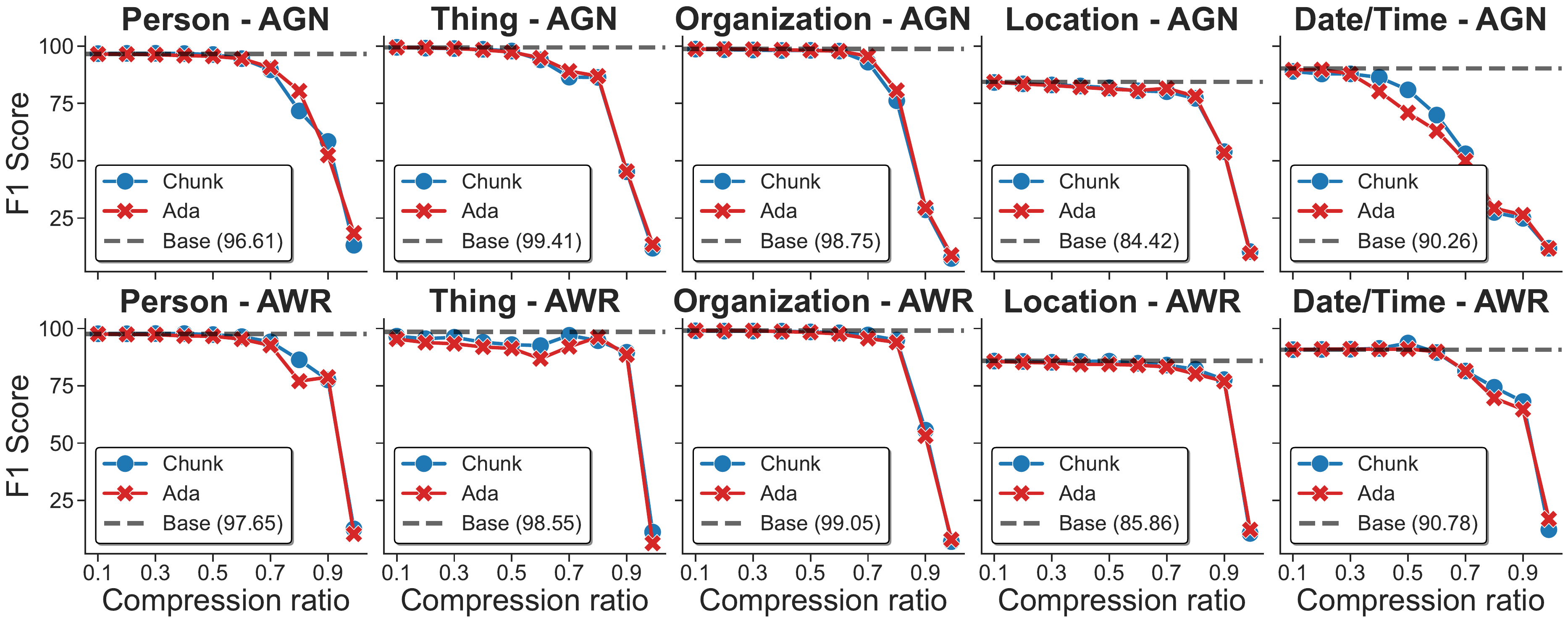}
    }\\
    \subfloat[Qwen-2.5 7B Instruct]{
    \includegraphics[width=\linewidth]{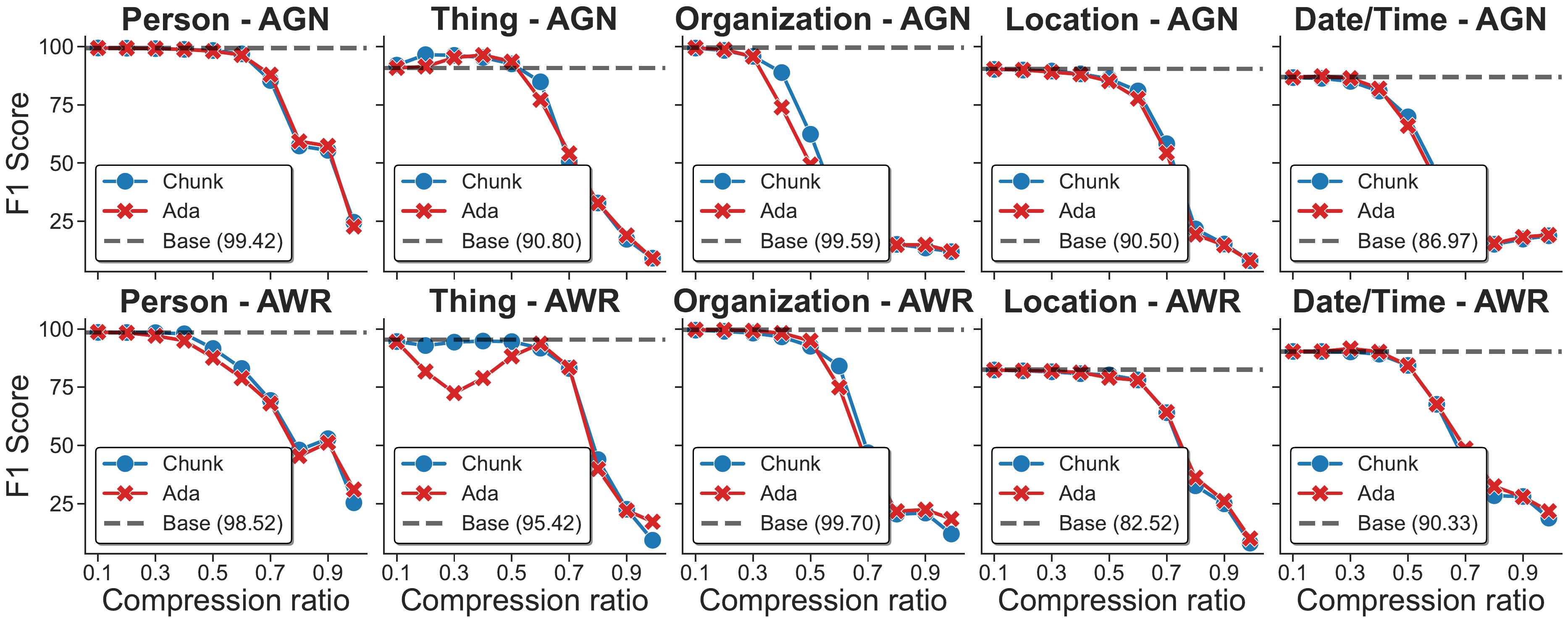}
    }
    \caption{Knowledge manipulation answer-type tag-level behavior.}
    \label{fig:KnowledgeManTextPerf}
\end{figure*}

\begin{figure*}[!htb]
    \centering
    \subfloat[LLaMA-3 8B Instruct]{
    \includegraphics[width=0.49\linewidth]{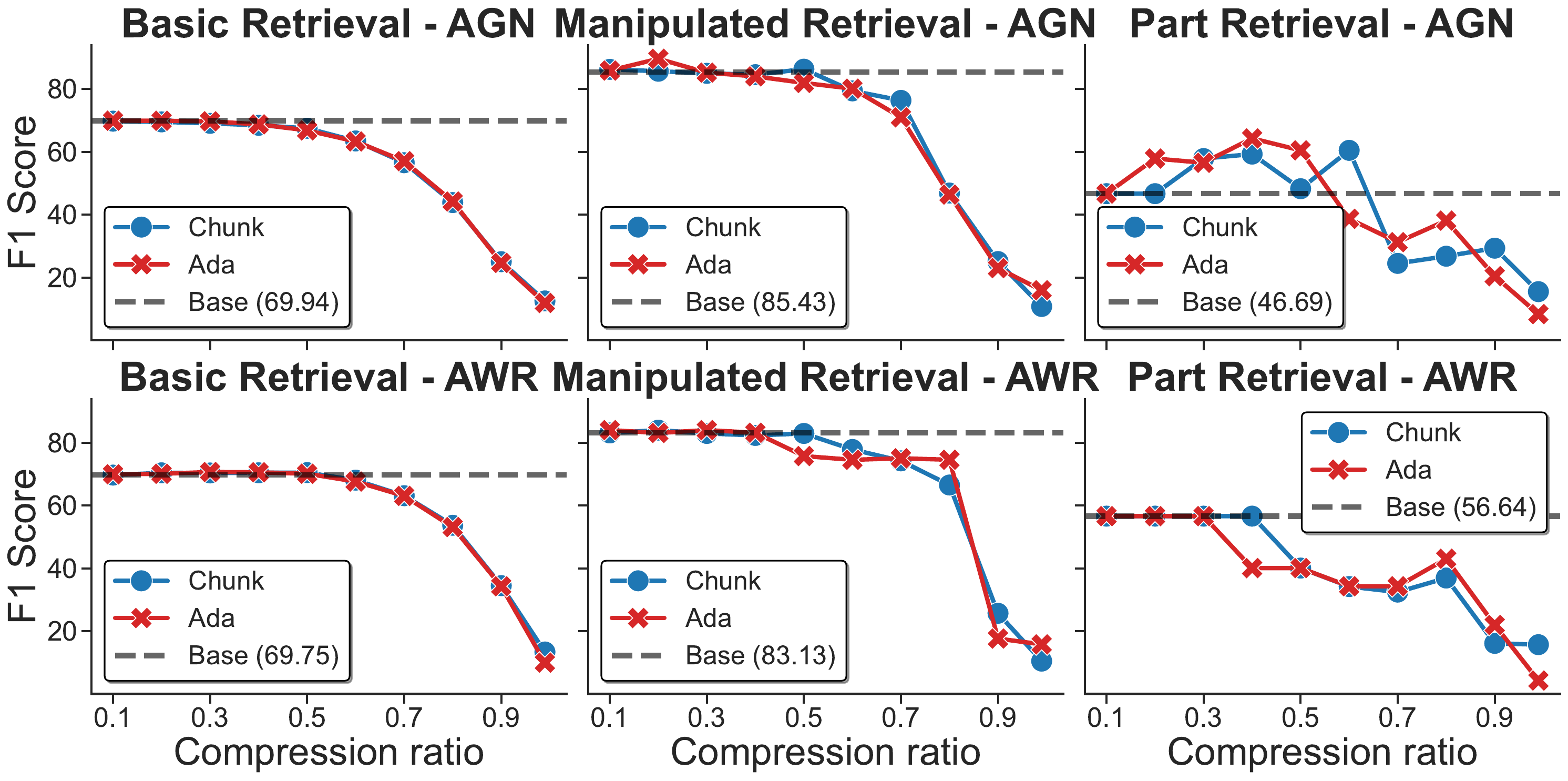}
    }
    \subfloat[Qwen-2.5 7B Instruct]{
    \includegraphics[width=0.49\linewidth]{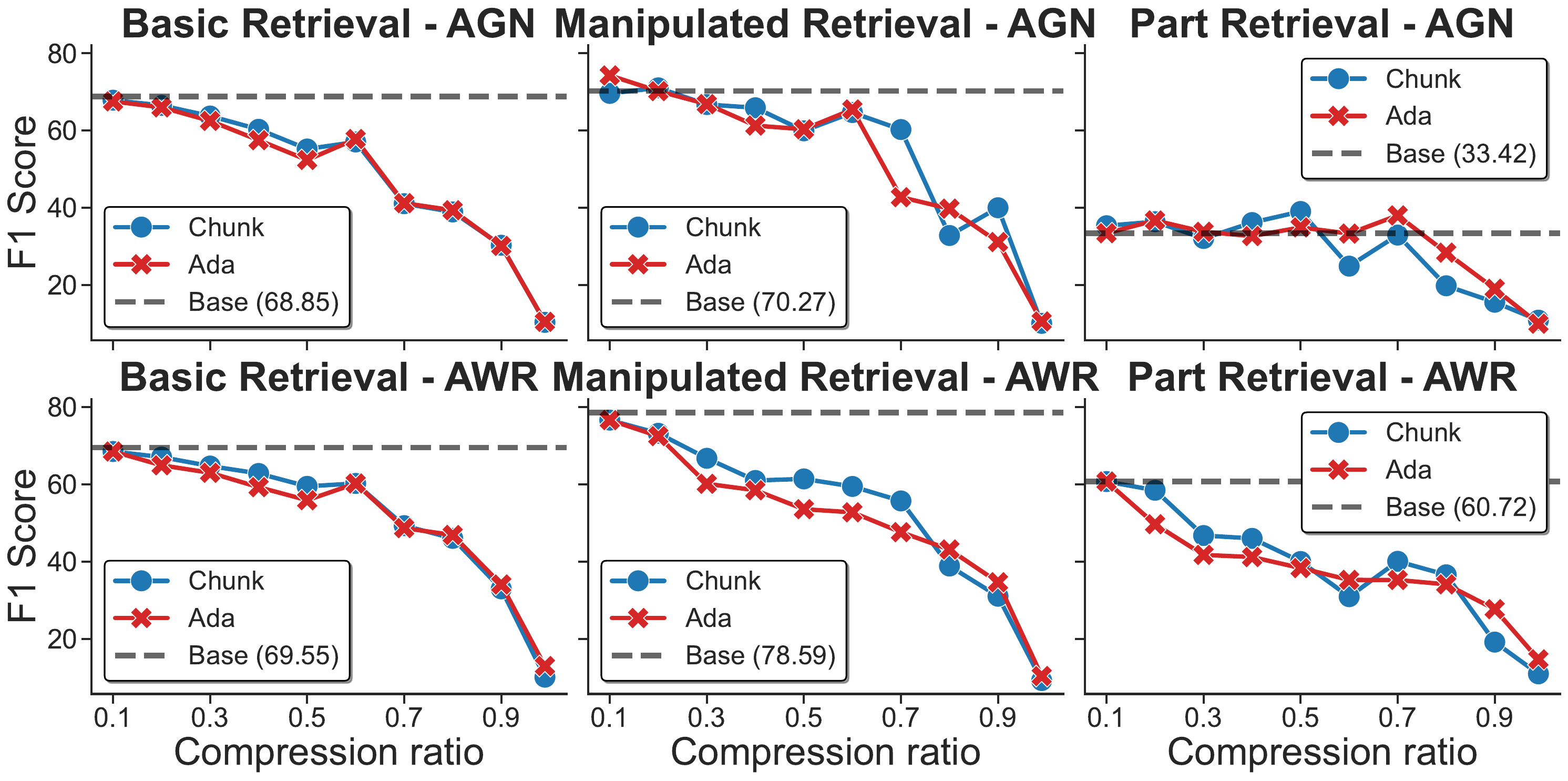}
    }
    \caption{Question-type tags for Base task. Corresponding results for LLaMA-3.2 3B, Qwen-2.5 3B, and Qwen-2.5 14B are shown in Figure \ref{fig:BaseTaskQuestionApp} of Appendix~\ref{appx:result}.}
    \label{fig:BaseTaskQuestionPerf}
\end{figure*}

\begin{figure*}[!htb]
    \centering
    \subfloat[LLaMA-3 8B Instruct]{
    \includegraphics[width=0.49\linewidth]{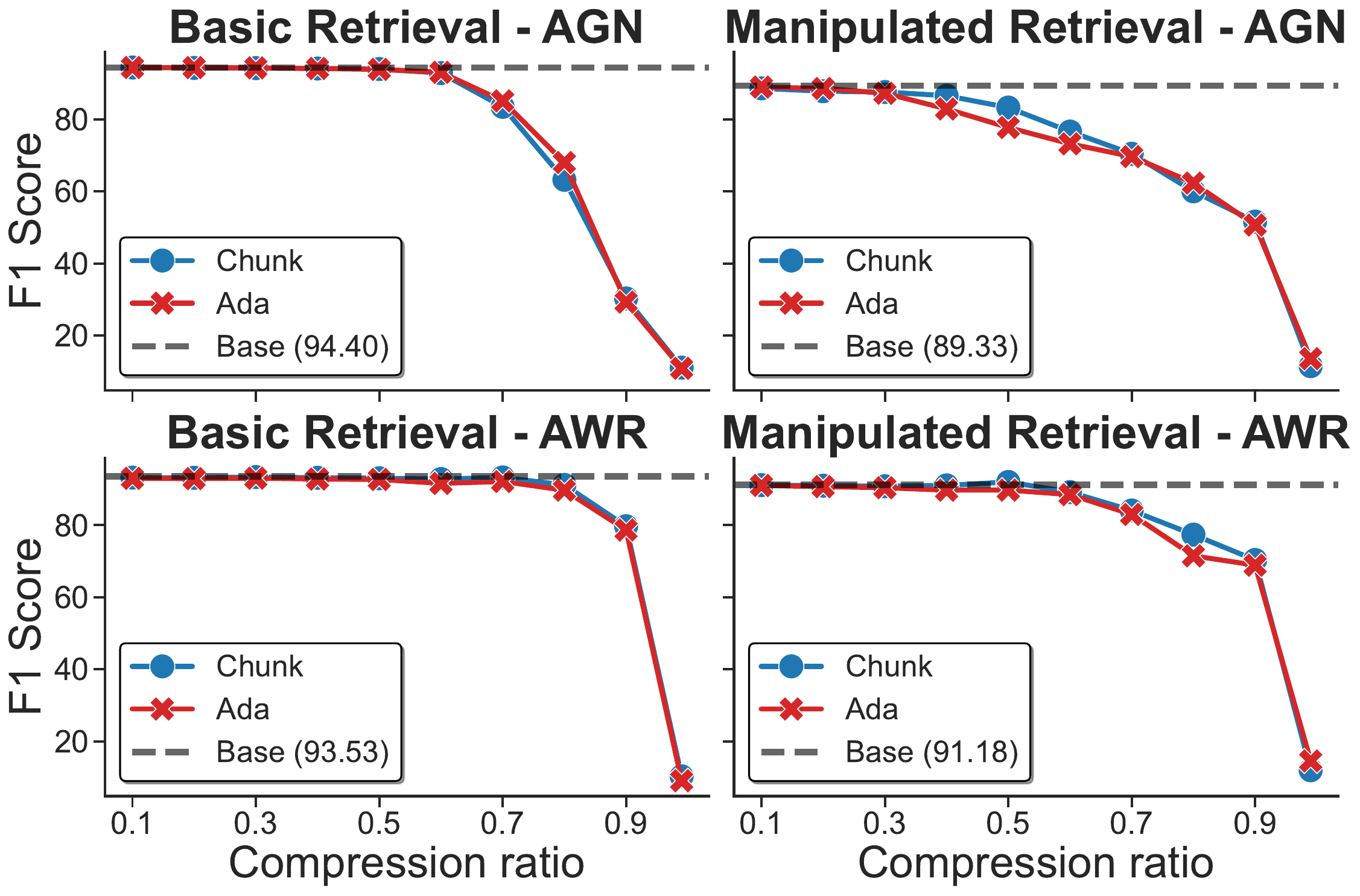}
    }
    \subfloat[Qwen-2.5 7B Instruct]{
    \includegraphics[width=0.49\linewidth]{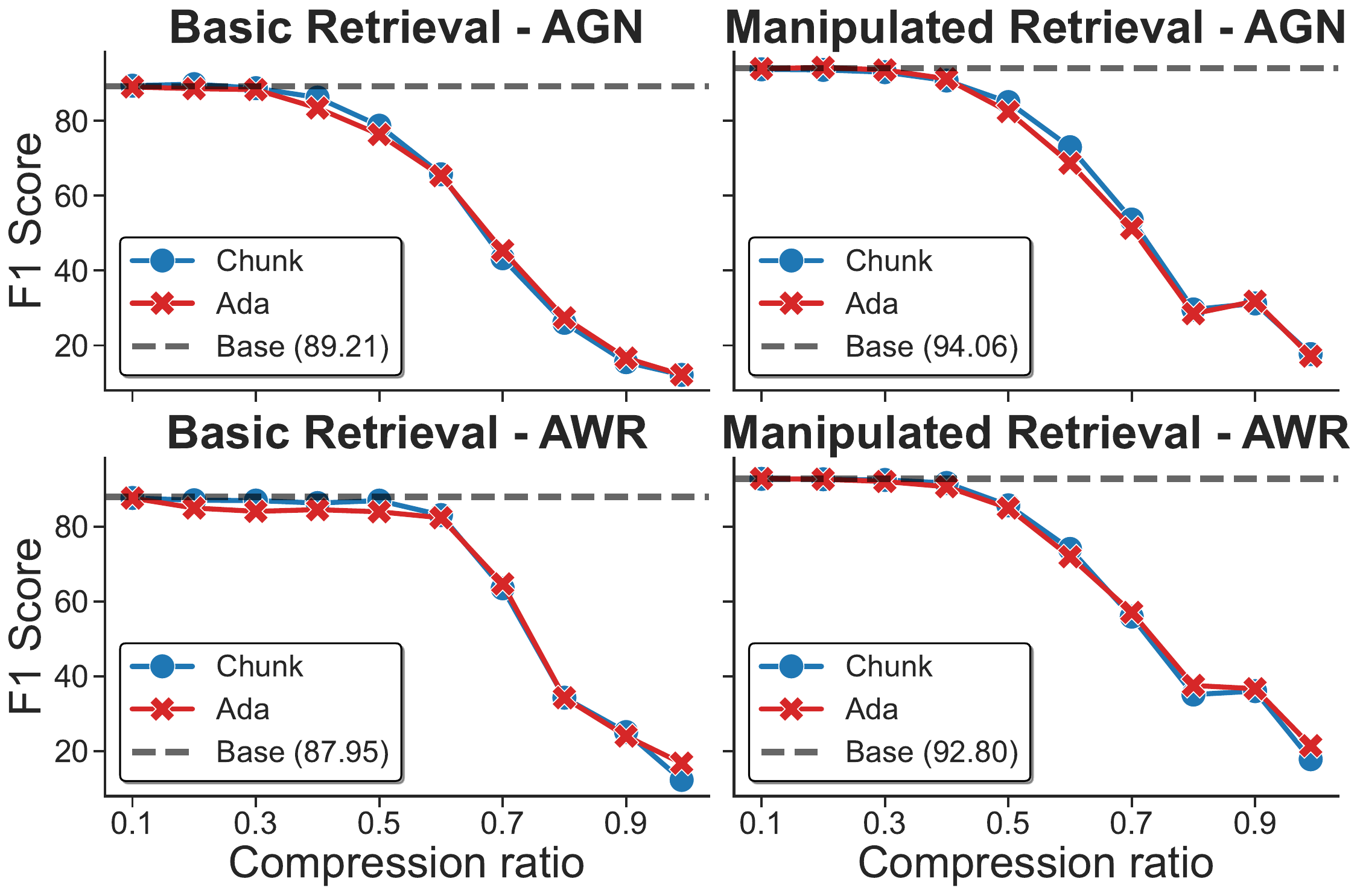}
    }
    \caption{Knowledge manipulation question-type analysis.}
    \label{fig:KnowledgeManQuestionPerf}
\end{figure*}

\begin{figure*}[!htb]
    \centering
    \subfloat[LLaMA-3 8B Instruct]{
    \includegraphics[width=0.49\linewidth]{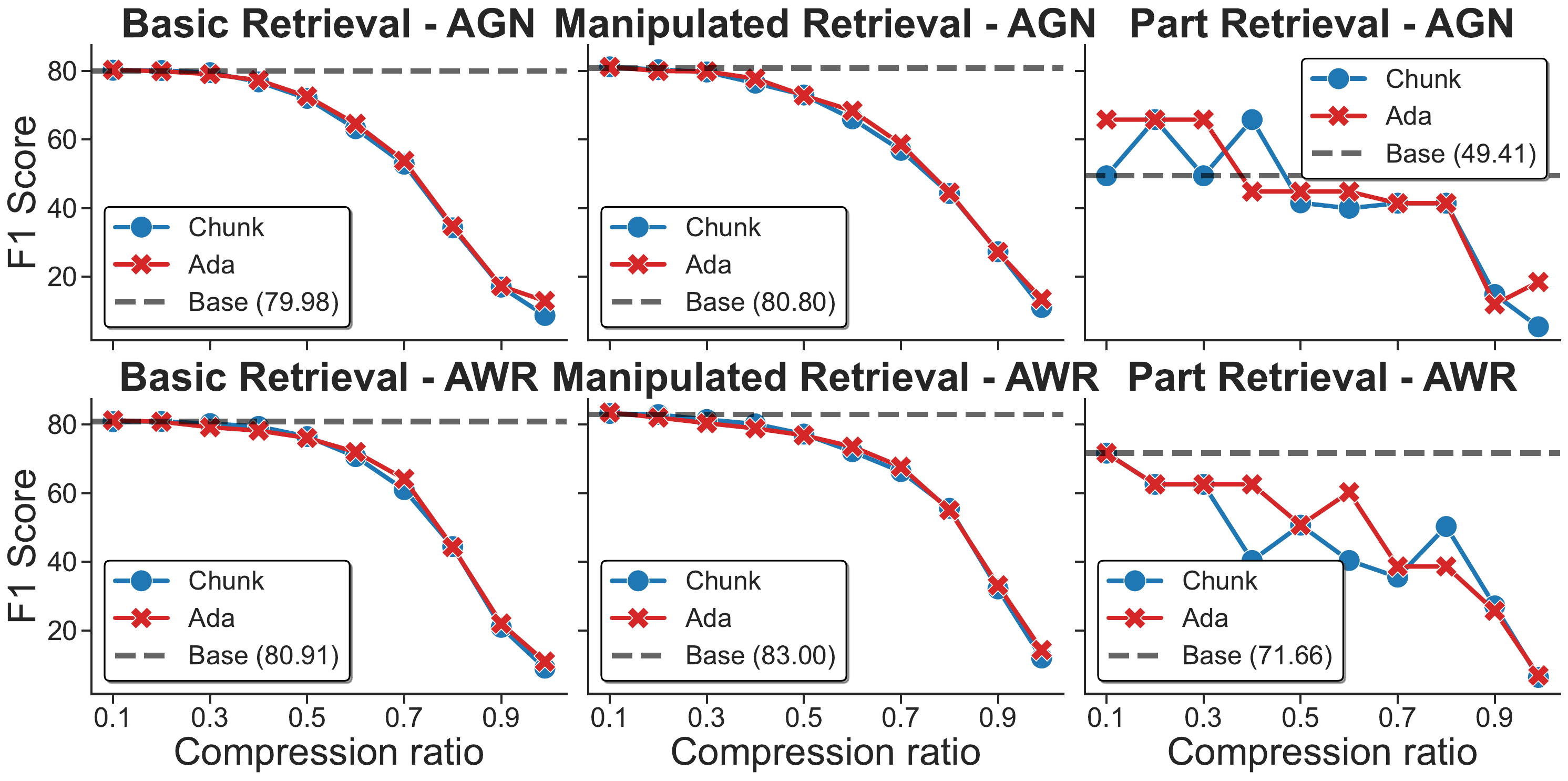}
    }
    \subfloat[Qwen-2.5 7B Instruct]{
    \includegraphics[width=0.49\linewidth]{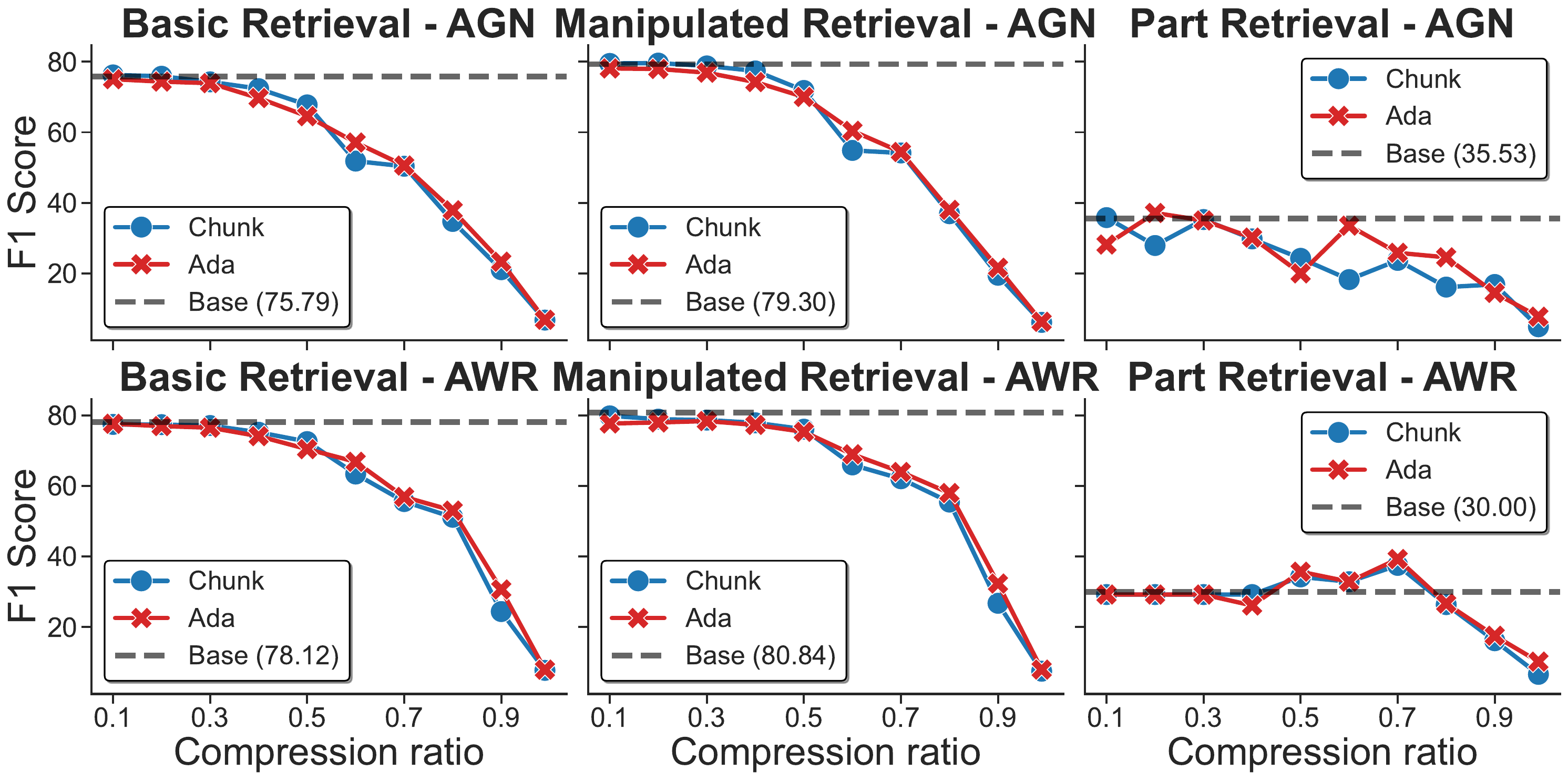}
    }
    \caption{Multi entity question-type behavior shows a similar trend to Base task but with much smoother curves in question agnostic. The question aware setup has a far more jagged curve with distinct points of significant drop off. Corresponding results for LLaMA-3.2 3B, Qwen-2.5 3B, and Qwen-2.5 14B are shown in Figure \ref{fig:MultiEntityQuestionApp} of Appendix~\ref{appx:result}.}
    \label{fig:MultiEntityQuestionPerf}
\end{figure*}

The \textit{Organization} tag reveals a distinct but related weakness. Organizational entities frequently encode nested relational structure, including sub-entities, roles, or hierarchical relationships. In the question-aware setup, both models show sharper declines for Organization-tagged answers compared to the question-agnostic condition. This suggests that query-conditioned pruning may collapse hierarchical distinctions into coarse labels. Under compression, structured representations are simplified, and fine-grained relational links are among the first to degrade. The failure pattern indicates not merely loss of tokens, but loss of structural decomposition.

Multi-entity tag results  reinforce this interpretation, as shown in Figure~\ref{fig:MultiEntityTextPerf}. The \textit{Thing} and \textit{Organization} tags display more irregular degradation curves, especially under moderate-to-high compression, while \textit{Person} and \textit{Location} remain comparatively stable. This asymmetry suggests that compression preferentially harms categories requiring cross-token relational binding rather than those represented by isolated lexical anchors.

Knowledge manipulation further clarifies architectural divergence. As shown in Figure~\ref{fig:KnowledgeManTextPerf}, Qwen exhibits differentiated degradation across tags, whereas LLaMA shows more synchronized decline. This suggests that Qwen allocates representational capacity unevenly across semantic domains, preserving certain categories longer under compression, while LLaMA maintains more uniform retention but collapses more abruptly when compression becomes extreme.

\paragraph{Question-Type Tags.}

Question-type tags reveal how compression interacts with cognitive demands. In the Base task (Figure~\ref{fig:BaseTaskQuestionPerf}), \textit{Standard} retrieval questions are relatively stable until moderate compression. These tasks require direct extraction of explicitly stated facts and thus depend primarily on token survival.

\textit{Manipulated} questions, which require transformation or reinterpretation of context, expose architectural differences. Qwen maintains closer alignment between AGN and AWR settings, indicating robust instruction-conditioning. LLaMA, while strong in raw retrieval, exhibits sharper divergence under AWR in manipulation-heavy tasks.

\textit{Part} retrieval questions degrade earliest and most irregularly. These queries require integrating scattered phrases across different regions of the context. Even when individual tokens survive eviction, the relational pathways connecting them become fragile. Performance curves for Part questions exhibit jagged transitions rather than smooth decline, indicating intermittent route failure rather than uniform capacity reduction.

Knowledge manipulation question tags (Figure~\ref{fig:KnowledgeManQuestionPerf}) highlight Qwen's strength in \textit{Manipulated} queries, where it outperforms LLaMA more consistently under compression. However, in pure \textit{Standard} retrieval, LLaMA retains an advantage. This reinforces the interpretation that the two architectures allocate representational resources differently: one optimized for instruction-conditioned flexibility, the other for dense storage and direct recall.

In Multi entity (Figure~\ref{fig:MultiEntityQuestionPerf}), \textit{Part} retrieval again exhibits the most instability. Performance oscillates across compression levels, indicating a threshold phenomenon where distributed semantic cues become intermittently unreachable. This instability aligns with the notion of representational rigidity: tokens remain present, but routing flexibility diminishes, preventing consistent multi-span integration.

\paragraph{Synthesis of Tag-Level Patterns.}

Across answer-type and question-type dimensions, three consistent patterns emerge. First, relational and hierarchical categories (\textit{Event}, \textit{Organization}, \textit{Part}) degrade earlier than atomic entity categories. Second, compression interacts with architecture-specific biases: Qwen preserves manipulation robustness longer, whereas LLaMA maintains stronger direct retention. Third, degradation often manifests as instability rather than smooth decline, suggesting phase transitions in token-route connectivity rather than linear capacity reduction.

These findings reinforce the interpretation that KV compression erodes semantic connectivity before eliminating token storage. Tag-level results, therefore, provide empirical support for viewing compression as a structural perturbation of token-route graphs, with relational categories serving as early indicators of route collapse.

\section{Discussion}

We interpret KV caching as a \emph{routing substrate}: compression perturbs not only memory size, but the existence and exploitability of token-level routes that carry evidence to the decoder. Across our analyses, two quantities dominate: (i) \emph{reachability} -- whether answer-critical tokens remain accessible through \emph{any} surviving head-wise pathway, and (ii) \emph{adaptivity} -- whether the model retains enough head diversity to re-route attention when the context becomes sparse. We organize the discussion around these mechanisms and connect them to architectural depth profiles, the universal safety cliff, and probing-based evidence of a representation--behavior gap.

\subsection{Structural Metrics for KV Compression}

Compression can cause failure either by \emph{erasing} necessary evidence or by leaving evidence present but \emph{poorly used}. To separate these effects, we track \textit{eviction} (survival) and \textit{consensus} (coordination). Let $T$ be the context token set and $H$ the head set. For token $t$, compression level $\alpha$, head $h$, and survived-token set $t^{(\alpha,h)}$, define
\[
S_h(t,\alpha)=
\begin{cases}
1 & \text{if } t\in t^{(\alpha,h)},\\
0 & \text{otherwise}.
\end{cases}
\qquad
\text{GlobalEvicted}(t,\alpha)=\mathbb{1}\Bigl[\sum_{h\in H} S_h(t,\alpha)=0\Bigr].
\]
The Eviction Rate is
\[
\text{EvictionRate}(\alpha)=\frac{1}{|T|}\sum_{t\in T}\text{GlobalEvicted}(t,\alpha),
\]
and the task-aware Global Eviction Ratio (GER) restricts to answer-relevant tokens $T_{\text{ans}}$:
\[
\text{GER}(\alpha)=\frac{1}{|T_{\text{ans}}|}\sum_{t\in T_{\text{ans}}}\text{GlobalEvicted}(t,\alpha).
\]
GER measures \emph{route deletion} at the evidence level: when GER is high, the model has no surviving access path to ground truth tokens, making hallucination structurally likely (Figure~\ref{fig:corr_ger_hal}). Survival alone is insufficient; however, the model must also \emph{coordinate} its remaining capacity. Let $A_{\ell,h}(t)$ be normalized attention weight from head $h$ at layer $\ell$ to token $t$, with top-attended token
\[
t^*_{\ell,h}=\arg\max_t A_{\ell,h}(t).
\]
We measure layer-wise head consensus as
\[
\text{Consensus}(\ell)=\frac{\bigl|\{t^*_{\ell,h}\}_{h\in H}\bigr|}{|H|},
\]
where lower values indicate stronger agreement (many heads focus on the same token) and higher values indicate diversity. Together, eviction and consensus let us distinguish \emph{representational erasure} (routes destroyed) from \emph{representational rigidity} (routes present but insufficiently adaptable).

\subsection{What does KV compression reveal about how different architectures allocate computation across depth?}

\begin{figure*}[!htb]
    \centering
    \subfloat[LLaMA-3 8B Instruct]{
    \includegraphics[width=0.49\linewidth]{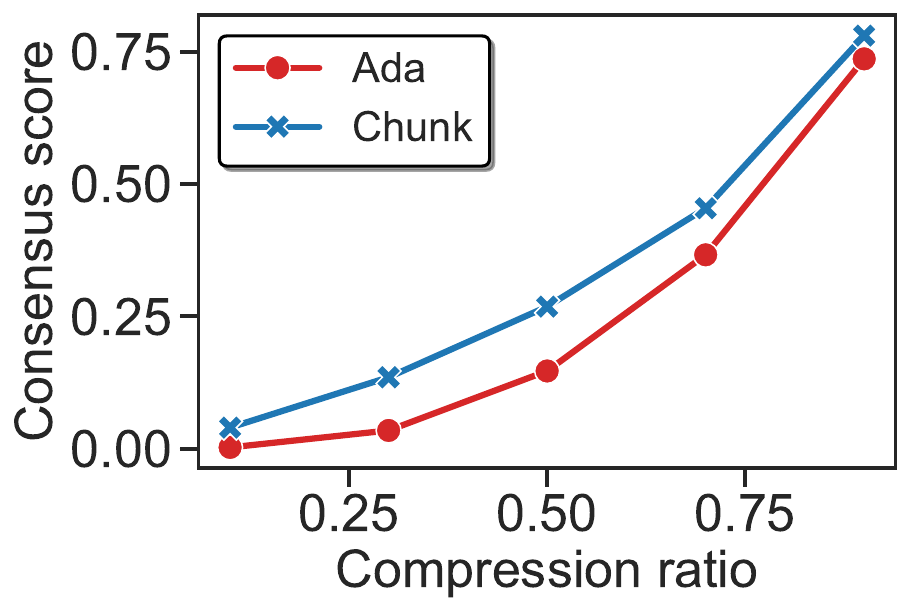}
    }
    \subfloat[Qwen-2.5 7B Instruct]{
    \includegraphics[width=0.49\linewidth]{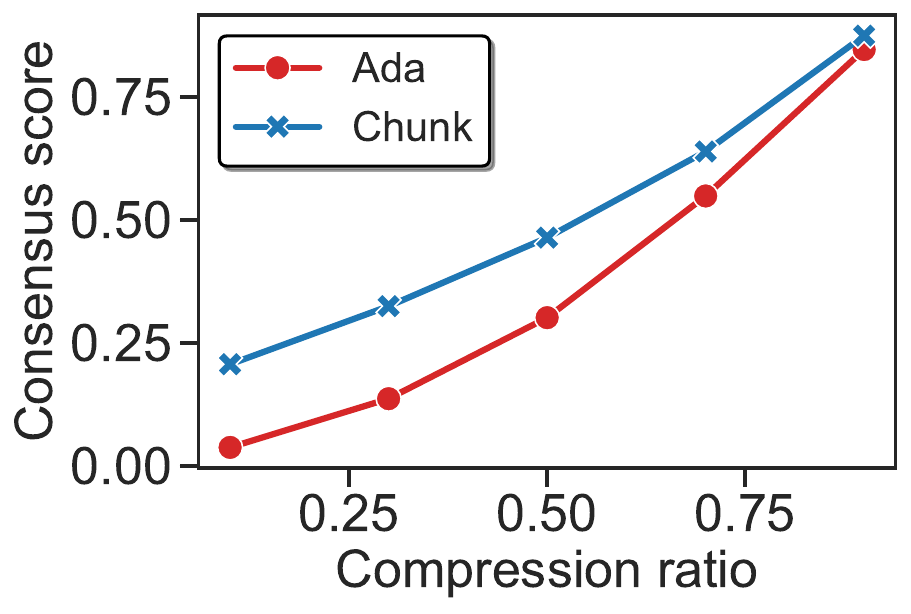}
    }
    \caption{Consensus across layers. LLaMA exhibits early stabilization and late specialization; Qwen exhibits early exploration and late consolidation. Corresponding results for LLaMA-3.2 3B, Qwen-2.5 3B, and Qwen-2.5 14B are shown in Figure \ref{fig:ConsensusScoresApp} of Appendix~\ref{appx:result}.}
    \label{fig:ConsensusScores}
\end{figure*}

\begin{figure*}[!htb]
    \centering
    \subfloat[LLaMA-3 8B Instruct 10\% \\Compression]{
    \includegraphics[width=0.33\linewidth]{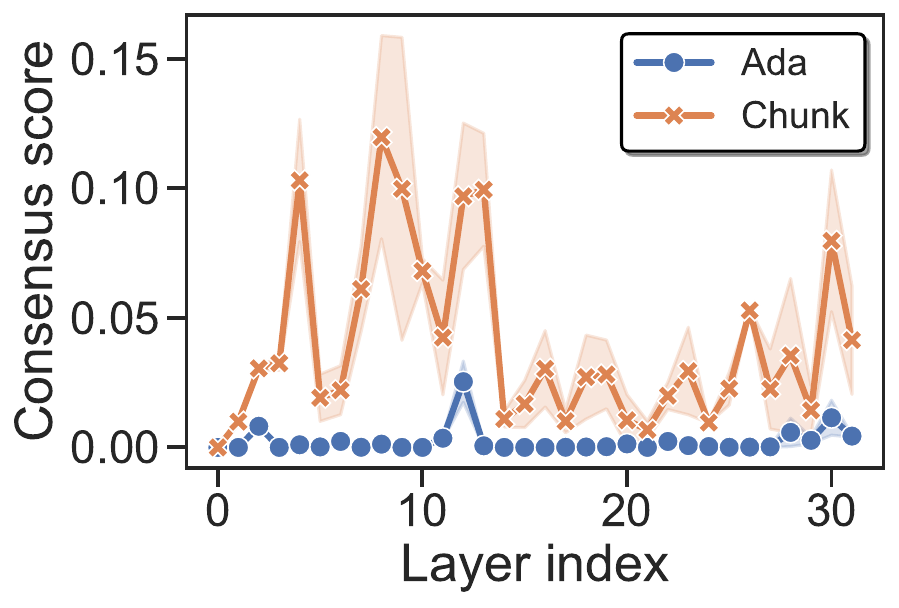}
    }
    \subfloat[LLaMA-3 8B Instruct 50\% \\Compression]{
    \includegraphics[width=0.33\linewidth]{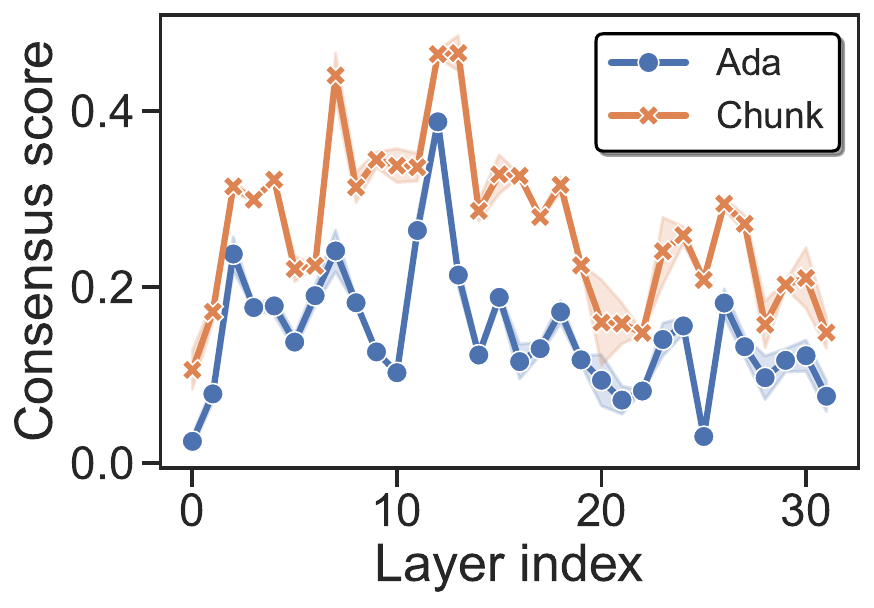}
    }
    \subfloat[LLaMA-3 8B Instruct 90\% \\Compression]{
    \includegraphics[width=0.33\linewidth]{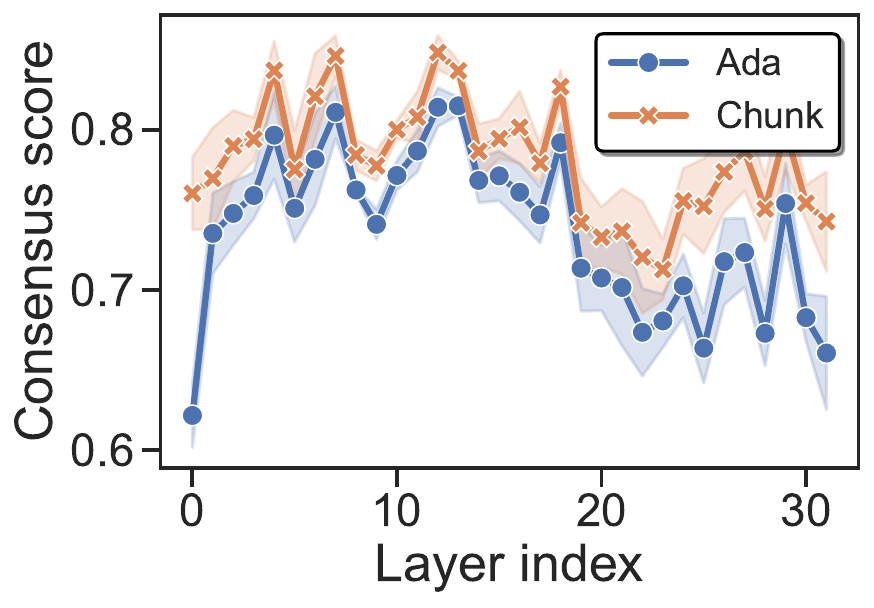}
    }
    \caption{LLaMA layerwise consensus under compression. Early agreement persists, with depth-wise diversification supporting parallel feature channels. Corresponding results for LLaMA-3.2 3B are shown in Figure \ref{fig:LLaMAConsensusLayerApp} of Appendix~\ref{appx:result}.}
    \label{fig:LLaMALayerConsensus}
\end{figure*}

\begin{figure*}[!htb]
    \centering
    \subfloat[Qwen-2.5 7B Instruct 10\% \\Compression]{
    \includegraphics[width=0.33\linewidth]{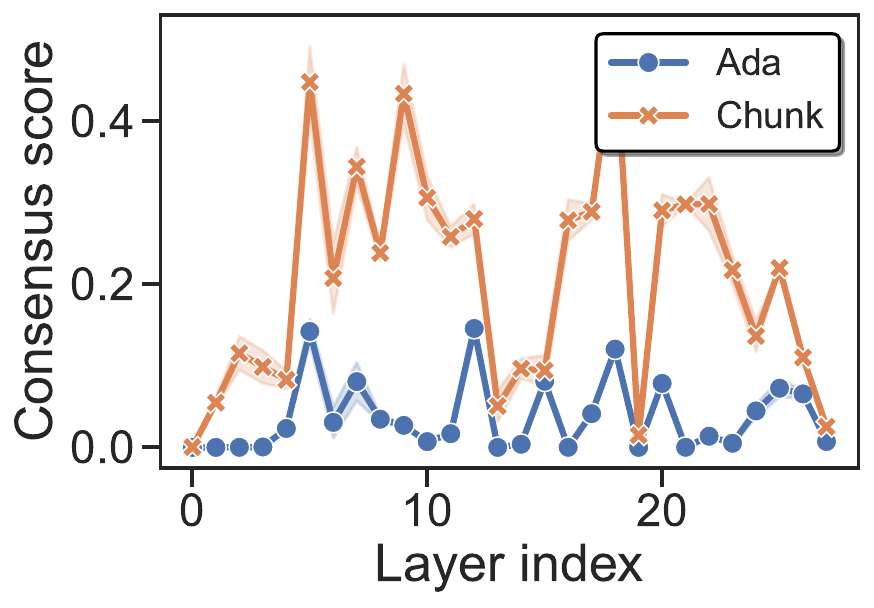}
    }
    \subfloat[Qwen-2.5 7B Instruct 50\% \\Compression]{
    \includegraphics[width=0.33\linewidth]{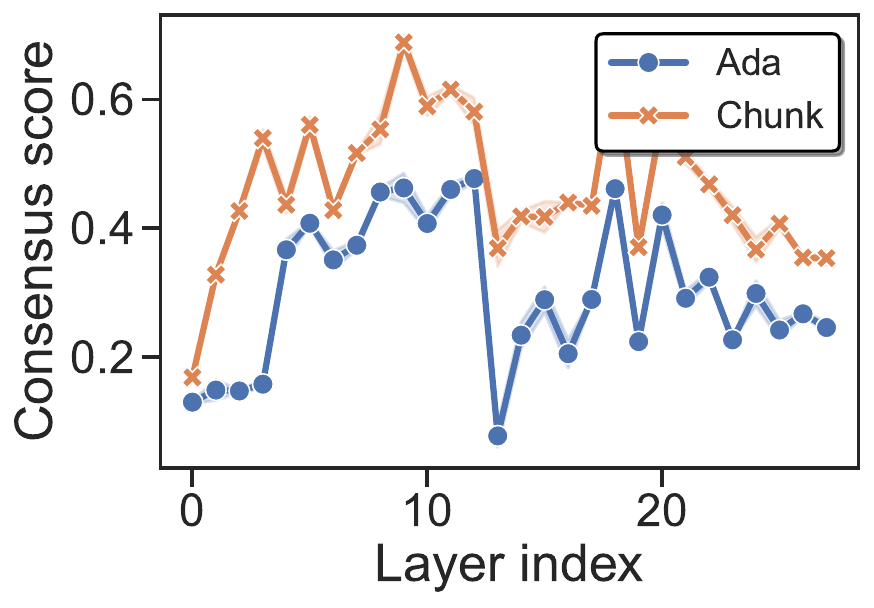}
    }
    \subfloat[Qwen-2.5 7B Instruct 90\% \\Compression]{
    \includegraphics[width=0.33\linewidth]{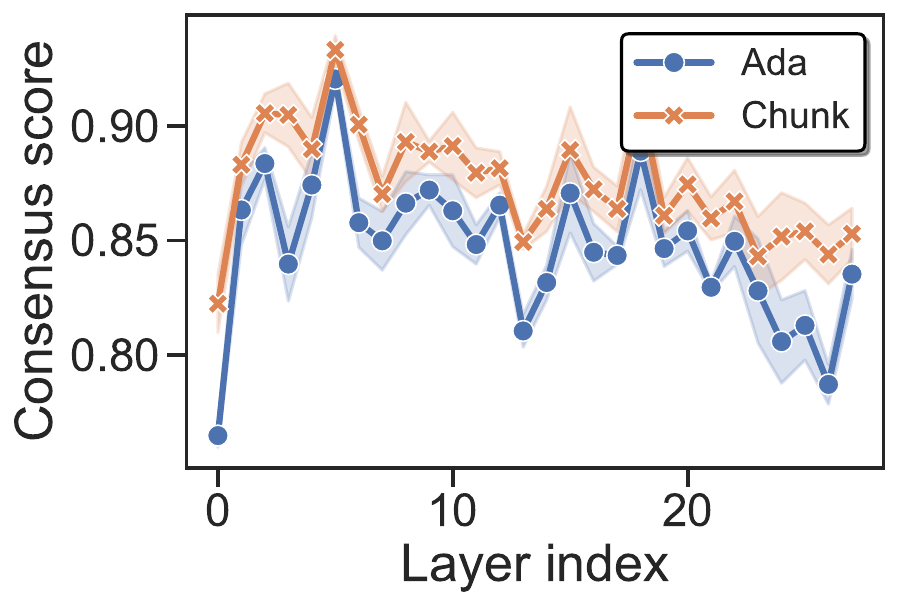}
    }
    \caption{Qwen layerwise consensus under compression. Diversity persists deeper and collapses late, consistent with a consolidation-heavy tail. Corresponding results for Qwen-2.5 3B and Qwen-2.5 14B are shown in Figure \ref{fig:QwenConsensusLayerApp} of Appendix~\ref{appx:result}.}
    \label{fig:QwenLayerConsensus}
\end{figure*}

A consistent architectural inversion appears between the LLaMA and Qwen families: LLaMA tends to stabilize early and diversify later, whereas Qwen tends to explore early and consolidate late. This is visible in layer-wise consensus trends (Figure~\ref{fig:ConsensusScores}) and in how consensus patterns evolve under increasing compression (Figures~\ref{fig:LLaMALayerConsensus} and~\ref{fig:QwenLayerConsensus}). Mechanistically, these profiles imply different ``decision depths'': in LLaMA, early-layer agreement can act like normalization, supporting later specialization; in Qwen, deep-layer consolidation can act like a late-stage decision cascade, concentrating routing onto a narrow token set.

\begin{figure*}[!htb]
    \centering
    \subfloat[10\% Compression]{
    \includegraphics[width=0.8\linewidth]{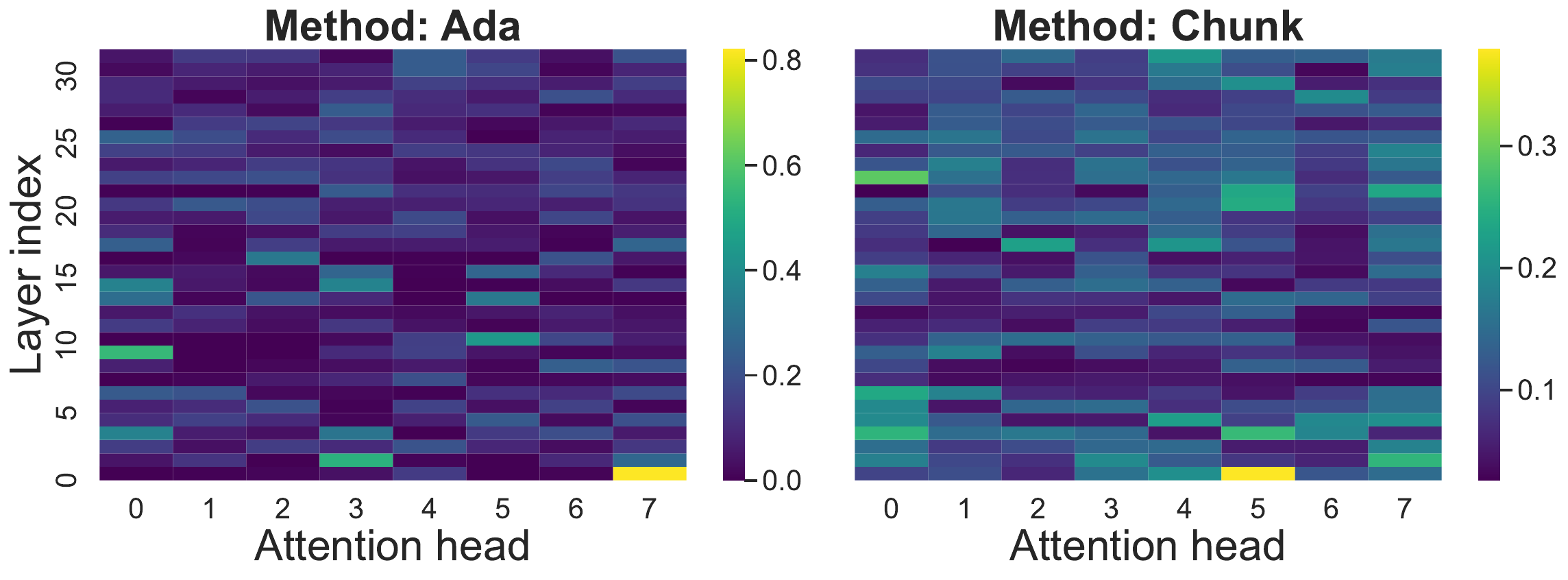}
    }\\
    \subfloat[90\% Compression]{
    \includegraphics[width=0.8\linewidth]{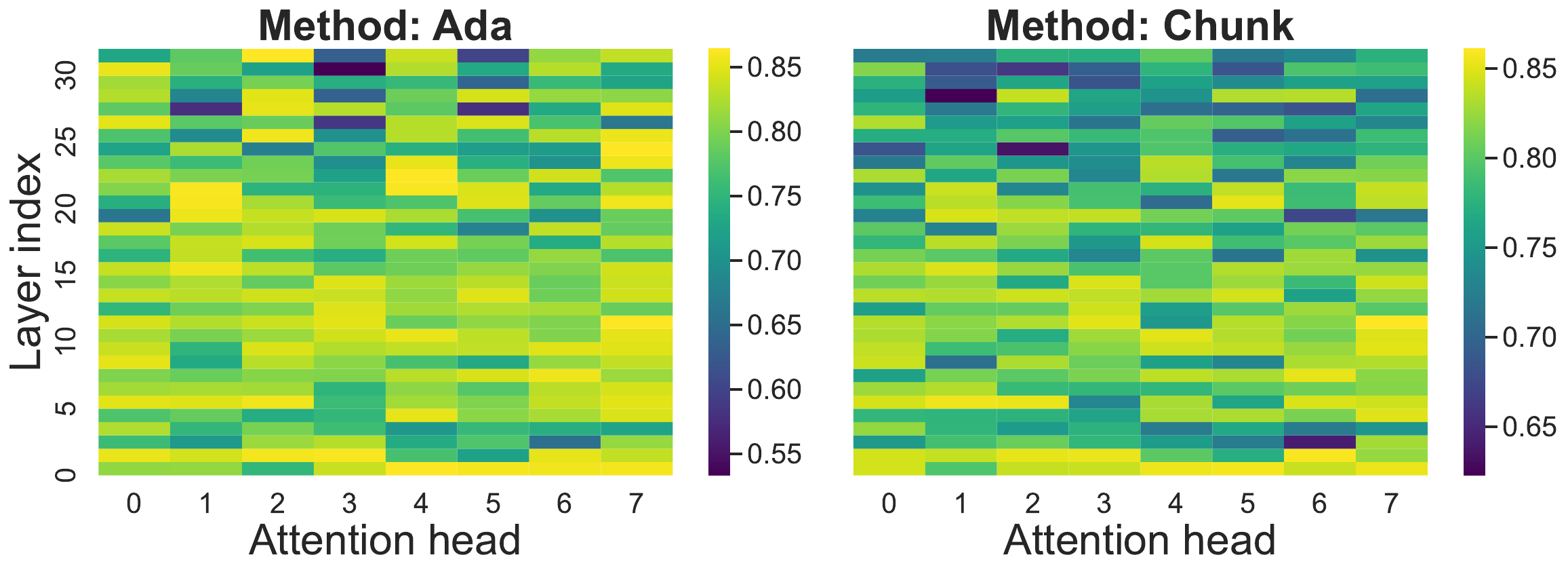}
    }
    \caption{LLaMA attention heatmaps under compression. Chunk pruning removes contiguous segments; Ada pruning is more irregular. At 0.9, global route destruction dominates. Corresponding results for LLaMA-3.2 3B are shown in Figure \ref{fig:LLaMA3BAttnHeatmapsApp} of Appendix~\ref{appx:result}.}
    \label{fig:LLaMA8BAttnHeatmaps}
\end{figure*}

\begin{figure*}[!htb]
    \centering
    \subfloat[10\% Compression]{
    \includegraphics[width=0.8\linewidth]{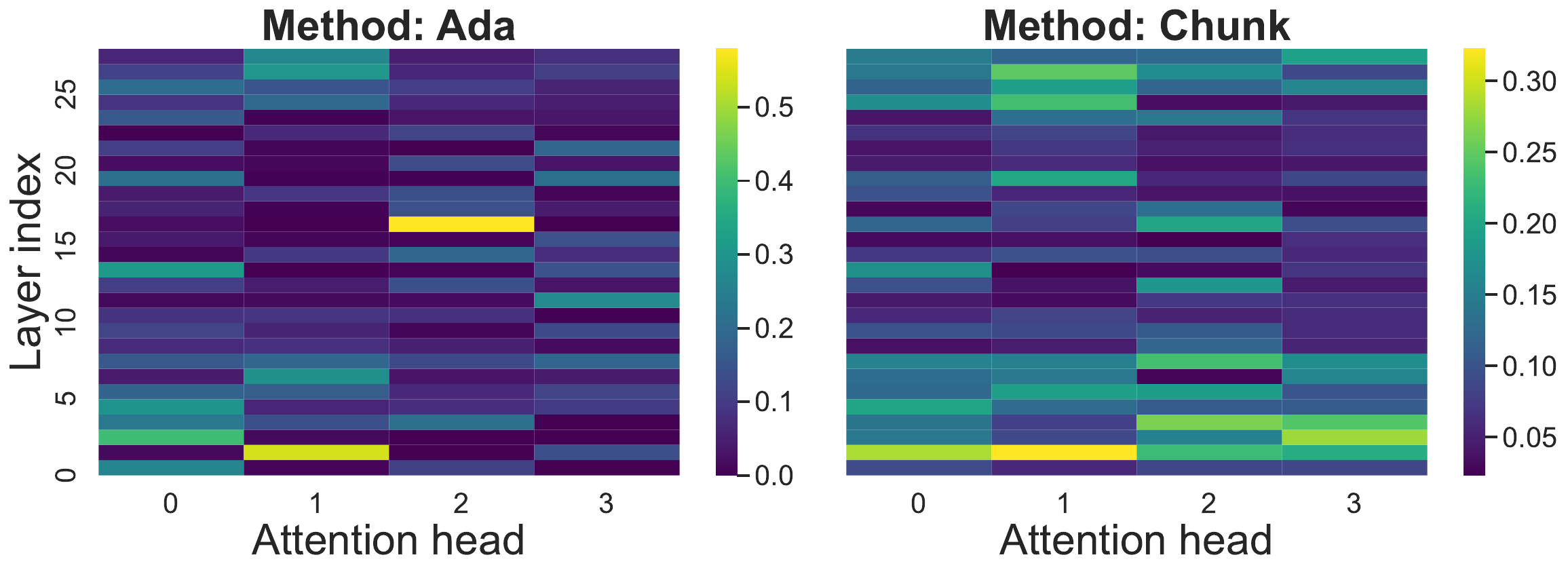}
    }\\
    \subfloat[90\% Compression]{
    \includegraphics[width=0.8\linewidth]{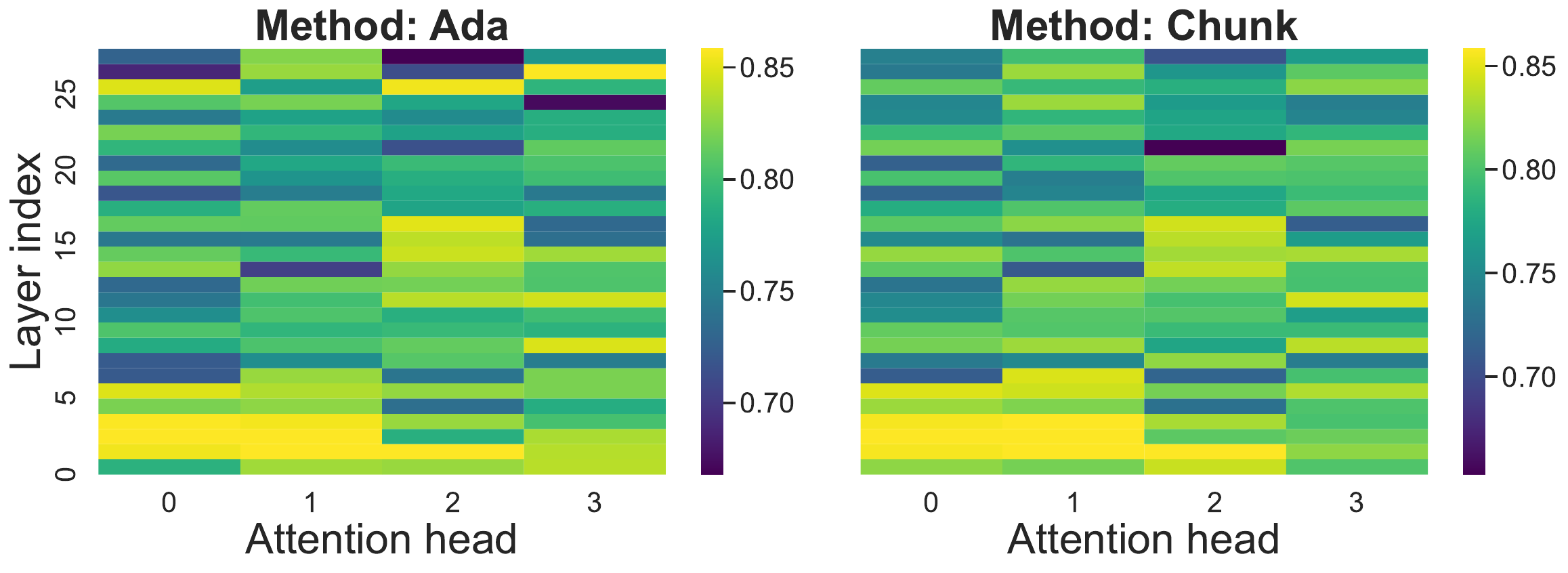}
    }
    \caption{Qwen attention heatmaps under compression. Policy differences are visible at low compression but are overwhelmed at 0.9 by large-scale pruning. Corresponding results for Qwen-2.5 3B, and Qwen-2.5 14B are shown in Figure \ref{fig:Qwen3BAttnHeatmapsApp} and Figure \ref{fig:Qwen14BAttnHeatmapsApp} of Appendix~\ref{appx:result}.}
    \label{fig:Qwen7BAttnHeatmaps}
\end{figure*}

This depth allocation has direct implications for compression design: policies that assume a universal ``pyramid'' (monotone entropy decrease with depth) may transfer poorly across families, since the fragile computation can occur at different depths. The attention heatmaps reinforce that pruning policies perturb token routes differently at moderate compression (chunk-style contiguous removal vs.\ Ada-style non-contiguous removal), but converge to similar large-scale route destruction at extreme compression (Figures~\ref{fig:LLaMA8BAttnHeatmaps} and~\ref{fig:Qwen7BAttnHeatmaps}).

\begin{tcolorbox}[colback=white,colframe=blue!50,title=Observation: Compression must be architecture-aware]
Figures~\ref{fig:ConsensusScores}--\ref{fig:QwenLayerConsensus} indicate that the layers most responsible for stabilization vs.\ consolidation differ by family, so the same ``depth prior'' for pruning can be safe for one model and brittle for another.
\end{tcolorbox}

\begin{algorithm}[t]
\caption{Linear Probing Under KV Compression}
\label{alg:probing}
\begin{algorithmic}[1]
\Require Dataset $\mathcal{D}=\{(x_i, y_i)\}_{i=1}^N$; model $f$; compression $\alpha$; press config $\Pi$; probed layers $\mathcal{L}$; train/val indices $\mathcal{I}_{tr},\mathcal{I}_{va}$
\For{$i=1$ to $N$}
    \State Run $f$ on $x_i$ with KVPress $(\Pi,\alpha)$; record $\{h_{i,\ell}\}_{\ell\in\mathcal{L}}$
    \State $z_i\gets \mathrm{Pool}(\{h_{i,\ell}\}_{\ell\in\mathcal{L}})$
\EndFor
\State Train linear $g(z)=\mathrm{softmax}(Wz+b)$ on $\{(z_i,y_i)\}_{i\in\mathcal{I}_{tr}}$ with $f$ frozen
\State Evaluate macro-F1 on $\{(z_i,y_i)\}_{i\in\mathcal{I}_{va}}$
\end{algorithmic}
\end{algorithm}

\subsection{Why does ``safe'' KV compression suddenly fail at extreme eviction?}
\label{sec:discussion_cliff}

\begin{figure*}[!htb]
    \centering
    \subfloat[LLaMA-3 8B Instruct]{
    \includegraphics[width=0.4\linewidth]{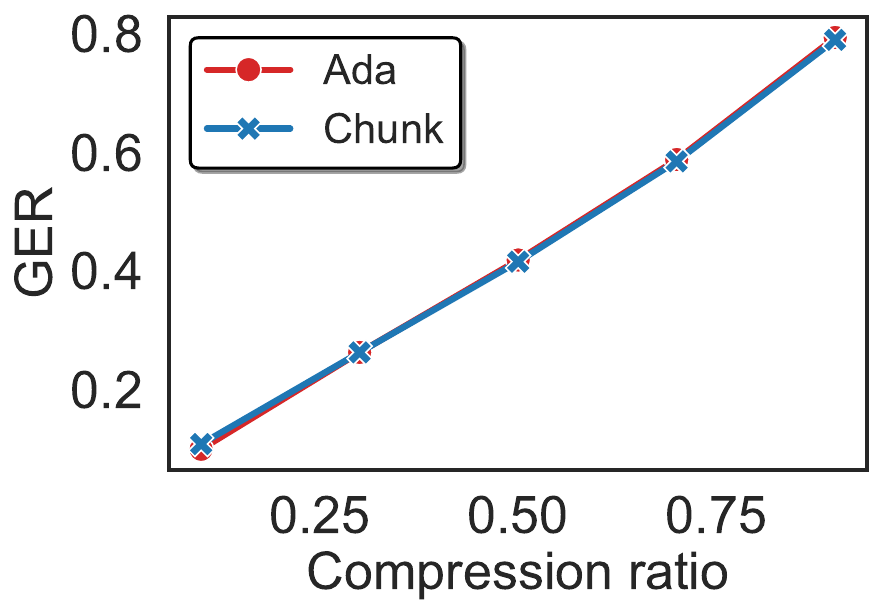}
    }
    \subfloat[Qwen-2.5 7B Instruct]{
    \includegraphics[width=0.4\linewidth]{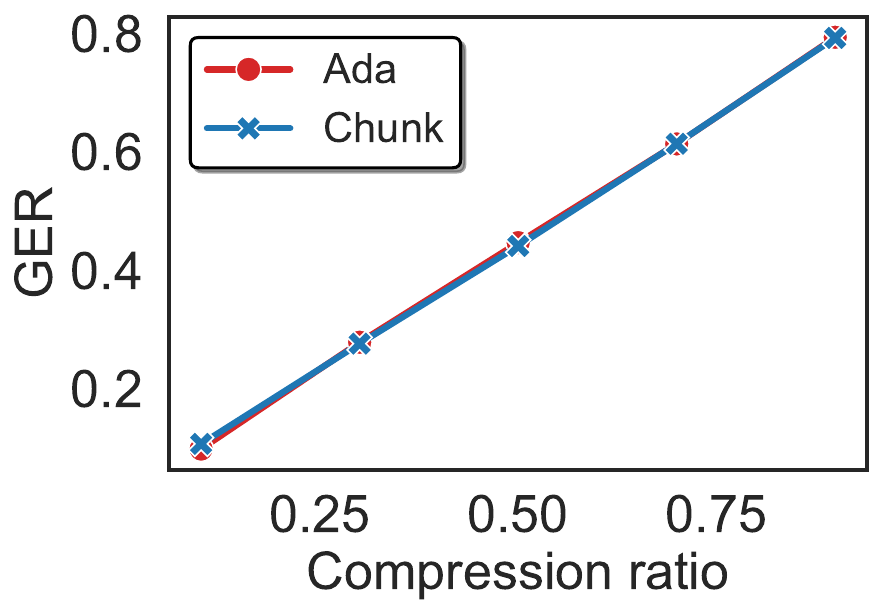}
    }
    \caption{Eviction rates vs.\ compression. Pruning is broadly question-agnostic unless explicitly conditioned. Corresponding results for LLaMA-3.2 3B, Qwen-2.5 3B, and Qwen-2.5 14B are shown in Figure \ref{fig:EvictionRateApp} of Appendix~\ref{appx:result}.}
    \label{fig:EvictionRate}
    \vspace{-5mm}
\end{figure*}

\begin{figure*}[!htb]
    \centering
    \subfloat[LLaMA-3 8B Instruct]{
    \includegraphics[width=0.8\linewidth]{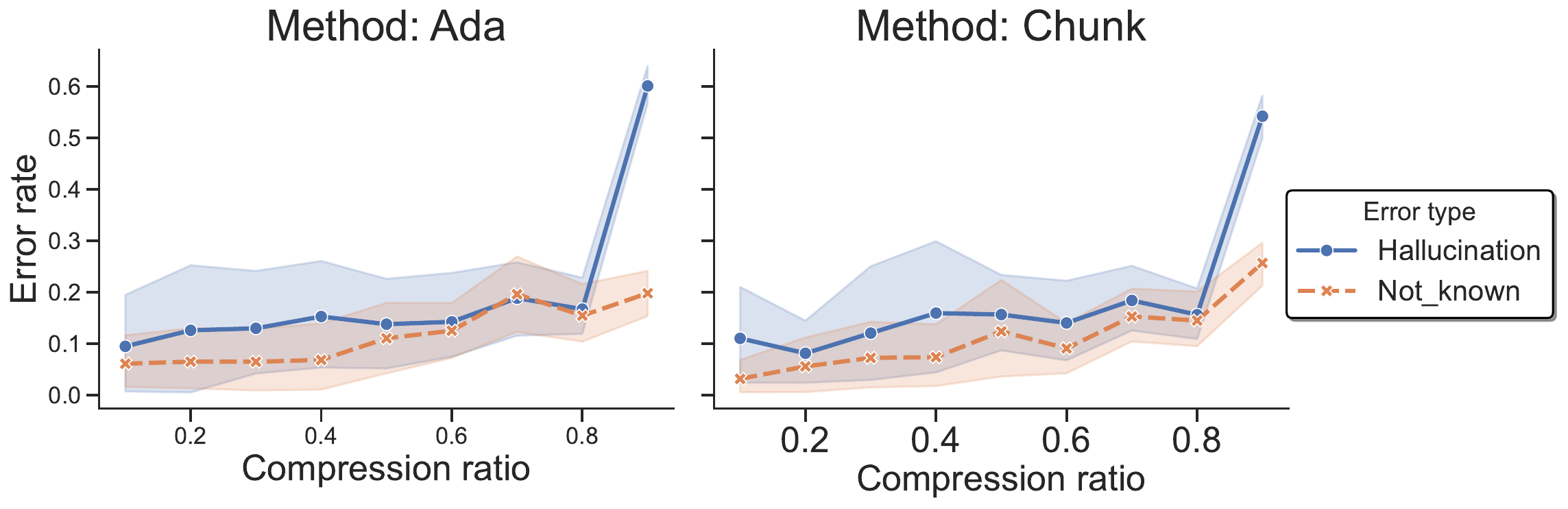}
    }\\
    \subfloat[Qwen-2.5 7B Instruct]{
    \includegraphics[width=0.8\linewidth]{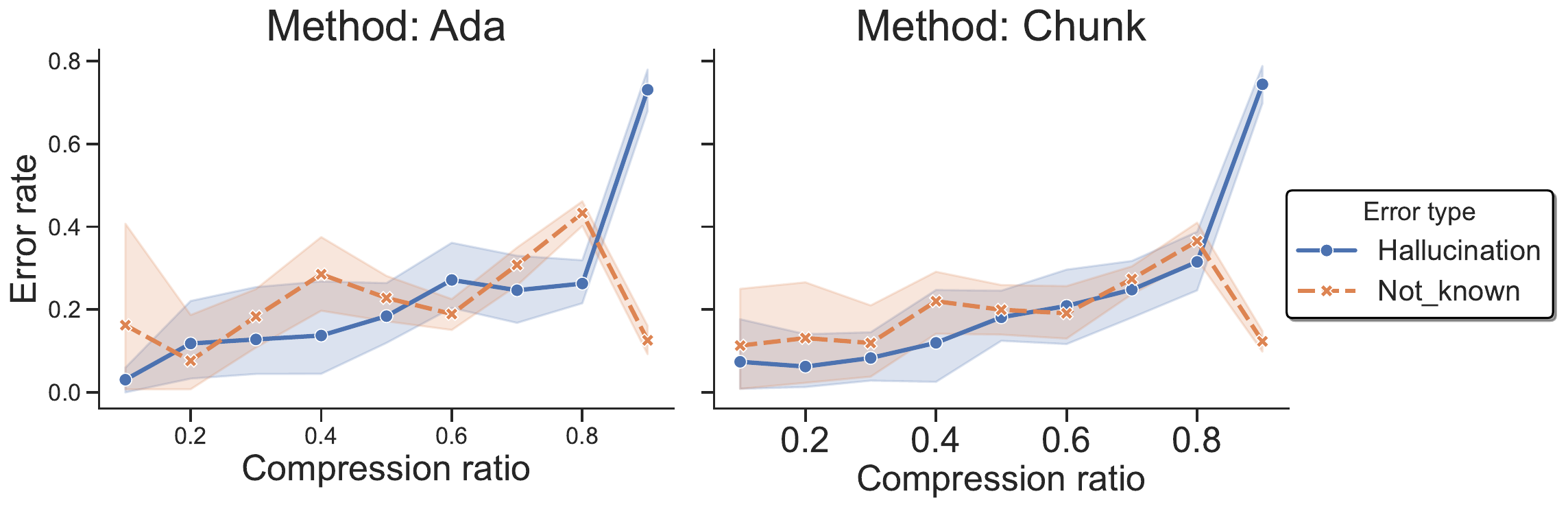}
    }
    \caption{Error rates vs.\ compression. Hallucination exhibits a cliff near $\alpha\approx 0.9$, with ``unknown'' behavior more erratic under AdaKV. Corresponding results for LLaMA-3.2 3B, Qwen-2.5 3B, and Qwen-2.5 14B are shown in Figure \ref{fig:ErrorRateApp} of Appendix~\ref{appx:result}.}
    \label{fig:ErrorRate}
    \vspace{-5mm}
\end{figure*}

\begin{figure*}[!htb]
\centering
\subfloat[LLaMA-3 8B Instruct]{
    \includegraphics[width=0.8\linewidth]{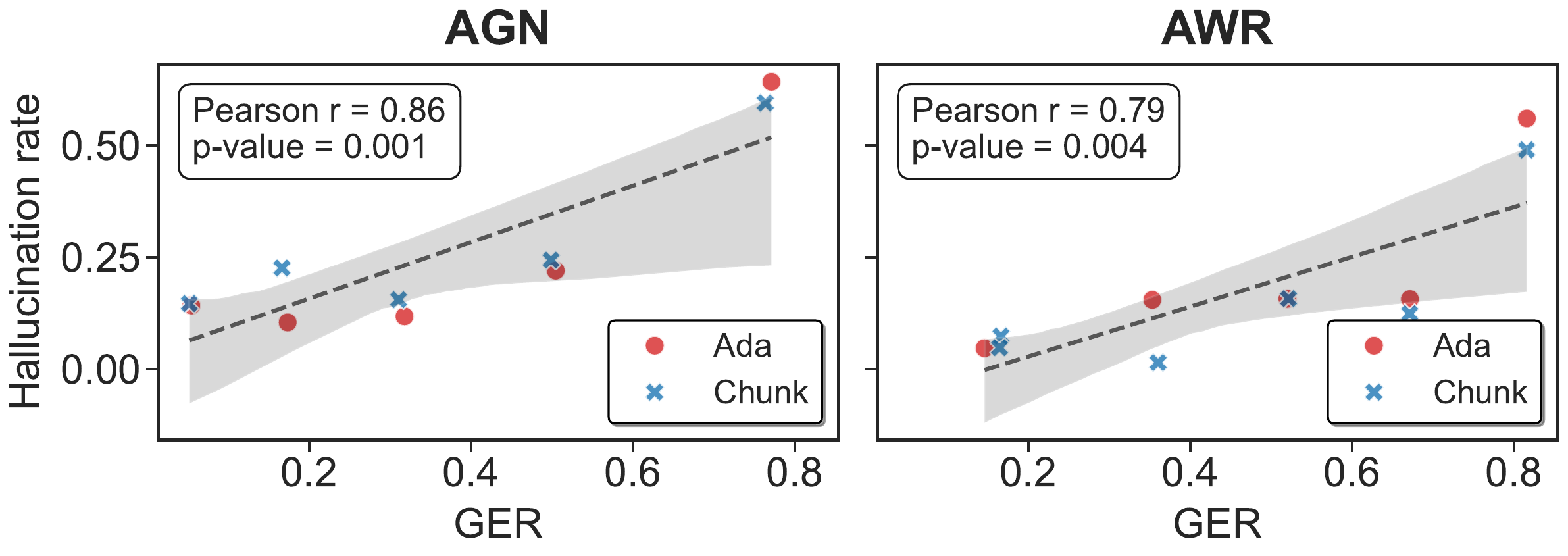}
    } \\
    \subfloat[Qwen-2.5 7B Instruct]{
    \includegraphics[width=0.8\linewidth]{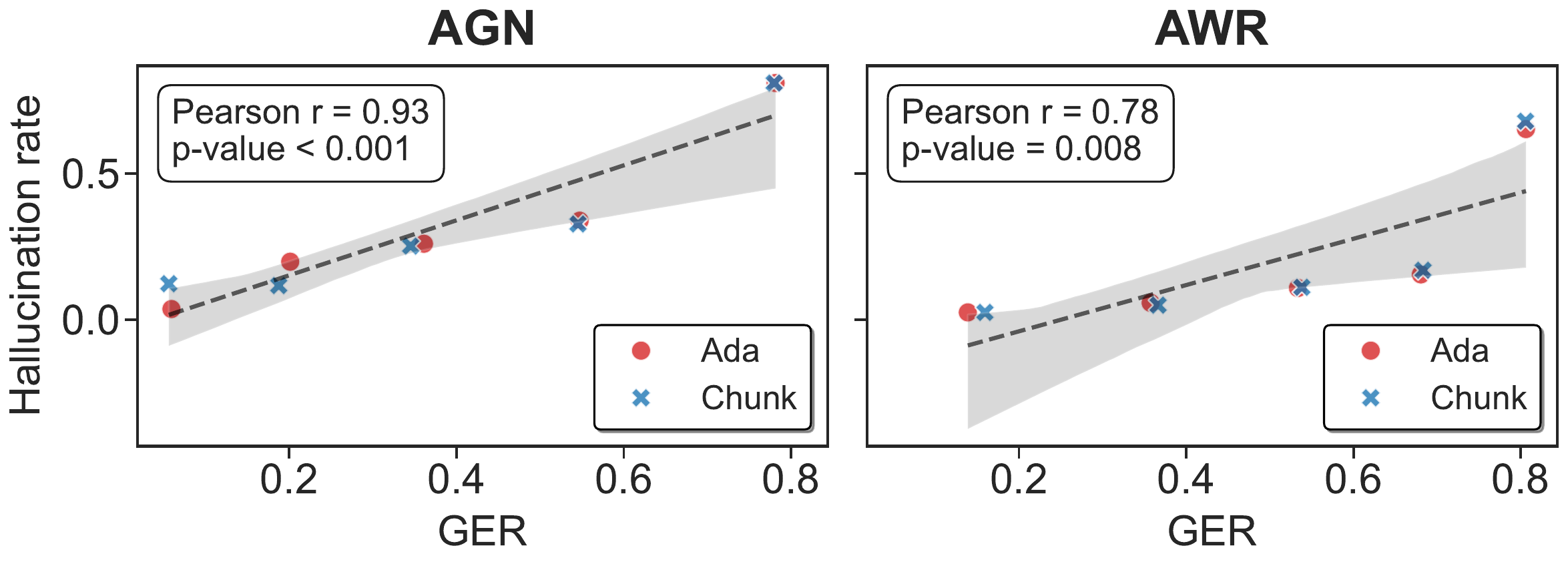}
    }
    \caption{GER correlates strongly with hallucination rate, indicating that global route deletion is a primary driver of catastrophic failure. Corresponding results for LLaMA-3.2 3B, Qwen-2.5 3B, and Qwen-2.5 14B are shown in Figure \ref{fig:corr_ger_hal_all} of Appendix~\ref{appx:result}.}
    \label{fig:corr_ger_hal}
\end{figure*}

Across models and presses, we observe a universal safety cliff near $\alpha\approx 0.9$: hallucination rates spike once roughly 90\% of KV entries are removed (Figure~\ref{fig:ErrorRate}). The key transition is not a smooth decline in representational ``quality'' but a sharp increase in the probability of \emph{global route deletion}: answer-relevant tokens become simultaneously evicted across all heads, leaving no remaining pathway to evidence. This explains why eviction curves can remain comparatively stable across $\alpha$ (Figure~\ref{fig:EvictionRate}) while error rates exhibit a sharp nonlinearity (Figure~\ref{fig:ErrorRate}), and why GER correlates strongly with hallucination (Figure~\ref{fig:corr_ger_hal}).

\begin{figure*}[!htb]
    \centering
    \subfloat[LLaMA-3 8B Instruct]{
    \includegraphics[width=0.4\linewidth]{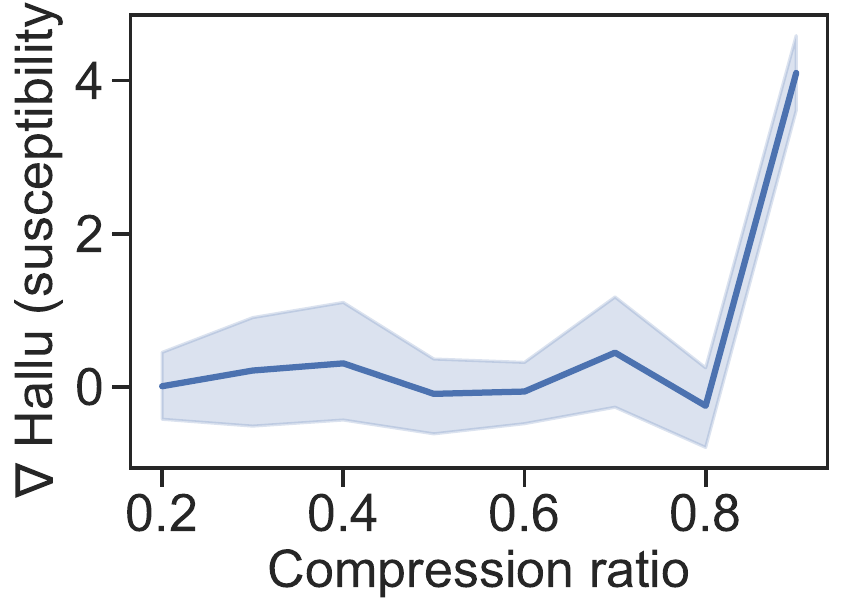}
    }
    \subfloat[Qwen-2.5 7B Instruct]{
    \includegraphics[width=0.4\linewidth]{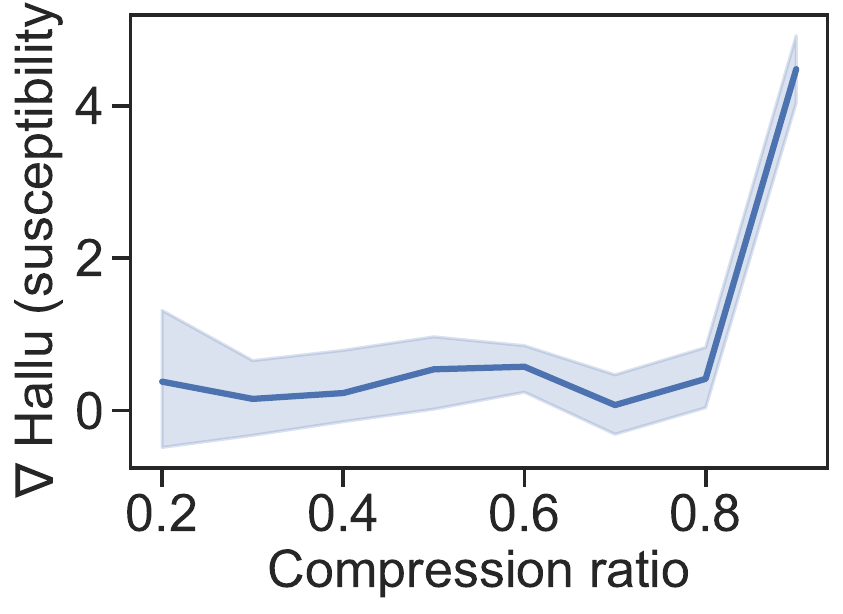}
    }
    \caption{Compression susceptibility $\chi=\frac{\partial H}{\partial \alpha}$ peaks sharply near $\alpha\approx 0.9$, consistent with a critical transition in the error landscape.}
    \label{fig:phase_stats}
\end{figure*}

At the same time, some failures occur even when GER is low, pointing to \emph{rigidity}: if attention consolidates too strongly onto a narrow focal set, the model may not re-route effectively when that set is pruned or when the remaining evidence is weakly aligned. This is most plausible in architectures with strong late consolidation (Figures~\ref{fig:ConsensusScores} and~\ref{fig:QwenLayerConsensus}), where token survival can coexist with under-utilization.

\begin{tcolorbox}[colback=white,colframe=blue!50,title=Observation: Two mechanisms explain the safety cliff]
Figures~\ref{fig:corr_ger_hal} and~\ref{fig:ConsensusScores} together support two failure modes: (i) \textbf{erasure} (high GER removes all evidence routes), and (ii) \textbf{rigidity} (routes survive but consolidation limits re-routing), both of which manifest sharply near $\alpha\approx 0.9$ (Figure~\ref{fig:ErrorRate}).
\end{tcolorbox}

\subsection{What survives compression inside the model, and why does survival not imply use?}

\begin{figure*}[!htb]
    \centering
    \subfloat[LLaMA-3 8B Instruct]{
    \includegraphics[width = 0.38\linewidth]{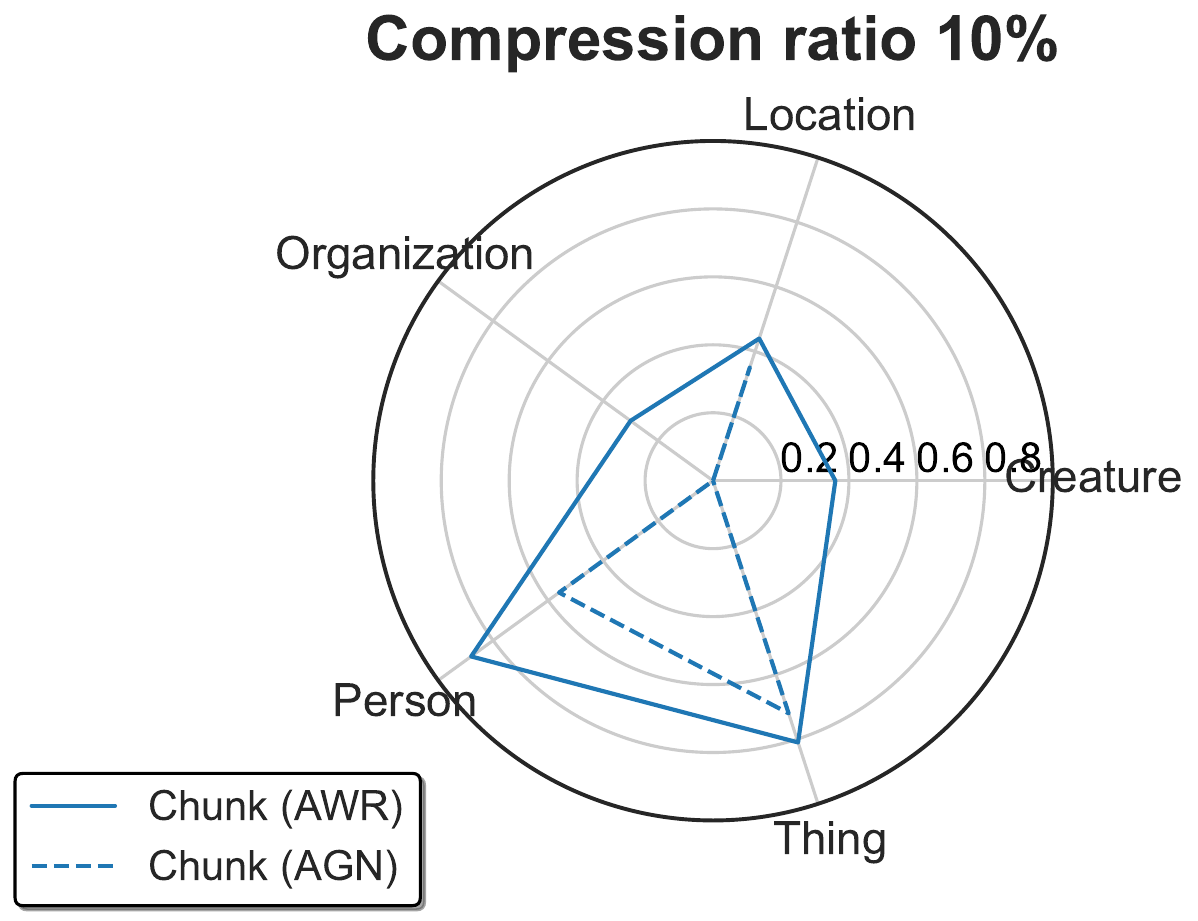}
    \includegraphics[width = 0.31\linewidth]{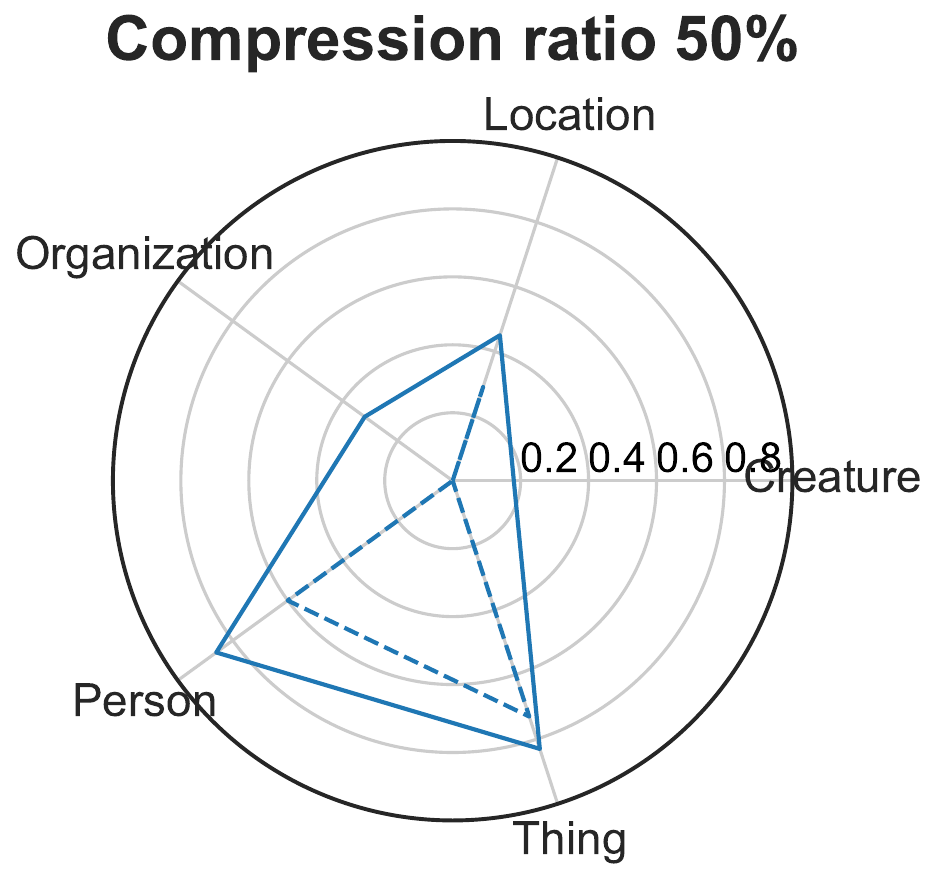}
    \includegraphics[width = 0.31\linewidth]{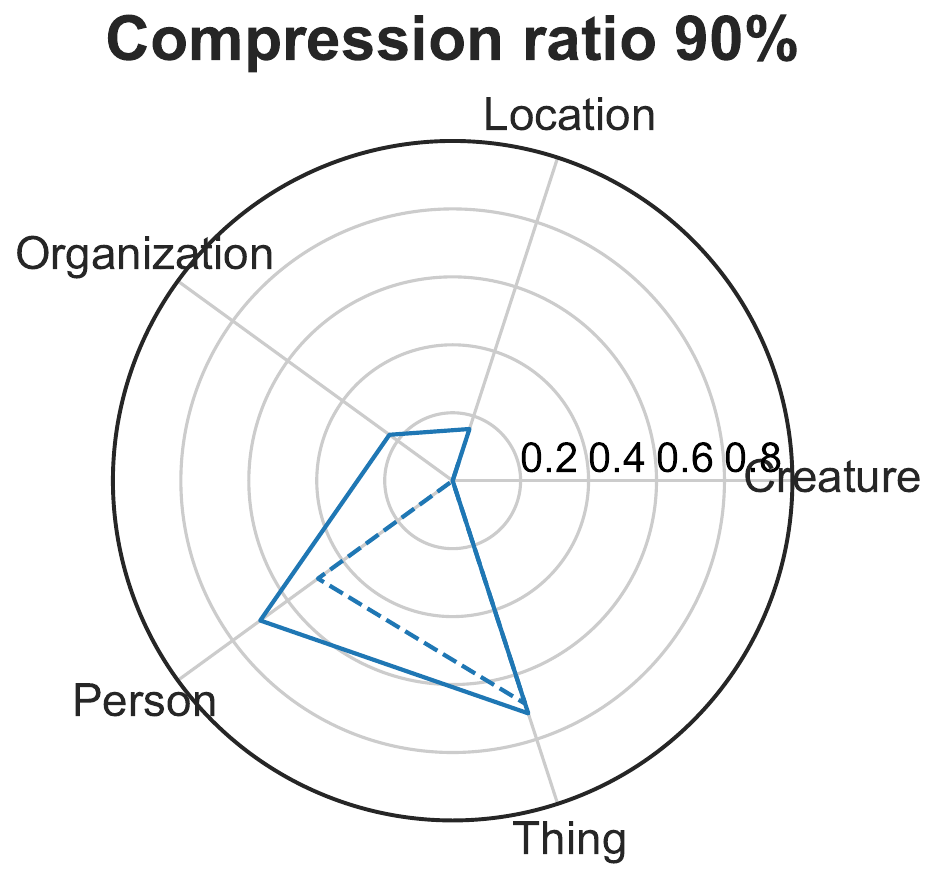}
    }\\
    \subfloat[Qwen-2.5 7B Instruct]{
    \includegraphics[width = 0.38\linewidth]{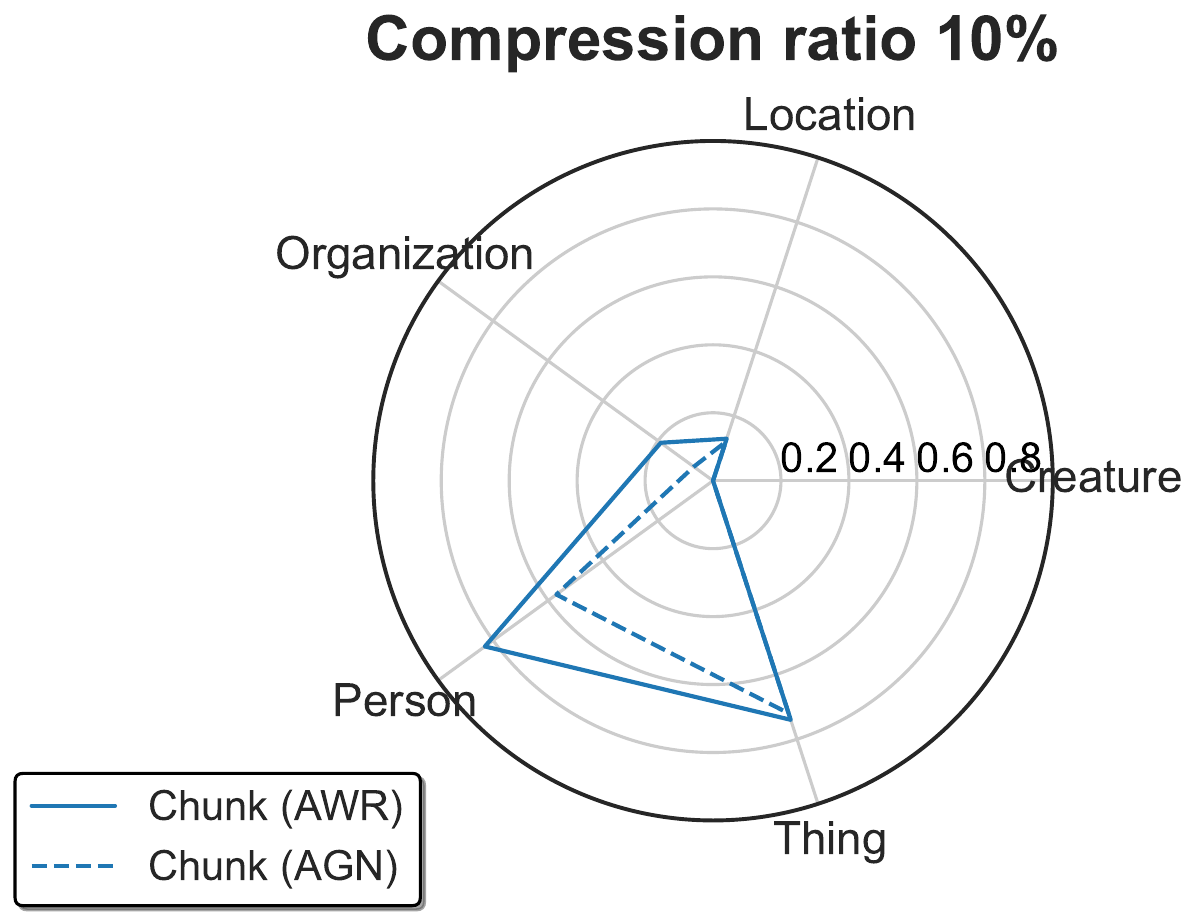}
    \includegraphics[width = 0.31\linewidth]{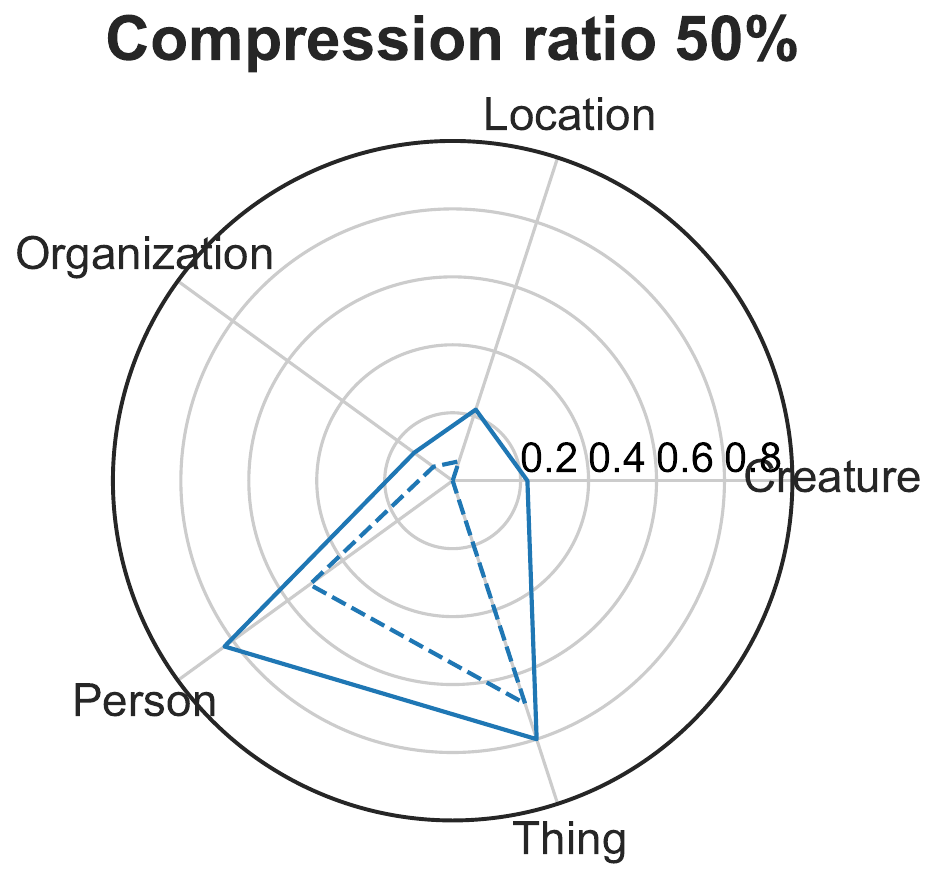}
    \includegraphics[width = 0.31\linewidth]{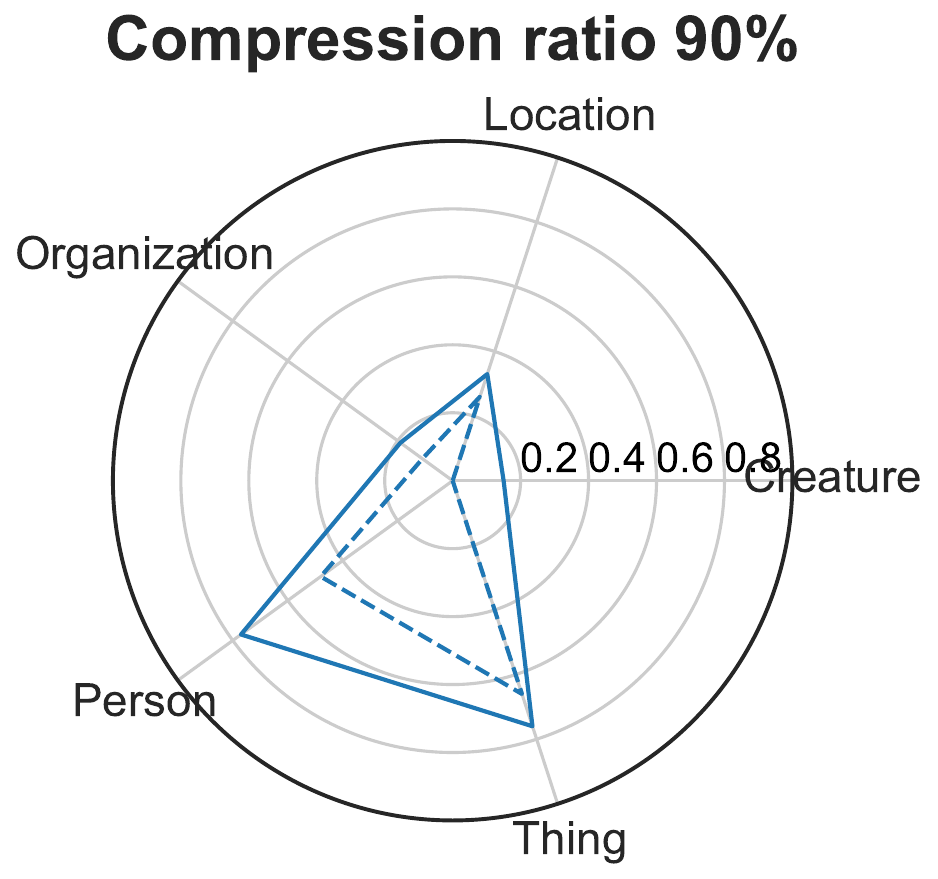}
    }\\
    \caption{Base task probing across compression. Some tags remain decodable; others collapse early and may partially recover under question-aware setups. Corresponding results for LLaMA-3.2 3B, Qwen-2.5 3B, and Qwen-2.5 14B are shown in Figure \ref{fig:BaseTaskProbingRadarApp} of Appendix~\ref{appx:result}.}
    \label{fig:BaseTaskProbingRadar}
\end{figure*}

\paragraph{Probing mechanism.}
To measure \emph{what remains linearly accessible} in the residual stream after KV compression, we train linear probes on frozen hidden states under the same press configuration used for generation. For each model, dataset, and compression ratio $\alpha$, we cache hidden states and train a lightweight linear classifier to predict the gold answer's concept tag (e.g., \emph{Person}, \emph{Location}, \emph{Organization}), reporting macro-F1 (Algorithm~\ref{alg:probing}). Probing quantifies representational \emph{presence/accessibility} rather than \emph{use}: the decoder can still fail if routing collapses (high GER) or if coordination becomes rigid, which is why high probe scores can coexist with high hallucination near the cliff (Figures~\ref{fig:BaseTaskProbingRadar} and~\ref{fig:ErrorRate}).

\begin{figure*}[!htb]
    \centering
    \subfloat[LLaMA-3 8B Instruct]{
    \includegraphics[width = 0.38\linewidth]{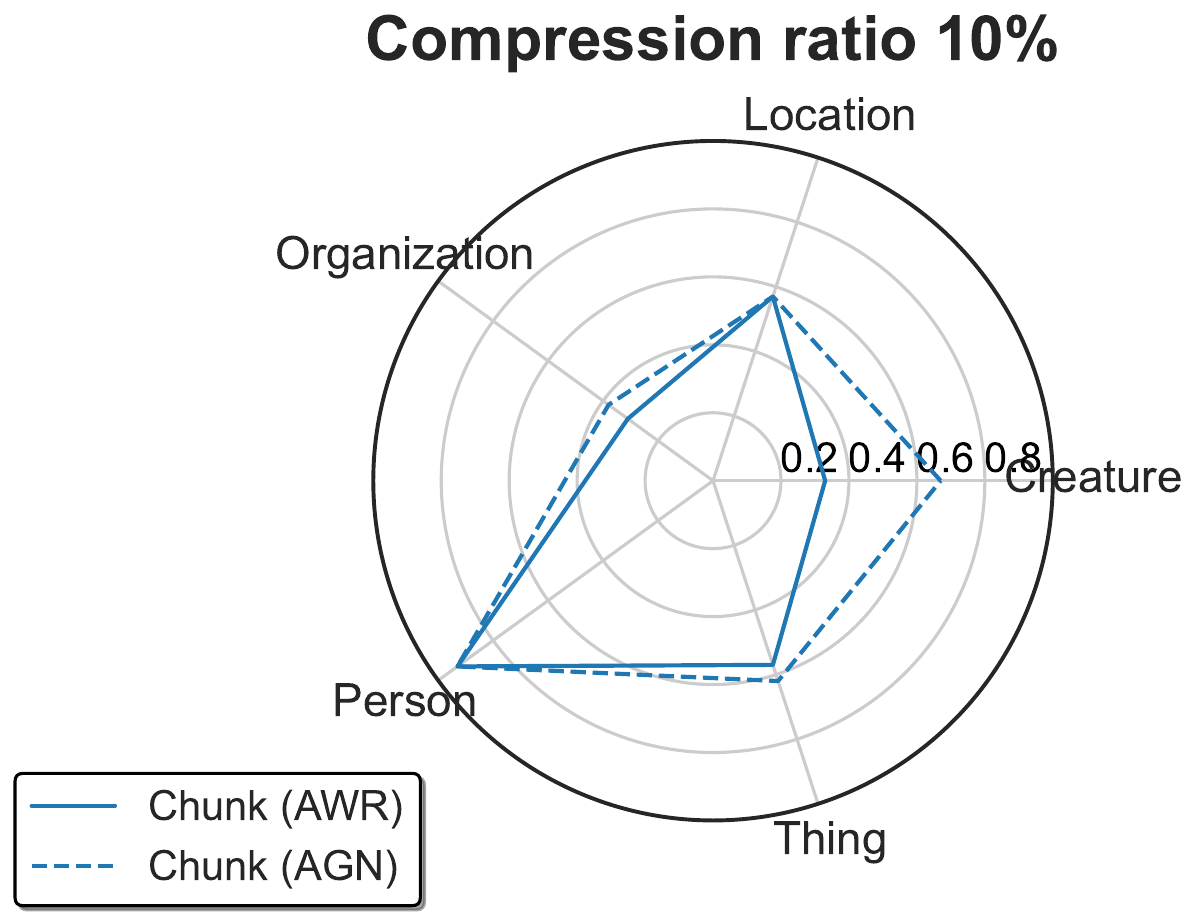}
    \includegraphics[width = 0.31\linewidth]{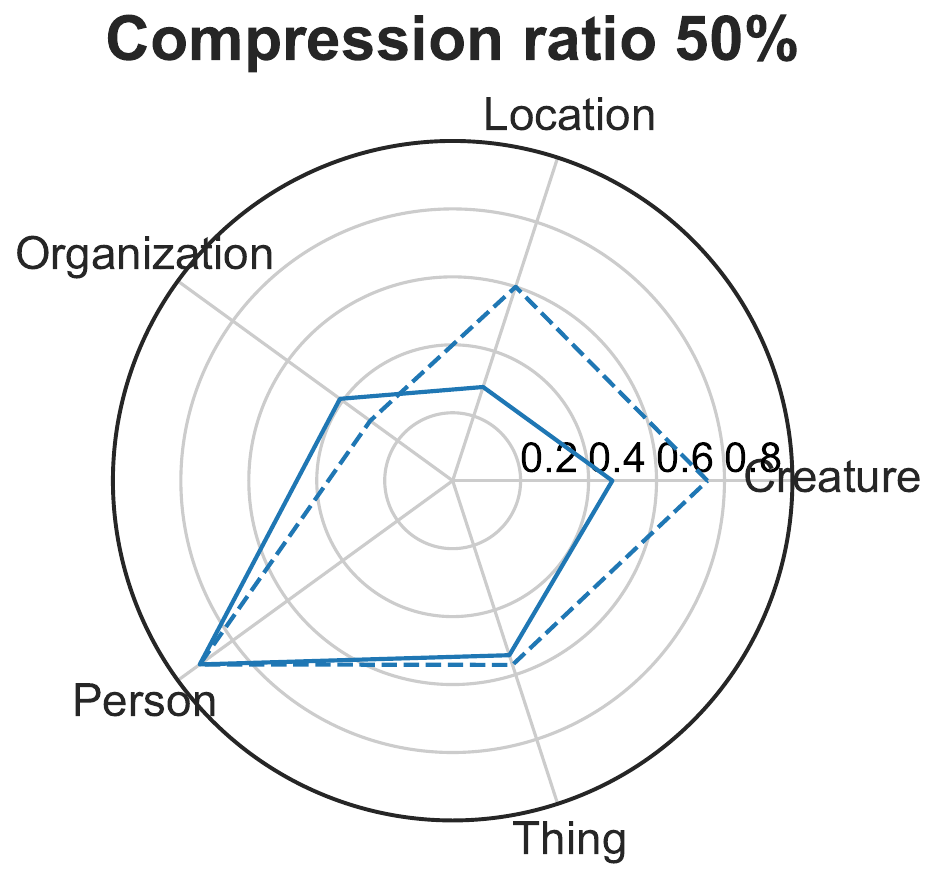}
    \includegraphics[width = 0.31\linewidth]{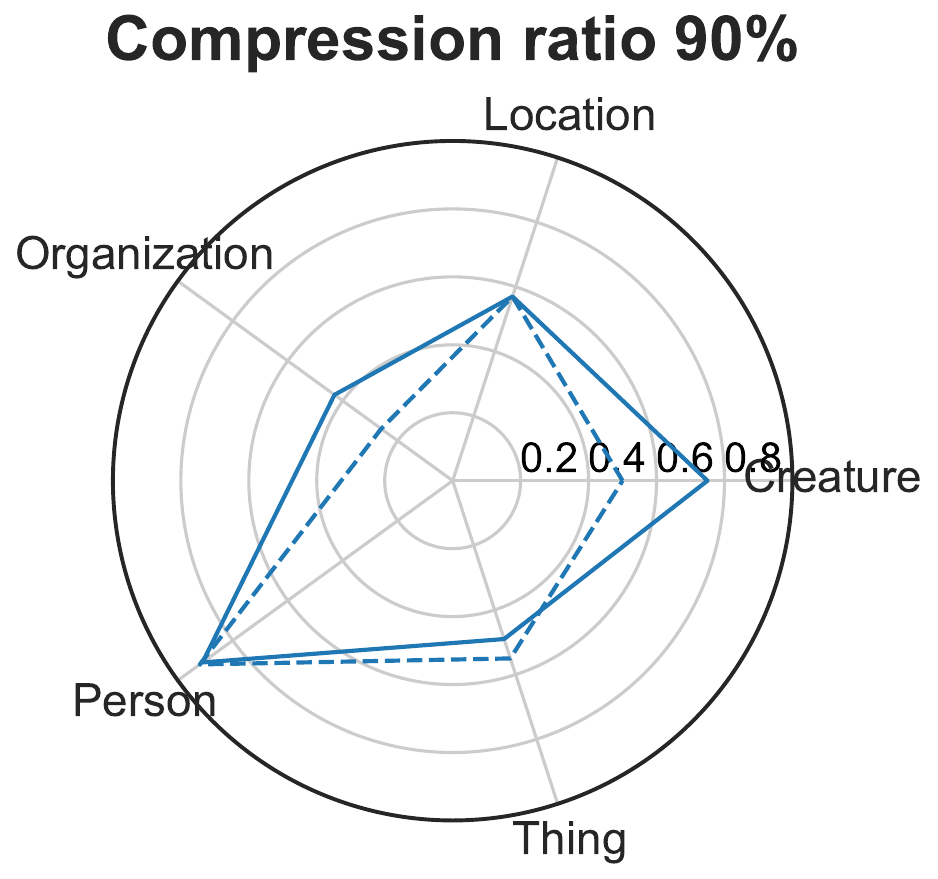}
    }\\
    \subfloat[Qwen-2.5 7B Instruct]{
    \includegraphics[width = 0.38\linewidth]{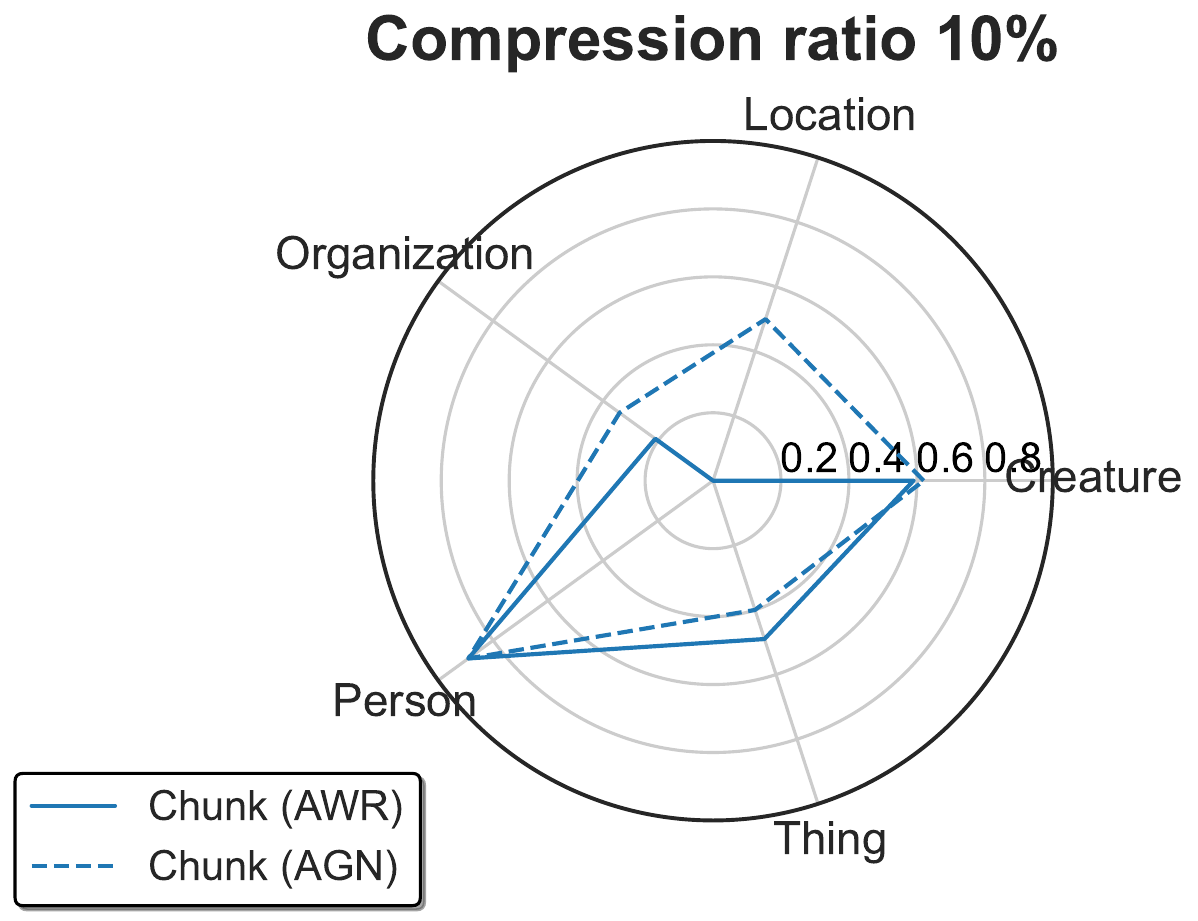}
    \includegraphics[width = 0.31\linewidth]{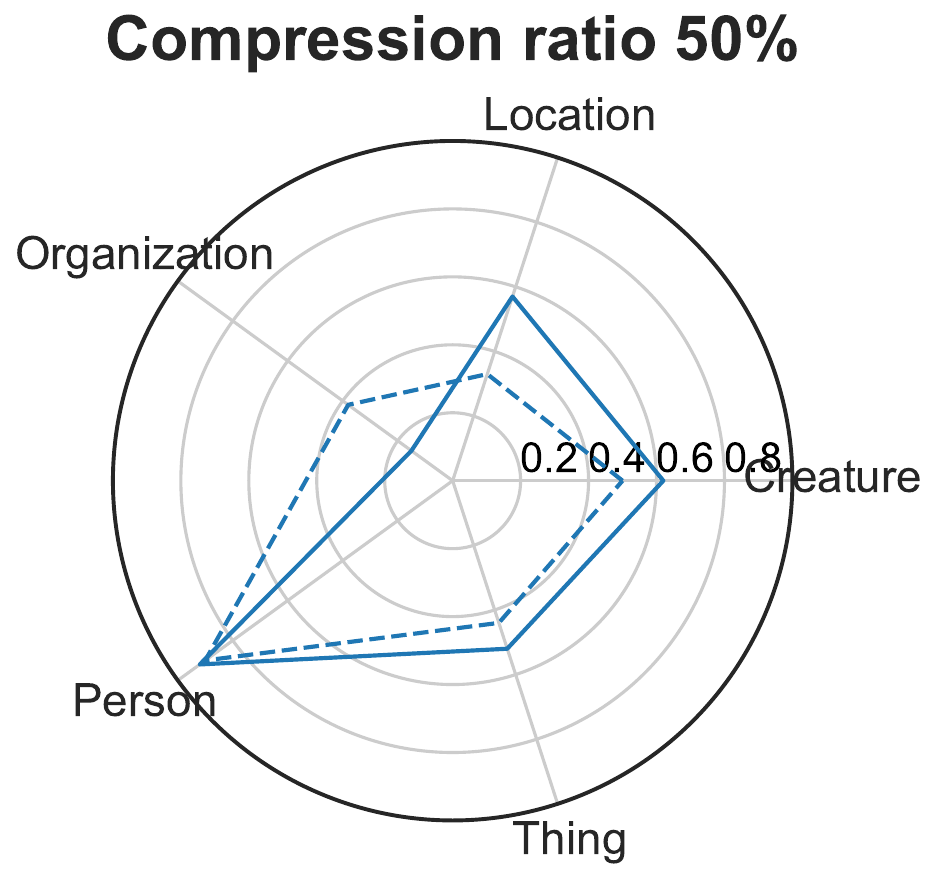}
    \includegraphics[width = 0.31\linewidth]{figures/Qwen_2.5_7B_Instruct_probing_radar_0.9.pdf}
    }
    \caption{Multi-Entity probing across compression. Question awareness helps selectively rather than uniformly, suggesting interaction between representational geometry and pruning alignment. Corresponding results for LLaMA-3.2 3B, Qwen-2.5 3B, and Qwen-2.5 14B are shown in Figure \ref{fig:MultiEntityProbingRadarApp} of Appendix~\ref{appx:result}.}
    \label{fig:MultiEntityProbingRadar}
\end{figure*}

\begin{tcolorbox}[colback=white,colframe=blue!50,title=Observation: Retention $\neq$ utilization]
Probing can remain high while generation fails: concept vectors may stay decodable (Figures~\ref{fig:BaseTaskProbingRadar}--\ref{fig:MultiEntityProbingRadar}) even as hallucination spikes near the cliff (Figure~\ref{fig:ErrorRate}), indicating that routing availability and flexibility, not just representational presence, limit reliability.
\end{tcolorbox}

\subsection{When does question-aware compression help, and why does it help some models more than others?}

Question-aware compression can be seen as injecting an early discriminative bias: pruning retains tokens that best separate the answer entity from competitors, which is consistent with improved probe separability in several regimes (Figures~\ref{fig:BaseTaskProbingRadar} and~\ref{fig:MultiEntityProbingRadar}). Its impact is larger when deep layers amplify early selection signals, as in late-consolidating architectures (Figures~\ref{fig:ConsensusScores} and~\ref{fig:QwenLayerConsensus}). Conversely, tasks that require latent bridge tokens (e.g., multi-hop or discrepancy detection) can remain fragile under query-centric heuristics if relevance is not locally aligned with query overlap, suggesting that robust question-aware pruning must preserve \emph{routes}, not just query-similar tokens.

\begin{tcolorbox}[colback=white,colframe=blue!50,title=Observation: Query signals improve discriminative preservation]
Question-aware pruning tends to preserve tokens that maintain decision boundaries for the correct answer, improving probe separability (Figures~\ref{fig:BaseTaskProbingRadar}--\ref{fig:MultiEntityProbingRadar}) and mitigating catastrophic failure near $\alpha\approx 0.9$ (Figure~\ref{fig:ErrorRate}), especially in late-consolidating models (Figure~\ref{fig:ConsensusScores}).
\end{tcolorbox}

\subsection{Is KV compression best understood as gradual representational decay or as a phase transition?}

Finally, the mismatch between probing and generation trajectories suggests that KV compression is not well-modeled as smooth representational decay: probe scores can drop early (Figure~\ref{fig:BaseTaskProbingRadar}) while generation stays stable until the cliff (Figure~\ref{fig:ErrorRate}). This is more consistent with a reachability-driven transition: performance remains robust while at least one viable route to answer evidence survives, and collapses when routes are globally deleted (high GER) or become too rigid to exploit (high consolidation), aligning with stable eviction trends (Figure~\ref{fig:EvictionRate}) and sharp error nonlinearity (Figure~\ref{fig:phase_stats}).

\begin{tcolorbox}[colback=white,colframe=blue!50,title=Observation]
Figures~\ref{fig:EvictionRate}--\ref{fig:ErrorRate} indicate that reliability depends on whether the token-route graph still contains viable paths to answer evidence (Figure~\ref{fig:corr_ger_hal}) and whether routing remains sufficiently flexible (Figure~\ref{fig:ConsensusScores}), supporting a transition-like view of extreme KV compression.
\end{tcolorbox}

\section{Theoretical Implications of Token-Route Sparsity}
\label{sec:theory}

\subsection{From Parameter Lottery Tickets to Token-Route Subgraphs}

Recent theory gives strong \emph{parameter-level} sparsity guarantees for multi-head self-attention. In particular, the Strong Lottery Ticket Hypothesis (SLTH) for attention~\citep{otsuka2025strong} shows that, when an attention module is sufficiently wide, there exist sparse \emph{parameter masks} whose restricted attention operator can approximate a target attention behavior without retraining. This is a statement about \emph{weights}: a subset of parameters can implement (approximately) the same mapping as the dense module. KV cache compression is qualitatively different. It leaves the parameters untouched but removes \emph{rows} of the key--value memory at inference time (i.e., removes tokens from the set that attention can read), thereby changing which token-to-token interactions are even possible. This motivates a complementary notion of sparsity defined on the \emph{token connectivity} induced by attention during a forward pass.

\paragraph{Definition 1 (Token-Attention Graph).}
Fix an input sequence of length $S$ and consider a transformer with layers $\ell\in\{0,\dots,L\}$ and (KV) heads $h\in\{1,\dots,H\}$. Let $V=\{(\ell,i): \ell\in\{0,\dots,L\},\, i\in\{1,\dots,S\}\}$ denote token-nodes indexed by layer and position. For each layer transition $\ell\to \ell+1$ and head $h$, let $A_{\ell,h}\in\mathbb{R}^{S\times S}$ be the (row-stochastic) attention matrix used to compute the head output at layer $\ell$. For a fixed threshold $\varepsilon\ge 0$, define directed edges
\[
E \;=\;\bigl\{\, (\ell,i)\to(\ell,j)\ :\ \exists h \text{ s.t. } A_{\ell,h}(i,j)>\varepsilon \,\bigr\}.
\]
The resulting directed layered graph $G=(V,E)$ summarizes which token positions at layer $\ell$ can directly route information from token $j$ into token $i$ at the same layer.\footnote{Edges are drawn within each layer because attention aggregates values from source positions $j$ into destination positions $i$ before the residual update. Any equivalent convention (e.g., connecting $(\ell,j)$ to $(\ell+1,i)$) yields the same reachability notions up to relabeling.}

\paragraph{Definition 2 (Compressed Token-Attention Graph).}
At compression level $\alpha$, a press method selects, for each head $h$ and layer $\ell$, a subset of \emph{surviving} key--value positions $U_{\ell,h}^{(\alpha)}\subseteq\{1,\dots,S\}$. Define the compressed graph $G_\alpha=(V,E_\alpha)$ by removing all edges that route through pruned KV positions:
\[
E_\alpha\;=\;\bigl\{\, (\ell,i)\to(\ell,j)\in E\ :\ \exists h \text{ s.t. } j\in U_{\ell,h}^{(\alpha)} \ \wedge\ A_{\ell,h}(i,j)>\varepsilon \,\bigr\}.
\]
Equivalently, $G_\alpha$ is the routing graph induced by the attention operator after zeroing all columns $j\notin U_{\ell,h}^{(\alpha)}$ (and renormalizing) in each $A_{\ell,h}$.

\paragraph{Definition 3 (Token-Route Lottery Ticket, TR-LT).}
Let $q$ be a designated query token position (typically the question token region) and let $T_{\mathrm{ans}}\subseteq\{1,\dots,S\}$ be the set of answer-relevant context positions.\footnote{In our experiments, $T_{\mathrm{ans}}$ is obtained from the gold answer span / annotation used for evaluation and probing.} A TR-LT at compression $\alpha$ is a subgraph $H_\alpha\subseteq G_\alpha$ such that (i) there exists $t\in T_{\mathrm{ans}}$ with a directed path from $(\ell_q,q)$ to $(\ell,t)$ in $H_\alpha$ for some layers $\ell_q,\ell$, and (ii) restricting attention routing to the edges of $H_\alpha$ is sufficient (in the sense of preserving the model's correct output on that instance) to support correct generation. Intuitively, TR-LTs are \emph{token-level} sparse routing backbones that remain functional under compression, complementing SLTH's \emph{parameter-level} sparse backbones.

\subsection{Reachability, Redundancy, and When Erasure Forces Hallucination}

The key structural question under KV compression is whether answer evidence remains reachable through the compressed routing graph. Let $R_\alpha(q)\subseteq\{1,\dots,S\}$ denote the set of token positions reachable from $q$ in $G_\alpha$ (dropping layer indices for notational simplicity). Define the reachability event
\[
\mathcal{C}_\alpha(q)\;:=\;\bigl(T_{\mathrm{ans}}\cap R_\alpha(q)\neq\emptyset\bigr).
\]
If $\mathcal{C}_\alpha(q)$ fails, then no answer-relevant token is reachable via attention routing, so the model cannot condition its final states on the answer evidence through self-attention paths.

\paragraph{Proposition 1 (Redundant head-wise routes yield robustness).}
Assume that for each $t\in T_{\mathrm{ans}}$, there exist at least $k$ head-disjoint directed paths from $q$ to $t$ in the uncompressed graph $G$.\footnote{Head-disjoint means that the paths can be realized through disjoint head-specific edge sets, formalizing cross-head redundancy.} Suppose that, for each head, the KV token $t$ survives (i.e., is not pruned in that head wherever it would be used) independently with probability at least $p$. Then
\[
\Pr\!\bigl[t\in R_\alpha(q)\bigr] \;\ge\; 1-(1-p)^k.
\]
\emph{Interpretation.} If answer evidence is redundantly accessible through multiple heads, the probability that \emph{all} routes are destroyed decays exponentially in the redundancy $k$. This explains why moderate compression can preserve behavior: even if individual routes are fragile, \emph{at least one} tends to survive.

\paragraph{Proposition 2 (Erasure implies loss of contextual grounding).}
Assume that the model has not already encoded the answer content into the reachable residual stream \emph{before} the KV entries corresponding to $T_{\mathrm{ans}}$ are removed (i.e., the only way to use answer evidence is to route to surviving answer tokens).\footnote{This assumption rules out degenerate cases where answer information is perfectly ``copied'' into other surviving tokens prior to eviction; empirically, the GER--hallucination correlation (Figure~\ref{fig:corr_ger_hal}) suggests such leakage is limited at the safety cliff.} If there exists $\alpha^\star$ such that
\[
T_{\mathrm{ans}}\cap R_{\alpha^\star}(q)=\emptyset,
\]
then no self-attention routing mechanism can condition the model's hidden states on answer evidence at compression $\alpha^\star$. Consequently, correct generation cannot be guaranteed from context and the model must rely on parametric priors, making hallucination structurally likely.

\emph{Proof sketch.} Each attention head output at every layer is a weighted linear combination of \emph{surviving} value vectors. If no answer token is reachable from the query token in $G_{\alpha^\star}$, then no sequence of attention aggregations can incorporate any value vector originating from $T_{\mathrm{ans}}$ into the query-conditioned hidden states. Under the no-leakage assumption, the decoder has no contextual evidence for the answer and must default to priors. \hfill $\square$

Proposition~2 makes explicit what GER diagnoses: not merely ``how many'' tokens remain, but whether the routing graph still contains \emph{any} path to answer evidence. This connects directly to the empirical cliff: generation stays stable while reachability holds and collapses when global route deletion becomes common (Figures~\ref{fig:EvictionRate},~\ref{fig:ErrorRate},~\ref{fig:corr_ger_hal}).

\subsection{Representational Rigidity: Survive-but-not-used Failures}

Reachability is necessary but not sufficient: we observe regimes where answer tokens survive (low GER) yet performance deteriorates, consistent with \emph{rigidity} in routing. Formally, suppose that at layer $\ell$ many heads place their maximal attention on the same token $t^\star$, leaving little diversity in which tokens are actively read.

\paragraph{Proposition 3 (High agreement reduces re-routing capacity).}
Fix a layer $\ell$. Let $t^*_{\ell,h}=\arg\max_t A_{\ell,h}(t)$ be the top-attended token of head $h$ and define the agreement fraction
\[
\rho_\ell \;:=\; \max_{t}\ \frac{1}{H}\bigl|\{h:\ t^*_{\ell,h}=t\}\bigr|.
\]
If the maximizer token $t^\star$ (achieving $\rho_\ell$) is pruned in the relevant heads under compression, then at least a $\rho_\ell$ fraction of heads must shift their primary attention to secondary tokens. If those secondary tokens systematically exclude $T_{\mathrm{ans}}$, then the effective probability of routing to answer evidence at layer $\ell$ is reduced by a factor proportional to $\rho_\ell$.

\emph{Interpretation.} When routing is highly concentrated, compression can remove a small set of ``consensus'' tokens and force many heads to re-route simultaneously. If the remaining attention mass is not diverse enough to rediscover answer evidence, the model can fail despite token survival elsewhere. This is the structural sense in which consensus can create \emph{representational rigidity}: the model retains evidence in memory, but lacks routing flexibility to \emph{use} it.

\subsection{Failure Case Analysis: A Minimal Decomposition}

We can express failure probability as a mixture of an \emph{erasure} mechanism and a \emph{rigidity} mechanism. Let $\mathcal{F}_\alpha$ denote generation failure at compression level $\alpha$. Let $\mathcal{E}_\alpha$ denote the event that answer-relevant tokens are globally evicted across heads (the task instance is ``route-deleted'' in the sense measured by GER). Conditioning on $\mathcal{E}_\alpha$ gives
\[
\Pr(\mathcal{F}_\alpha)
=
\Pr(\mathcal{F}_\alpha \mid \mathcal{E}_\alpha)\Pr(\mathcal{E}_\alpha)
+
\Pr(\mathcal{F}_\alpha \mid \neg \mathcal{E}_\alpha)\Pr(\neg \mathcal{E}_\alpha).
\]
The first term captures \emph{route existence}: when $\mathcal{E}_\alpha$ holds, contextual grounding is unavailable, so $\Pr(\mathcal{F}_\alpha \mid \mathcal{E}_\alpha)$ is close to $1$ in practice. The second term captures \emph{route usability}: even when answer tokens survive in principle, failure can occur if routing collapses onto inflexible patterns. In our setting, $\Pr(\mathcal{E}_\alpha)$ is estimated by $\mathrm{GER}(\alpha)$ (Figure~\ref{fig:corr_ger_hal}), while head-level consensus trends (Figure~\ref{fig:ConsensusScores}) provide a measurable proxy for rigidity. This decomposition aligns with the empirical pattern that the safety cliff is driven primarily by global route deletion at extreme compression, while intermediate degradations can reflect survive-but-not-used effects in consolidation-heavy regimes.

\subsection{Sparse Token-Routes as the Inference-Time Counterpart of SLTH}

SLTH~\citep{otsuka2025strong} implies that attention contains  \emph{sparse parameter} subnetworks capable of approximating dense attention behavior. Our empirical results identify a complementary inference-time phenomenon: robustness under KV compression is governed by the survival of sparse \emph{token-route} subnetworks (TR-LTs) within the compressed token-attention graph. In this view, KV compression is a controlled perturbation that reveals the minimal routing backbone required for correct generation. Extreme compression induces failure when (i) cross-head reachability collapses (high GER, Proposition~2) or (ii) routing becomes too rigid to exploit surviving evidence (high agreement/low diversity, Proposition~3). Together, these results connect theoretical sparsity guarantees for attention to a concrete mechanism of inference-time sparsification: it is not the \emph{amount} of KV memory retained that governs reliability, but whether the compressed routing graph preserves at least one viable, usable token-route to answer evidence.

\section{Limitations and Future Work}

Although our framework provides mechanistic insight into KV cache compression, several natural extensions remain. First, our controlled synthetic datasets are intentionally constructed to isolate routing-sensitive behaviors such as multi-hop chaining, entity disambiguation, and coreference consistency. While this design enables causal attribution of failure modes, it abstracts away from the full heterogeneity of natural language and large-scale real-world corpora. Future work should investigate whether the same reachability–rigidity decomposition persists in long-document QA, retrieval-augmented generation, code reasoning, and multimodal contexts. In particular, studying compression effects in systems with external memory or tool usage may reveal additional forms of routing resilience or fragility not captured by purely internal KV pruning.

Second, our theoretical treatment focuses on structural metrics -- Global Eviction Ratio and head-level consensus -- as explanatory variables for compression-induced phase transitions. While these metrics show strong empirical alignment with performance collapse, a more refined analysis of attention operators under structured token removal could yield tighter guarantees. For example, spectral characterizations of attention graphs, connectivity thresholds in layered routing structures, or formal bounds on multi-head redundancy may sharpen the theoretical link between token-route sparsity and robustness. On the modeling side, an important direction is the co-design of sparsity mechanisms and architecture: rather than post-hoc pruning, training objectives could explicitly encourage redundant cross-head routing or controlled consensus to maintain effective route capacity under compression. Such approaches may lead to principled efficient-attention designs that preserve structural reachability while reducing memory cost.

\section{Conclusion}

In this work, we reconceptualized KV cache compression as a structural intervention on the routing geometry of self-attention rather than a mere memory-reduction technique. Through controlled synthetic benchmarks, layer-wise routing analysis, and formalization of reachability and rigidity metrics, we demonstrated that compression-induced failures arise from two distinct mechanisms: global erasure of answer-relevant tokens and collapse of routing diversity despite token survival. The sharp performance cliff observed at extreme compression corresponds to a structural phase transition in semantic reachability, revealing that a sparse token-route backbone governs inference-time robustness. By connecting these empirical findings to sparsity theory in attention, we extend the intuition of lottery tickets from parameter subnetworks to dynamic token-route structures. Our results suggest that future long-context and efficient-attention systems must preserve not only token representations but also minimal routing capacity across heads and layers, aligning efficiency with structural integrity rather than local importance heuristics.

\FloatBarrier

\newpage

\appendix
\FloatBarrier

\section{Dataset Descriptions}
\label{appx:data}

\begin{table}[htbp]
\centering
\resizebox{\columnwidth}{!}{
\begin{tabular}{p{3cm} p{8cm} p{4.5cm} p{4.5cm}}
\toprule
\textbf{Task} & \textbf{Context} & \textbf{Question} & \textbf{Response} \\
\midrule
Base task & Gonzalo Batistuta, an Argentine forward, is widely regarded as one of the greatest football players of all time... & What is Gonzalo Batistuta's nationality? & Argentine \\
\cdashline{1-4}
Knowledge man. & Cora Delaine was born on December 11, 1954. They studied in Daegu Global Science University... & What is Cora Delaine's first name? & Cora \\
\cdashline{1-4}
Multi presence & The Astralis Spire stands tall as one of the most remarkable achievements of modern architecture and engineering. Rising high above the skyline of its bustling city... & What was the main reason for constructing Astralis Spire? & To improve television and radio broadcasting disrupted by urban development \\
\cdashline{1-4}
Multi entity & Harland Kane is one of the most influential authors of contemporary horror, suspense, and supernatural fiction. Over a career spanning decades, Kane has... & Which novel by Kane is set in a haunted hotel? & The Shadow of the Pines \\
\cdashline{1-4}
Long context & During the mid-Cretaceous period, the Riverback Sailbacks were formidable semi-aquatic predators, inhabiting river systems, floodplains, and coastal wetlands. Known for their elongated jaws, conical teeth, and large... & Where was Kaelin Vireo from? & The riverine fossil station of Thaloris \\
\cdashline{1-4}
Coreference & Julian Foster was born on September 7, 1962 in Aleppo, Aleppo Governorate, Syria. He is male. He studied in Tianjin Harbor University in Podgorica, Podgorica, Montenegro... & What does he work as? & Tour Guide \\
\cdashline{1-4}
\multirow{4}{*}{Hops} 
& \multirow{4}{7cm}{Long before colonial records traced the legends of the high Mexican plateau, the City of Teotilac rose from volcanic plains as a geometric marvel of light, shadow, and faith.Built by the enigmatic Itzaca people, Teotilac's avenues and pyramids were laid out to capture the solar zenith, with precise alignments connecting celestial motion to civic life...} 
& What was the City of Teotilac, and what made it distinctive? (Hop 0) 
& A sacred city built by the Itzaca, aligned with celestial cycles (Hop 0) \\
\cdashline{3-4}
& 
& How did Teotilac's design integrate cosmic and civic order? (Hop 1) 
& Its pyramids and avenues encoded solar and temporal order (Hop 1) \\
\cdashline{3-4}
& 
& What was the Prism of Yulnah project? (Hop 2) 
& A modern optical archaeology project at Teotilac (Hop 2) \\
\cdashline{3-4}
& 
& Who led the Prism of Yulnah and what modern tools were employed? (Hop 3)
& Led by Dr. Lira Montoya using spectral drones and mirrors (Hop 3) \\
\bottomrule
\end{tabular}
}
\caption{Sample contexts and inputs for each task along with the ground truth answer.}
\label{tab:ContextSamples}
\end{table}

\begin{table}[bhtp]
\centering
\resizebox{\columnwidth}{!}{
\begin{tabular}{p{2cm} p{7cm} p{4.5cm} p{4.5cm}}
\toprule
\textbf{Tag} & \textbf{Context} & \textbf{Question} & \textbf{Response} \\
\midrule
Person & Joao Fernandes, widely considered one of the greatest footballers of all time... & What is the name of the person described in the passage? & Joao Fernandes \\
\cdashline{1-4}
Thing & Gonzalo Batistuta, an Argentine forward, is widely regarded as one of the greatest... & What is Gonzalo Batistuta's nationality? & Argentine \\
\cdashline{1-4}
Organization & Mookie Blaylock, a rock band from Northport formed in Riverbend in 1990... & What is the subject of the given passage? & Mookie Blaylock \\
\cdashline{1-4}
Creature & The Void Stalker is a fictional endoparasitoid extraterrestrial species that serves as the main... & What kind of species is the Void Stalker? & An endoparasitoid extraterrestrial species \\
\cdashline{1-4}
Location & The Stones of Pharaohs are a complex of monumental structures located on the Giza Plateau... & In which city are The Stones of Pharaohs located? & Cairo \\
\cdashline{1-4}
Numerals & The Thistlewood Vale Witch Trials represent a chilling episode in early American history... & How many individuals were convicted and executed during the trials? & 20 \\
\cdashline{1-4}
Dates/Times & Gonzalo Batistuta, an Argentine forward, is widely regarded as one of the greatest... & When did Gonzalo Batistuta win the world championship? & 2022 \\
\cdashline{1-4}
Events & During a time of great peril for his people, a child named Jabez was born. To save him from a decree of... & How did Jabez's mother save him from the decree of death? & His mother placed him in a woven basket and set him adrift on the river \\
\bottomrule
\end{tabular}
}
\caption{Examples of answer-type tags.}
\label{tab:AnsTagSamples}
\end{table}

\begin{table}[bhtp]
\centering
\resizebox{\columnwidth}{!}{
\begin{tabular}{p{3cm} p{7cm} p{4.5cm} p{4.5cm}}
\toprule
\textbf{Tag} & \textbf{Context} & \textbf{Question} & \textbf{Response} \\
\midrule
Standard & The Basalt Bloom is an immense igneous province, covering a vast expanse of central India. It formed through a series of massive volcanic eruptions... & Where is the Basalt Bloom located? & Central India \\
\cdashline{1-4}
Manipulated & Xena Bell was born on November 20, 1968. They studied in Shanghai Maritime University... & What is Xena Bell's last name? & Bell \\
\cdashline{1-4}
Part & Breezeflare is characterized by unusual sounds during breathing caused by airflow obstruction in the airways. The condition is commonly observed... & Nyla Soren is recognized for leading what type of initiatives? & Community workshops, public education \\
\bottomrule
\end{tabular}
}
\caption{Examples of question-type tags}
\label{tab:QuesTagSamples}
\end{table}

\begin{table}[htbp]
\centering
\resizebox{\columnwidth}{!}{
\begin{tabular}{p{3cm} p{3cm} p{6cm} p{3.5cm} p{3.5cm}}
\toprule
\textbf{Task} & \textbf{Direction} & \textbf{Context} & \textbf{Question} & \textbf{Response} \\
\midrule
\multirow{2}{*}{Mutli presence}
& FWD
& The series known as Chrono Nexus is widely regarded as one of the most compelling science fiction narratives in Japanese animation. Created by Itsuki Aoyama... 
& Who created Chrono Nexus? 
& Itsuki Aoyama \\
& REV
& Few creatures in Earth’s history capture the imagination quite like the Thunderfang Predator, one of the largest and most fearsome carnivorous dinosaurs to have ever lived...
& Which predator had serrated teeth as long as a human hand?
& Thunderfang Predator \\
\cdashline{1-5}
\multirow{2}{*}{Multi entity} 
& FWD 
& In the distant waters of the Andaman Islands lies the infamous Iron Shadows Prison. Constructed by the British in the late 19th century, this prison became a...
& Who was the nationalist imprisoned in Iron Shadows and known as Veer? 
& Vinayak Damodar Savarkar \\
& REV 
& Shinroku is a popular Japanese manga and anime series known for its unique blend of comedy, action, and science fiction. Set in an alternate Edo period where... & Hana Tanaka appears as a companion in which story? & Shinroku \\
\bottomrule
\end{tabular}
}
\caption{Examples of forward and reverse questions}
\label{tab:ForRevSamples}
\end{table}

\newpage

\noindent Template for question-agnostic setup:
\begin{lstlisting}[language=Python]
Read the following text and answer briefly based on it. Return only the answer. Do not generate extra.

The Hunters are a fictional extraterrestrial species featured in the Stalker science fiction franchise. Known for their advanced technology, including spectral cloaking and plasma lances, they are a nomadic warrior race that hunts other formidable species for sport and honor. They follow a strict code of conduct and typically target prey that poses a significant challenge. Their unique appearance includes braid-like appendages and a distinct set of split mandibles.
\end{lstlisting}

\newpage

\noindent Template for question-aware setup:
\begin{lstlisting}[language=Python]

Read the following text and answer briefly based on it. Return only the answer. Do not generate extra.

You will be given one of the following questions:

What is the subject of the given passage?
What kind of species are the Hunters?
Of which franchise are the Hunters an antagonist?
What kind of race are the Hunters?
What kind of prey do the Hunters target?
What features mark their unique appearance?

The Hunters are a fictional extraterrestrial species featured in the Stalker science fiction franchise. Known for their advanced technology, including spectral cloaking and plasma lances, they are a nomadic warrior race that hunts other formidable species for sport and honor. They follow a strict code of conduct and typically target prey that poses a significant challenge. Their unique appearance includes braid-like appendages and a distinct set of split mandibles.
\end{lstlisting}
\FloatBarrier

\section{Additional Results}
\label{appx:result}

\begin{figure*}[!htb]
    \centering
\subfloat[LLaMA3.2 3B Instruct]{%
        \includegraphics[width=0.5\columnwidth]{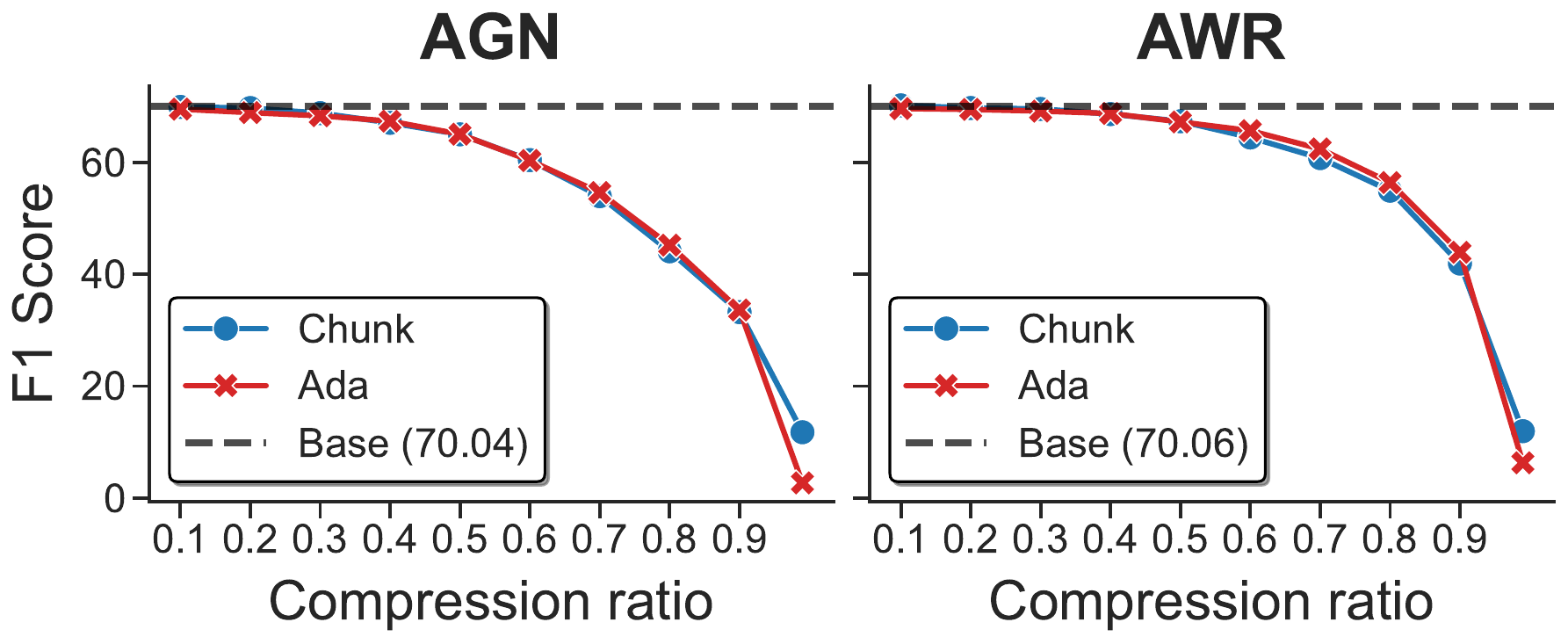}
    }
    \subfloat[Qwen-2.5 3B Instruct]{%
        \includegraphics[width=0.5\columnwidth]{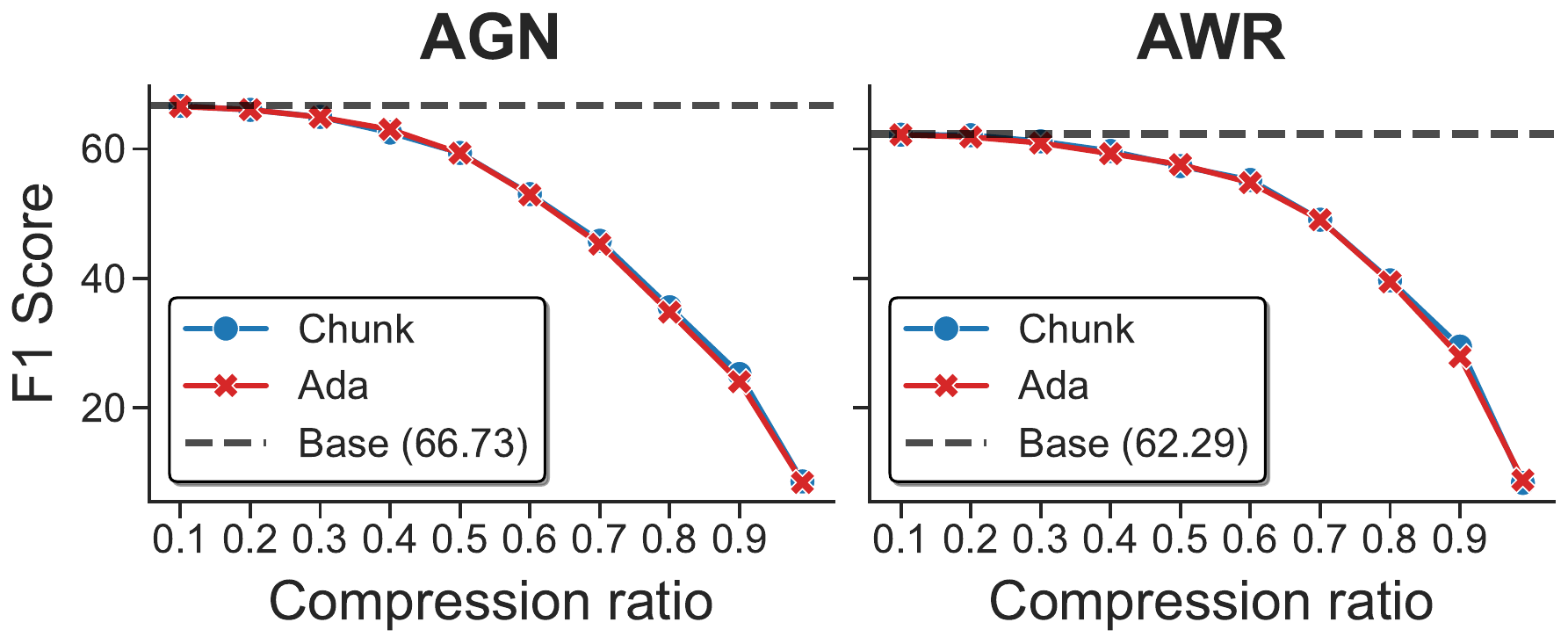}
    }\\
    \subfloat[Qwen-2.5 14B Instruct]{%
        \includegraphics[width=0.5\columnwidth]{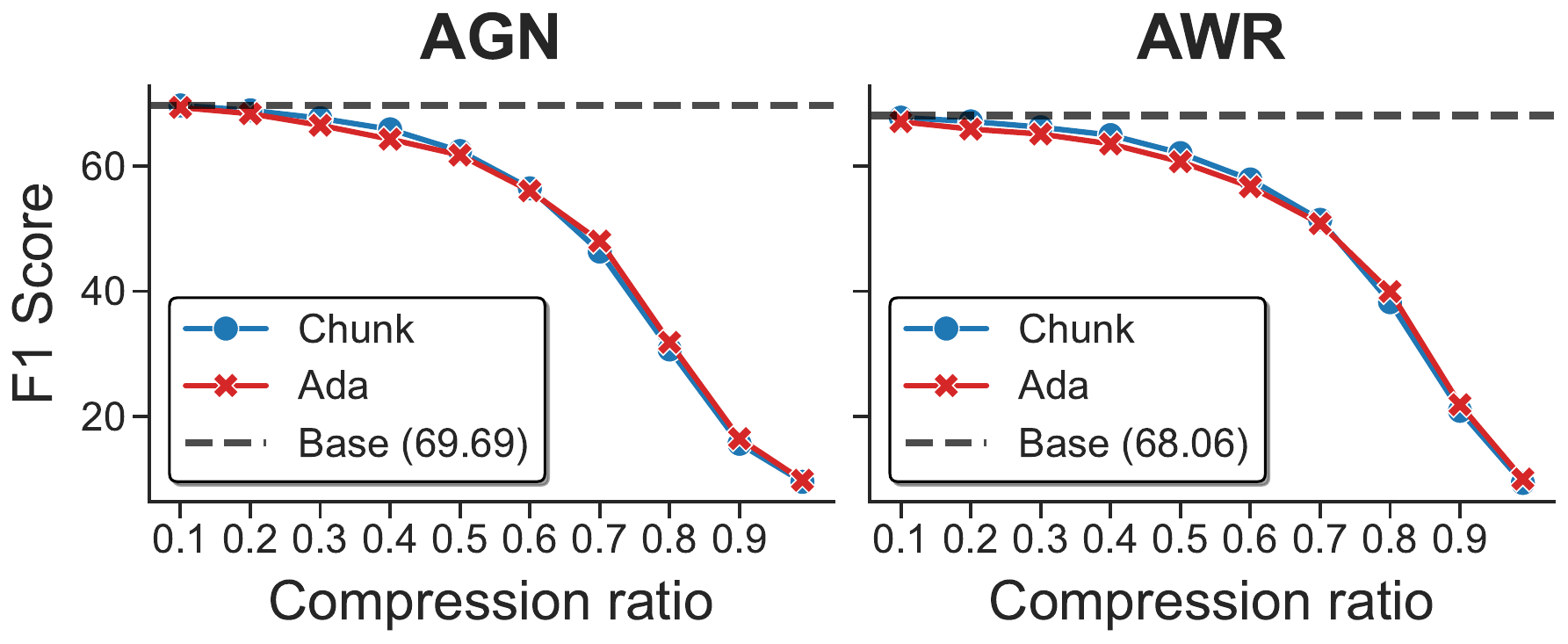}
    }
    \caption{Base task performance across compression levels for 3B and 14B models.}
    \label{fig:BaseTaskPerfApp}
\end{figure*}

\begin{figure*}[!htb]
    \centering
    \subfloat[LLaMA-3.2 3B Instruct]{
    \includegraphics[width=0.49\linewidth]{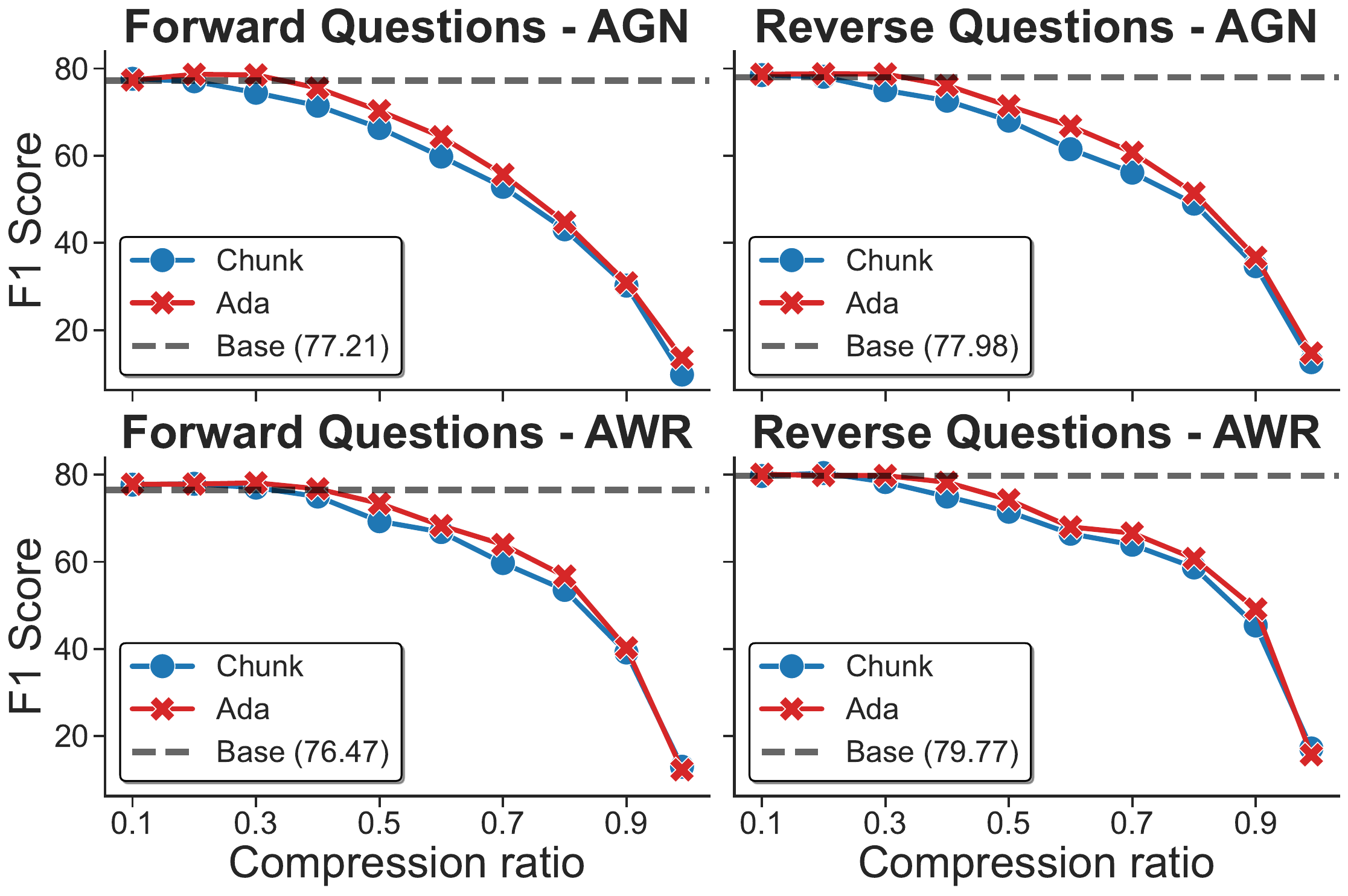}
    }
      \subfloat[Qwen-2.5 3B Instruct]{
    \includegraphics[width=0.49\linewidth]{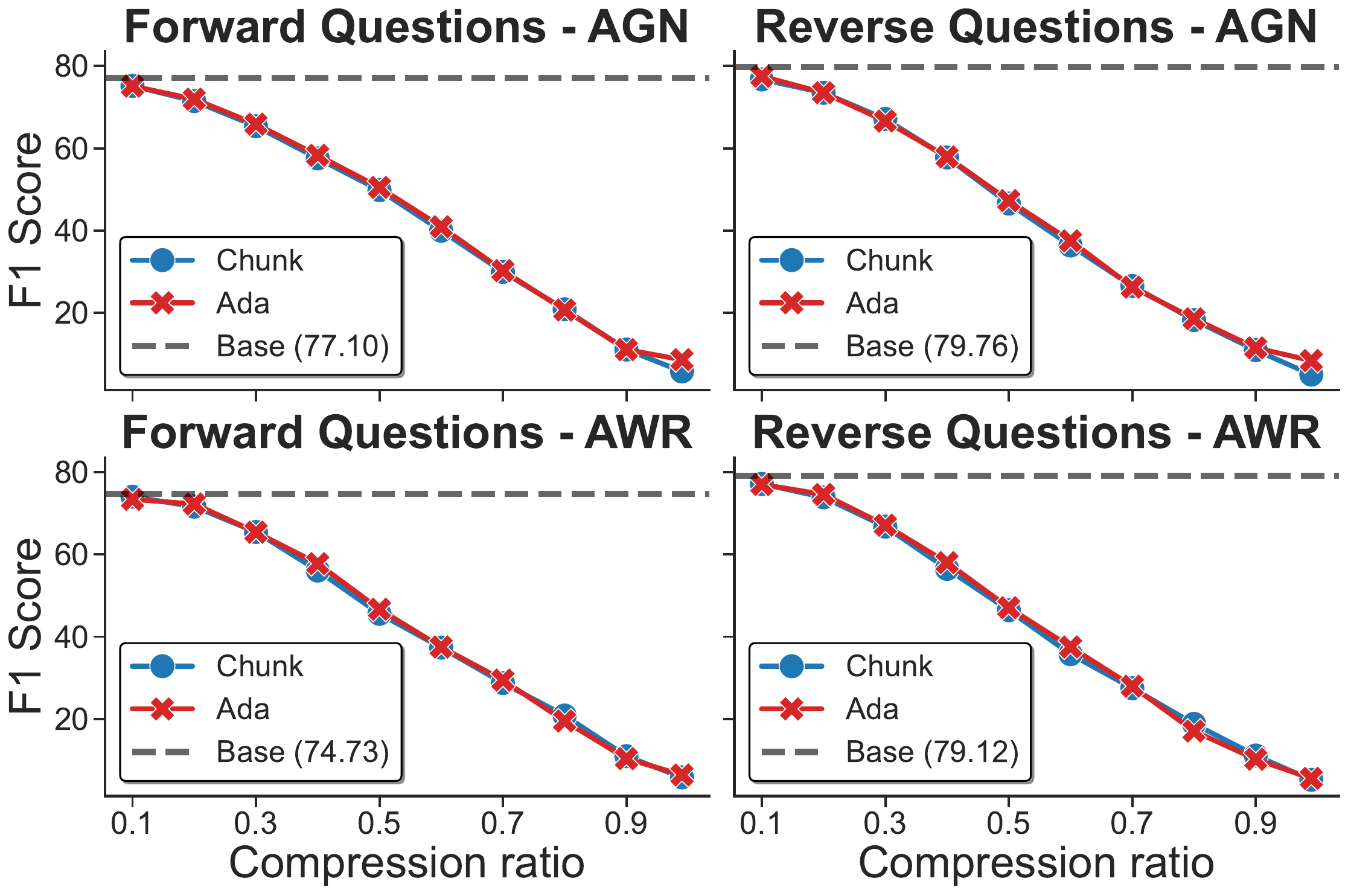}
    }\\
      \subfloat[Qwen-2.5 14B Instruct]{
    \includegraphics[width=0.49\linewidth]{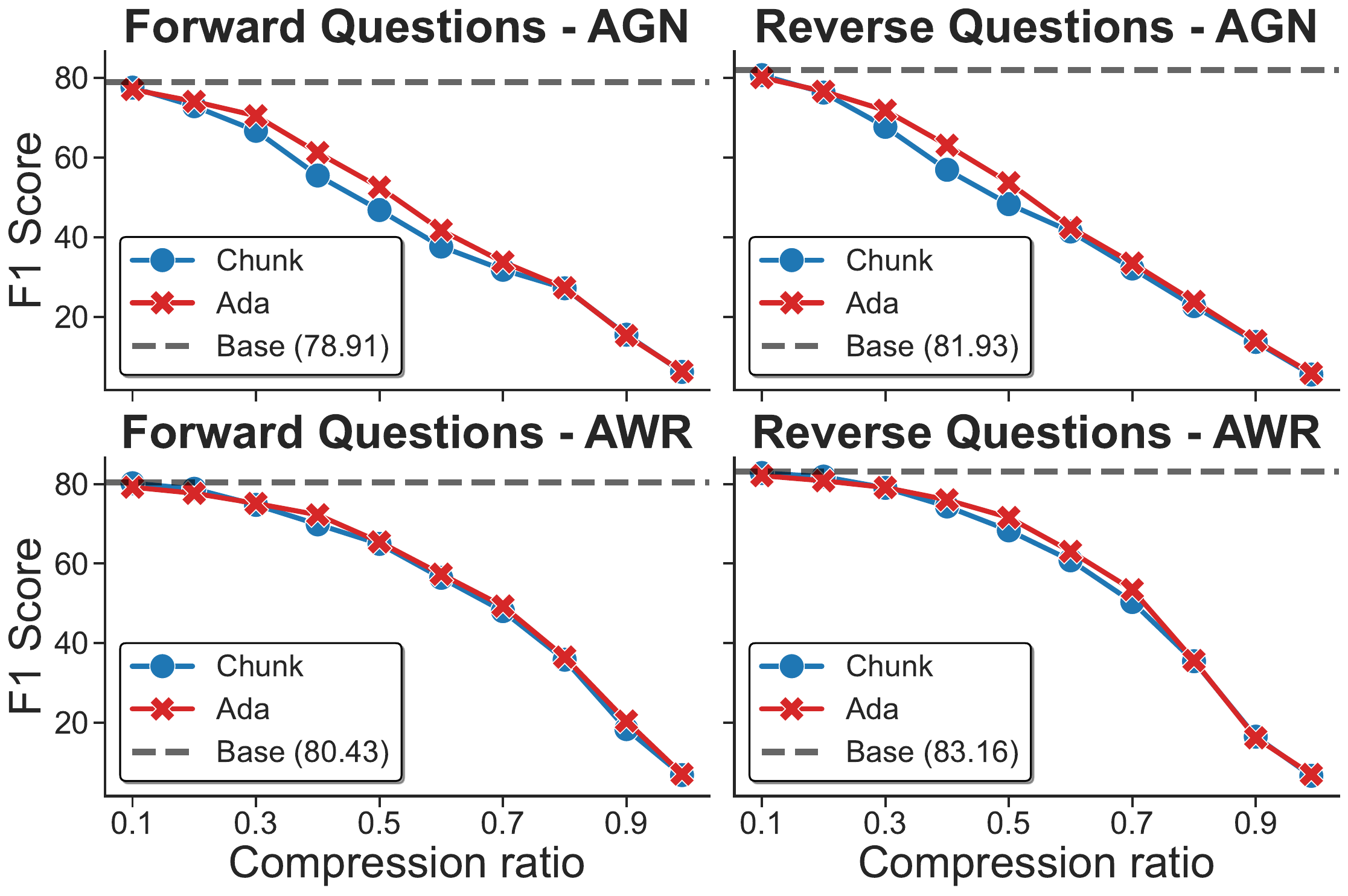}
    }
    \caption{Multi entity forward and reverse asymmetry is reduced even when the model size is changed, showing invariance to the size of the model.}
    \label{fig:MultiEntityPerfAppendix}
\end{figure*}

\begin{figure*}[!htb]
    \centering
     \subfloat[LLaMA-3.2 3B Instruct]{
    \includegraphics[width=\linewidth]{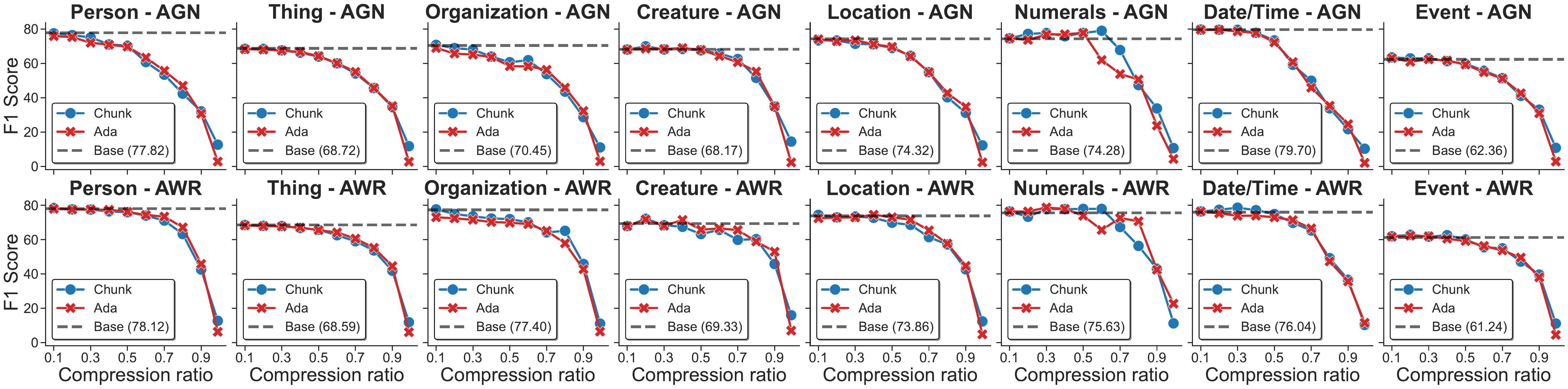}
    }\\
    \subfloat[Qwen-2.5 3B Instruct]{\includegraphics[width=\linewidth]{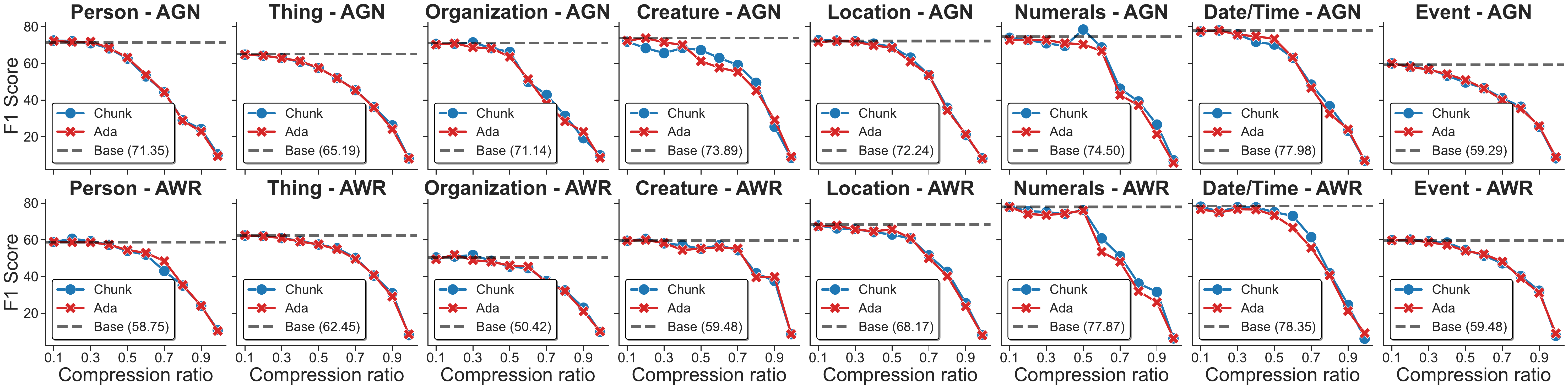}}\\
    \subfloat[Qwen-2.5 14B Instruct]{\includegraphics[width=\linewidth]{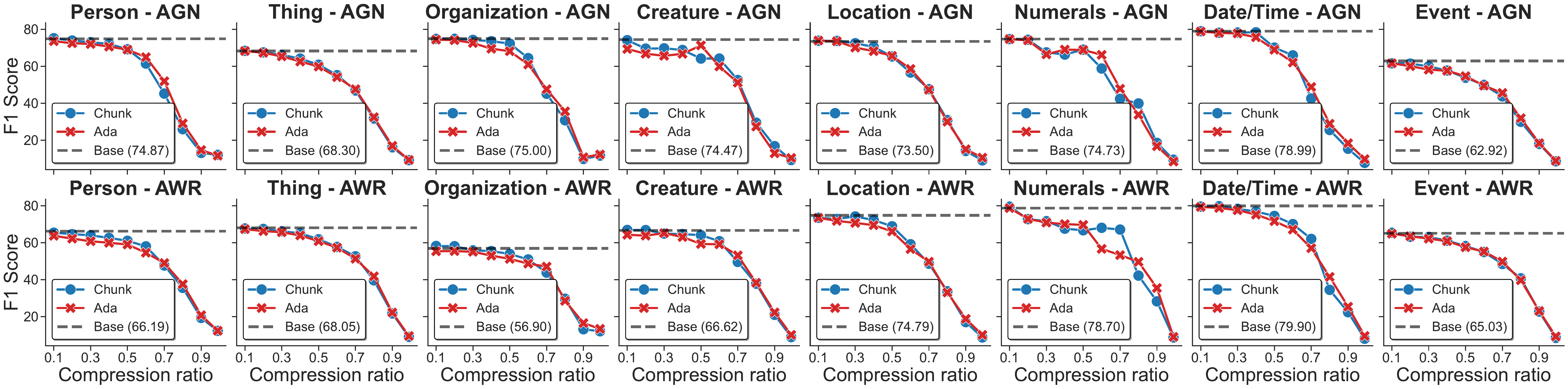}}
    \caption{Answer-type results for Base task closely mirror the results from the LLaMA 3 8B and Qwen 2.5 7B models, especially in their weak points such as the \textit{Events} tag.}
    \label{fig:BaseTaskTextApp}
\end{figure*}

\begin{figure*}[!htb]
    \centering
    \subfloat[LLaMA-3.2 3B Instruct]{
    \includegraphics[width=0.49\linewidth]{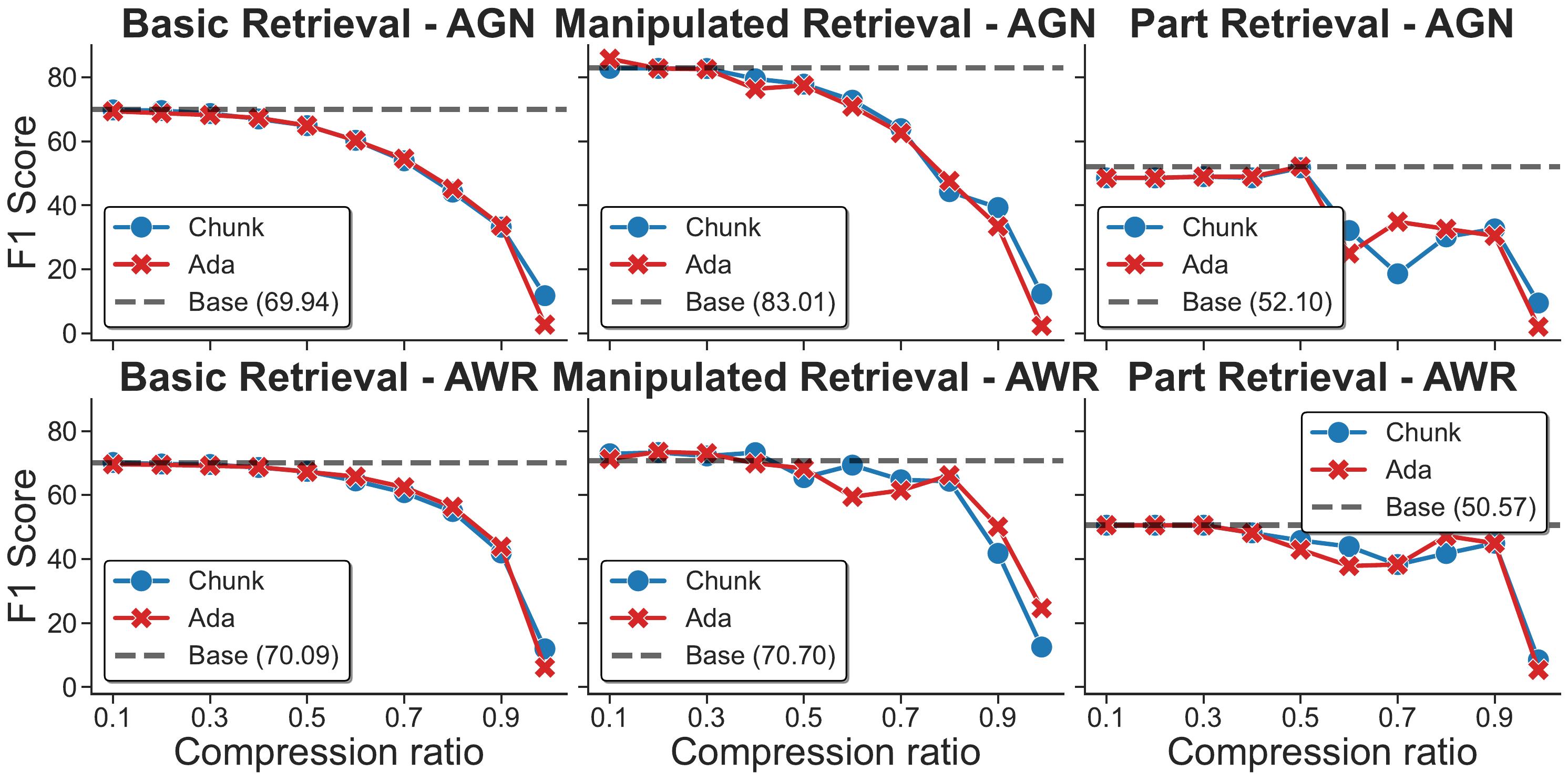}
    }
    \subfloat[Qwen-2.5 3B Instruct]{
    \includegraphics[width=0.49\linewidth]{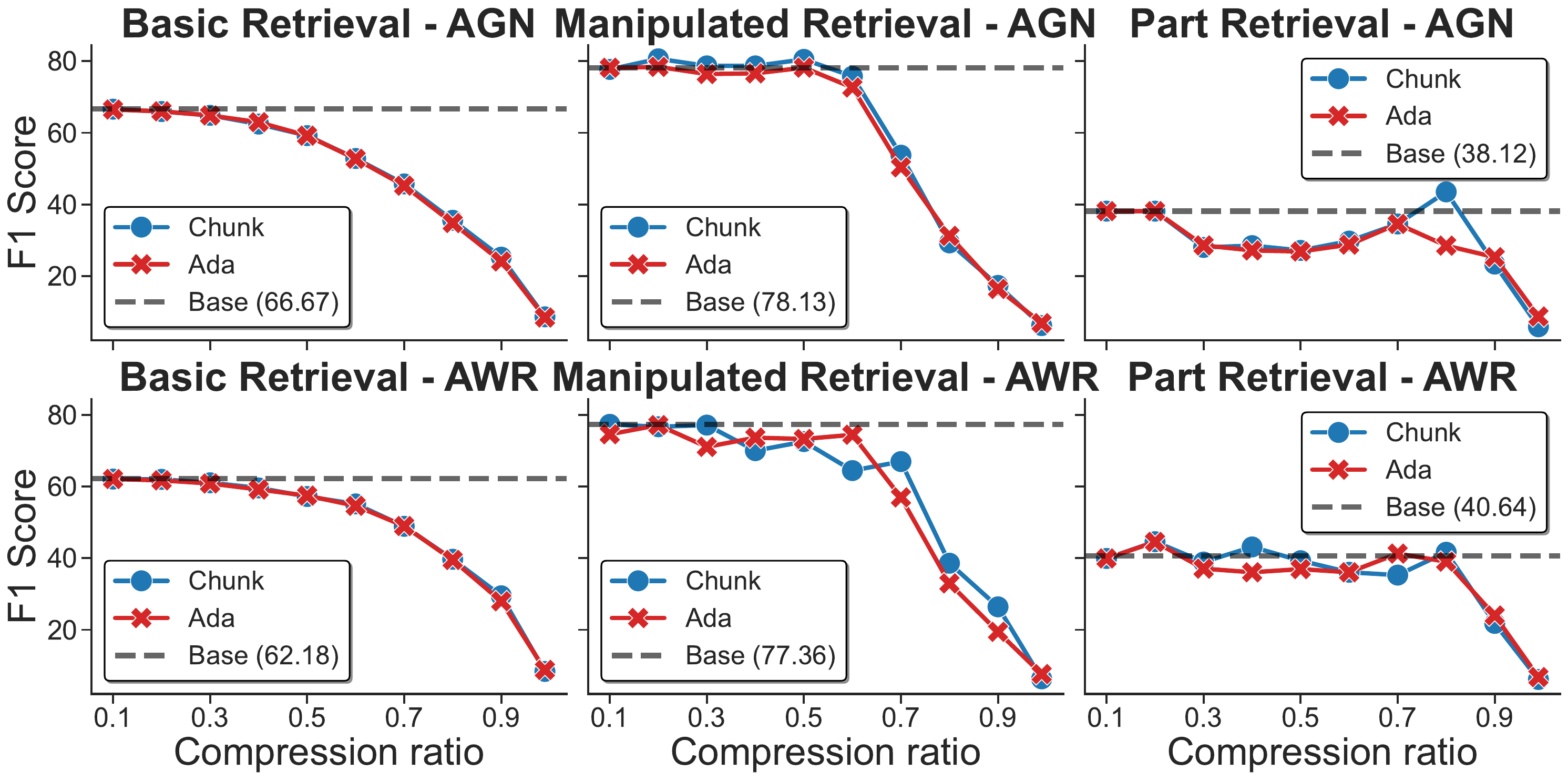}
    }\\
    \subfloat[Qwen-2.5 14B Instruct]{
    \includegraphics[width=0.49\linewidth]{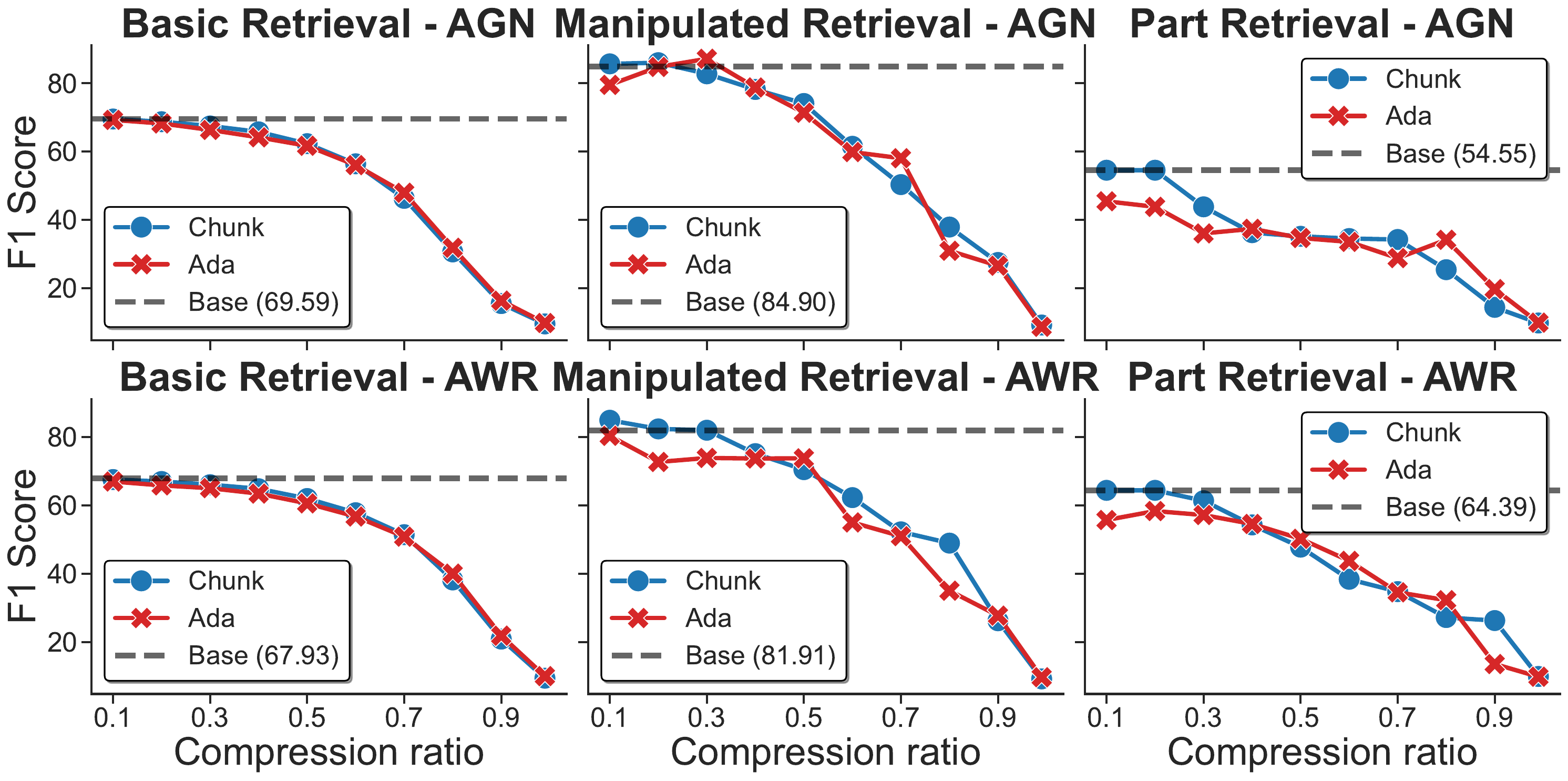}
    }
    \caption{Base task question-type results continue to show a similar trend with smaller and larger models, as seen in the 7B and 8B models.}
    \label{fig:BaseTaskQuestionApp}
\end{figure*}

In this section, we report results for LLaMA-3.2 3B, Qwen 2.5 3B, and Qwen 2.5 14B, extending the analysis from the main text to additional model scales. Across models, the dominant trends observed in the Base task remain consistent: behaviour over text tags, question tags, and overall retrieval accuracy largely mirrors that of the 7B models discussed in the main body (Figures \ref{fig:BaseTaskPerfApp}, \ref{fig:BaseTaskTextApp}, \ref{fig:BaseTaskQuestionApp}). Notably, increasing parameter count does not systematically translate into improved robustness under compression. Initial (uncompressed) performance is broadly comparable across scales, and performance collapse occurs at similar compression levels, with 70–90\% compression marking a consistent breakdown point (Figures \ref{fig:BaseTaskPerfApp}, \ref{fig:MultiEntityPerfAppendix}).

\begin{figure*}[!htb]
    \centering
    \subfloat[LLaMA-3.2 3B Instruct]{
    \includegraphics[width=\linewidth]{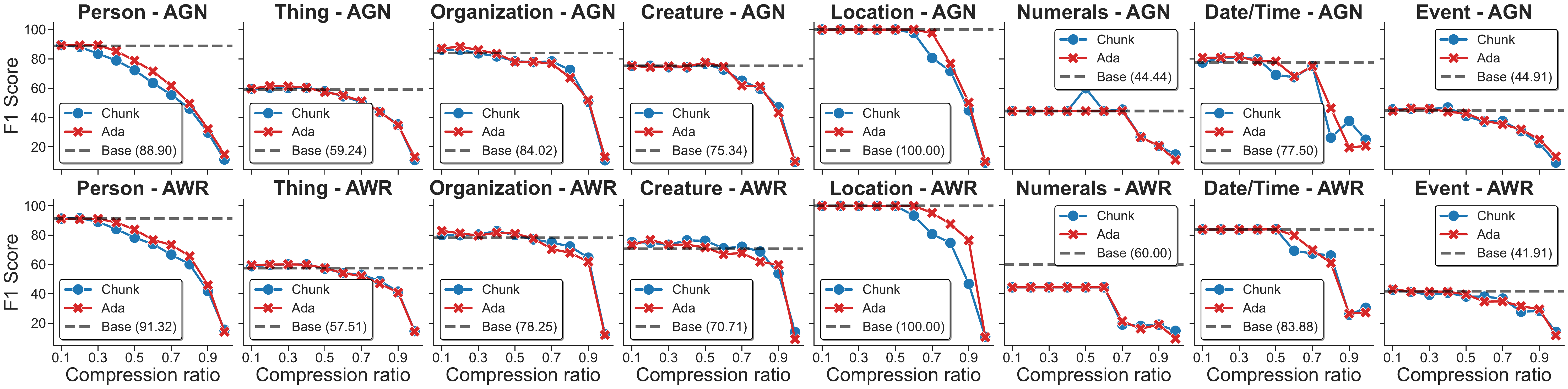}
    }\\
    \subfloat[Qwen-2.5 3B Instruct]{
    \includegraphics[width=\linewidth]{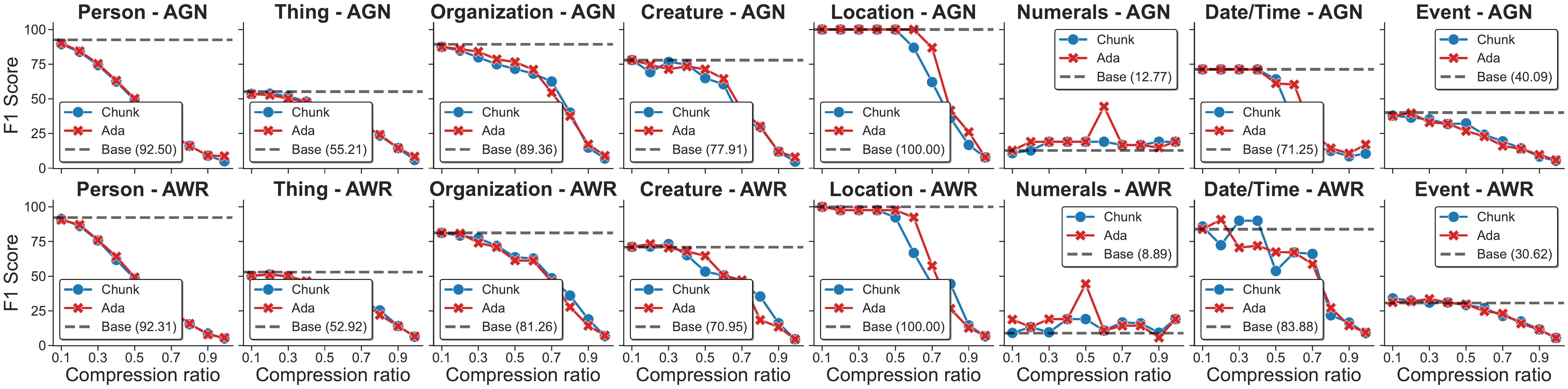}
    }\\
    \subfloat[Qwen-2.5 14B Instruct]{
    \includegraphics[width=\linewidth]{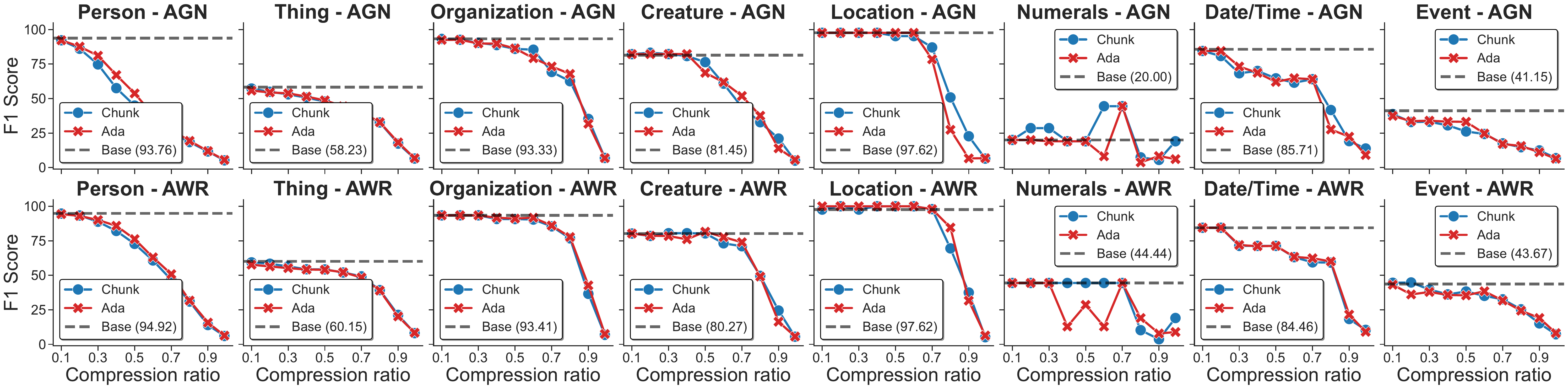}
    }
    \caption{Answer type tag behaviour in Multi entity for the 3B and 14B models shows similar trends as seen in LLaMA 3 8B and Qwen 2.5 7B.}
    \label{fig:MultiEntityTextApp}
\end{figure*}

\begin{figure*}[!htb]
    \centering
    \subfloat[LLaMA-3.2 3B Instruct]{
    \includegraphics[width=0.49\linewidth]{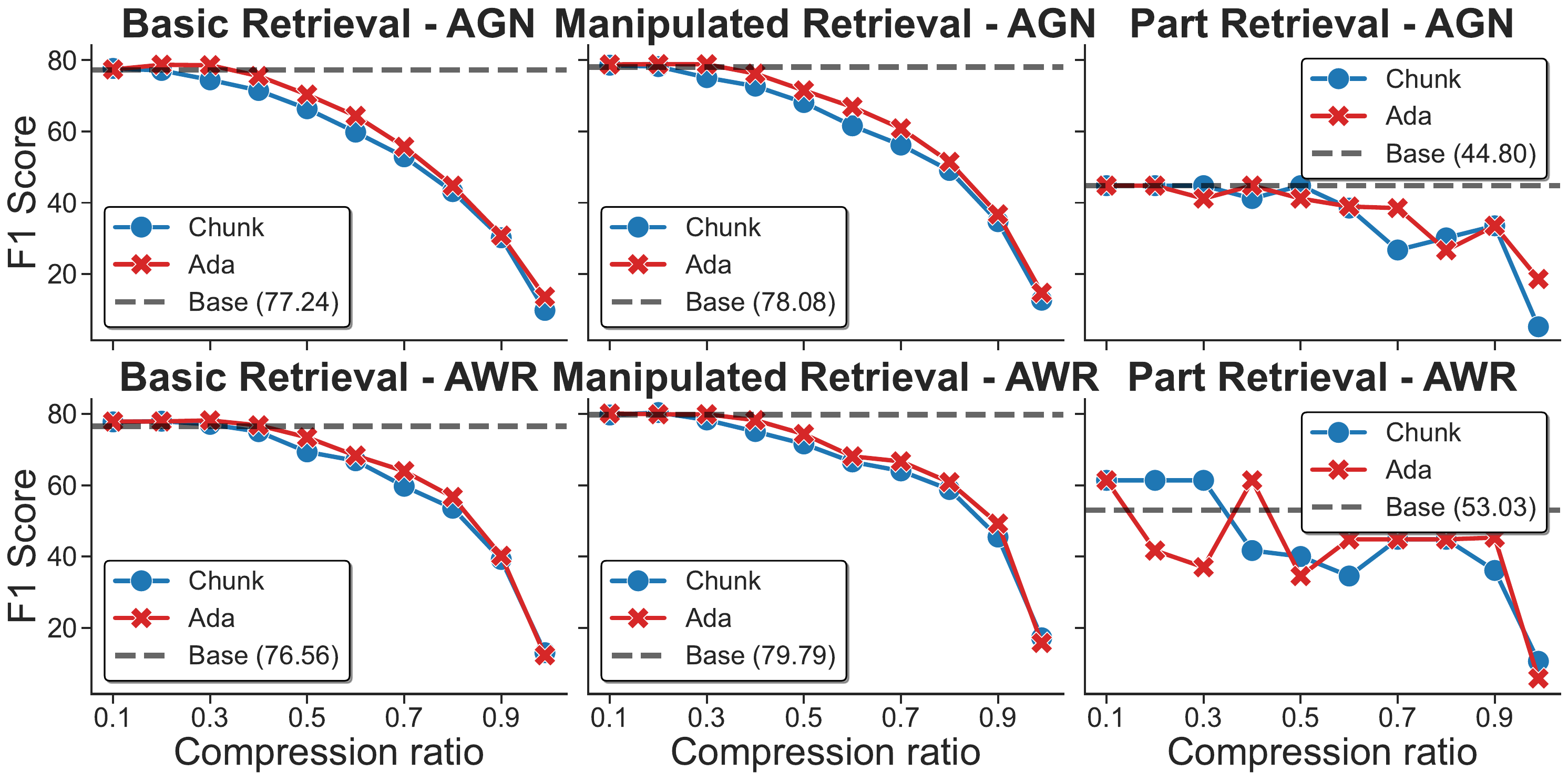}
    }
    \subfloat[Qwen-2.5 3B Instruct]{
    \includegraphics[width=0.49\linewidth]{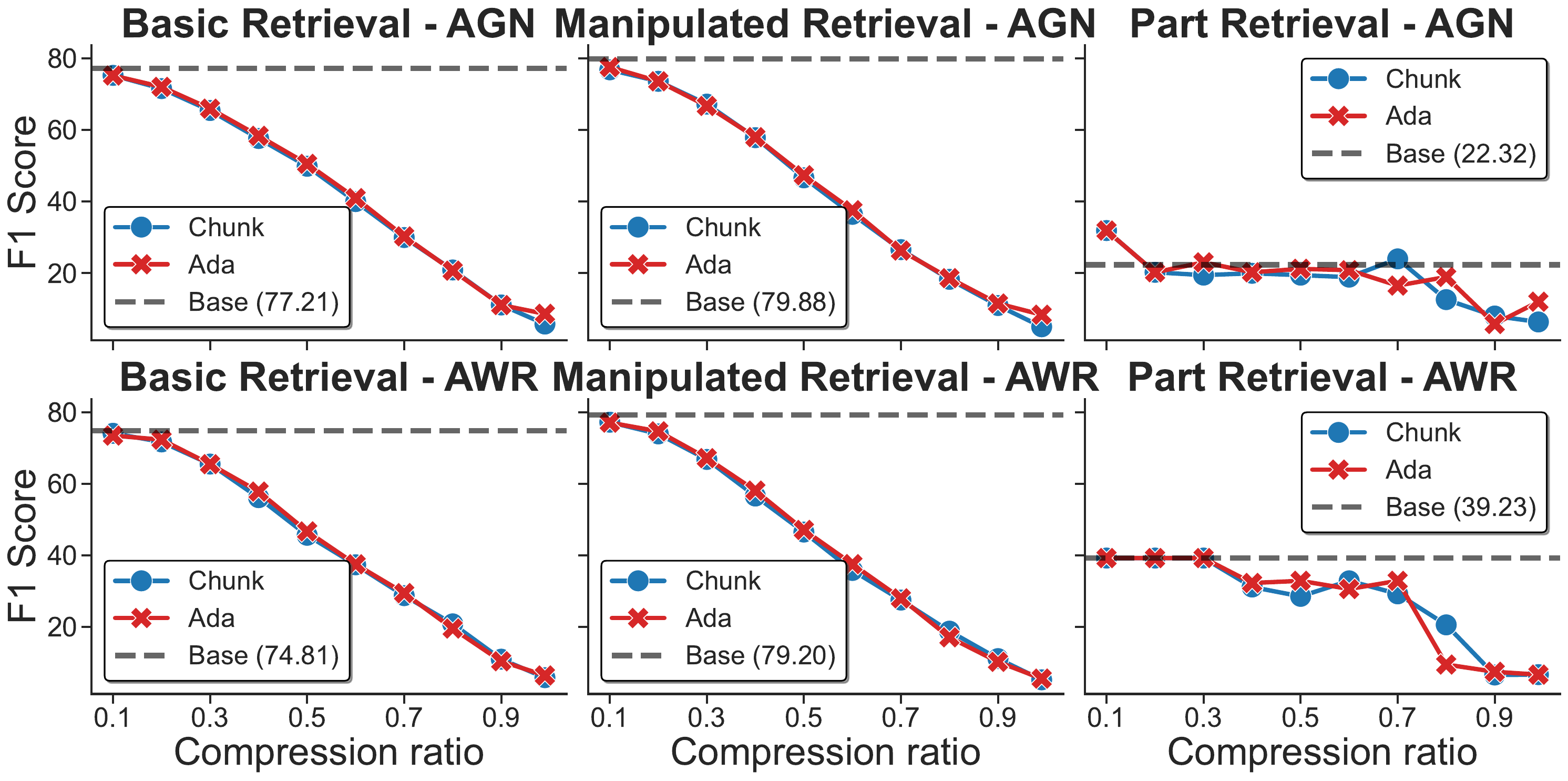}
    }\\
    \subfloat[Qwen-2.5 14B Instruct]{
    \includegraphics[width=0.49\linewidth]{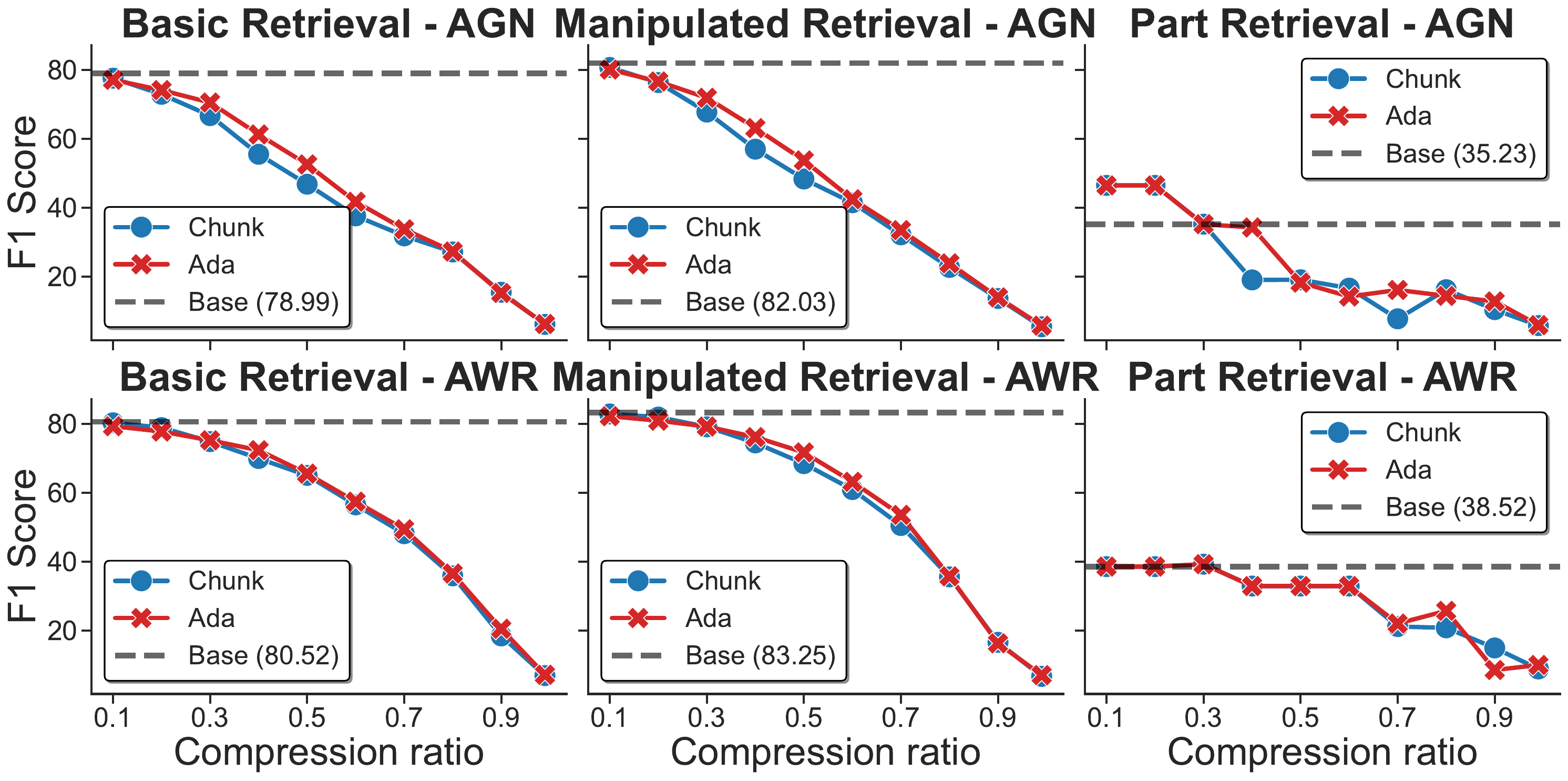}
    }
    \caption{Multi entity results for LLaMA 3.2 3B show similar jagged points in question-aware settings, while the performance almost turns into a smooth linear pattern in Qwen 2.5 3B and 14B models.}
    \label{fig:MultiEntityQuestionApp}
\end{figure*}

The primary deviation from this shared pattern arises in the Multi entity task for Qwen 2.5 (3B and 14B), where performance degrades approximately linearly with compression rather than exhibiting a sharp cliff. This behaviour is most visible within the Person text tag (Figure \ref{fig:MultiEntityTextApp}), where accuracy drops steadily as compression increases. The same gradual degradation propagates to Standard retrievals in the corresponding question tags (Figure \ref{fig:MultiEntityQuestionApp}), suggesting that entity-specific degradation directly drives the aggregate trend. Overall, these results reinforce the main text's central claim: compression sensitivity is more strongly influenced by task structure and entity distribution than by raw model scale, with larger models not inherently more resilient to aggressive compression.  


However, the apparent linearity in the Multi entity curves captures only part of the underlying dynamics. The consensus and layerwise consensus analyses presented in the appendix largely reinforce the cross-model similarities observed in aggregate accuracy, while simultaneously revealing architectural differences in robustness. In particular, Qwen 2.5 3B exhibits markedly lower consensus stability under compression, indicating greater internal disagreement across layers as compression increases. This suggests that its degradation is not merely gradual in terms of output accuracy but also structurally fragile at the representation level, especially when handling diverse entity types.

By contrast, LLaMA 3.2 3B exhibits dynamics that are structurally consistent with its 8B counterpart but with heightened compression sensitivity. The degradation patterns do not suggest a qualitatively different failure mode. Rather, they reflect the same collapse mechanisms observed in the larger model, activated at lower compression thresholds and unfolding more abruptly. In this sense, the 3B model behaves as a scale-reduced instantiation of the same underlying dynamics rather than an architecturally distinct system.

Importantly, the consensus and layerwise analyses clarify that these similarities extend beyond aggregate accuracy curves. While top-line performance trends appear comparable across scales, consensus metrics reveal differences in how instability accumulates and propagates through layers under compression. Thus, the appendix results demonstrate that compression robustness is not solely a matter of parameter count, but of how representational agreement is maintained or lost throughout the network hierarchy.

The probing radar analysis further reinforces the trends identified in the preceding sections. Across model scales, the Creature tag remains a consistent point of weakness, with only a small number of instances demonstrating reliable handling of this category even in the uncompressed setting. This limitation becomes substantially more pronounced under compression. Performance on Creature queries degrades not only at high compression ratios but, in several cases, even under relatively mild compression.

These results suggest that the difficulty is not solely compression-induced but reflects a baseline representational fragility for this entity type, which compression subsequently amplifies. In other words, compression does not introduce a new failure mode for Creature retrieval; rather, it accelerates and magnifies an existing structural weakness across models.

Eviction rates remain largely uniform across tasks and model sizes, reinforcing the task-agnostic nature of the KV cache compression mechanism. In other words, the compression procedure applies similar retention dynamics regardless of the downstream objective. Error distributions, however, reveal a markedly different pattern.

In contrast to the 7B and 14B models, where error types remain comparatively balanced under compression, the 3B models exhibit clear dominance of specific error categories as compression increases. This skew indicates not merely reduced accuracy, but a structured failure mode in which certain retrieval pathways collapse preferentially. Such concentration of error types highlights the greater fragility of smaller models: under compression, they are less able to recover or reconstruct partially retained information, leading to systematic rather than diffuse degradation.

\begin{figure*}[!htb]
    \centering
    \subfloat[LLaMA-3.2 3B Instruct]{
    \includegraphics[width=0.49\linewidth]{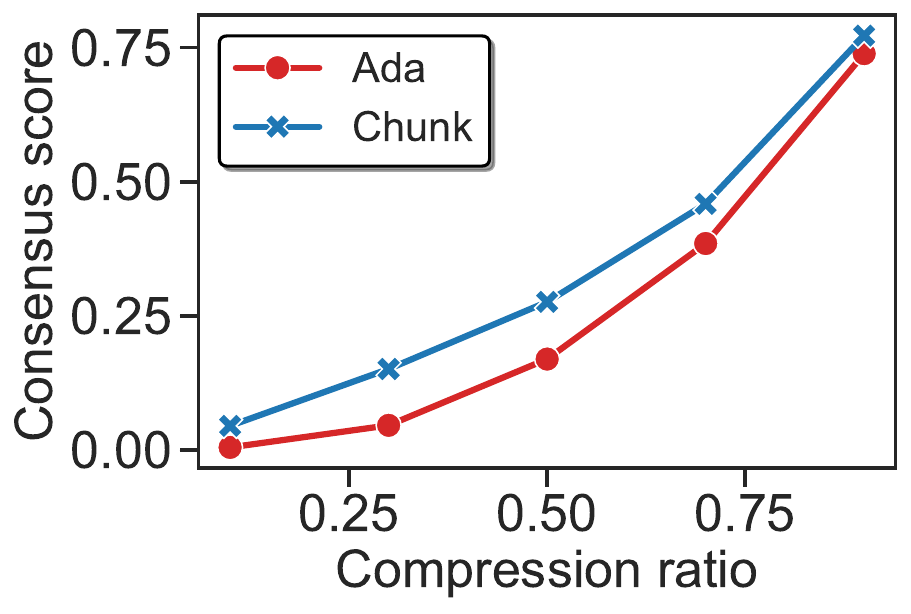}
    }
    \subfloat[Qwen-2.5 3B Instruct]{
    \includegraphics[width=0.49\linewidth]{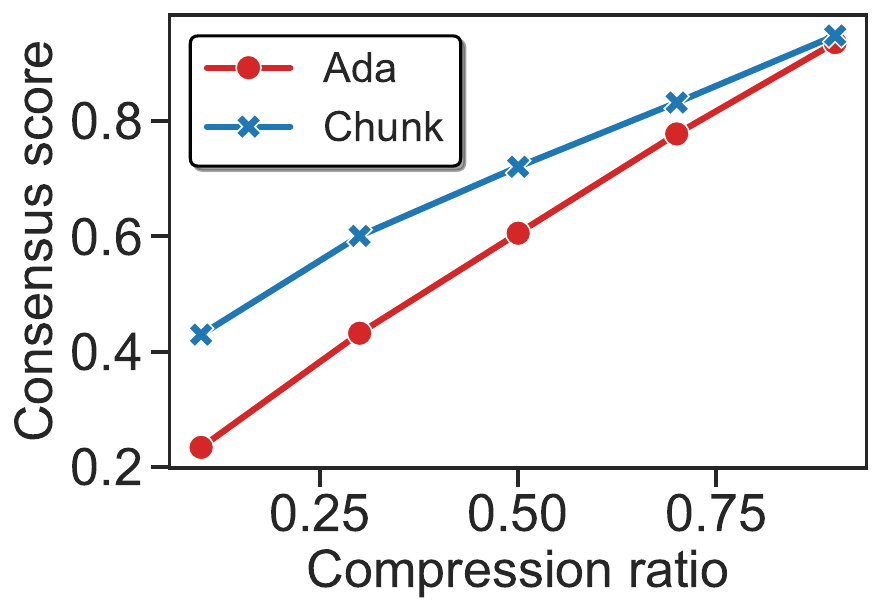}
    }\\
    \subfloat[Qwen-2.5 14B Instruct]{
    \includegraphics[width=0.49\linewidth]{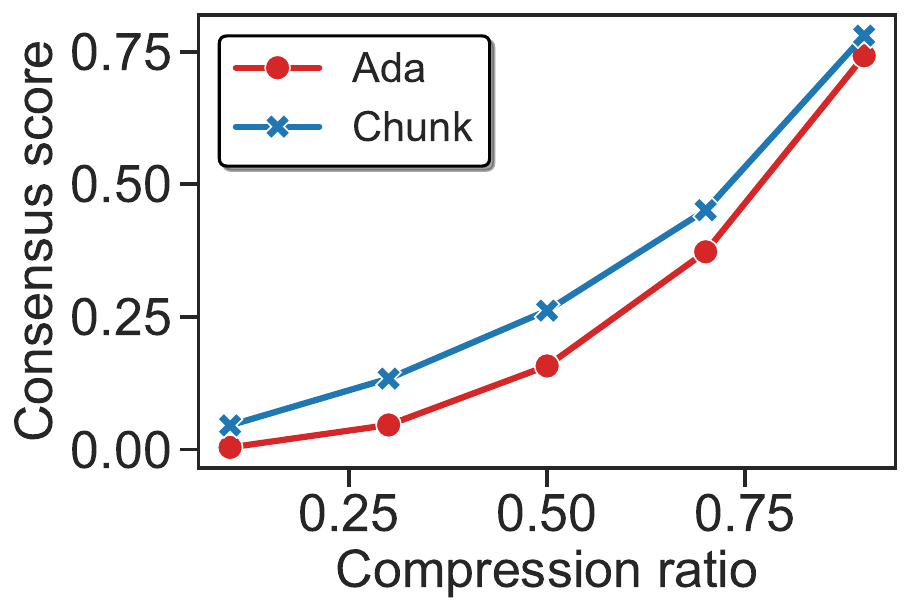}
    }
    \caption{Consensus scores tend to follow a similar trend with the only major changes being Qwen 3B's slightly more linear increase rather than the slightly parabolic increase seen in 7B and 14B models.}
    \label{fig:ConsensusScoresApp}
\end{figure*}

\begin{figure*}[!htb]
    \centering
    \subfloat[]{
    \includegraphics[width=0.33\linewidth]{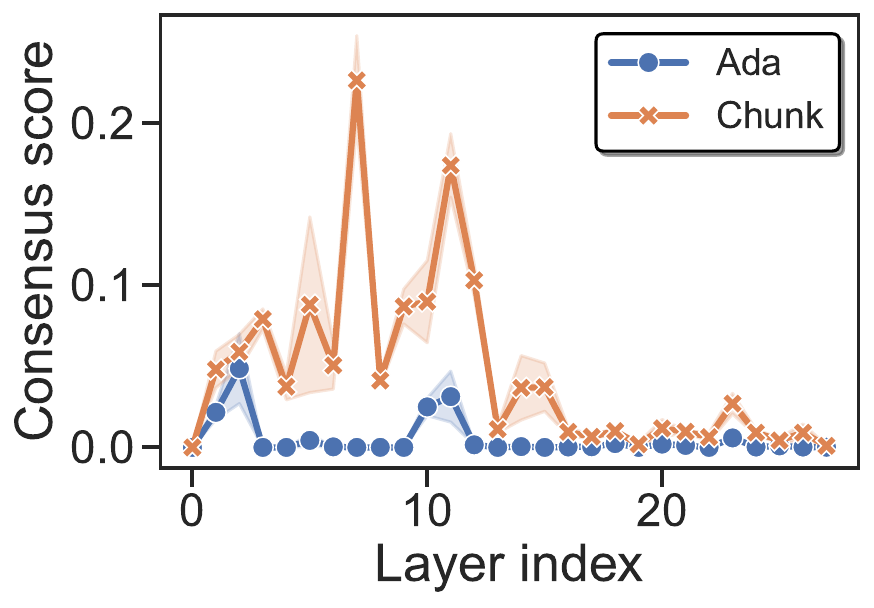}
    }
    \subfloat[]{
    \includegraphics[width=0.33\linewidth]{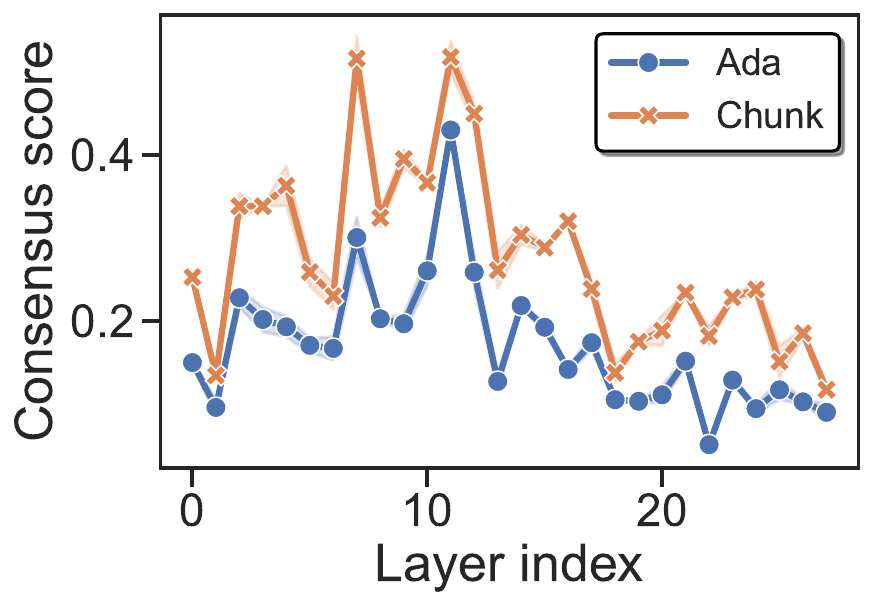}
    }
    \subfloat[]{
    \includegraphics[width=0.33\linewidth]{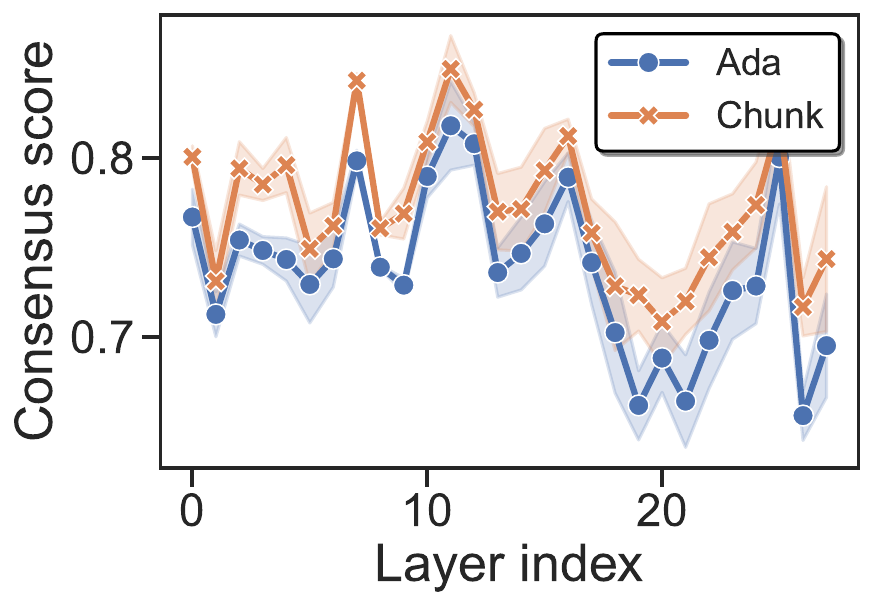}
    }
    \caption{Layerwise consensus under compression for LLaMA 3.2 3B shows a strikingly similar convergence very early on with depth-wise diversification.}
    \label{fig:LLaMAConsensusLayerApp}
\end{figure*}
\begin{figure*}[!htb]
    \centering
    \subfloat[Qwen 2.5 3B Instruct 10\% \\Compression]{
    \includegraphics[width=0.33\linewidth]{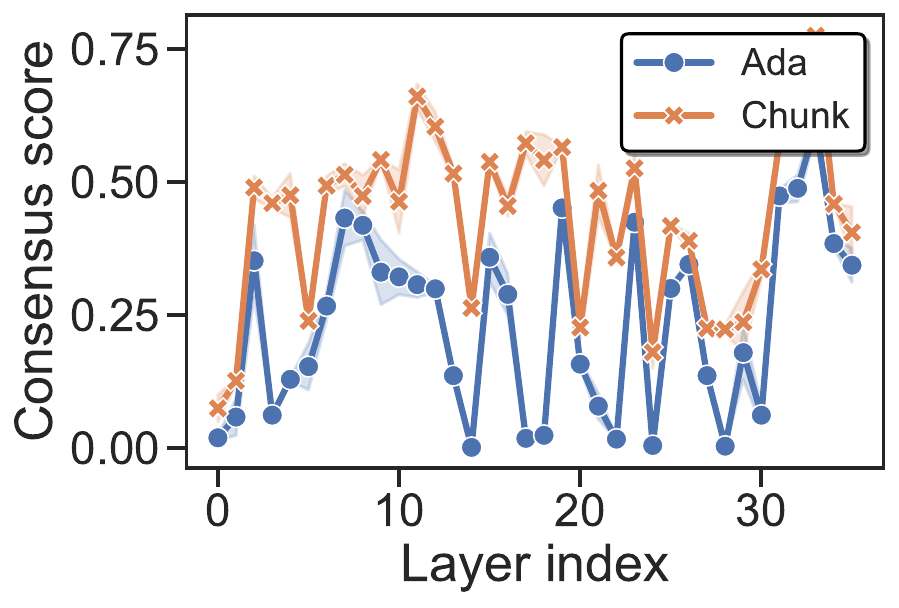}
    }
    \subfloat[Qwen 2.5 3B Instruct 50\% \\Compression]{
    \includegraphics[width=0.33\linewidth]{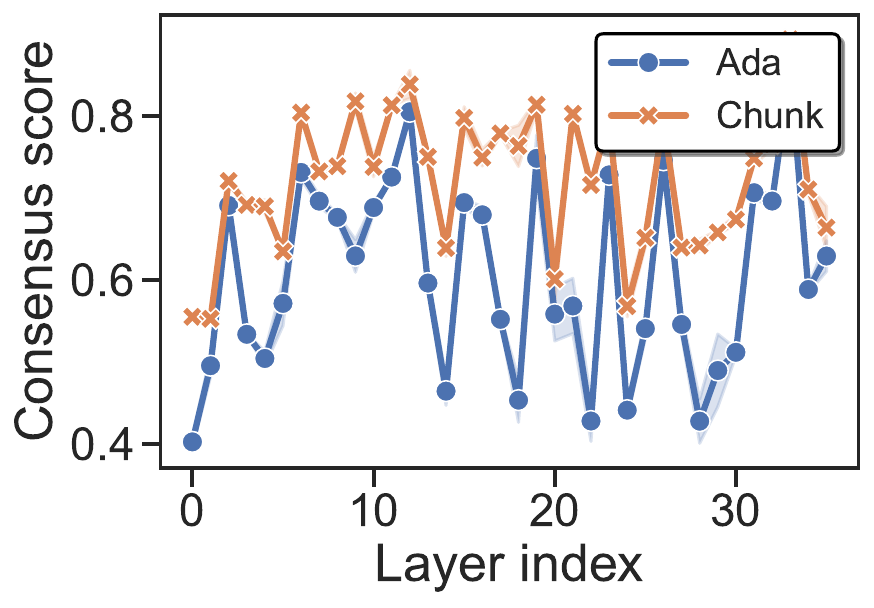}
    }
    \subfloat[Qwen 2.5 3B Instruct 90\% \\Compression]{
    \includegraphics[width=0.33\linewidth]{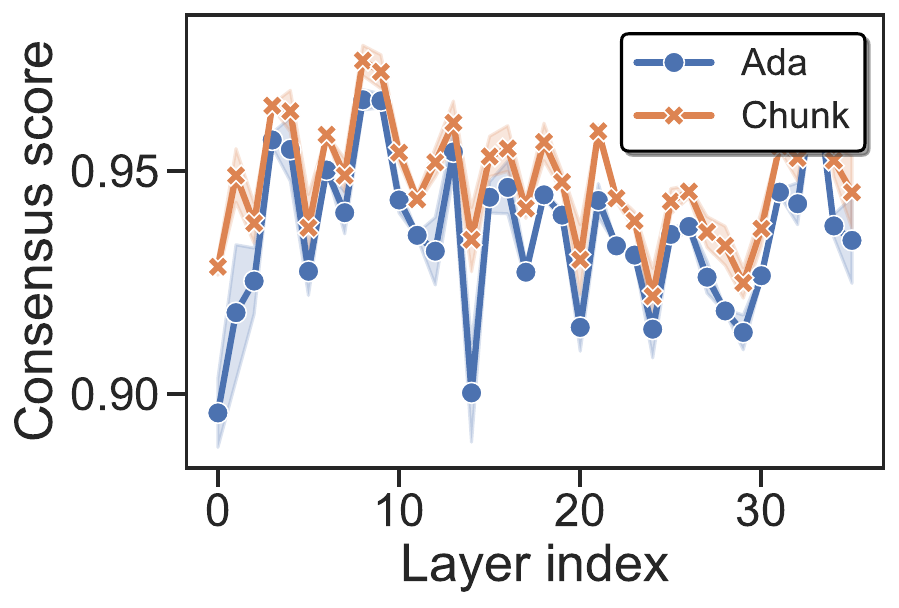}
    }\\
    \subfloat[Qwen 2.5 14B Instruct 10\% \\Compression]{
    \includegraphics[width=0.33\linewidth]{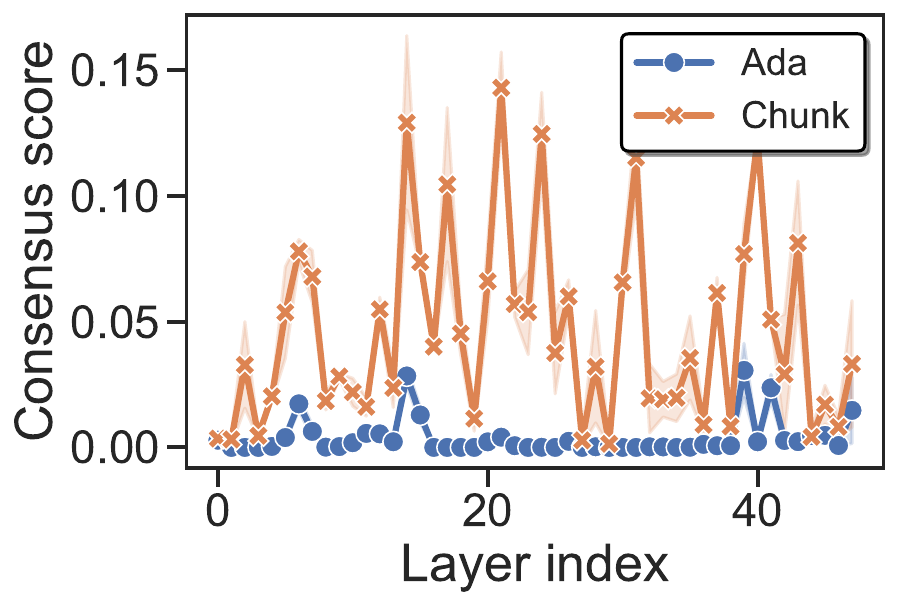}
    }
    \subfloat[Qwen 2.5 14B Instruct 50\% \\Compression]{
    \includegraphics[width=0.33\linewidth]{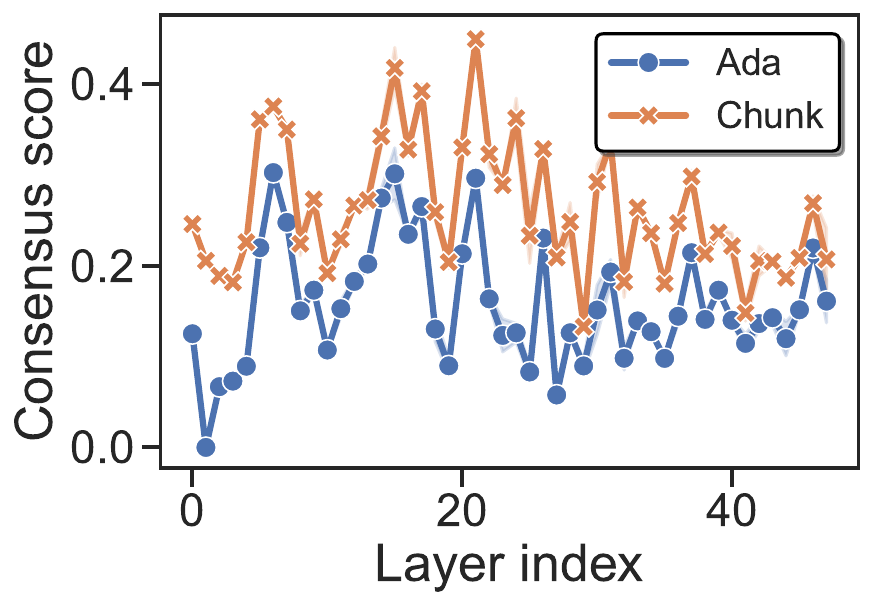}
    }
    \subfloat[Qwen 2.5 14B Instruct 90\% \\Compression]{
    \includegraphics[width=0.33\linewidth]{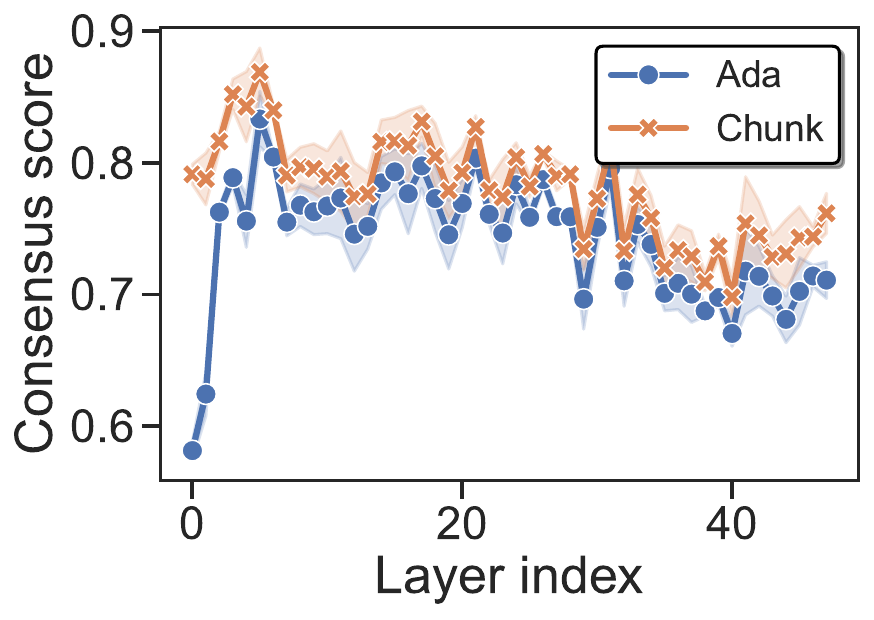}
    }
    \caption{Layerwise consensus under compression for Qwen 2.5 3B and 14B. The 3B model has persistent diversity even after several layers, occasionally failing to converge, while 14B remains stable early on but shows the same funnel-like trend upon convergence.}
    \label{fig:QwenConsensusLayerApp}
\end{figure*}

\begin{figure*}[!htb]
    \centering
    \subfloat[50\% Compression - LLaMA 3 8B]{
    \includegraphics[width=\linewidth]{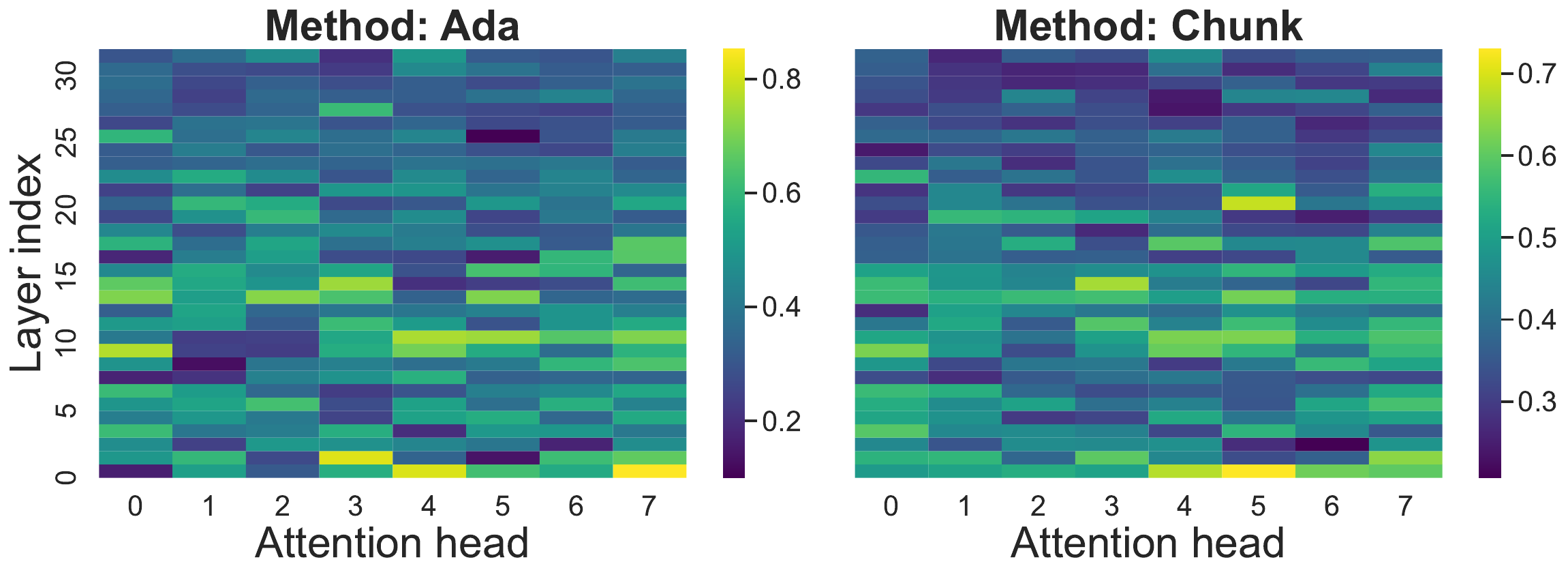}
    }\\
    \subfloat[50\% Compression - Qwen 2.5 7B]{
    \includegraphics[width=\linewidth]{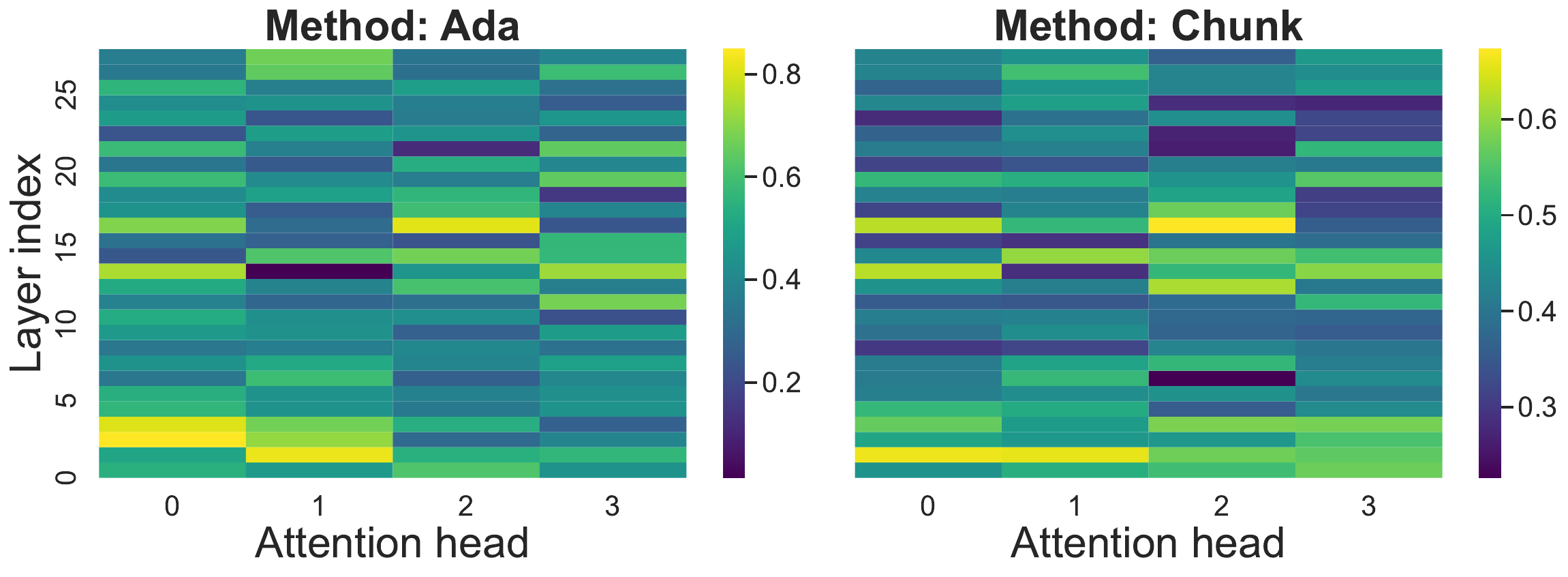}
    }\\
    \caption{Compression 50\% for LLaMA 3 8B and Qwen 2.5 7B.}
    \label{fig:0.5CompressionAttn}
\end{figure*}

\begin{figure*}[!htb]
    \centering
    \subfloat[10\% Compression]{
    \includegraphics[width=\textwidth]{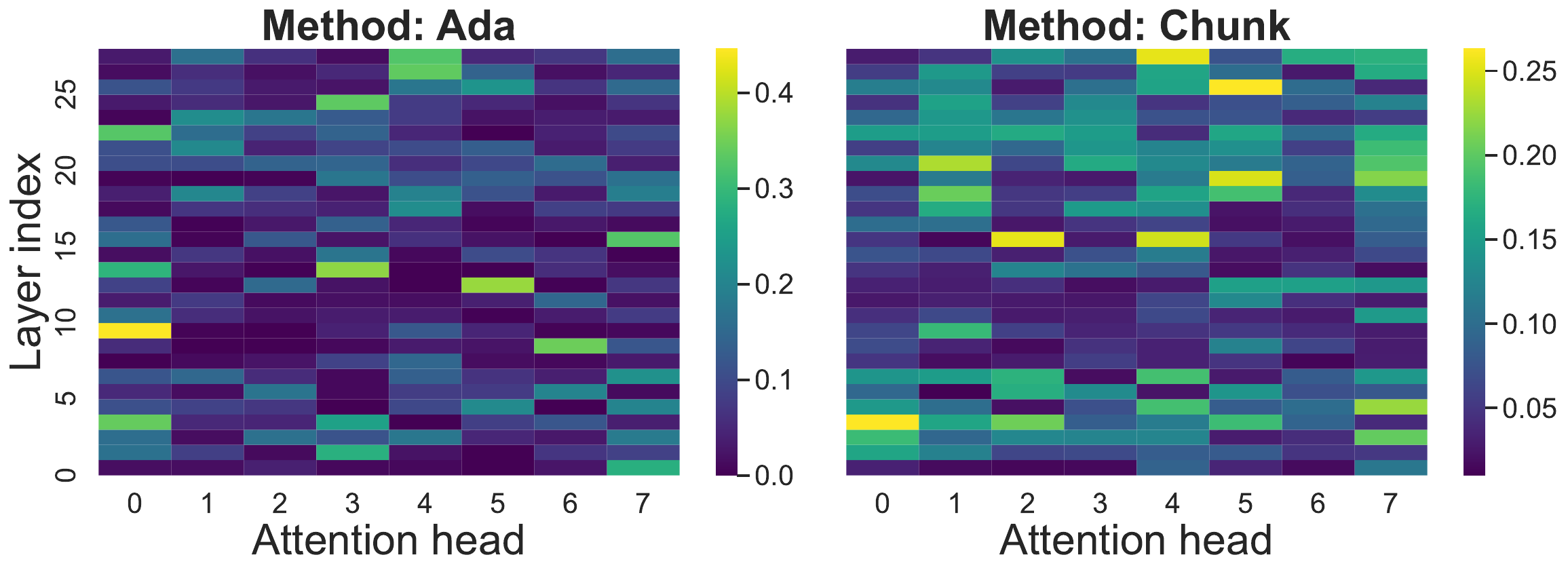}
    }\\
    \subfloat[50\% Compression]{\includegraphics[width=\textwidth]{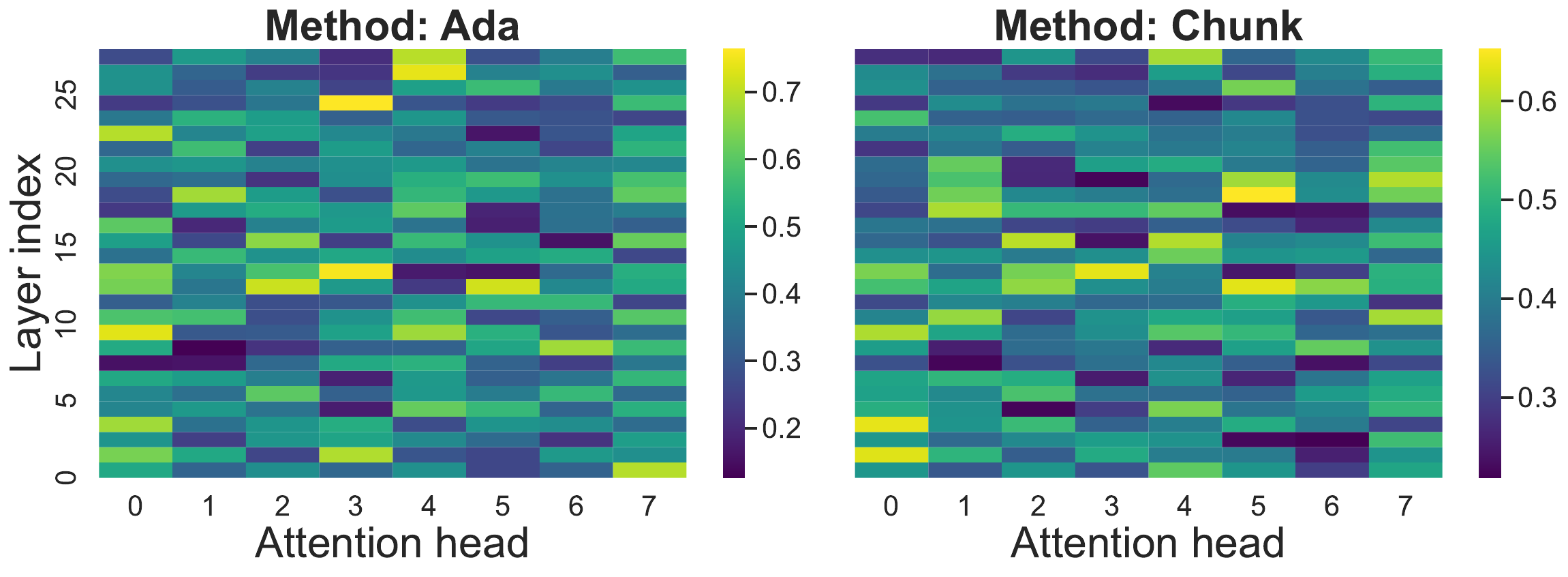}
    }\\
    \subfloat[90\% Compression]{
    \includegraphics[width=\textwidth]{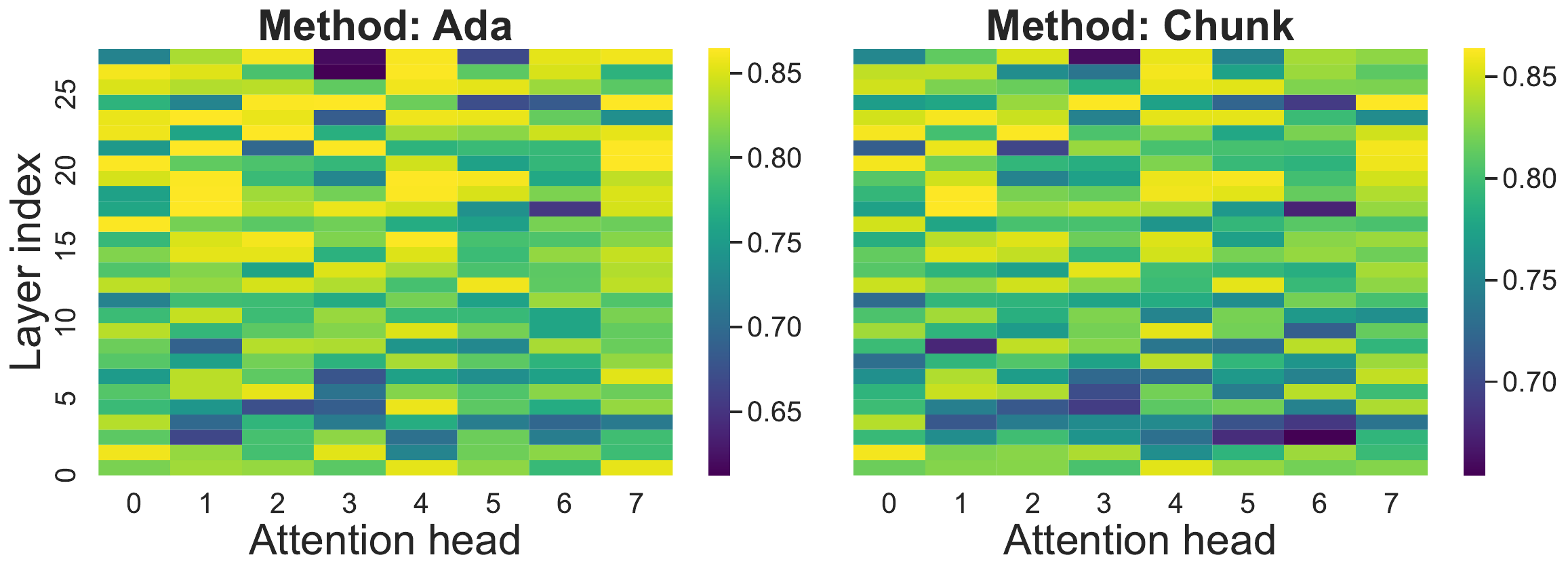}
    }\\
    \caption{Similar smooth pruning in Chunk and irregular removal in Ada is seen in LLaMA 3.2 3B as well.}
    \label{fig:LLaMA3BAttnHeatmapsApp}
\end{figure*}

\begin{figure*}[!htb]
    \centering
    \subfloat[Compression 10\%]{
    \includegraphics[width=\textwidth]{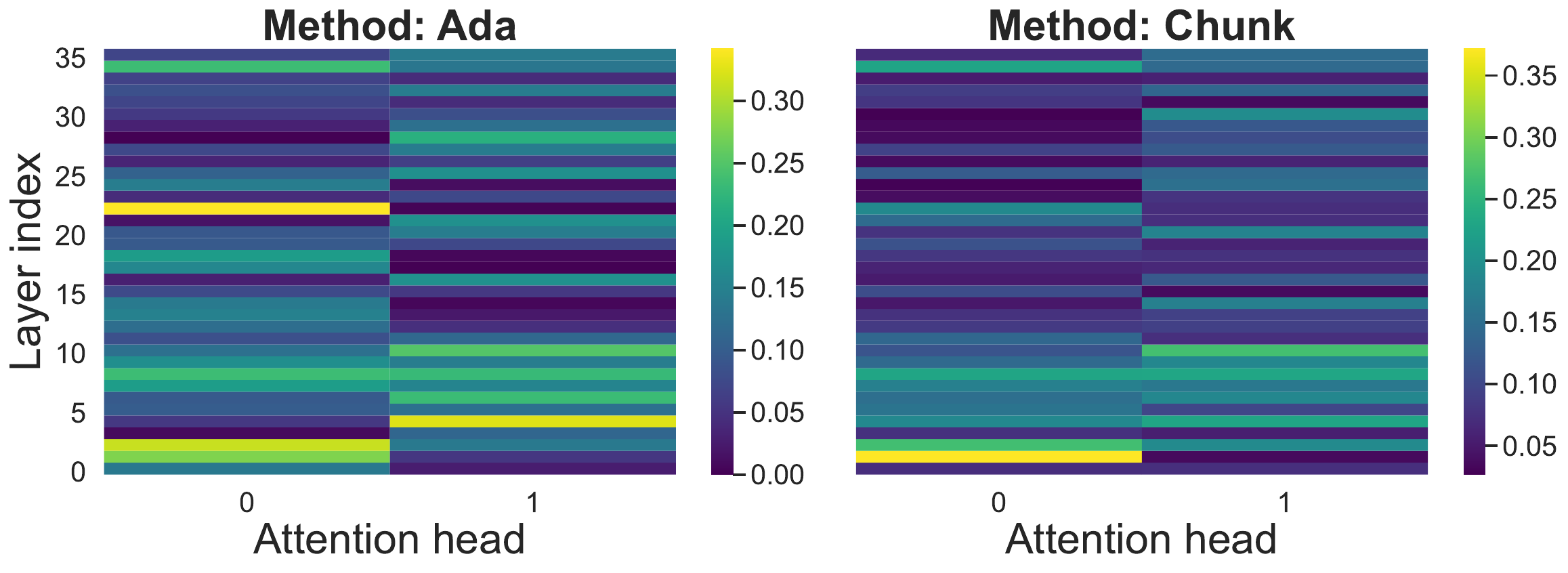}
    }\\
    \subfloat[Compression 50\%]{
    \includegraphics[width=\textwidth]{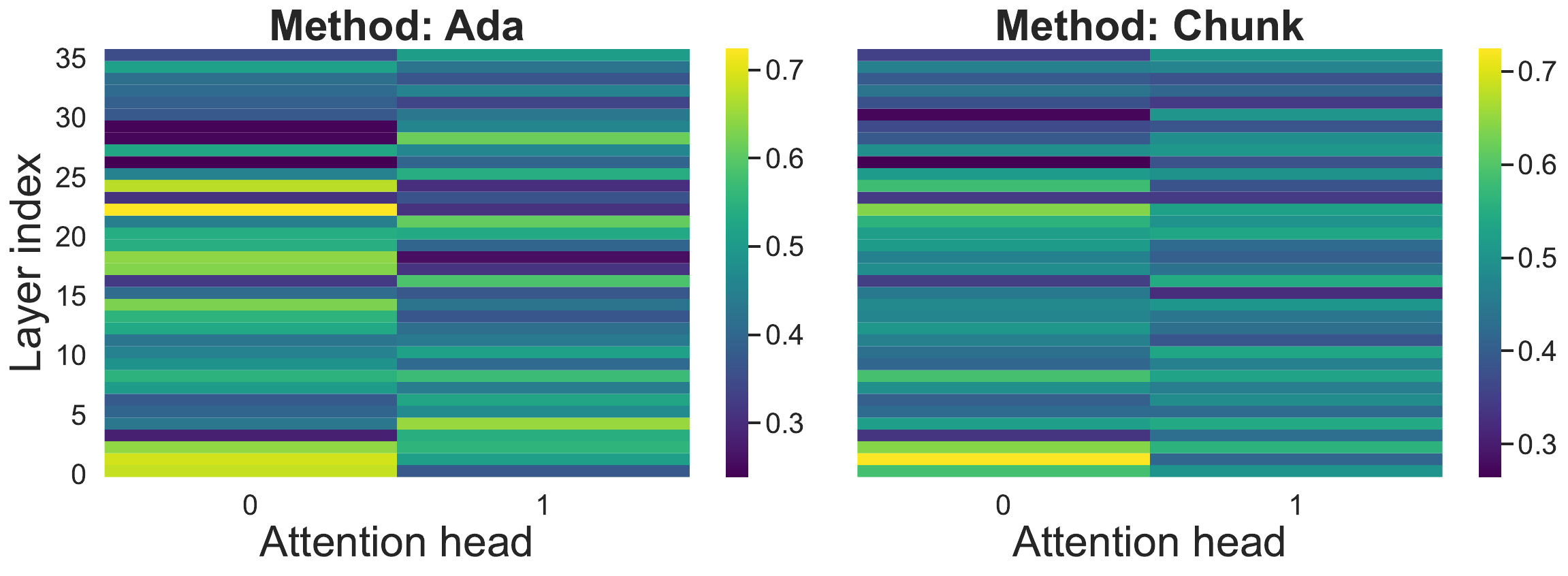}
    }\\
    \subfloat[Compression 90\%]{
    \includegraphics[width=\textwidth]{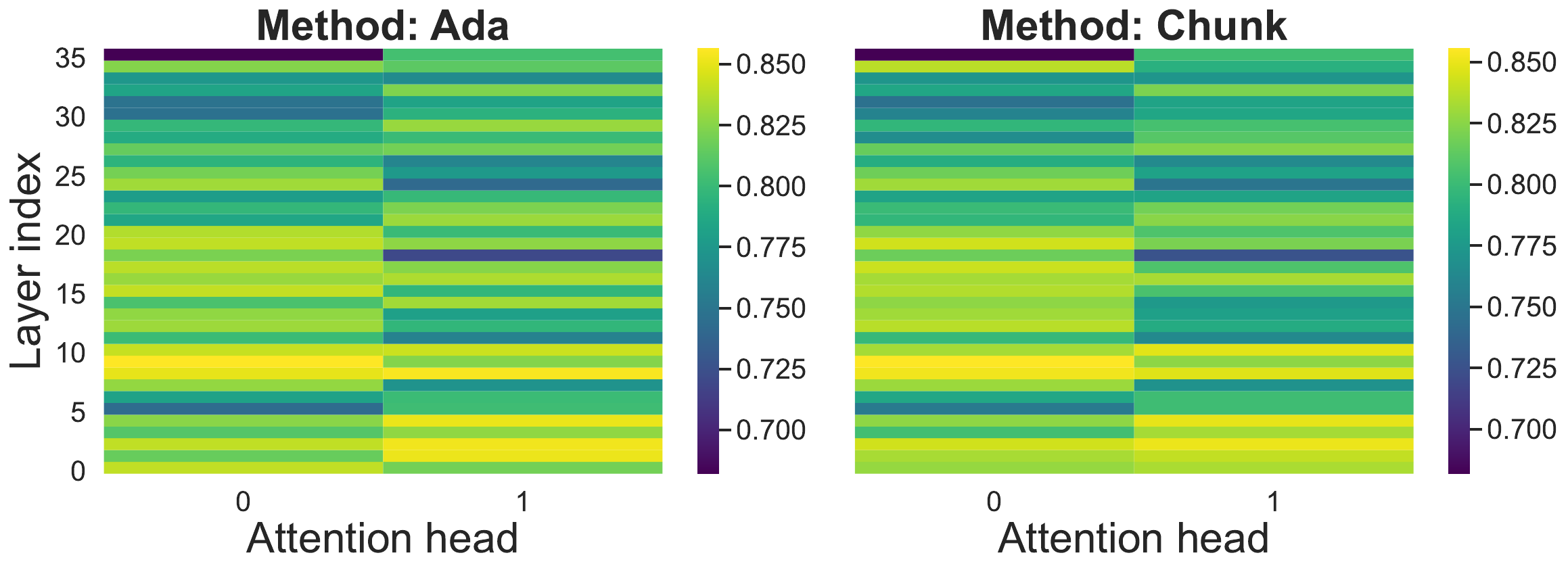}
    }
    \caption{The non-contiguous nature of pruning is far less diminished in the Qwen 2.5 3B model due to its use of only two heads across all layers.}
    \label{fig:Qwen3BAttnHeatmapsApp}
\end{figure*}

\begin{figure*}[!htb]
    \centering
    \subfloat[Compression 10\%]{
    \includegraphics[width=\textwidth]{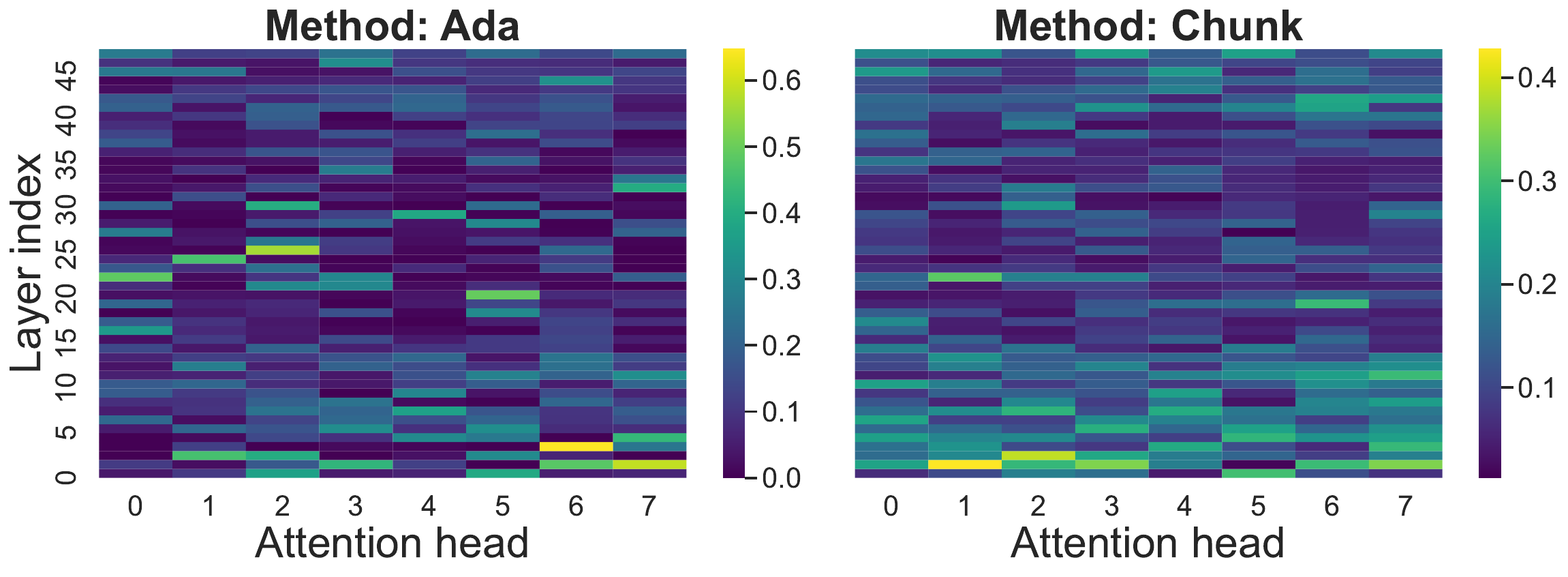}
    }\\
    \subfloat[Compression 50\%]{
    \includegraphics[width=\textwidth]{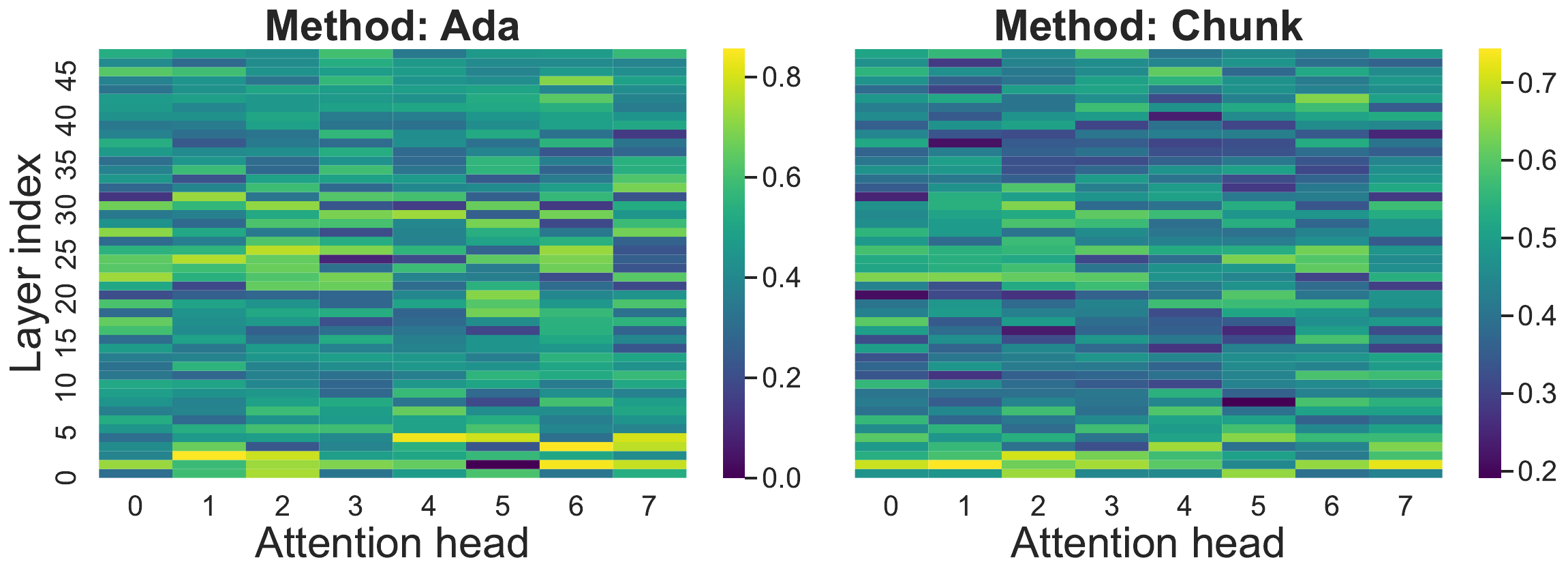}
    }\\
    \subfloat[Compression 90\%]{
    \includegraphics[width=\textwidth]{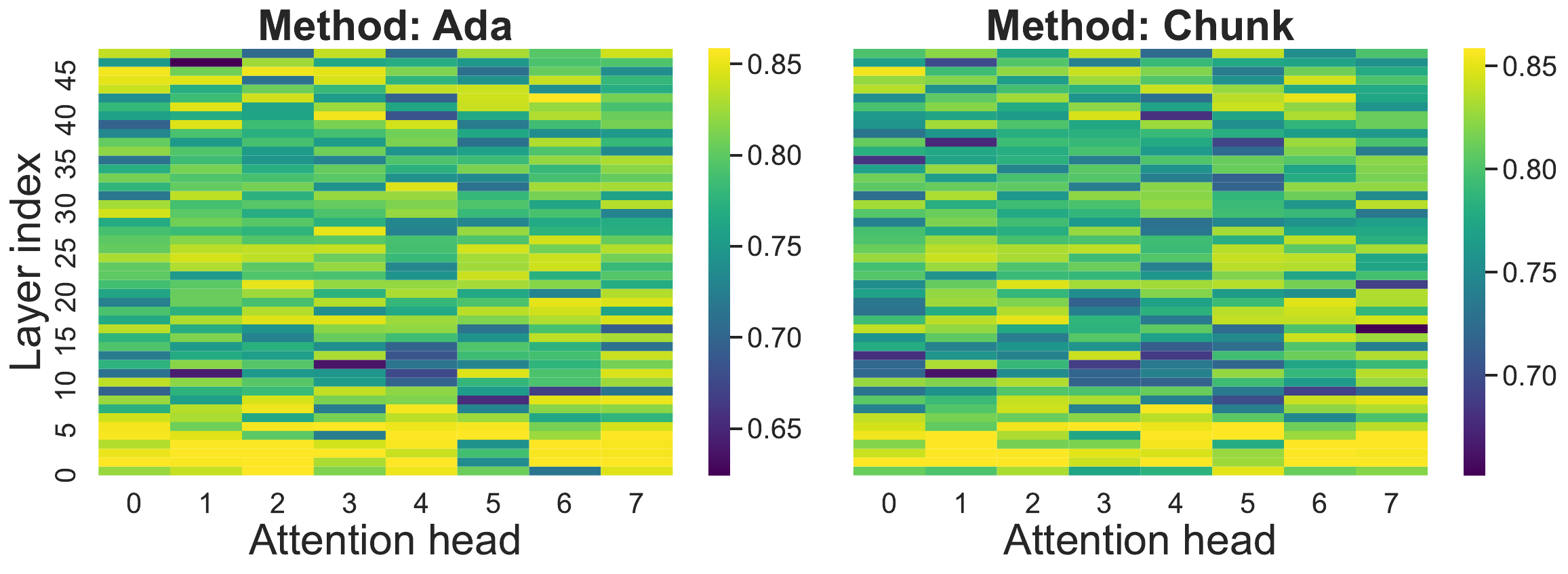}
    }
    \caption{The use of eight heads in Qwen 2.5 14B shows similar trends to the LLaMA 3 family of models, with eviction patterns being fairly similar in their nature.}
    \label{fig:Qwen14BAttnHeatmapsApp}
\end{figure*}

\begin{figure*}[!htb]
    \centering
    \subfloat[LLaMA-3.2 3B Instruct]{
    \includegraphics[width=0.38\linewidth]{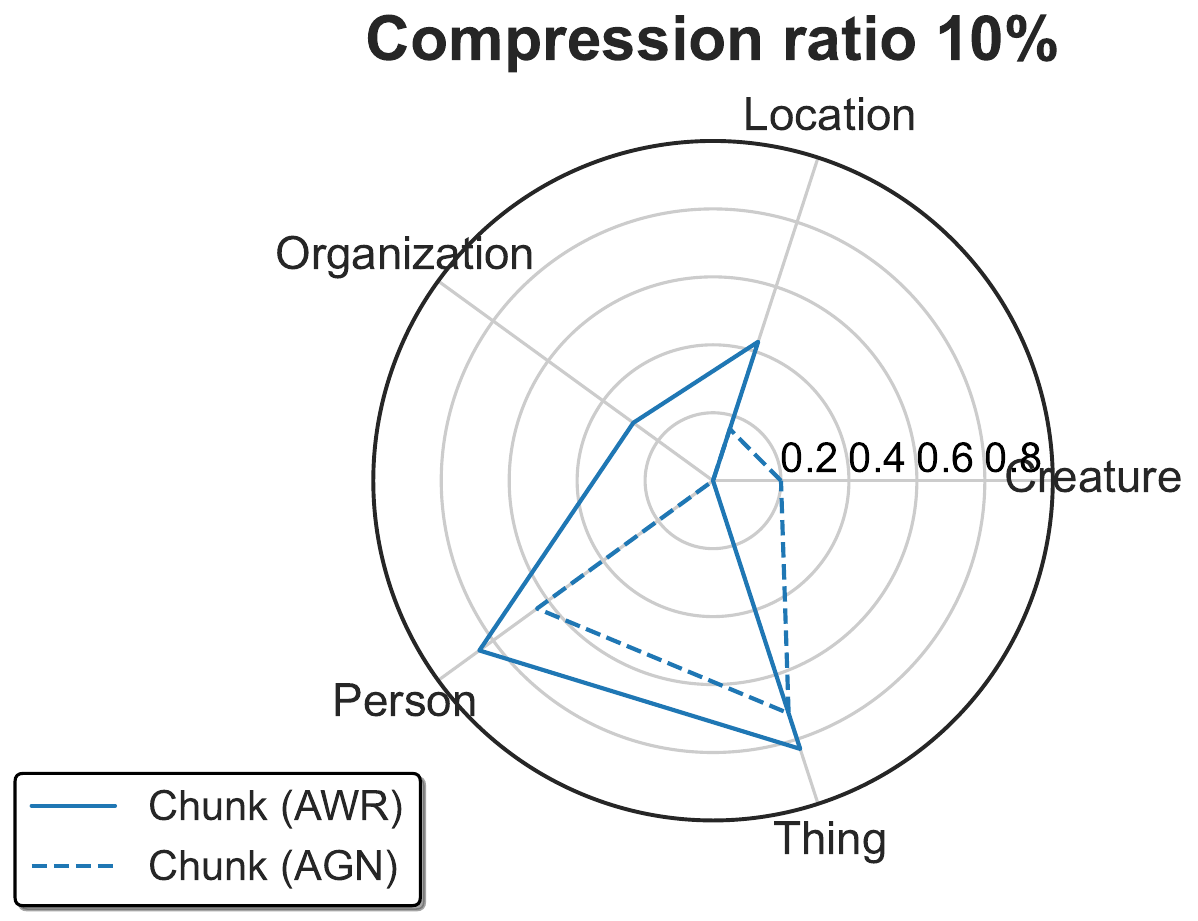}

    \includegraphics[width=0.31\linewidth]{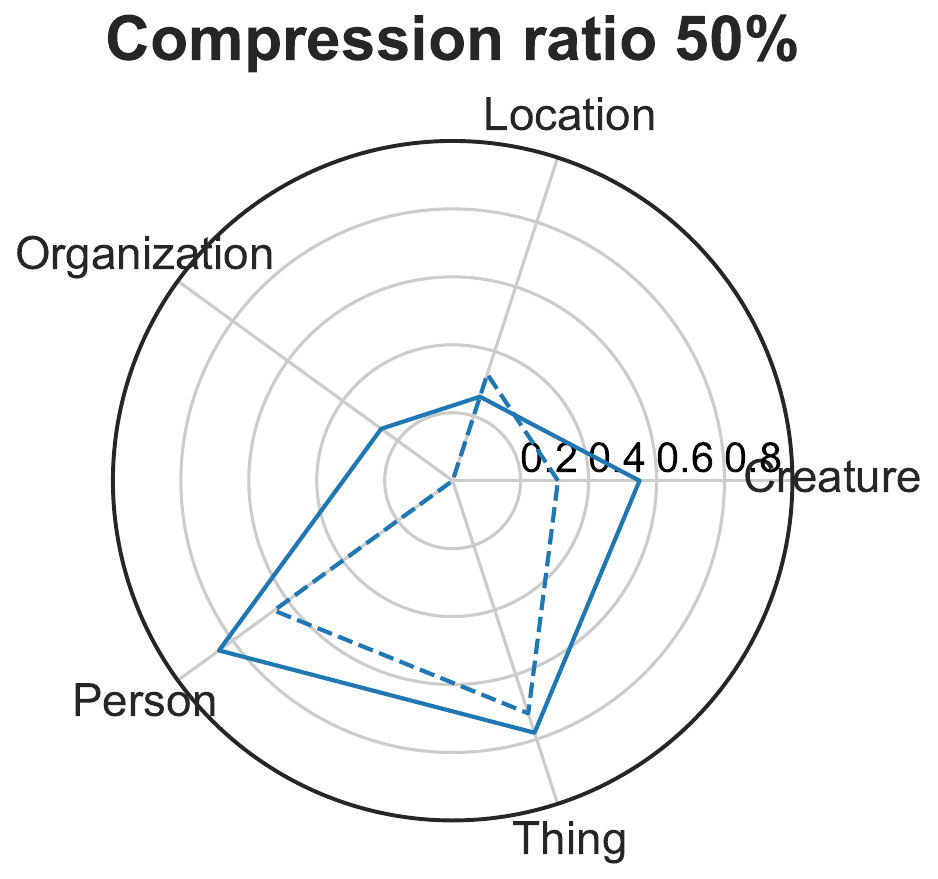}

    \includegraphics[width=0.31\linewidth]{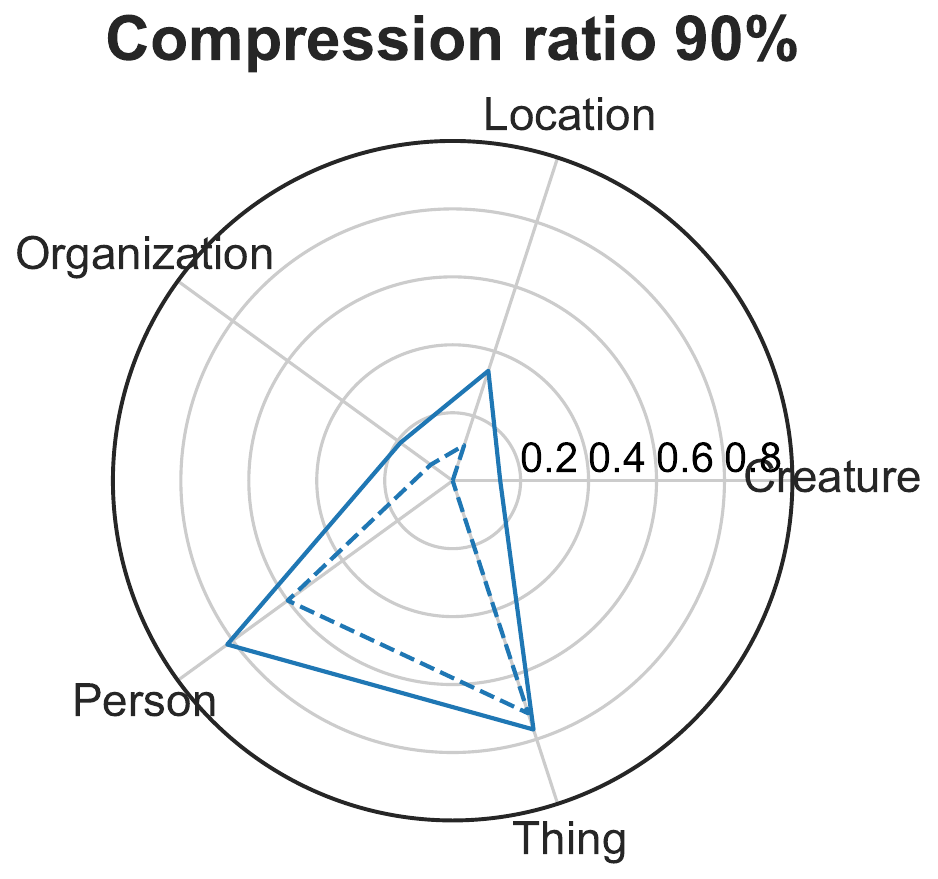}
    }\\
    \subfloat[Qwen-2.5 3B Instruct]{
    \includegraphics[width=0.38\linewidth]{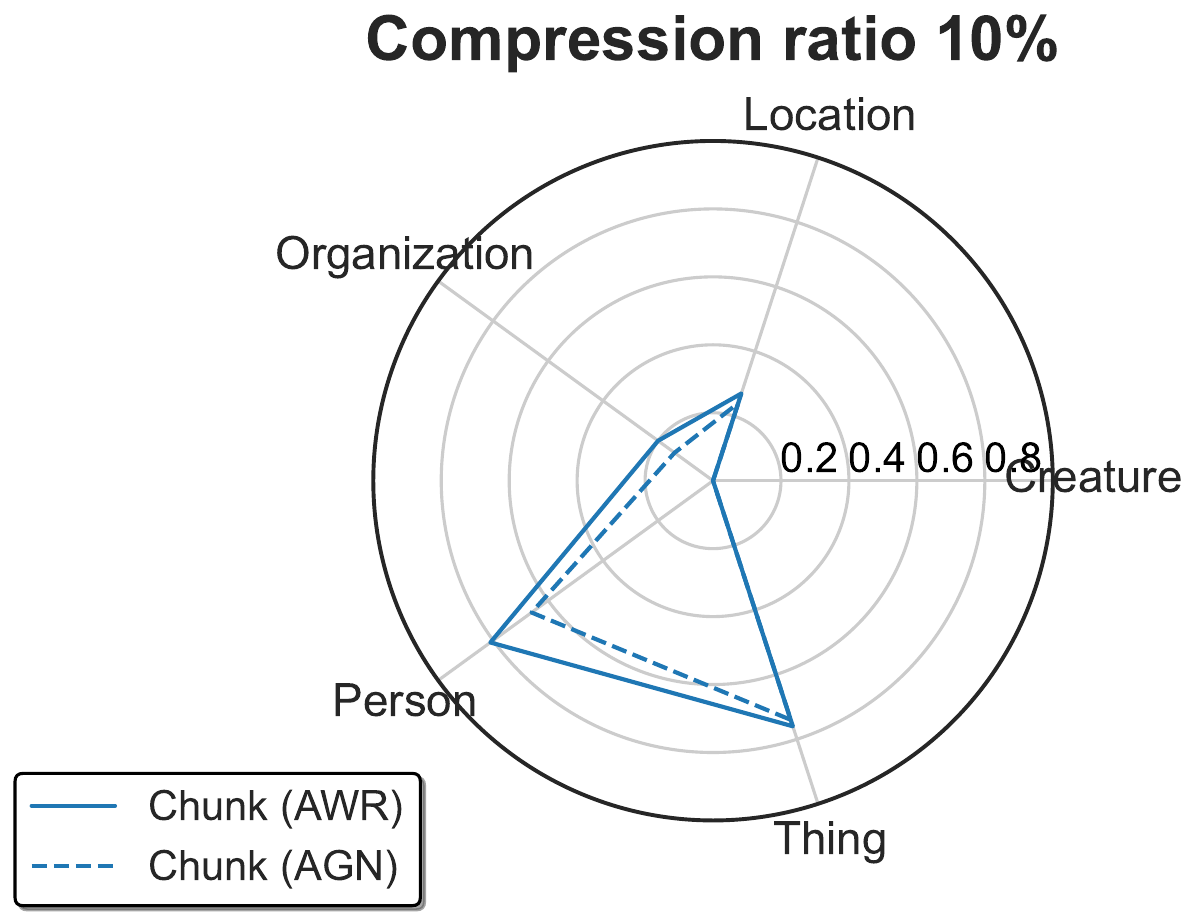}
    \includegraphics[width=0.31\linewidth]{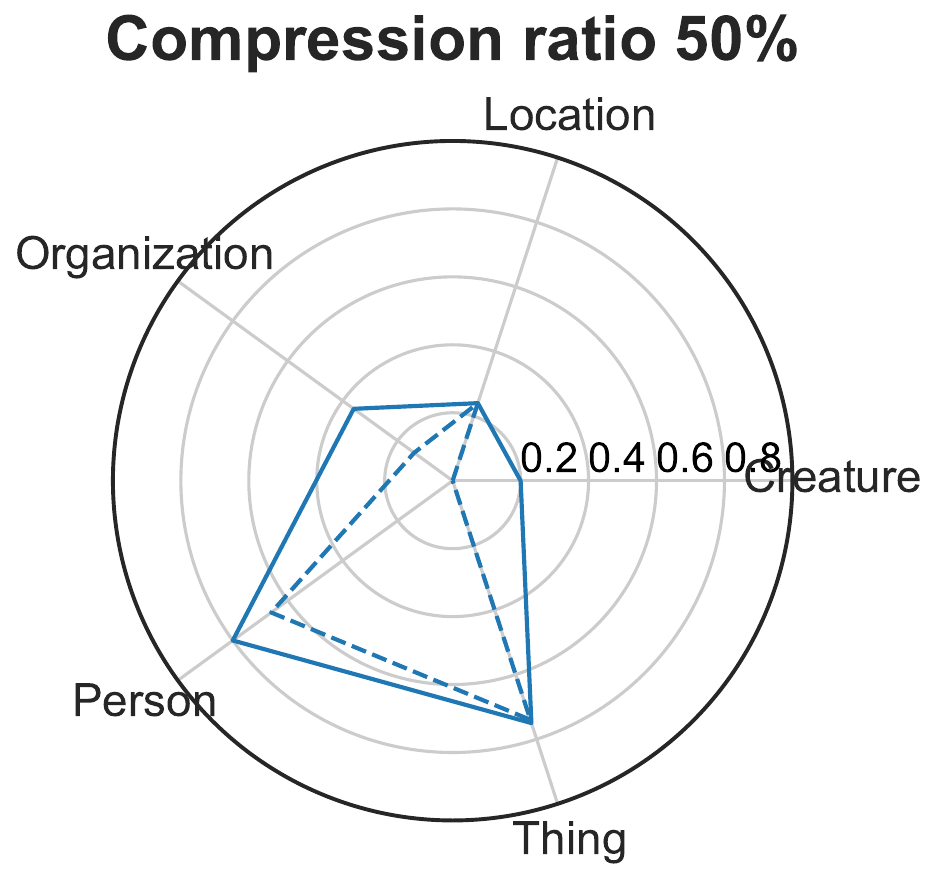}
    \includegraphics[width=0.31\linewidth]{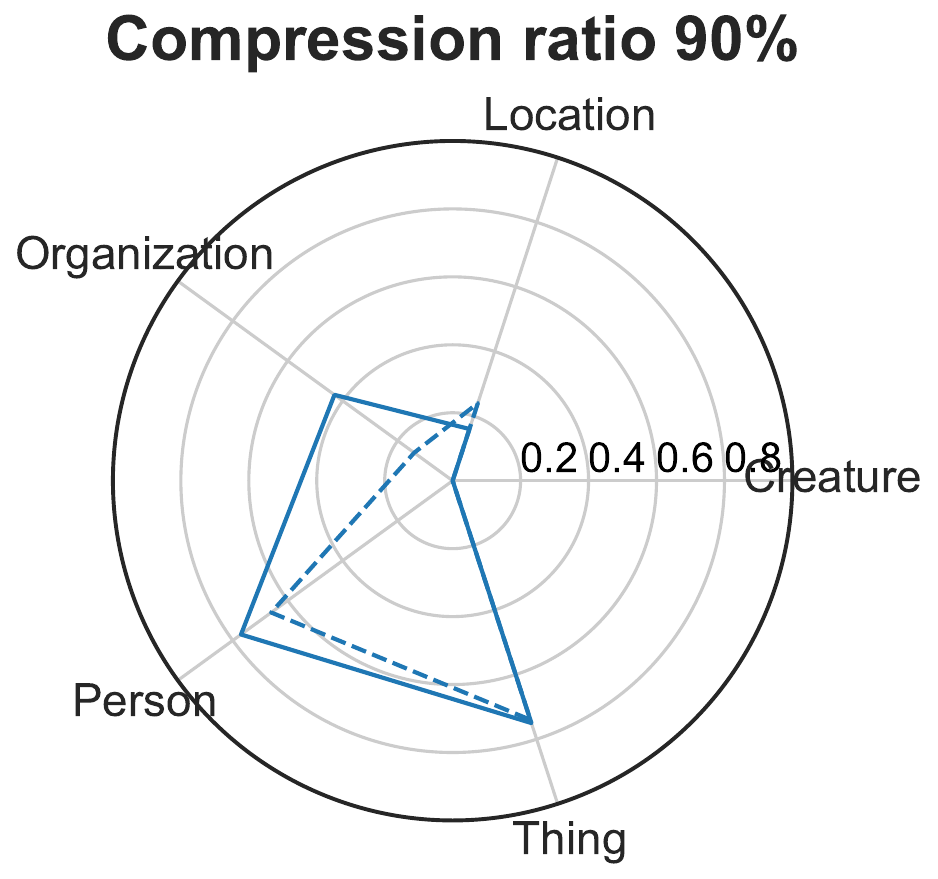}
    }\\
    \subfloat[Qwen-2.5 14B Instruct]{
    \includegraphics[width=0.38\linewidth]{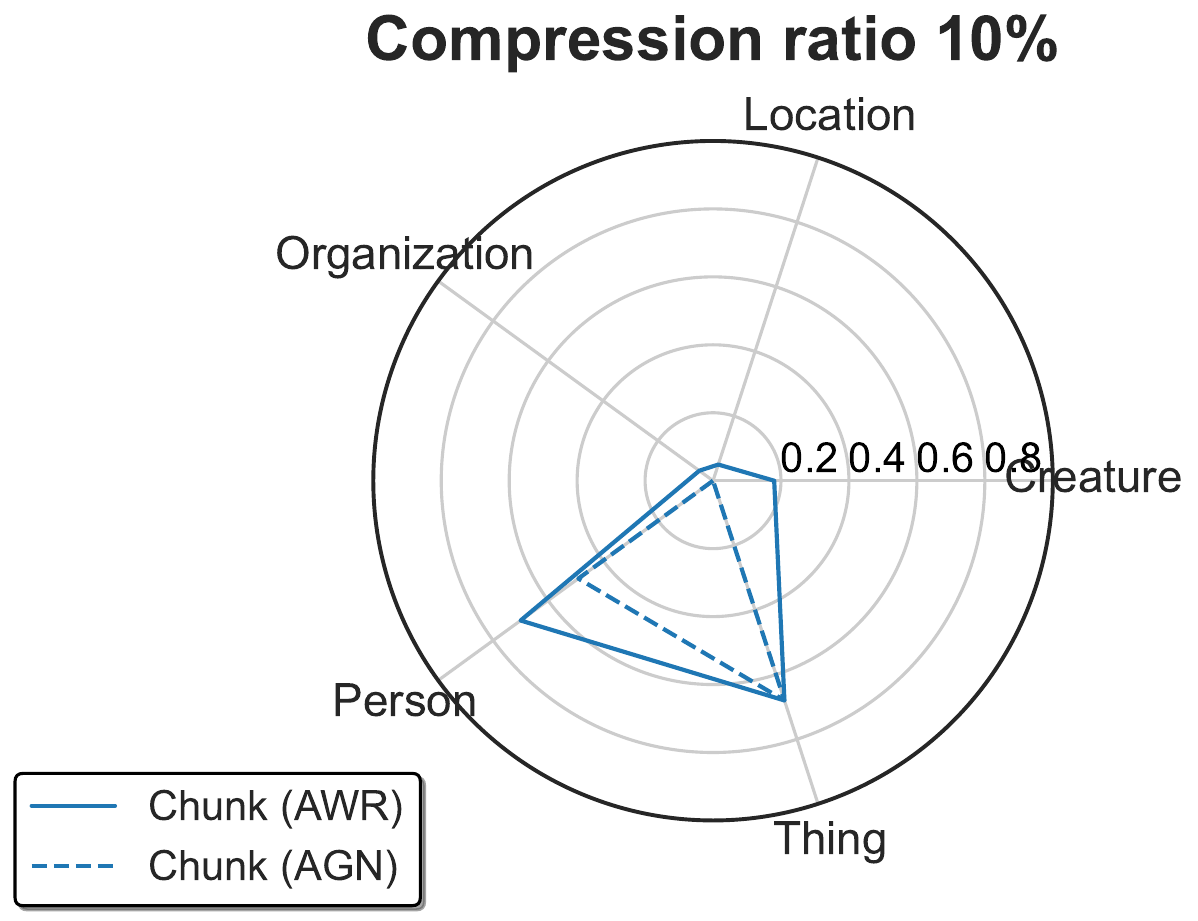}
    \includegraphics[width=0.31\linewidth]{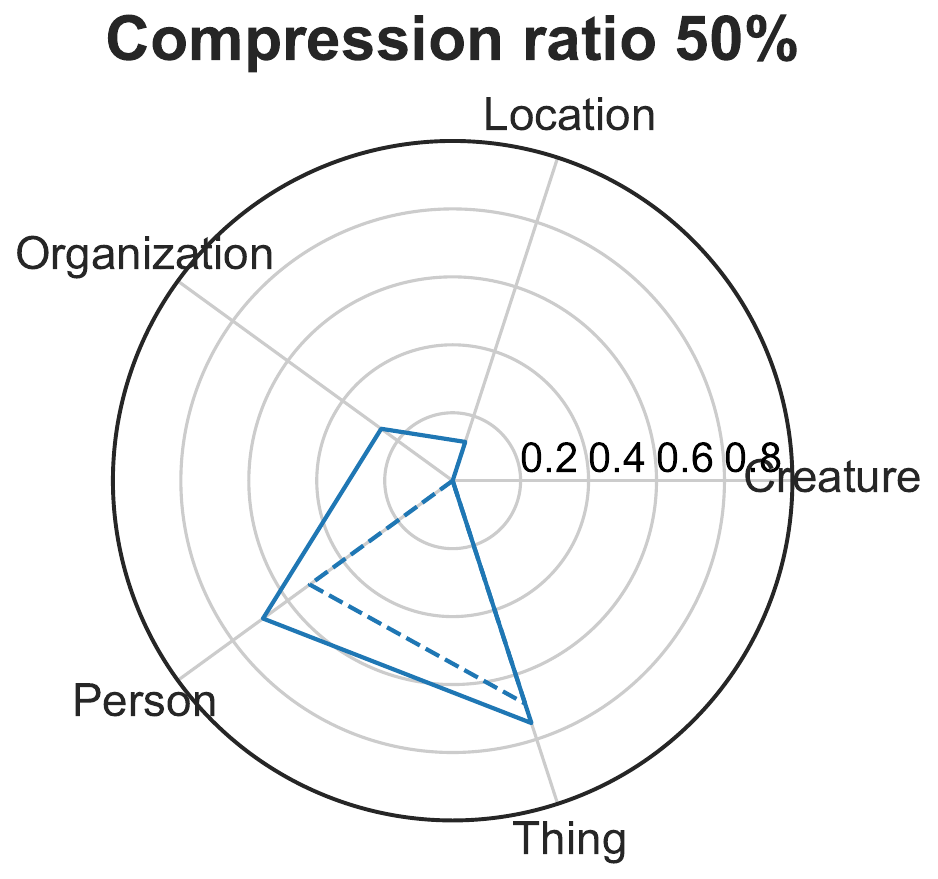}
    \includegraphics[width=0.31\linewidth]{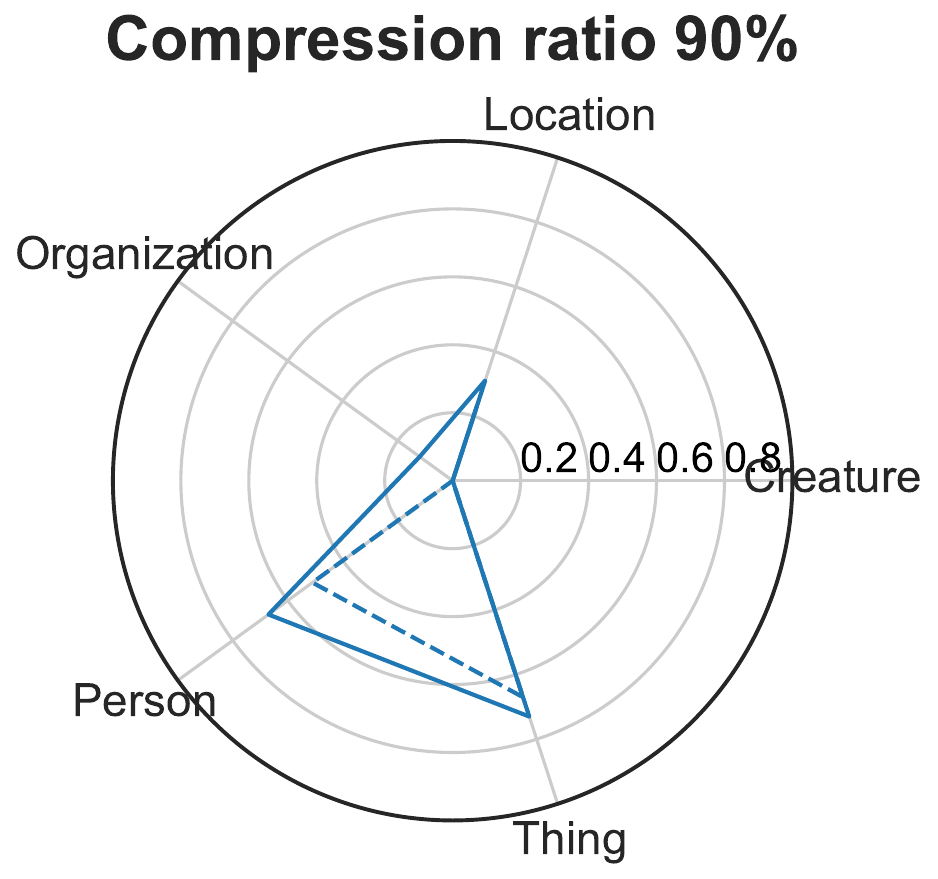}
    }
    \caption{Probing results still show drastic inconsistency as compression as the organization and creature tags still suffer.}
    \label{fig:BaseTaskProbingRadarApp}
\end{figure*}
\begin{figure*}[!htb]
    \centering
     \subfloat[LLaMA-3.2 3B Instruct]{
    \includegraphics[width=0.38\linewidth]{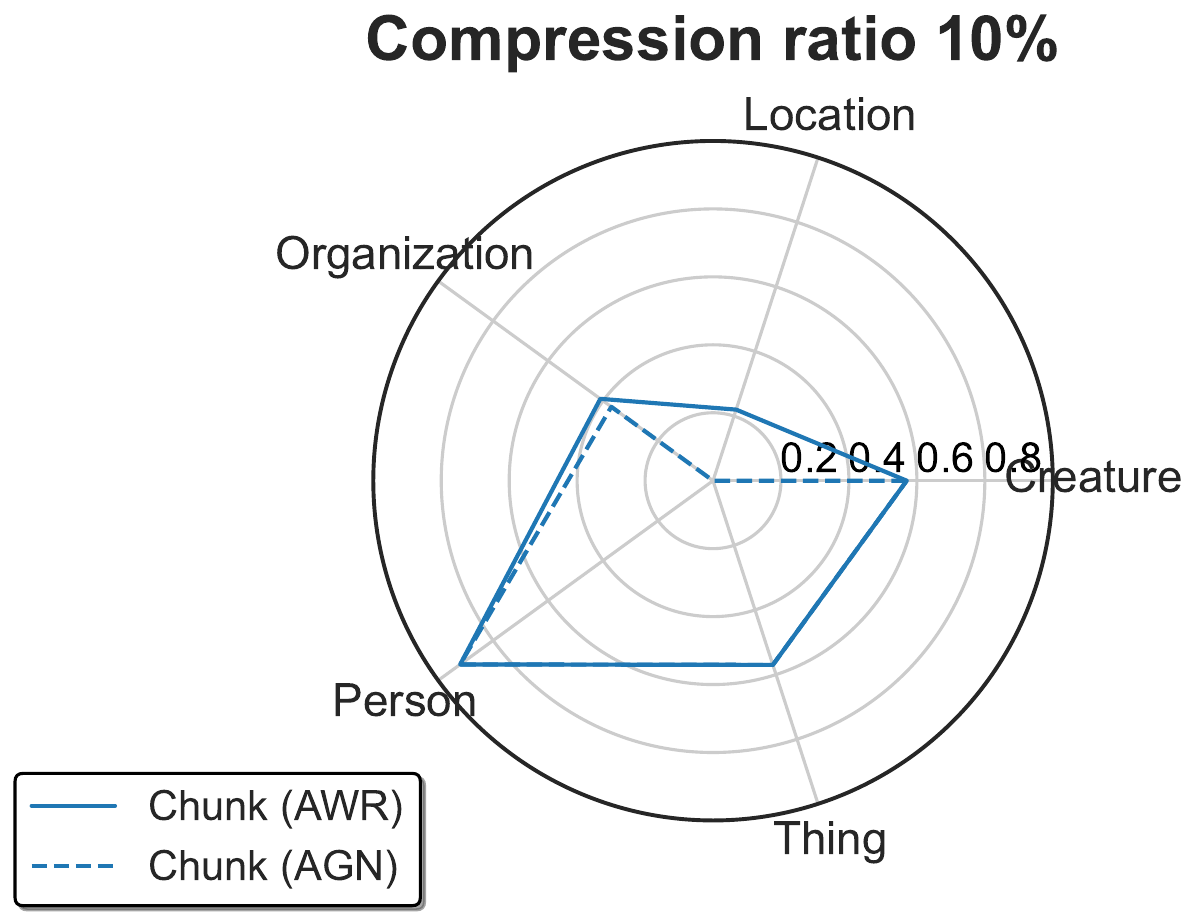}

    \includegraphics[width=0.31\linewidth]{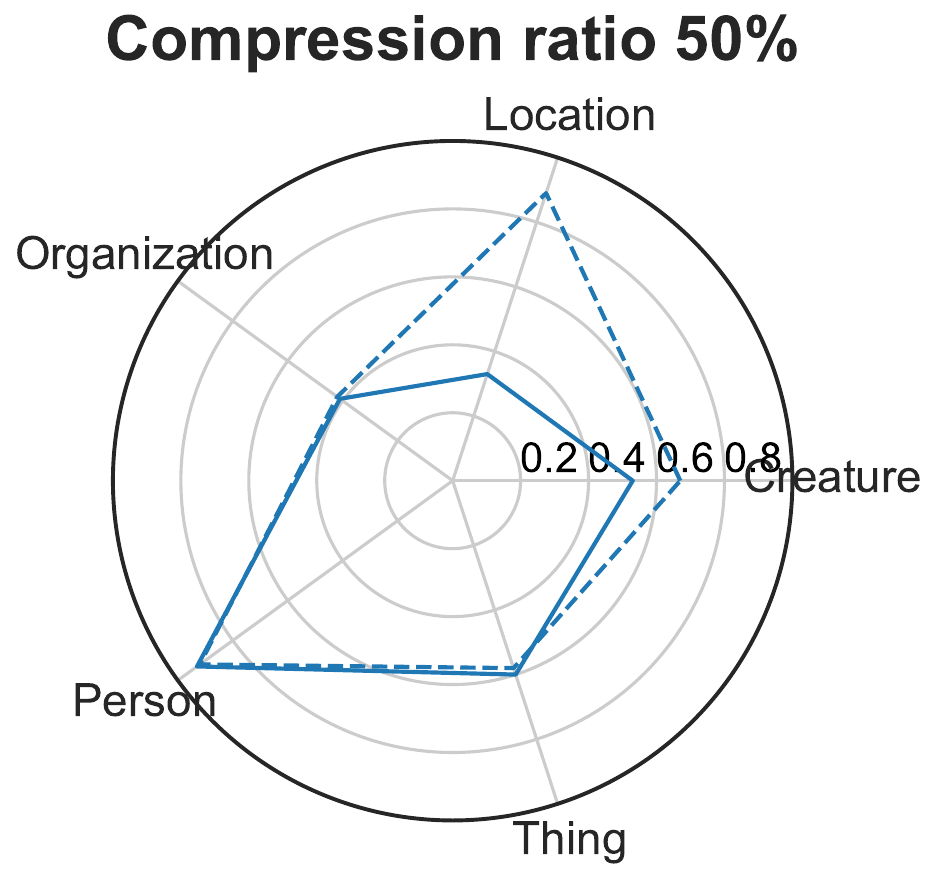}

    \includegraphics[width=0.31\linewidth]{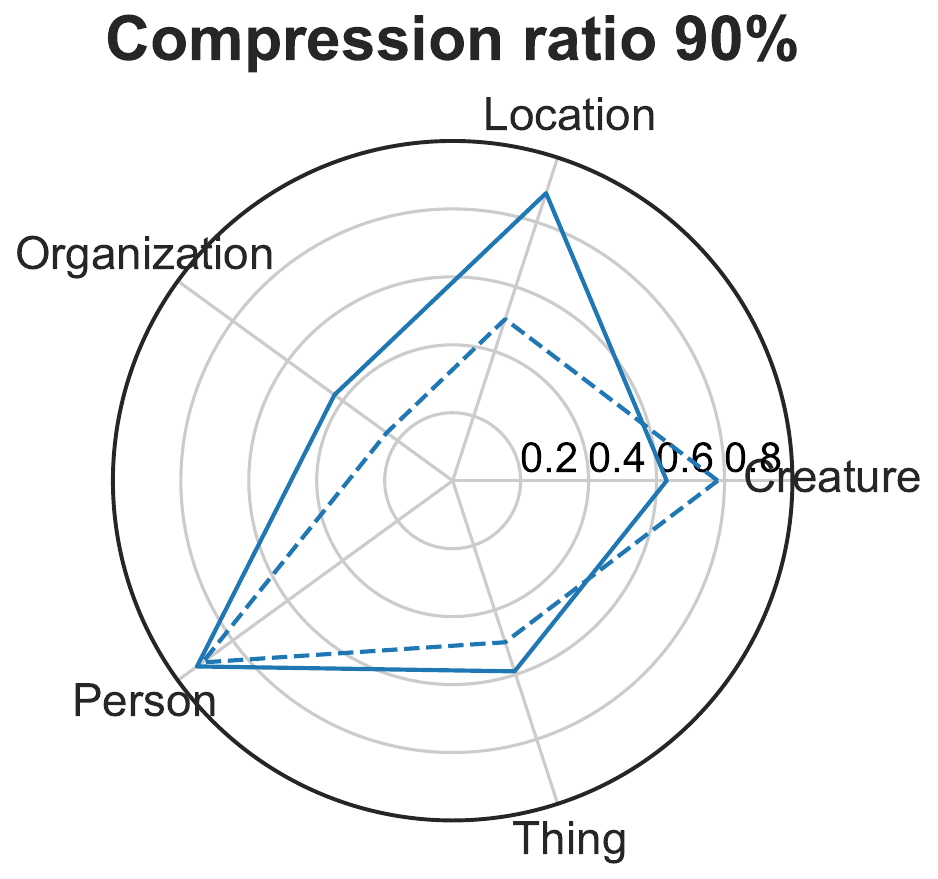}
    }\\
    \subfloat[Qwen-2.5 3B Instruct]{
    \includegraphics[width=0.38\linewidth]{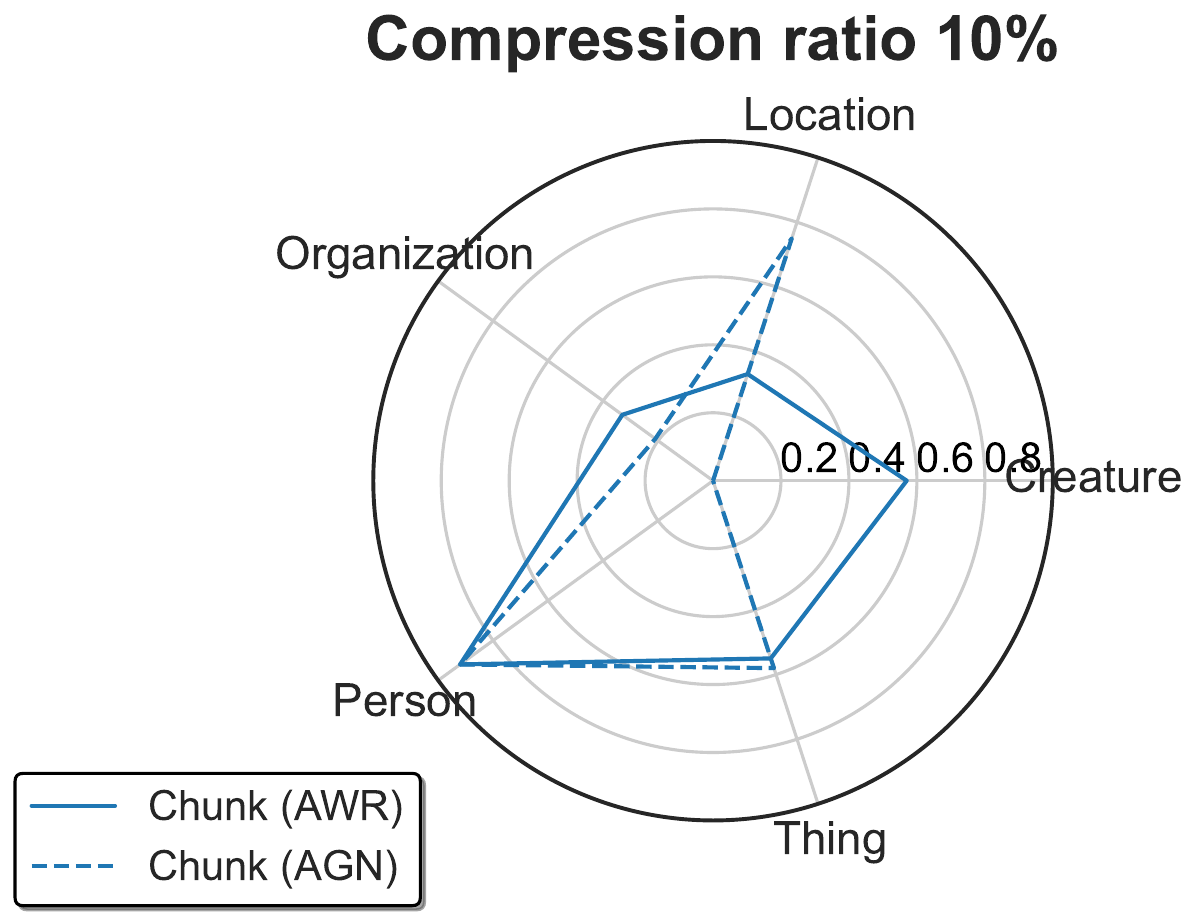}

    \includegraphics[width=0.31\linewidth]{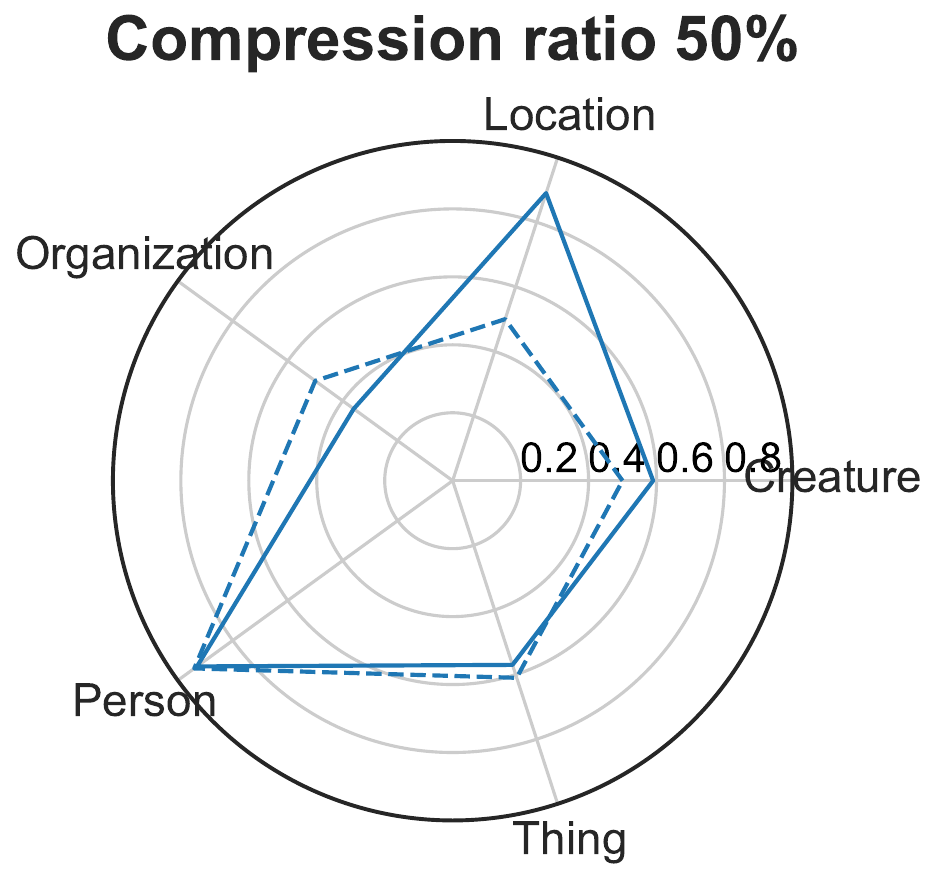}
    \includegraphics[width=0.31\linewidth]{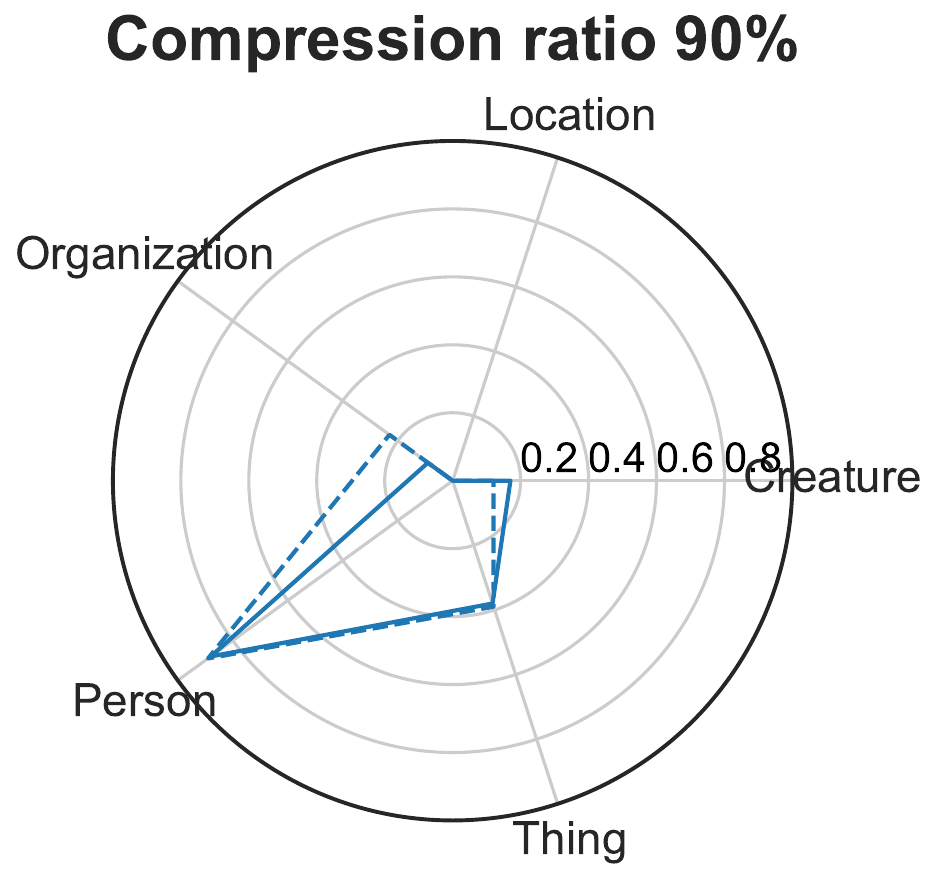}
    }\\
    \subfloat[Qwen-2.5 14B Instruct]{
    \includegraphics[width=0.38\linewidth]{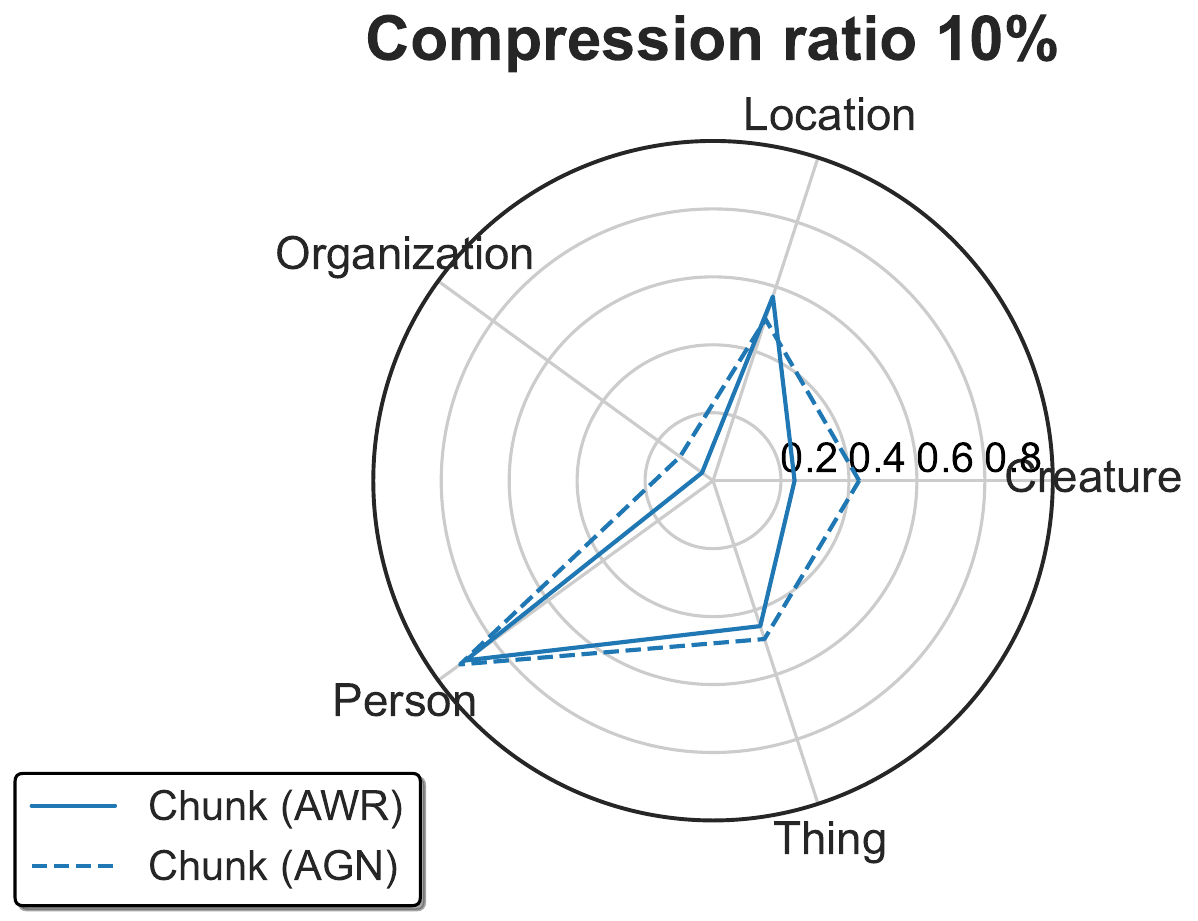}

    \includegraphics[width=0.31\linewidth]{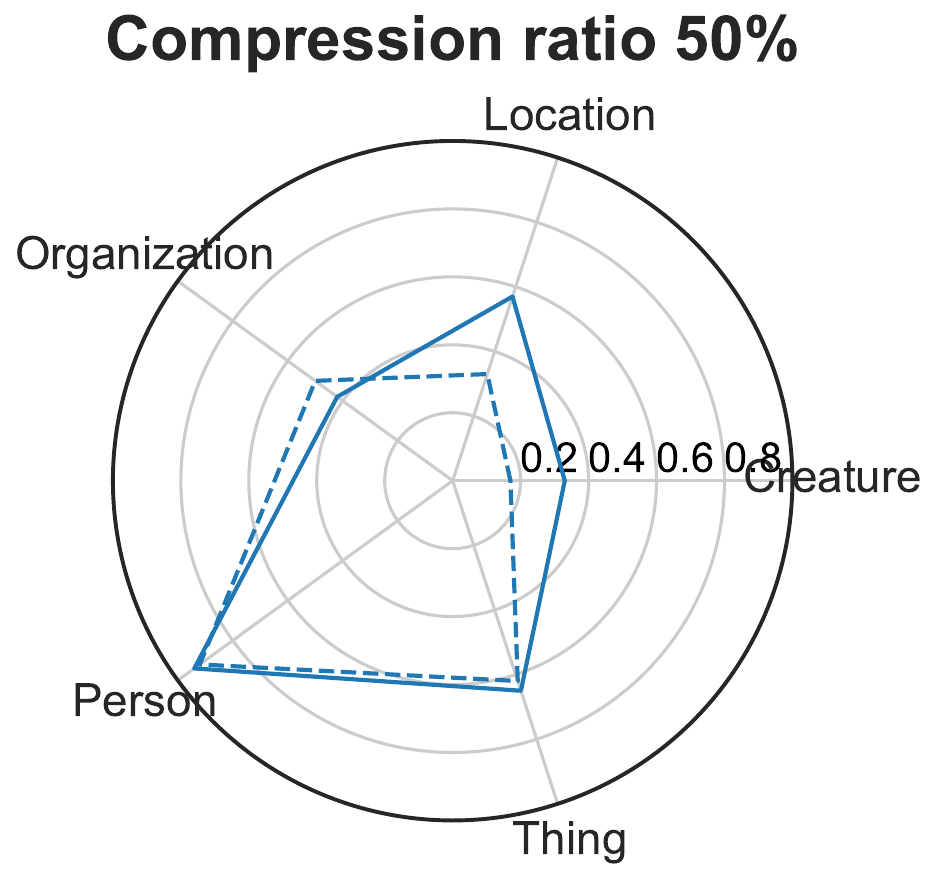}
    \includegraphics[width=0.31\linewidth]{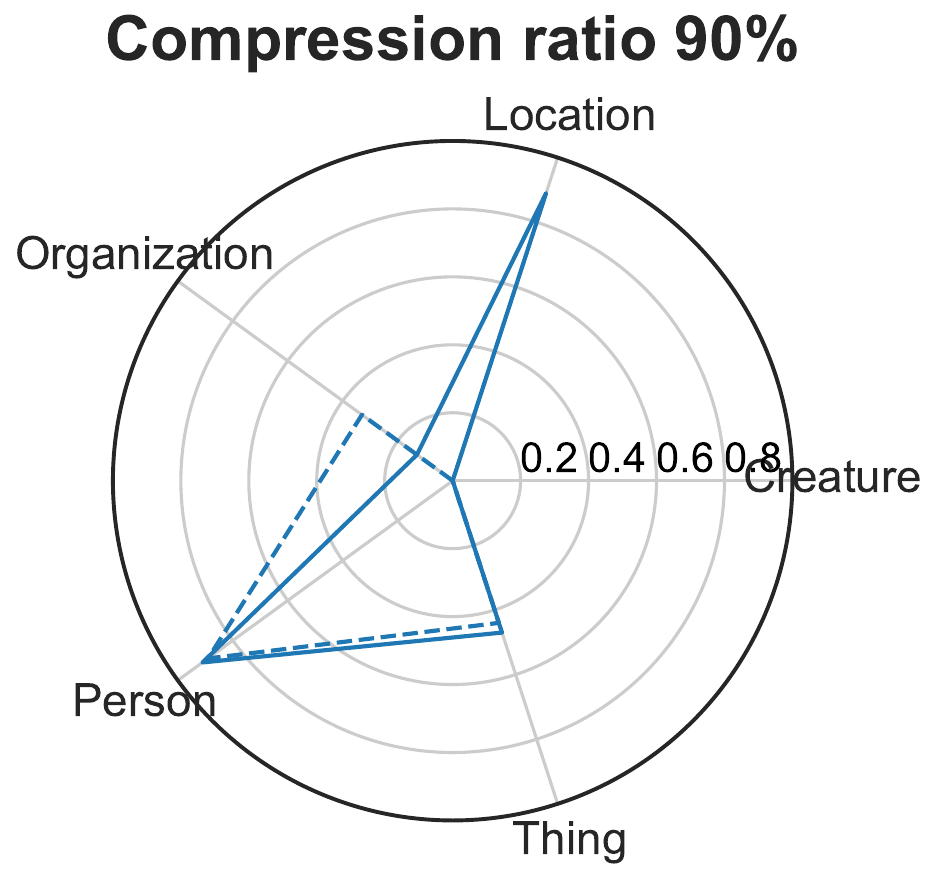}
    }\\
    \caption{Question-aware and agnostic compression still induce fairly different understandings despite the different model sizes and architectures involved.}
    \label{fig:MultiEntityProbingRadarApp}
\end{figure*}

\begin{figure*}[!htb]
    \centering
    \subfloat[LLaMA-3.2 3B Instruct]{
    \includegraphics[width=0.4\linewidth]{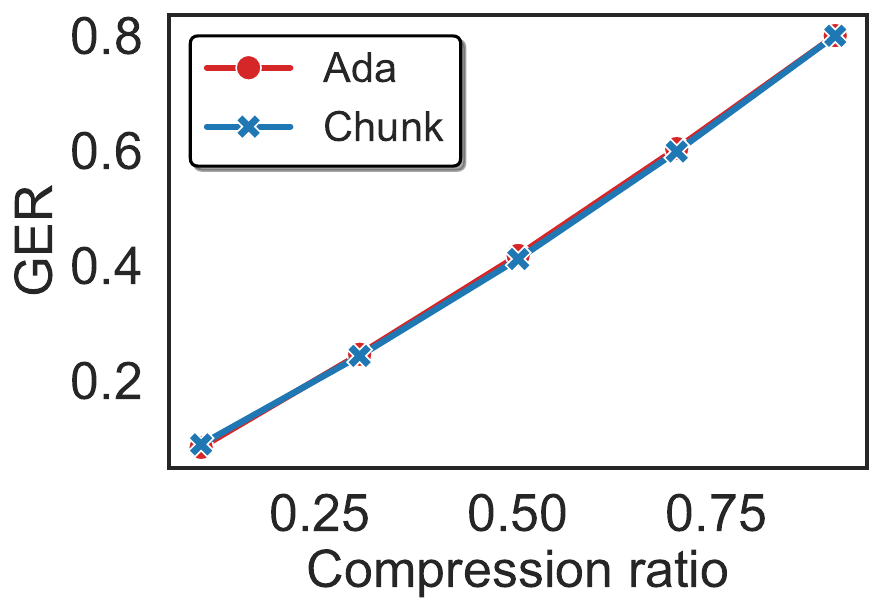}
    }
    \subfloat[Qwen-2.5 3B Instruct]{
    \includegraphics[width=0.4\linewidth]{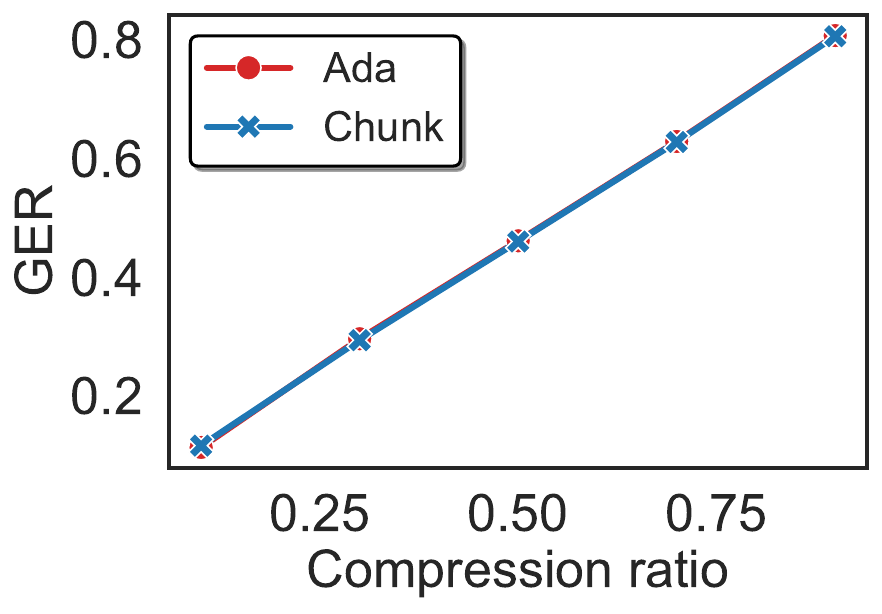}
    }\\
    \subfloat[Qwen-2.5 14B Instruct]{
    \includegraphics[width=0.4\linewidth]{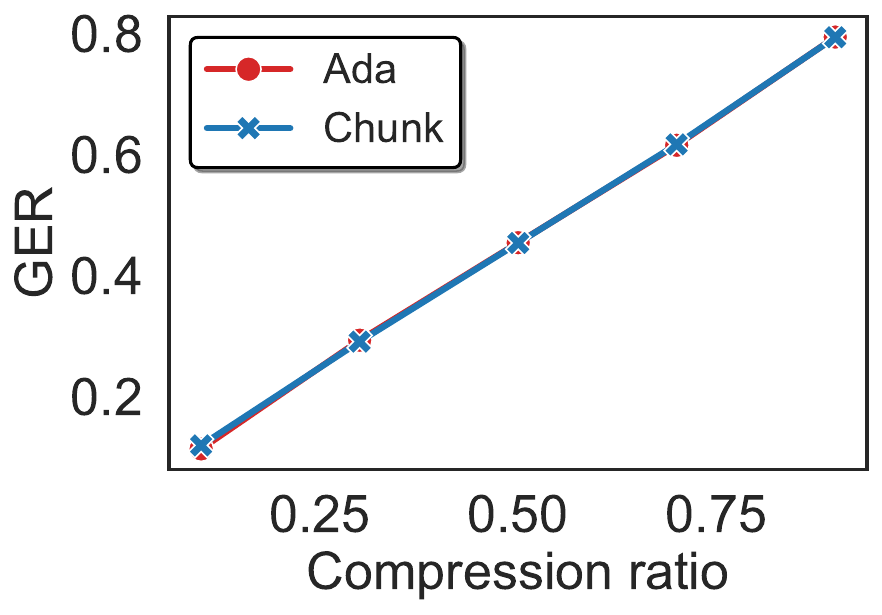}
    }
    \caption{Eviction trends still remain agnostic to downward tasks even when changes to the scale of the model are seen.}
    \label{fig:EvictionRateApp}
\end{figure*}

\begin{figure*}[!htb]
    \centering
    \subfloat[LLaMA-3.2 3B Instruct]{
    \includegraphics[width=0.8\linewidth]{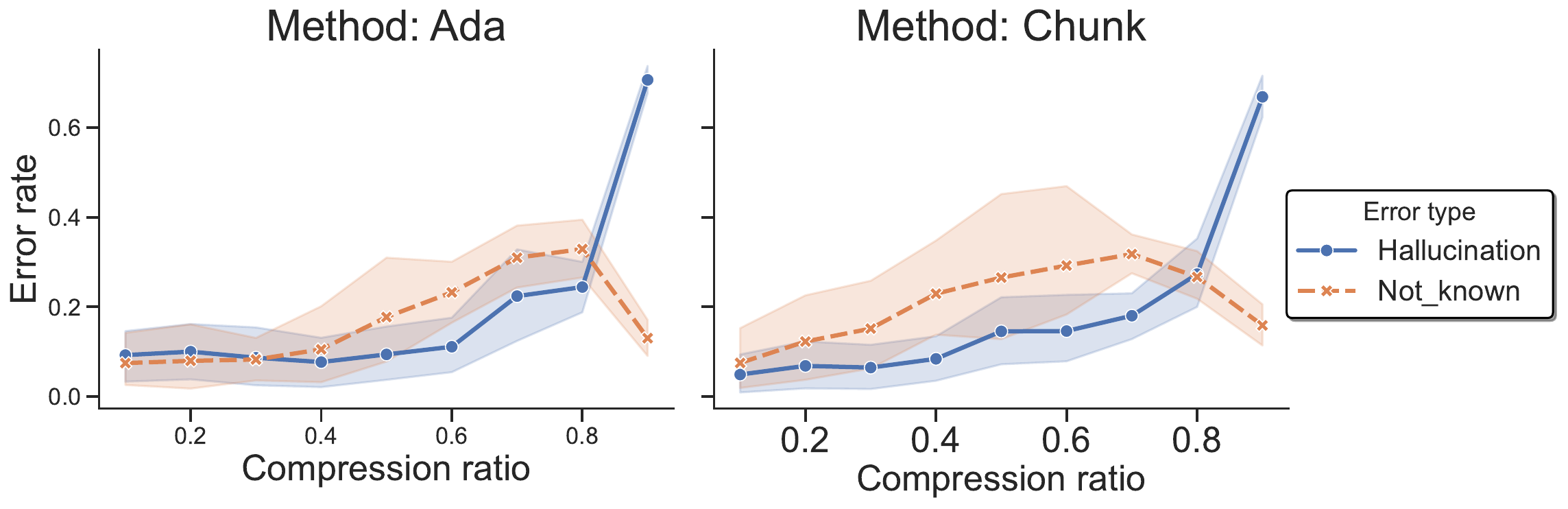}
    }\\
    \subfloat[Qwen-2.5 3B Instruct]{
    \includegraphics[width=0.8\linewidth]{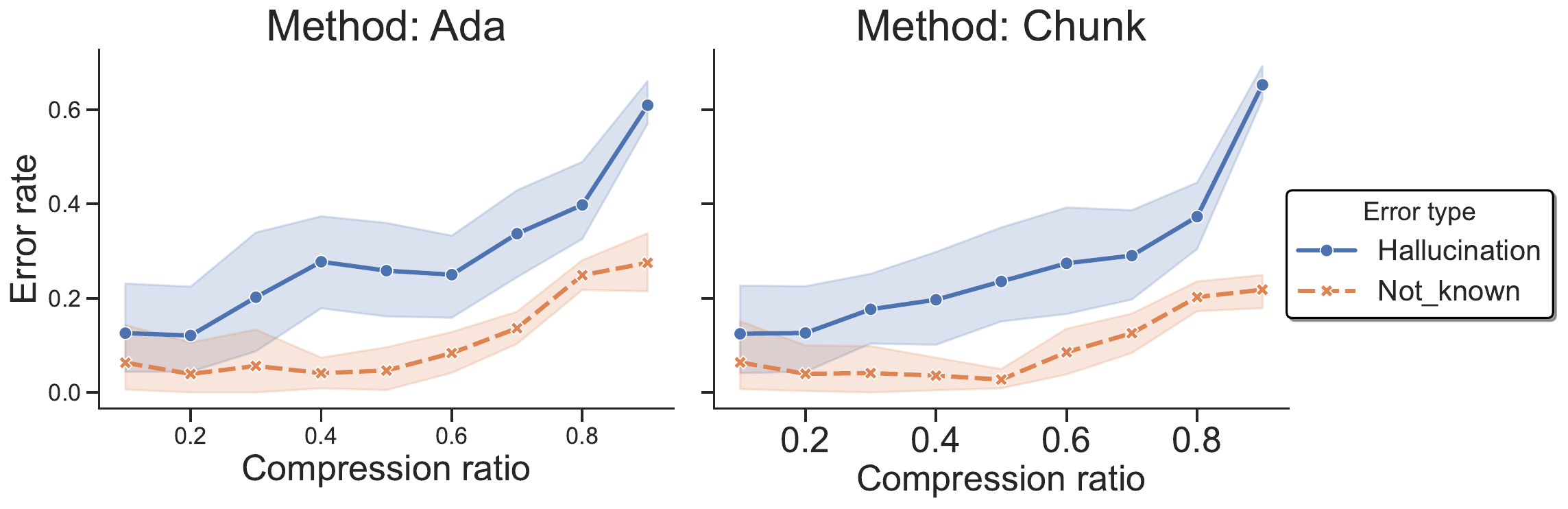}
    }\\
    \subfloat[Qwen-2.5 14B Instruct]{
    \includegraphics[width=0.8\linewidth]{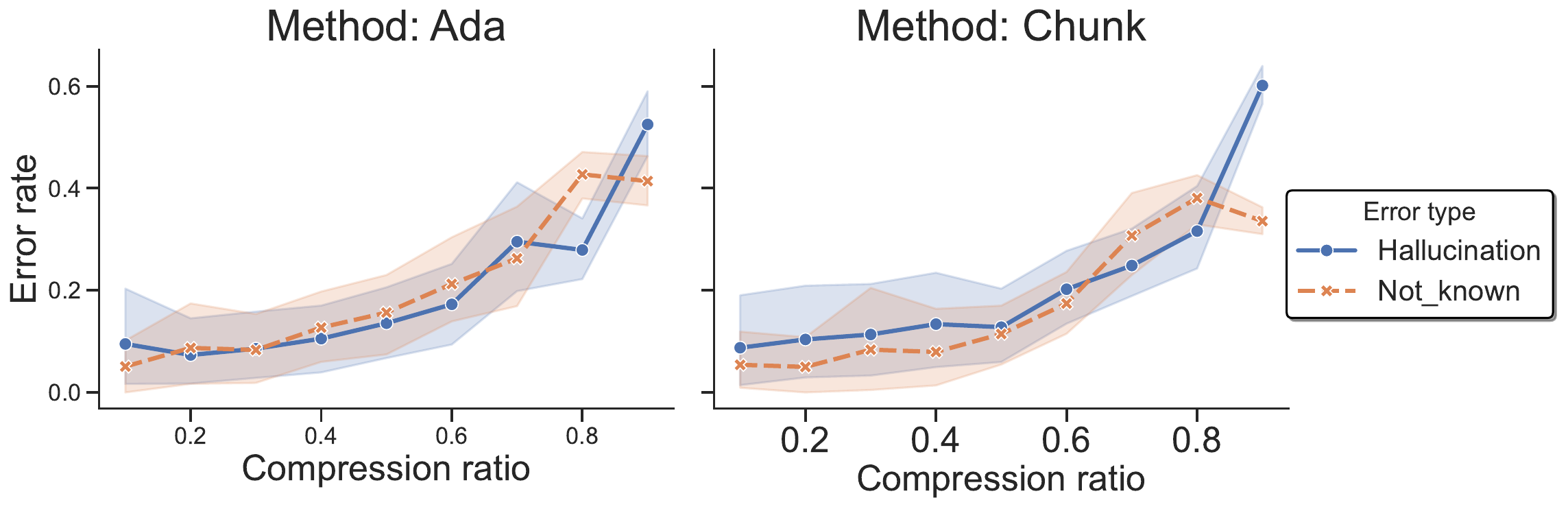}
    }
    \caption{Smaller models (increasing unknown errors in AdaKV persist fragility but retain the similar hallucination cliff as compared to larger models (Qwen 2.5 14B). The trend of increased unknown errors in AdaKV is still present.}
    \label{fig:ErrorRateApp}
\end{figure*}

\begin{figure*}[!htb]
\centering
\subfloat[LLaMA-3.2 3B Instruct]{
    \includegraphics[width=0.8\linewidth]{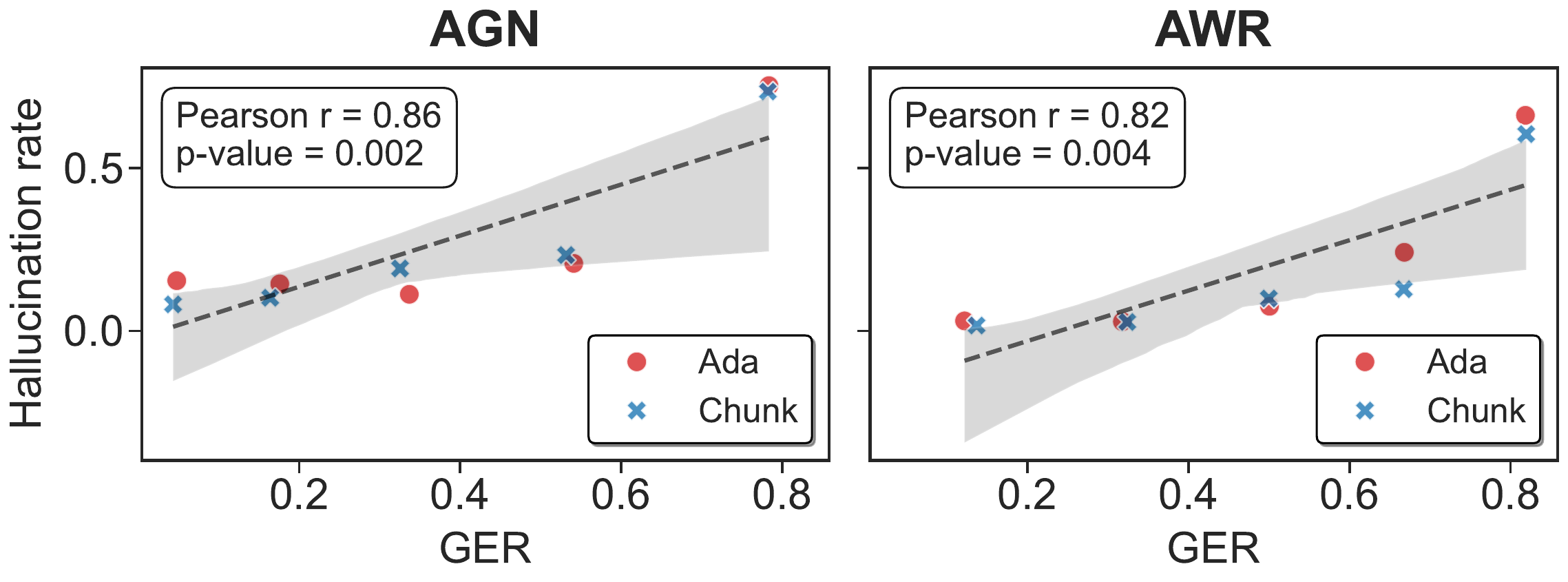}
    }\\
    \subfloat[Qwen-2.5 3B Instruct]{
    \includegraphics[width=0.8\linewidth]{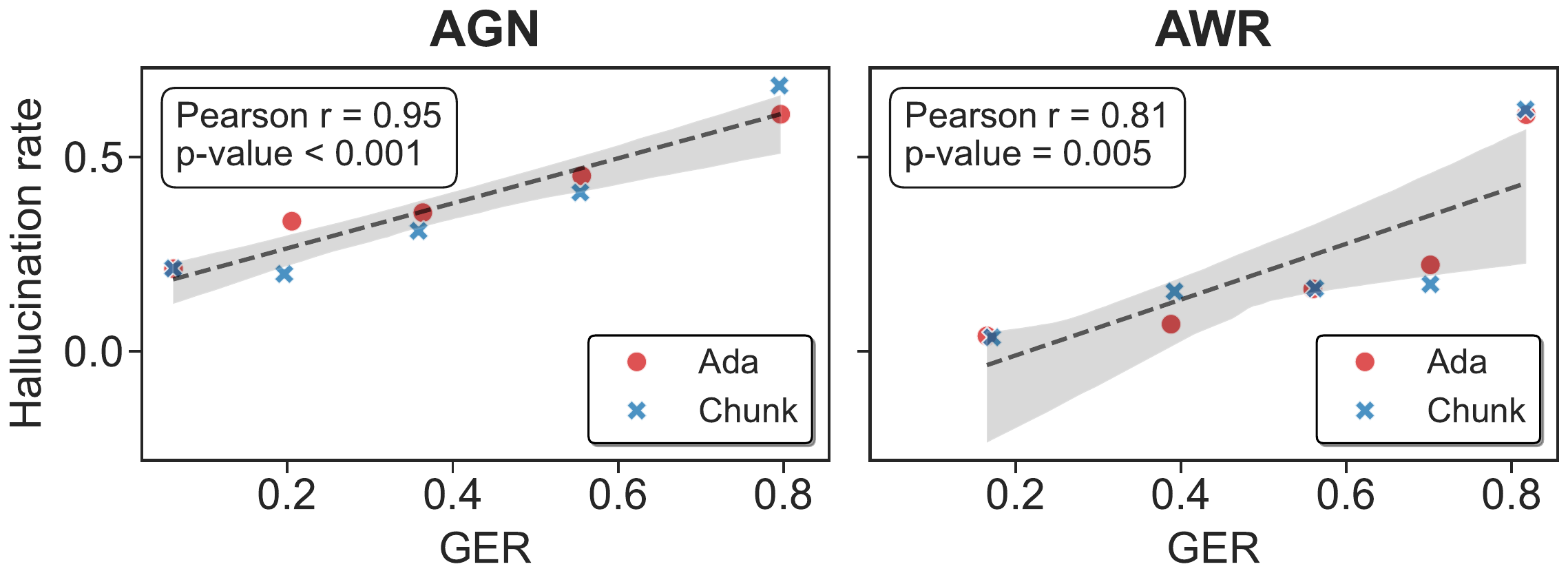}
    } \\
    \subfloat[Qwen-2.5 14B Instruct]{
    \includegraphics[width=0.8\linewidth]{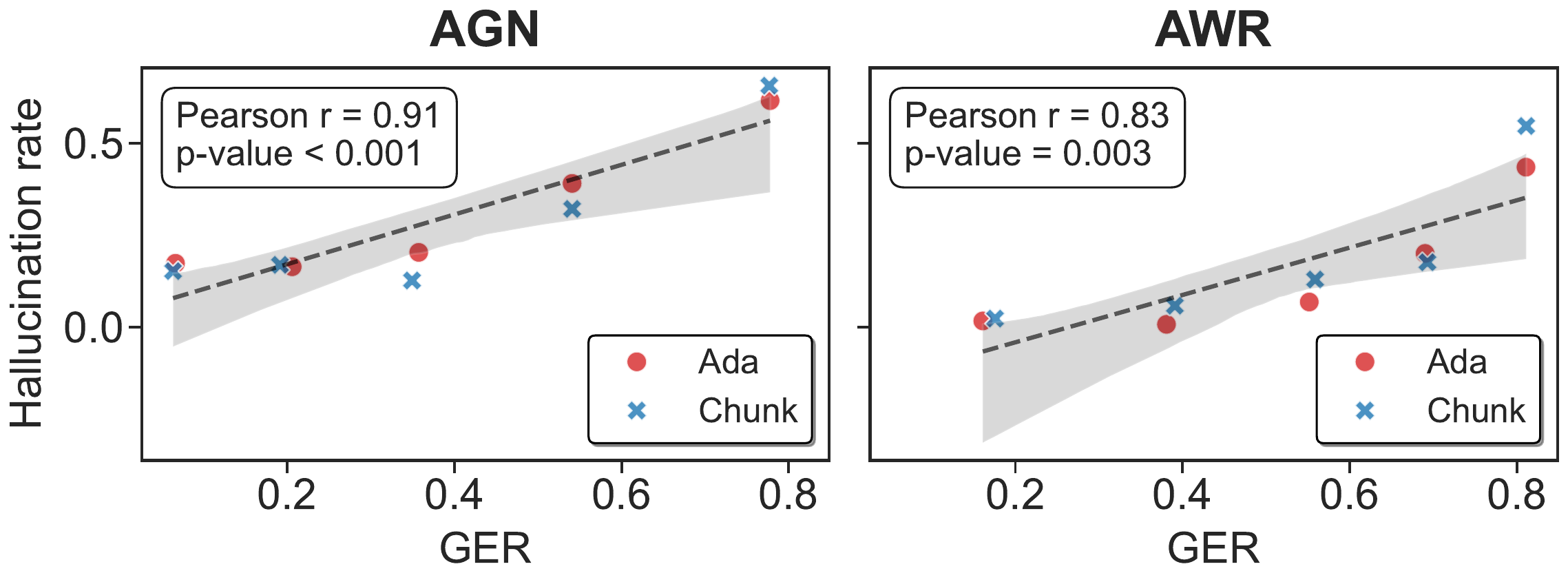}
    }
    \caption{Correlation between GER and hallucination rate for all models.}
    \label{fig:corr_ger_hal_all}
\end{figure*}

\FloatBarrier

\vskip 0.2in
\bibliography{sample}

\end{document}